\documentclass
[	
	pdftex,	 	
	paper				= a4,				
	toc					= listof,			
	toc					= bib,				
	titlepage			= true,	 			
	numbers 			= noendperiod, 		
	draft 				= false,			
	twoside 			= false,	 		
	headinclude			= true,				
	footinclude			= false,			
	fontsize 			= 11pt, 			
]
{scrreprt}

\newif\iftikzexternal
\newif\ifgerman

\newcommand{\documenttitle}{Uncertainty Calibration and its Application to Object Detection}
\newcommand{\documentauthor}{\textbf{Fabian Thomas K\"uppers}}

\usepackage{datetime}
\newdateformat{monthyeardate}{%
    \monthname[\THEMONTH] \THEYEAR
}%
\newdateformat{daymonthyeardate}{%
    \twodigit{\THEDAY}. \monthname[\THEMONTH] \THEYEAR
}

\usepackage{graphicx}
\graphicspath{{images/}}

\tikzexternaltrue

\usepackage{makecell}
\usepackage{ifthen}
\usepackage{stackengine}
\usepackage{subcaption}
\usepackage{textgreek}

\usepackage[onehalfspacing]{setspace}
\usepackage[utf8]{inputenc}
\usepackage{listings}
\usepackage{tabularx}
\usepackage{bbm}
\usepackage{abstract}
\usepackage{multirow}
\usepackage{amsmath}
\usepackage{amssymb}
\usepackage{dsfont}
\usepackage[usenames, dvipsnames]{xcolor}
\usepackage{framed}
\usepackage{booktabs}
\usepackage{acronym}
\usepackage{caption}
\usepackage{hyperref}
\hypersetup{pdfcreator={Fabian Küppers}}
\usepackage[intoc, noprefix]{nomencl}
\makenomenclature

\ifgerman
    \usepackage[ngerman]{babel}
    
\fi

\usepackage{tikz}
\usepackage{overpic}

\iftikzexternal
    \usetikzlibrary{external}
    \tikzset{external/system call={pdflatex \tikzexternalcheckshellescape -extra-mem-top=10000000 -extra-mem-bot=10000000 -halt-on-error -interaction=batchmode -jobname "\image" "\texsource"}}
\fi

\usepackage{pgfplots}
\pgfplotsset{compat=1.15}

\usepackage[
    paper=a4paper,
    left=25mm,
    right=20mm,
    top=30mm,
    bottom=40mm
]{geometry}

\usepackage[headsepline=0.5pt]{scrlayer-scrpage}
\usepackage{scrhack}

\clearpairofpagestyles

\automark{chapter}

\ohead{\rightmark}

\setkomafont{pageheadfoot}{\normalfont\sffamily\bfseries\small}

\setkomafont{pagenumber}{\normalfont\sffamily}

\makeatletter

\let\origsubsubsection\subsubsection
\renewcommand\subsubsection{\@ifstar{\starsubsubsection}{\nostarsubsubsection}}

\newcommand\nostarsubsubsection[1]
{\subsubsectionprelude\origsubsubsection{#1}\subsubsectionpostlude}

\newcommand\starsubsubsection[1]
{\subsubsectionprelude\origsubsubsection*{#1}\subsubsectionpostlude}

\newcommand\subsubsectionprelude{%
    \vspace{0em}
}

\newcommand\subsubsectionpostlude{%
    \vspace{-1em}
}

\AtBeginDocument	
{																
    			
}

\makeatother													

\hypersetup{					
    pdftitle			= 	{\documenttitle},   			
    pdfauthor			= 	{\documentauthor},     			
    pdfsubject			=	{\documenttitle},   			
    bookmarksnumbered	= 	true, 							
    colorlinks			= 	true,       					
    linkcolor			= 	Black,          				
    citecolor			= 	Black,        					
    urlcolor			= 	Black,    						
}

\ExplSyntaxOn%
\DeclareExpandableDocumentCommand{\arabicnumeral}{m}
{
    \int_from_roman:n { #1 }
}
\ExplSyntaxOff

\newcolumntype{Y}{>{\centering\arraybackslash}X}

\newcommand{\chapref}[1]{Chap.~\ref{#1}}
\newcommand{\secref}[1]{Sec.~\ref{#1}}
\newcommand{\figref}[1]{Fig.~\ref{#1}}
\newcommand{\tabref}[1]{Tab.~\ref{#1}}

\newcommand{\camocopyright}[1]{\textcopyright { #1} camo.nrw. Reprinted, with permission.}
\newcommand{\acmcopyright}[1]{\textcopyright { #1} ACM. Reprinted, with permission.}
\newcommand{\ieeecopyright}[1]{\textcopyright { #1} IEEE. Reprinted, with permission.}
\newcommand{\springercopyright}[1]{\textcopyright { #1} Springer Nature. Reprinted, with permission.}

\newcommand{\fasterrcnn}[1][]{%
	\ifthenelse{\equal{#1}{}}{\texttt{Faster R-CNN}}{\texttt{Faster R-CNN R#1-FPN}}%
}

\newcommand{\fasterrcnnx}[1][]{%
	\ifthenelse{\equal{#1}{}}{\texttt{Faster R-CNN}}{\texttt{Faster R-CNN X#1-FPN}}%
}

\newcommand{\retinanet}[1][]{%
	\ifthenelse{\equal{#1}{}}{\texttt{RetinaNet}}{\texttt{RetinaNet R#1-FPN}}%
}

\newcommand{\maskrcnn}[1][]{%
	\ifthenelse{\equal{#1}{}}{\texttt{Mask R-CNN}}{\texttt{Mask R-CNN R#1-FPN}}%
}

\newcommand{\pointrend}[1][]{\texttt{PointRend}}
\newcommand{\hrnet}[1][]{\texttt{HRNet}}
\newcommand{\deeplab}[1][]{\texttt{DeepLabv#1}}
\newcommand{\deeplabp}[1][]{\texttt{DeepLabv3+}}

\newcommand{\quantilecalibration}{quan\-tile cal\-i\-bra\-tion}
\newcommand{\quantilecalibrated}{quan\-tile-cal\-i\-bra\-ted}
\newcommand{\distributioncalibration}{dis\-tri\-bution cal\-i\-bra\-tion}
\newcommand{\distributioncalibrated}{dis\-tri\-bution-cal\-i\-bra\-ted}
\newcommand{\variancecalibration}{vari\-ance cal\-i\-bra\-tion}
\newcommand{\variancecalibrated}{vari\-ance-cal\-i\-bra\-ted}

\makeatletter
\newsavebox\myboxA
\newsavebox\myboxB
\newlength\mylenA

\newcommand*\xoverline[2][0.75]{%
    \sbox{\myboxA}{$\m@th#2$}%
    \setbox\myboxB\null
    \ht\myboxB=\ht\myboxA%
    \dp\myboxB=\dp\myboxA%
    \wd\myboxB=#1\wd\myboxA
    \sbox\myboxB{$\m@th\overline{\copy\myboxB}$}
    \setlength\mylenA{\the\wd\myboxA}
    \addtolength\mylenA{-\the\wd\myboxB}%
    \ifdim\wd\myboxB<\wd\myboxA%
    \rlap{\hskip 0.5\mylenA\usebox\myboxB}{\usebox\myboxA}%
    \else
    \hskip -0.5\mylenA\rlap{\usebox\myboxA}{\hskip 0.5\mylenA\usebox\myboxB}%
    \fi}
\makeatother

\newcommand*\circled[1]{\tikz[baseline=(char.base)]{\node[shape=circle,draw,inner sep=1pt] (char) {#1};}}
\DeclareMathOperator*{\argmin}{argmin}
\DeclareMathOperator*{\argmax}{argmax}

\definecolor{confidence}{rgb}{0.12, 0.46, 0.71}
\definecolor{ece}{rgb}{0.84, 0.15, 0.16}

\newcommand{\intdigits}{\mathbb{Z}}
\newcommand{\realdigits}{\mathbb{R}}
\newcommand{\realdigitspositive}{\mathbb{R}_{>0}}

\newcommand{\pdf}{f}
\newcommand{\cdf}{F}
\newcommand{\ppf}{F^{-1}}
\newcommand{\prob}{\mathbb{P}}
\newcommand{\probscore}{p}
\newcommand{\credibleinterval}{\hat{C}}
\newcommand{\variationaldist}{f^\ast}

\newcommand{\sampledfrom}{\sim}
\newcommand{\outerprod}{\otimes}
\newcommand{\kroneckerprod}{\outerprod}

\newcommand{\elementwiseprod}{\odot}
\newcommand{\distancemat}{\mathbf{D}}

\newcommand{\gp}{\text{gp}}
\newcommand{\kernel}{k}
\newcommand{\kernelmatrix}{\mathbf{K}}
\newcommand{\mvkernel}{\mathbf{k}}
\newcommand{\coregion}{\mathbf{B}}
\newcommand{\numinducing}{N^\ast}
\newcommand{\indexinducing}{u}
\newcommand{\gpscale}{\nu}
\newcommand{\gpbias}{\kappa}

\newcommand{\boundary}{a}
\newcommand{\boundarysec}{b}

\newcommand{\normaldistribution}{\mathcal{N}}

\newcommand{\betadistribution}{\text{Beta}}
\newcommand{\dirichletdistribution}{\text{Dir}}
\newcommand{\cauchydistribution}{\text{Cauchy}}
\newcommand{\bernoullidistribution}{\text{Bern}}
\newcommand{\categoricaldistribution}{\text{Cat}}

\newcommand{\expectation}{\mathbb{E}}
\newcommand{\distvariance}{\mathrm{Var}}
\newcommand{\distcovariance}{\mathrm{Cov}}

\newcommand{\factorial}{!}

\newcommand{\gammafunc}{\Gamma}
\newcommand{\betafunc}{\text{B}}
\newcommand{\diracfunc}{\delta}
\newcommand{\betafunca}{\alpha}
\newcommand{\betafuncb}{\beta}
\newcommand{\allbetafunca}{\boldsymbol{\betafunca}}

\newcommand{\betaparama}{a}
\newcommand{\betaparamb}{b}
\newcommand{\betaparamc}{c}
\newcommand{\allbetaparama}{\mathbf{\betaparama}}
\newcommand{\allbetaparamb}{\mathbf{\betaparamb}}

\newcommand{\betaratio}{\lambda}

\newcommand{\mean}{\mu}
\newcommand{\meanvec}{\boldsymbol{\mu}}
\newcommand{\stddev}{\sigma}
\newcommand{\variance}{\sigma^2}
\newcommand{\cov}{\boldsymbol{\Sigma}}
\newcommand{\correlation}{\rho}

\newcommand{\decomposed}{\mathbf{L}}
\newcommand{\diagonal}{\mathbf{D}}

\newcommand{\quantile}{\tau}
\newcommand{\identity}{\mathbf{I}}

\newcommand{\cauchymode}{\vartheta}
\newcommand{\cauchyscale}{\gamma}

\newcommand{\cauchyscalevec}{\boldsymbol{\cauchyscale}}

\newcommand{\T}{\top}

\newcommand{\diff}{\mathrm{d}}

\newcommand{\inputvariate}{X}
\newcommand{\outputvariate}{Y}
\newcommand{\predoutputvariate}{\hat{\outputvariate}}
\newcommand{\matchedvariate}{\hat{M}}
\newcommand{\probvariate}{P}
\newcommand{\predconfidencevariate}{\hat{P}}
\newcommand{\groundtruthconfidencevariate}{\bar{P}}
\newcommand{\bboxvariate}{R}
\newcommand{\predbboxvariate}{\hat{R}}
\newcommand{\collectvariate}{\hat{S}}
\newcommand{\distvariate}{\Pi}

\newcommand{\allinputvariates}{\mathbf{\inputvariate}}
\newcommand{\alloutputvariates}{\mathbf{\outputvariate}}

\newcommand{\allbboxvariates}{\mathbf{\bboxvariate}}
\newcommand{\allpredbboxvariates}{\mathbf{\hat{\bboxvariate}}}
\newcommand{\allcollectvariates}{\mathbf{\collectvariate}}

\newcommand{\inputset}{\mathcal{\inputvariate}}
\newcommand{\outputset}{\mathcal{Y}}

\newcommand{\probset}{[0, 1]}
\newcommand{\bboxset}{\mathcal{\bboxvariate}}

\newcommand{\distset}{\mathcal{P}}
\newcommand{\dataset}{\mathcal{D}}
\newcommand{\collectset}{\mathcal{S}}

\newcommand{\singleinput}{x}
\newcommand{\singleoutput}{y}
\newcommand{\allsingleinput}{\mathbf{\singleinput}}

\newcommand{\predoutput}{\hat{\singleoutput}}
\newcommand{\groundtruthoutput}{\bar{\singleoutput}}
\newcommand{\matched}{\hat{m}}
\newcommand{\confidence}{\probscore}
\newcommand{\predconfidence}{\hat{\probscore}}

\newcommand{\bbox}{r}
\newcommand{\predbbox}{\hat{\bbox}}
\newcommand{\allbboxes}{\mathbf{\bbox}}
\newcommand{\allpredbboxes}{\mathbf{\predbbox}}
\newcommand{\allgroundtruthbboxes}{\mathbf{\bar{\bbox}}}
\newcommand{\collect}{\hat{s}}
\newcommand{\dist}{\pi}
\newcommand{\allcollect}{\mathbf{\collect}}

\newcommand{\groundtruthvariate}{\bar{\outputvariate}}

\newcommand{\groundtruthbboxvariate}{\bar{\bboxvariate}}
\newcommand{\allgroundtruthbboxvariates}{\bar{\allbboxvariates}}
\newcommand{\singlegroundtruth}{\bar{\singleoutput}}
\newcommand{\gtset}{\outputset}
\newcommand{\singlegroundtruthbbox}{\bar{\bbox}}

\newcommand{\numcollectedvariates}{K}

\newcommand{\numpixel}{J}
\newcommand{\indexpixel}{j}
\newcommand{\pixelset}{\mathcal{\numpixel}}

\newcommand{\numobjects}{V}
\newcommand{\indexobjects}{v}
\newcommand{\objectset}{\mathcal{\numobjects}}

\newcommand{\numdims}{K}
\newcommand{\indexdims}{k}
\newcommand{\indexdimsalt}{j}
\newcommand{\numclasses}{\numdims}
\newcommand{\indexclasses}{\indexdims}

\newcommand{\numbboxdims}{L}
\newcommand{\indexbboxdims}{l}

\newcommand{\numbins}{I}
\newcommand{\indexbins}{i}
\newcommand{\bin}{B}
\newcommand{\mvbin}{\mathbf{B}}

\newcommand{\numsamples}{N}
\newcommand{\indexsamples}{n}
\newcommand{\indexset}{\mathcal{M}}

\newcommand{\numstochastic}{T}
\newcommand{\indexstochastic}{t}

\newcommand{\numhidden}{L}
\newcommand{\indexhidden}{l}

\newcommand{\numneurons}{M}

\newcommand{\numproposals}{U}
\newcommand{\indexproposals}{u}

\newcommand{\logit}{z}
\newcommand{\logitvariate}{Z}

\newcommand{\parameter}{\theta}

\newcommand{\allparameters}{\boldsymbol{\parameter}}
\newcommand{\parameterset}{\Theta}
\newcommand{\allestimatedparameters}{\hat{\allparameters}}
\newcommand{\allestimatedoutputparameters}{\hat{\boldsymbol{\vartheta}}}
\newcommand{\hbparameter}{\theta}
\newcommand{\allhbparameter}{\boldsymbol{\theta}}

\newcommand{\allpredhbparameter}{\hat{\allhbparameter}}
\newcommand{\variationalparameter}{\boldsymbol{\omega}}

\newcommand{\learningrate}{\eta}
\newcommand{\layerfunction}{\mathcal{V}}
\newcommand{\convolutionoutput}{o}

\newcommand{\convolutionwidth}{W_k}
\newcommand{\convolutionheight}{H_k}

\newcommand{\loss}{\mathcal{L}}
\newcommand{\entropy}{\mathcal{H}}
\newcommand{\kldivergence}{D_{\text{KL}}}

\newcommand{\nees}{\epsilon}

\newcommand{\loglikelihoodratio}{\ell r}

\newcommand{\complexity}{\mathcal{O}}

\newcommand{\model}{d}
\newcommand{\calmodel}{h}
\newcommand{\linkfunction}{h'}

\newcommand{\calibrated}{\hat{q}}
\newcommand{\calibratedvariate}{\hat{Q}}

\newcommand{\pdfcalibrated}{g}
\newcommand{\cdfcalibrated}{G}

\newcommand{\diag}{\text{diag}}
\newcommand{\determinant}{\text{det}}
\newcommand{\ind}{\mathds{1}}

\newcommand{\centerx}{c_x}
\newcommand{\centery}{c_y}
\newcommand{\width}{w}
\newcommand{\height}{h}
\newcommand{\channels}{c}
\newcommand{\Width}{W}
\newcommand{\Height}{H}
\newcommand{\Channels}{C}

\newcommand{\sigmoid}{\Phi}
\newcommand{\softmax}{\Phi_{\text{SM}}}
\newcommand{\scaleweight}{w}
\newcommand{\scalevec}{\mathbf{w}}
\newcommand{\scalebias}{\delta}
\newcommand{\scaleweightmat}{\mathbf{W}}
\newcommand{\scaleweightset}{\mathcal{W}}

\newcommand{\ece}{\text{ECE}}
\newcommand{\dece}{\text{D-ECE}}

\newcommand{\mqce}{\text{M-QCE}}
\newcommand{\cqce}{\text{C-QCE}}
\newcommand{\uce}{\text{UCE}}
\newcommand{\ence}{\text{ENCE}}
\newcommand{\picp}{\text{PICP}}

\newcommand{\acc}{\text{acc}}

\newcommand{\precision}{\text{prec}}
\newcommand{\conf}{\text{conf}}

\newcommand{\positive}{+}
\newcommand{\negative}{-}
\newcommand{\meansub}{\eta}
\newcommand{\scalesub}{\gamma}

\newcommand{\hiddenstate}{\widetilde{\mathbf{c}}}
\newcommand{\hiddenstatevariate}{\widetilde{\mathbf{C}}}
\newcommand{\hiddenstateset}{\mathcal{C}}

\newcommand{\hiddenbbox}{\widetilde{\allbboxes}}
\newcommand{\hiddenbboxvariate}{\widetilde{\allbboxvariates}}

\newcommand{\numsteps}{T}
\newcommand{\timestep}{t}

\newcommand{\trackconfidencevariate}{\widetilde{P}}

\newcommand{\trackmatchedvariate}{\widetilde{M}}
\newcommand{\trackmatched}{\widetilde{m}}

\newcommand{\observationtransition}{\mathbf{H}}
\newcommand{\observationnoise}{\boldsymbol{\varrho}}
\newcommand{\observationnoisecov}{\boldsymbol{\Lambda}}

\newcommand{\statetransition}{\mathbf{F}}
\newcommand{\statenoise}{\boldsymbol{\varepsilon}}
\newcommand{\statenoisecov}{\boldsymbol{\Psi}}

\newcommand{\numstatedims}{\numbboxdims^{\ast}}

\begin{document}


\pagenumbering{Roman}

\thispagestyle{empty}

\begin{titlepage}{%
	\begin{minipage}[t]{0.5\textwidth}
	\flushleft{\includegraphics[height = 2cm]{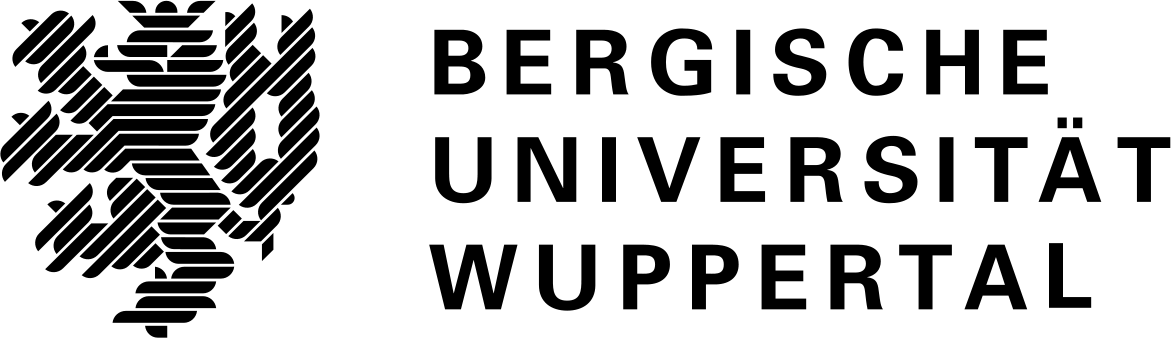}}
	\end{minipage}%
	\begin{minipage}[t]{0.5\textwidth}
		\flushright{\includegraphics[height = 2cm]{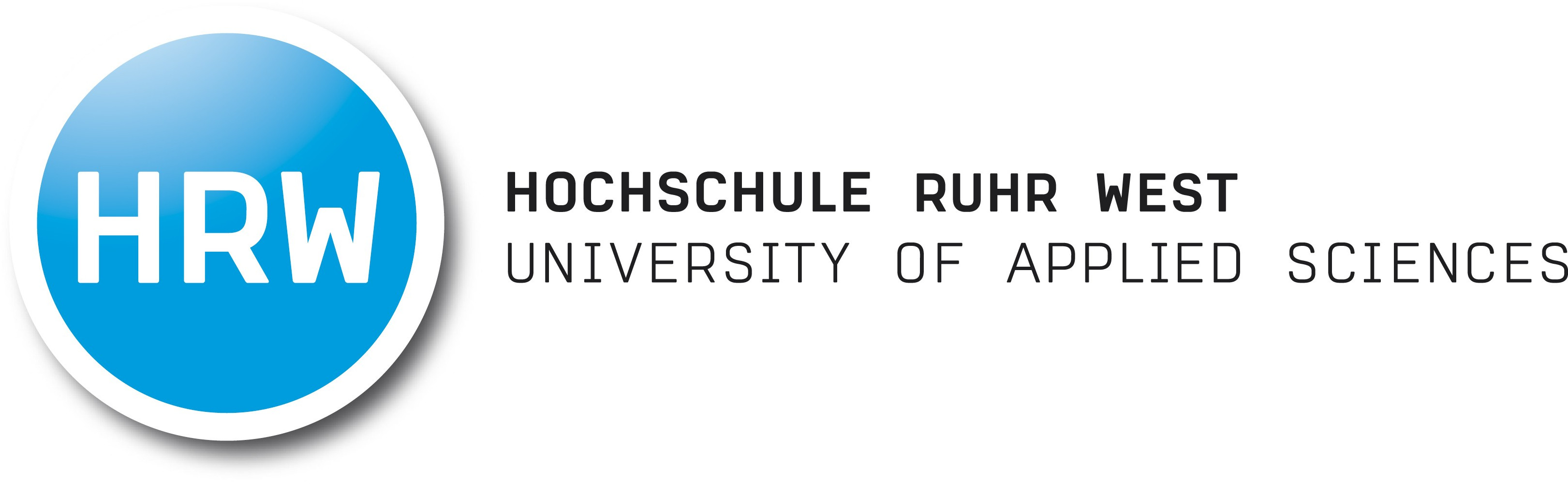}}
	\end{minipage}%
	
	\centering
	\vspace{1.0cm}
	
	\begingroup
	\fontsize{22pt}{22pt}\selectfont
	\textbf{\documenttitle}
	
	\endgroup
	
	\vspace{5em}
	
	\begingroup
	\fontsize{18pt}{16pt}\selectfont
	\textbf{DISSERTATION}\\

    \fontsize{14pt}{16pt}\selectfont
	\vspace{3em}
	zur Erlangung des Grades\\
	\textbf{Doctor rerum naturalium} \\
	\vspace{0.5em}
    im Fach \textbf{Informatik} \\
	an der \textbf{Fakultät für Mathematik und Naturwissenschaften} \\
	der \textbf{Bergischen Universität Wuppertal}
	
	\vspace{1cm}
	von\\
	
	\begingroup
	\fontsize{15pt}{16pt}\selectfont
	\documentauthor\\
	\endgroup
	
	\vspace{2cm}
	
	\endgroup
	\begingroup
	\fontsize{13pt}{15pt}\selectfont
	
	\begin{tabularx}{0.95\textwidth}{X X}
	Erstprüfer & Prof. Dr. rer. nat. Hanno Gottschalk \\
	& Bergische Universität Wuppertal\\
	& \\
	Zweitprüfer & Prof. Dr.-Ing. Anselm Haselhoff\\
	& Hochschule Ruhr West
	\end{tabularx}
	
	\vspace{\fill}
	
	Mülheim an der Ruhr, Januar 2023
	
	\endgroup

}\end{titlepage}

\newpage

\ifgerman
\else
    \chapter*{Affidavit}
\fi
\thispagestyle{empty}

\vspace{3cm}

\parbox{12cm}{%
    \leftskip1cm
    \ifgerman
\else
        I hereby declare that I have\\
        written this thesis by myself,\\
        only using the sources and\\
        materials as indicated.
    \fi
}

\vspace{3cm}
\setlength{\unitlength}{1cm}

\begin{picture}(16,1)
   \put(1,0){\line(1,0){7}}
\end{picture}

\par

\parbox{12cm}
{
   \leftskip1.0cm
   Mülheim an der Ruhr, \daymonthyeardate\today\\
}

\clearpage

\setcounter{page}{1}


\chapter*{Abstract}
Image-based environment perception is an important component especially for driver assistance systems or autonomous driving.
In this scope, modern neuronal networks are used to identify multiple objects as well as the according position and size information within a single frame.
The performance of such an object detection model is important for the overall performance of the whole system.
However, a detection model might also predict these objects under a certain degree of uncertainty.
In this context, we distinguish between epistemic model uncertainty (e.g., due to a lack of knowledge) and aleatoric data uncertainty (e.g., due to random effects in the input data).
These uncertainties might have a large impact on the safety of the whole system.

In this work, we examine the semantic uncertainty (which object type?) as well as the spatial uncertainty (where is the object and how large is it?).
We evaluate if the predicted uncertainties of an object detection model match with the observed error that is achieved on real-world data.
In the first part of this work, we introduce the definition for confidence calibration of the semantic uncertainty in the context of object detection, instance segmentation, and semantic segmentation.
We integrate additional position information in our examinations to evaluate the effect of the object's position on the semantic calibration properties.
Besides measuring calibration, it is also possible to perform a post-hoc recalibration of semantic uncertainty that might have turned out to be miscalibrated.
Thus, we derive new methods to perform a position-dependent recalibration of possibly uncalibrated uncertainty information.
Based on these methods, we introduce the concept of Bayesian confidence calibration which allows to determine the epistemic uncertainty that is inherent in the recalibration methods itself.
In this way, it is possible to add a new kind of uncertainty to the overall perception chain.

The second part of this work deals with the spatial uncertainty obtained by a probabilistic detection model.
In the scope of regression uncertainty, several definitions for the term of calibration exists which we review and relate to each other within this work.
Similar to semantic confidence calibration, it is possible to apply a post-hoc calibration of the spatial uncertainty.
We review and extend common calibration methods so that it is possible to obtain parametric uncertainty distributions for the position information in a more flexible way.

In the last part, we demonstrate a possible use-case for our derived calibration methods in the context of object tracking.
In contrast to object detection, an object tracker seeks to identify the same objects over a sequence of images over time.
We integrate our previously proposed calibration techniques and demonstrate the usefulness of semantic and spatial uncertainty calibration in a subsequent process.
We can show that uncertainty calibration leads to a significantly improved object tracking.

In conclusion, in this work we show that common object detection models tend to be miscalibrated.
This holds for the semantic as well as for the spatial uncertainty.
Our new calibration methods are useful techniques to correct miscalibrated uncertainty estimates and have shown to be a valuable contribution to the overall environment perception process.


\chapter*{Preface \& Acknowledgments}
The use of the 1st person plural has been the familiar style for writing my previous scientific publications and is also used throughout this thesis, even though the majority of the work has been carried out by myself.
This work was conducted in the context of the research project ``KI-Absicherung - Safe AI for Automated Driving'' at the Ruhr West University of Applied Sciences, Bottrop, Germany in cooperation with Elektronische Fahrwerksysteme GmbH, Gaimersheim, Germany, and Visteon Electronics Germany GmbH, Kerpen, Germany.

First and foremost, I would like to gratefully thank Prof. Dr.-Ing. Anselm Haselhoff from the Ruhr West University of Applied Sciences for giving me the opportunity to work at the Ruhr West University within the ``KI-Absicherung'' project, for his extensive support, and the fruitful discussions during the last three years.
Furthermore, I would like to thank our partner Elektronische Fahrwerksysteme GmbH and especially Jonas Schneider for the excellent collaboration.
We started the project ``KI-Absicherung'' in cooperation with Visteon Elektronics under the leadership of Dr. Andreas Wedel, whom I would also like to thank.
Finally, I would like to thank Prof. Dr. Hanno Gottschalk from the University of Wuppertal for his support and for the opportunity to graduate at the University of Wuppertal.

\vspace{3em}
Ich möchte mich bei meinen Freunden, meiner Familie und Kollegen für die großartige Unterstützung in meiner Promotionszeit bedanken.
Die Gespräche und die mir entgegengebrachte Unterstützung hat mir sehr geholfen, auch schwierigere Phasen in der Promotionszeit zu überstehen.
Ein großes Dankeschön geht an Helena, die sich bereiterklärt hat, mein ``yellow-from-the-egg''-Englisch zu korrigieren.

Mein langjähriges Ziel, die Promotion abzuschließen, erfordert ein hohes Maß an Durchhaltevermögen.
Ich möchte daher ausdrücklich Nathalie dafür danken, dass sie dieses Durchhaltevermögen mit mir hat.
Ohne deine Unterstützung wäre ich nicht so weit gekommen.

\chapter*{Publication Overview}
\begin{enumerate}
    \item Küppers, et al.: ``Multivariate Confidence Calibration for Object Detection,'' in \textit{Proceedings of the IEEE/CVF Conference on Computer Vision and Pattern Recognition Workshops}, 2020 \cite{Kueppers2020}.
    \item Schwaiger, et al.: ``From Black-box to White-box: Examining Confidence Calibration under different Conditions,'' in \textit{Proceedings of the Workshop on Artificial Intelligence Safety 2021 (SafeAI 2021) co-located with the Thirty-Fifth AAAI Conference on Artificial Intelligence (AAAI 2021)}, 2021 \cite{Schwaiger2021}.
    \item Küppers, et al.: ``Bayesian Confidence Calibration for Epistemic Uncertainty Modelling,'' in \textit{Pro\-ceedings of the 2021 IEEE Intelligent Vehicles Symposium (IV)}, 2021 \cite{Kueppers2021}.
    \item Küppers, et al.: ``Confidence Calibration for Object Detection and Segmentation,'' in: Fingscheidt, Houben, and Gottschalk (eds.): \textit{Deep Neural Networks and Data for Automated Driving - Robustness, Uncertainty Quantification, and Insights Towards Safety}, Springer Nature Switzerland, pp.~225-250, 2022 \cite{Kueppers2022a}.
    \item Küppers, et al.: ``Calibration of Neural Networks for Detection Models (German original: Kalibrierung von Neuronalen Netzen für Detektionsmodelle),'' in: VDI Wissensforum GmbH (eds.): \textit{Fahrer\-as\-sis\-tenz\-sys\-te\-me und automatisiertes Fahren}, VDI Verlag, pp.~49-69, 2022 \cite{Kueppers2022c}.
    \item Küppers, et al.: ``Parametric and Multivariate Uncertainty Calibration for Regression and Object Detection,'' in \textit{European Conference on Computer Vision Workshops}, Springer, 2022, \textit{in press} \cite{Kueppers2022b}.
\end{enumerate}

\tableofcontents \clearpage
\listoffigures \clearpage
\listoftables \clearpage

\ifgerman
\else
    \chapter*{List of Acronyms}
    \addcontentsline{toc}{chapter}{List of Acronyms}
\fi

\begin{acronym}[LONGEST]
	\acro{PDF}{Probability Density Function}
    \acro{PMF}{Probability Mass Function}
	\acro{CDF}{Cumulative Density Function}
	\acro{ECDF}{Empirical Cumulative Density Function}
	\acro{HPDI}{Highest Posterior Density Interval}
	\acro{HPDR}{Highest Posterior Density Region}
	
	\acro{ERM}{Empirical Risk Minimization}
	\acro{MLE}{Maximum Likelihood Estimation}
	\acro{MAP}{Maximum a Posteriori}
	\acro{SGD}{Stochastic Gradient Descent}
	\acro{SVI}{Stochastic Variational Inference}
	\acro{ELBO}{Evidence Lower Bound}
	\acro{MCMC}{Markov-Chain Monte-Carlo}
	\acro{IoU}{Intersection over Union}
	\acro{mIoU}{Mean Intersection over Union}
	\acro{mAP}{Mean Average Precision}
	\acro{AUPRC}{Area under Precision-Recall Curve}
	
	\acro{ReLU}{Rectified Linear Unit}
	\acro{CNN}{Convolutional Neural Network}
	\acro{RPN}{Region Proposal Network}
	\acro{FPN}{Feature Pyramid Network}
	
	\acro{ECE}{Expected Calibration Error}
	\acro{MCE}{Maximum Calibration Error}
	\acro{D-ECE}{Detection Expected Calibration Error}
	\acro{MMCE}{Maximum Mean Calibration Error}
	\acro{NLL}{Negative Log Likelihood}
    \acro{MSE}{Mean Squared Error}
    \acro{M-QCE}{Marginal Quantile Calibration Error}
	\acro{C-QCE}{Conditional Quantile Calibration Error}
	\acro{UCE}{Uncertainty Calibration Error}
	\acro{ENCE}{Expected Normalized Calibration Error}
    \acro{PICP}{Prediction Interval Coverage Probability}
    \acro{MPIW}{Mean Prediction Interval Width}
    \acro{SGV}{Standardized Generalized Variance}
    \acro{NEES}{Normalized Estimation Error Squared}
    \acro{NIS}{Normalized Innovations Squared}
    \acro{MOT}{Multiple Object Tracking}
	
    \acro{GP}{Gaussian Process}
    \acro{RBF}{Radial Basis Function}
    \acro{BBQ}{Bayesian Binning into Quantiles}
    \acro{ENIR}{Ensemble of Near Isotonic Regression}
\end{acronym}
 \clearpage
\renewcommand\nomgroup[1]{%
  \item[\bfseries
  \ifstrequal{#1}{A}{Statistical Symbols}{%
  \ifstrequal{#1}{C}{Probability Distributions}{%
  \ifstrequal{#1}{B}{Functions}{%
  \ifstrequal{#1}{N}{Notation for used Random Variables and their Realizations}{%
  \ifstrequal{#1}{O}{Tracking-specific Symbols}{%
  \ifstrequal{#1}{P}{Miscellaneous Symbols}{}}}}}}%
]}

\nomenclature[Aa]{$\inputvariate, \outputvariate$}{Random variables}
\nomenclature[Ab]{$\allinputvariates, \alloutputvariates$}{Multivariate random variables $\allinputvariates = (\inputvariate_1, \ldots, \inputvariate_\numdims)^\T$ and $\alloutputvariates = (\outputvariate_1, \ldots, \outputvariate_\numdims)^\T$ of size $\numdims$}
\nomenclature[Ad]{$\prob(\inputvariate=\singleinput)$}{Probability for $\inputvariate$ being $\singleinput$}
\nomenclature[Ae]{$\expectation[\inputvariate]$}{Expectation of $\inputvariate$}
\nomenclature[Af]{$\distvariance[\inputvariate]$}{Variance of $\inputvariate$}
\nomenclature[Ag]{$\distcovariance[\inputvariate, \outputvariate]$}{Covariance of $\inputvariate$ and $\outputvariate$}
\nomenclature[Ah]{$\mean_\inputvariate$}{Mean of $\inputvariate$}
\nomenclature[Ai]{$\variance_\inputvariate$, $\stddev_\inputvariate$}{Variance and standard deviation of $\inputvariate$}
\nomenclature[Aj]{$\cov_{\inputvariate,\outputvariate}$}{Covariance matrix of $\inputvariate$ and $\outputvariate$}
\nomenclature[Ak]{$\pdf_\inputvariate(\singleinput)$}{Probability density function (PDF) of $\inputvariate$}
\nomenclature[Al]{$\cdf_\inputvariate(\singleinput)$}{Cumulative density function (CDF) of $\inputvariate$}
\nomenclature[Al]{$\ppf_\inputvariate(\quantile)$}{Percent point function (inverse of CDF) of $\inputvariate$ for any quantile $\quantile \in (0, 1)$}

\nomenclature[Ba]{$\gammafunc(\betafunca)$}{Gamma function: $\gammafunc(\betafunca)=(\betafunca-1)\factorial$ for $\betafunca \in \realdigitspositive$}
\nomenclature[Bb]{$\betafunc(\betafunca, \betafuncb)$}{Beta function: $\betafunc(\betafunca, \betafuncb) = \frac{\gammafunc(\betafunca)\gammafunc(\betafuncb)}{\gammafunc(\betafunca + \betafuncb)}$ for $\betafunca, \betafuncb \in \realdigitspositive$}
\nomenclature[Bc]{$\betafunc(\allbetafunca)$}{Multivariate beta function: $\betafunc(\allbetafunca) = \frac{\prod^\numdims_{\indexdims=1}\gammafunc(\betafunca_\indexdims)}{\gammafunc(\sum^\numdims_{\indexdims=1}\betafunca_\indexdims)}$ for $\allbetafunca \in \realdigitspositive^{\numdims}$ with $\numdims$ dimensions}
\nomenclature[Bd]{$\diracfunc(\singleinput)$}{Dirac delta function: $\diracfunc(\singleinput) = +\infty$  if $\singleinput = 0$ else $\diracfunc(\singleinput) = 0$}
\nomenclature[Be]{$\sigmoid(\singleinput)$}{Sigmoid function: $\sigmoid(\singleinput) = \frac{\exp(\singleinput)}{1 + \exp(\singleinput)} = \frac{1}{1 + \exp(-\singleinput)}$ with $\singleinput \in \realdigits$}
\nomenclature[Bf]{$\softmax(\singleinput_\indexdims)$}{Softmax function: $\softmax(\singleinput_\indexdims) = \frac{\exp(\singleinput_\indexdims)}{\sum_{\indexdims^\ast=1}^{\numdims} \exp(\singleinput_{\indexdims^\ast})}$ with $\singleinput_\indexdims \in \realdigits$ and $\numdims$ dimensions}
\nomenclature[Bg]{$\ind(\cdot)$}{Indicator function. Returns $1$ if argument is true, else $0$}
\nomenclature[Bh]{$\diag(\singleinput_1, \ldots, \singleinput_\numdims)$}{Diagonal matrix operator with diagonal elements $\singleinput_1, \ldots, \singleinput_\numdims$.}
\nomenclature[Bi]{$\determinant(\cov)$}{Determinant of matrix $\cov$}

\nomenclature[Ca]{$\bernoullidistribution(\singleinput; \parameter)$}{Bernoulli distribution: $\bernoullidistribution(\singleinput; \parameter) = \parameter^\singleinput(1-\parameter)^{1-\singleinput}$ with $\singleinput \in \{0, 1\}$ and $\parameter \in [0, 1]$}
\nomenclature[Cb]{$\categoricaldistribution(\singleinput; \allparameters)$}{Categorical distribution: $\categoricaldistribution(\singleinput; \allparameters) = \prod_{\indexdims=1}^{\numdims} \parameter_\indexdims^{\ind(\singleinput = \indexdims)}$ with $\singleinput \in \{1, \ldots, \numdims\}$ and $\parameter_\indexdims \in [0, 1]$}
\nomenclature[Cb]{$\normaldistribution (\singleinput; \mean, \variance)$}{Univariate normal distribution: $\normaldistribution (\singleinput; \mean, \variance) = \frac{1}{\sqrt{2\pi\variance}} \exp\Big(\frac{-1}{2\variance}(\singleinput-\mean)^2\Big)$ where $\variance \in \realdigitspositive$}
\nomenclature[Cc]{$\normaldistribution(\allsingleinput ; \meanvec, \cov)$}{Mv. normal distribution: $\normaldistribution(\allsingleinput ; \meanvec, \cov) = \frac{1}{\sqrt{(2\pi)^\numdims \text{det}(\cov)}} \exp\Big(-\frac{1}{2} (\allsingleinput-\meanvec)^\T\cov^{-1}(\allsingleinput-\meanvec)\Big)$}
\nomenclature[Cd]{$\betadistribution(\singleinput ; \betafunca, \betafuncb)$}{Beta distribution: $\betadistribution(\singleinput ; \betafunca, \betafuncb) = \frac{1}{\betafunc(\betafunca, \betafuncb)}\singleinput^{\betafunca-1}(1-\singleinput)^{\betafuncb-1}$ where $\singleinput \in [0,1]$ and $\betafunca, \betafuncb \in \realdigitspositive$}
\nomenclature[Ce]{$\dirichletdistribution(\allsingleinput ; \allbetafunca)$}{Dirichlet distribution: $\dirichletdistribution(\allsingleinput ; \allbetafunca) = \frac{1}{\betafunc(\allbetafunca)} \prod^\numdims_{\indexdims=1} \singleinput_\indexdims^{\betafunca_\indexdims-1}$ for $\allbetafunca \in \realdigitspositive^{\numdims}$ with $\numdims$ dimensions}
\nomenclature[Cf]{$\cauchydistribution(\singleinput; \cauchymode, \cauchyscale)$}{Cauchy distribution: $\cauchydistribution(\singleinput; \cauchymode, \cauchyscale) = \frac{1}{\pi}\frac{\cauchyscale}{\cauchyscale^2 + (\singleinput - \cauchymode)^2}$ where $\cauchyscale \in \realdigitspositive$}

\nomenclature[Na]{$\allinputvariates, \allsingleinput \in \inputset$}{Input image in space $\inputset$}
\nomenclature[Nb]{$\outputvariate, \singleoutput \in \outputset$}{Label of an object in $\outputset = \{1, \ldots, \numclasses\}$ with $\numclasses$ classes}
\nomenclature[Nca]{$\predoutputvariate, \predoutput \in \outputset$}{Predicted label of an object in $\outputset$ obtained by an object detector}
\nomenclature[Ncb]{$\predoutputvariate^\ast_\indexpixel, \predoutput^\ast_\indexpixel \in \outputset$}{Predicted label for each pixel $\indexpixel \in \pixelset$ obtained by an instance/semantic segmentation model}
\nomenclature[Nda]{$\groundtruthvariate, \groundtruthoutput \in \outputset$}{Ground-truth label of an object in $\outputset$}
\nomenclature[Ndb]{$\groundtruthvariate^\ast_\indexpixel, \groundtruthoutput^\ast_\indexpixel \in \outputset$}{Ground-truth label for each pixel $\indexpixel \in \pixelset$}
\nomenclature[Ne]{$\probvariate, \confidence \in \probset$}{Semantic confidence information of an object}
\nomenclature[Nfa]{$\predconfidencevariate, \predconfidence \in \probset$}{Predicted confidence information of an object obtained by an object detector}
\nomenclature[Nfb]{$\predconfidencevariate^\ast_\indexpixel, \predconfidence^\ast_\indexpixel \in \probset$}{Predicted confidence information for each pixel $\indexpixel \in \pixelset$ obtained by an instance/semantic segmentation model}
\nomenclature[Nfc]{$\calibratedvariate, \calibrated \in \probset$}{Calibrated confidence information of an object obtained by a calibration function}
\nomenclature[Ng]{$\allbboxvariates, \allbboxes \in \bboxset$}{Position information of an object in $\bboxset$, commonly with $\allbboxvariates = (\centerx, \centery, \width, \height)^\T$ where $\centerx$, $\centery$ are the center $x$ and $y$ position and $\width$, $\height$ are the width and height, respectively, so that $\bboxset = \realdigits^\numbboxdims$ with $\numbboxdims$ as the size of the box encoding}
\nomenclature[Nha]{$\allpredbboxvariates, \allpredbboxes \in \bboxset$}{Predicted position information of an object in $\bboxset$ obtained by an object detector}
\nomenclature[Nhb]{$\allbboxvariates^\ast_\indexpixel, \allbboxes \in \bboxset^\ast$}{Position information for each pixel $\indexpixel \in \pixelset$ where $\bboxset^\ast = \realdigits^{\numbboxdims^\ast}$ with $\numbboxdims^\ast$ as the size of the position encoding for each pixel}
\nomenclature[Ni]{$\hiddenbboxvariate, \hiddenbbox \in \bboxset$}{Predicted position information of an object in $\bboxset$ obtained by an object tracker}
\nomenclature[Ni]{$\allgroundtruthbboxvariates, \allgroundtruthbboxes \in \bboxset$}{Ground-truth position information of an object in $\bboxset$}
\nomenclature[Nk]{$\matchedvariate, \matched \in \{0, 1\}$}{Indicator variable that a predicted object matches a ground-truth object}
\nomenclature[Nl]{$\allcollectvariates, \allcollect \in \collectset$}{Aggregated detector output in $\collectset$ with $\allcollectvariates = (\predconfidencevariate, \predoutputvariate, \allpredbboxvariates)^\T$}
\nomenclature[Nk]{$\logitvariate, \logit \in \realdigits$}{Logit of an object detector (output before sigmoid/softmax)}

\nomenclature[Oa]{$\observationtransition \in \realdigits^{\numbboxdims \times \numbboxdims^\ast}$}{Observation matrix to translate from tracker state space to detection space during Kalman filtering, with detector box size $\numbboxdims$ and tracker state size $\numbboxdims^\ast$}
\nomenclature[Ob]{$\statetransition \in \realdigits^{\numbboxdims^\ast \times \numbboxdims^\ast}$}{State transition matrix to predict the proceeding state during Kalman filtering}
\nomenclature[Oc]{$\observationnoise \sampledfrom \normaldistribution(0, \observationnoisecov)$}{Static observation/measurement noise with zero mean and covariance matrix $\observationnoisecov \in \realdigits^{\numbboxdims \times \numbboxdims}$}
\nomenclature[Od]{$\statenoise \sampledfrom \normaldistribution(0, \statenoisecov)$}{Static system noise with zero mean and covariance matrix $\statenoisecov \in \realdigits^{\numbboxdims^\ast \times \numbboxdims^\ast}$}

\nomenclature[Pa]{$\dataset$}{Data set with $\numsamples$ samples}
\nomenclature[Pb]{$\calmodel(\cdot)$}{Calibration function for semantic confidence or spatial uncertainty calibration}
\nomenclature[Pc]{$\bin_\indexbins$}{Single bin used within a binning scheme for Histogram Binning or ECE calculation with $\indexbins \in \{1, \ldots, \numbins\}$ where $\numbins$ is the total number of bins}
\nomenclature[Pda]{$\scaleweight \in \realdigitspositive$}{Scale weight used for calibration methods such as Logistic Calibration, Variance Scaling, or GP-Normal}
\nomenclature[Pdb]{$\scalevec\in \realdigitspositive^\numcollectedvariates$}{Scale weight vector used for multivariate calibration methods such as position-dependent Logistic Calibration or multivariate GP-Normal with $\numcollectedvariates$ dimensions}
\nomenclature[Pe]{$\scalebias \in \realdigits$}{Scale bias used for calibration methods such as Logistic Calibration or Beta Calibration}
\nomenclature[Pf]{$\gp\big(0, \kernel(\cdot, \cdot), \coregion\big)$}{Gaussian Process with zero mean, kernel function $\kernel(\cdot, \cdot)$ and coregionalization matrix $\coregion$}

\printnomenclature[2cm]%
\label{last-roman-page}%
\clearpage

\pagenumbering{arabic}


\chapter{Introduction}
\label{chapter:introduction}

\begin{figure}[b!]
    \centering
    \begin{overpic}[width=1.0\textwidth, tics=5, trim=0 -5em 0 -7.5em]{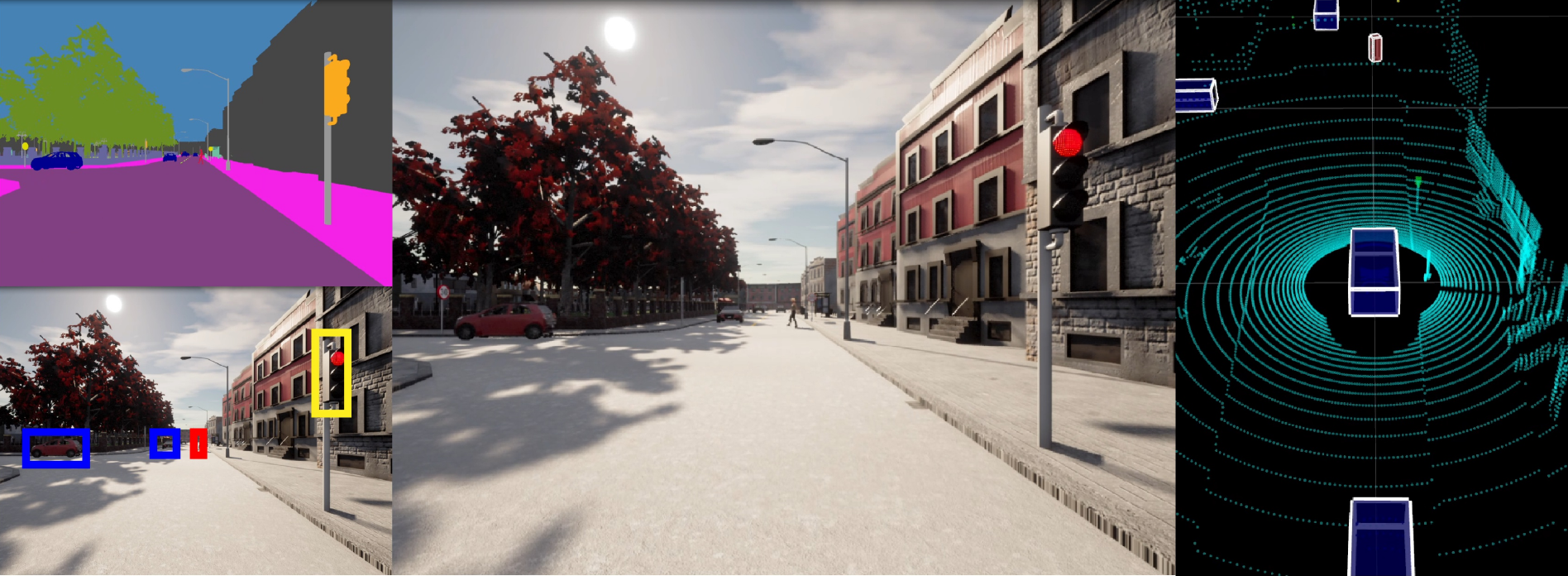}
        \put(1.5, 2){\scriptsize \textbf{Camera-based object detection}}
        \put(4, 42.5){\scriptsize \makecell[c]{\textbf{Camera-based}\\\textbf{semantic segmentation}}}
        \put(34.5, 42.5){\small \textbf{Scenario from camera perspective}}
        \put(80, 42.5){\scriptsize \makecell[c]{\textbf{LiDAR-based}\\\textbf{environment perception}}}
    \end{overpic}
    
    \vspace{-1em}
    \raggedleft{\scriptsize\camocopyright{2022}}
    
    \caption[Environment perception of an intelligent vehicle using different sensors and AI-based perception systems (simluated example).]{
        Environment perception of an intelligent vehicle using different sensors and AI-based perception systems (simluated example) \footnotemark.
        In this work, we focus on the camera-based environment perception with focus on the object detection task.
    }
    \label{fig:introduction:perception}
\end{figure}
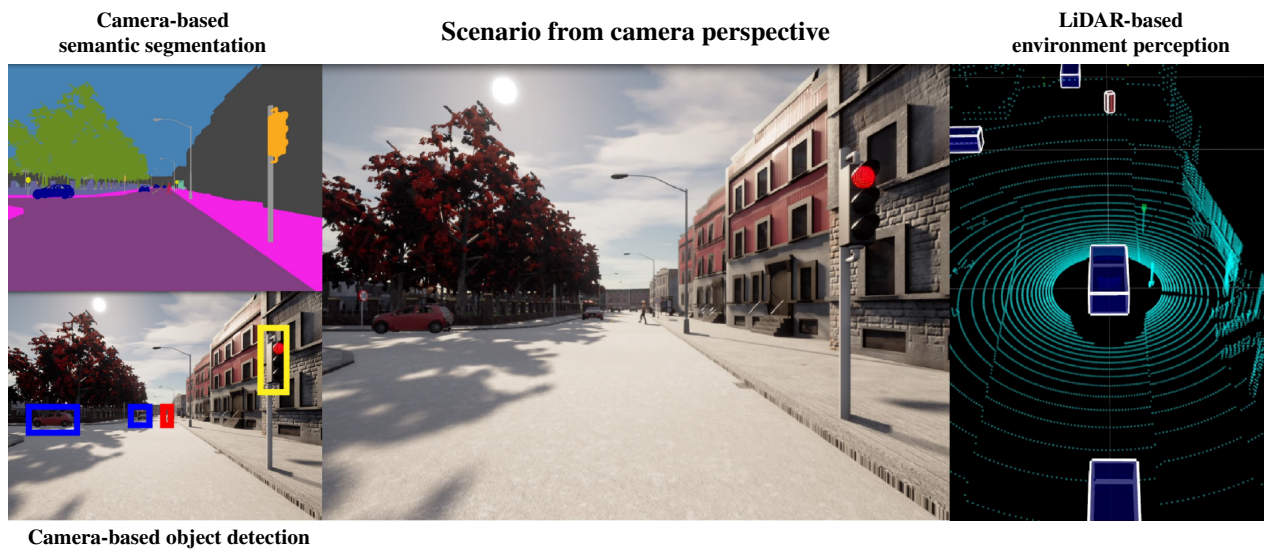

Machine learning models and especially neural networks are used in many applications nowadays, e.g., for image classification \cite{Sultana2018} or object detection \cite{He2016,Lin2017}.
Besides the advantages of using machine learning models, there are also some challenges that users and developers have to face.
In particular, the explainability \cite{Haselhoff2021} as well as the reliability \cite{Niculescu2005,Naeini2015,Guo2018} are relevant issues especially in the context of safety-critical applications such as autonomous driving \cite{Yang2018,Feng2019,Feng2021} or medical diagnosis \cite{Jiang2011,Laves2020,Mehrtash2020}.
In this scope, it is mandatory to be able to comprehend the decision-making process of a neural network especially within self-monitoring procedures to be able to identify critical situations or even induce appropriate fallback solutions.
If requested, a neural network will always return a decision on each input, regardless of its own uncertainty.
Especially in the case of safety-relevant decisions, it is necessary that the model can output a reliable self-assessment of its own uncertainty so that, in case of doubt, an adequate fallback solution can be applied, e.g., the decision-making can be handed over to a human.
In the scope of driver assistance systems and autonomous driving, environment perception is an important part to interact with the surrounding area.
Such a vehicle commonly uses multiple kinds of sensors such as cameras \cite{Banerjee2018}, radar \cite{Dickmann2016}, or LiDAR \cite{Lidar2020} to construct an environment model.
This process is schematically shown in \figref{fig:introduction:perception}.
In this thesis, we focus on the environment perception using camera-based object detection algorithms.
The process of object detection has rapidly developed in the last years.
Classical approaches such as a sliding window object detection using hand-crafted features and filters \cite{Dollar2011} have been replaced by more efficient deep learning based detection architectures such as Faster R-CNN \cite{He2016} or Single-Shot detectors \cite{Liu2016}.
The extended single-shot architectures such as YOLO \cite{Redmon2016} or RetinaNet \cite{Lin2017} provide further improvements in performance and speed.
These architectures are based on deep convolutinal neural networks \cite{LeCun1989,Goodfellow2016} which are neural networks with many intermediate convolutional operations.
A convolutional layer is nothing else but a filter mask that is applied to the preceding input to recognize several features, e.g., edges or textures.
The advantage of modern deep learning algorithms is that they do not require a hand-crafted design of the convolutional operations.
Instead, the models are able to autonomously learn the relevant aspects of the intermediate filters to detect objects within an image.
\footnotetext{Image from: Ruhr West University of Applied Sciences and University of Wuppertal: \textit{Automated Mobility: Overview of the technological fundamentals} (German original: \textit{Automatisierte Mobilität: Überblick über die technologischen Grundlagen}), p.~13, Fig.~5. \camocopyright{2022}}

First used for image classification \cite{LeCun1989,Krizhevsky2012}, deep learning architectures have been transferred to the more complex task of object detection \cite{Girshick2014,He2016,Liu2016,Redmon2016,Lin2017}.
The task of object detection is a joint task of classification (which type is the object?) and regression (where is the object located? which shape has the object?).
The initially proposed R-CNN architecture \cite{Girshick2014} utilizes a selective search algorithm to generate candidate boxes for possible objects in conjunction with a convolutional neural network to classify and refine these boxes.
However, the selective search still leads to a large amount of candidate boxes which makes the network training as well as the inference computational expensive.
The advanced architectures Fast R-CNN \cite{Girshick2015} and Faster R-CNN \cite{He2016} mitigate this problem by using a more flexible region proposal network to select candidate boxes.
In the object proposal stage, fixed image locations are used as prior information for possible object candidates and locations.
Afterwards, a region proposal network applies a first selection and refinement of candidate boxes.
These candidate boxes are finally used in the second stage of the object detector which applies a classification of the object category as well as a refinement of the object size and position using a separate neural network.
Similarly, the Single-Shot detector \cite{Liu2016}, YOLO \cite{Redmon2016}, and RetinaNet \cite{Lin2017} also use fixed image locations as prior information for object candidates and locations.
However, these architectures do not use a region proposal network but directly apply the selection, classification, and refinement step to the prior object candidates using a dedicated neural network.
These architectures use deep convolutional neural networks to extract the relevant image features.
This allows for a refinement and classification of the prior object locations to finally generate the object predictions.
Thus, these architectures are known as one-stage detection algorithms and have a very low computational runtime.

A related task to object detection is instance segmentation where the shape of individual objects is of interest in addition to their position and size.
With the success of modern object detection architectures, it has also been possible to develop  instance segmentation architectures.
For example, the Mask R-CNN architecture \cite{He2017} extends the Faster R-CNN \cite{He2016} detection model by an additional output that infers the object shape by classifying each pixel within the detected object boxes.
In contrast, with semantic segmentation, the association of a pixel with a particular class or segment is determined, but individual objects are not identified in this process.
Semantic segmentation architectures such as DeepLabv2 \cite{Chen2018} or DeepLabv3+ \cite{Chen2018a} utilize a deep convolutional neural network to extract relevant features within an input image.
The output of these segmentation models is a joint classification of all pixels in the input image that serves as an estimation of the relevant image segments.

Recent works have shown the superior performance of deep learning based object detectors and segmentation models over the classical approaches \cite{He2016,Liu2016,He2017,Lin2017,Chen2018a}.
However, driver assistance systems and especially autonomous driving are highly safety-relevant areas in which these detection algorithms need to operate.
As already stated, the explainability and the reliability of deep learning based methods are still open fields of current research \cite{Niculescu2005,Naeini2015,Guo2018,Haselhoff2021}.
A detection model needs to indicate in any case, if it is uncertain about a certain prediction.
For example, if a detection model needs to infer multiple objects within a frame under challenging conditions (e.g., rain or with sensors of low quality), the estimation either of the semantic class (what kind of object?) or of the position and size might be difficult and are subject to a high uncertainty.

Modern deep-learning based detection and segmentation architectures are able to estimate their prediction uncertainty during inference.
This uncertainty can be interpreted as a statistical measure for the estimation of the prediction error.
However, recent works have shown that modern neural networks tend to be too self-confident in their predictions \cite{Niculescu2005,Naeini2015,Guo2018,Kuleshov2018}, i.e., the estimated prediction uncertainty is too low compared with the observed error.
However, obtaining reliable uncertainty information is crucial especially for environment perception within the safety-relevant context of autonomous driving.

In the past, several works address the problem of unreliable uncertainty information by applying an additional calibration step during inference \cite{Platt1999,Zadrozny2001,Guo2018,Kuleshov2018}.
For classification, the authors in \cite{Platt1999} initially sought for a probabilistic output of support vector machines by utilizing the sigmoid function to transform an unrestricted output score to a score in the $[0, 1]$ interval that can be interpreted as an estimated probability of correctness \cite{Platt1999}.
In common literature, this method is known as logistic calibration or Platt scaling and can be used as a recalibration function to obtain calibrated probabilities \cite{Platt1999,Guo2018}.
As opposed to this approach, the authors in \cite{Zadrozny2001} proposed a binning scheme to convert uncalibrated probabilities to calibrated ones conditioned on the estimated uncertainty.
This approach is known as histogram binning.
An extension to this approach is \ac{BBQ} proposed by \cite{Naeini2015} which utilizes multiple weighted histogram binning schemes to obtain calibrated probabilities.
Furthermore, the authors in \cite{Naeini2015} proposed the \ac{ECE} and \ac{MCE} which are both metrics to quantify the misalignment between estimated and observed error.
Based on this research, the authors in \cite{Guo2018} found that modern network architectures in particular tend to be too self-confident in their predictions, leading to a high calibration error.
For multiclass classification tasks, the authors in \cite{Guo2018} proposed temperature scaling which is related to the logistic calibration function \cite{Platt1999} but only with a single rescaling parameter for all possible output classes.
Recently, the authors in \cite{Maag2020} and \cite{Schubert2021} investigated the calibration properties of modern segmentation and object detection architectures, respectively.
For both tasks, the authors studied the influence of miscellaneous factors (e.g., position and shape in object detection) to the calibration properties.

In contrast to classification, the task of regression is to estimate a continuous output score, e.g., the object location or shape when used within object detection.
In probabilistic regression, a model does not only estimate the requested regression score but also an additional uncertainty that indicates the belief of the estimator in the correctness of its prediction \cite{Kendall2017,He2019}.
Recent works found that these uncertainty scores also tend to miscalibration \cite{Kuleshov2018,Song2019,Levi2019,Laves2020} and proposed several methods such as Isotonic Regression \cite{Kuleshov2018}, Variance Scaling \cite{Levi2019,Laves2020}, and GP-Beta \cite{Song2019} that are applied after inference as an additional calibration step.
A detailled explanation of the calibration methods for probabilistic regression is given in \chapref{chapter:regression}.

\section{Research Question and Novelty}
In this thesis, we address the problem of reliable uncertainty information obtained by modern detection algorithms.
We seek to examine if the estimated uncertainty is trustworthy, i.e., if it matches the observed error in real-world applications.
We follow this research question and apply our examinations to the semantic (label) uncertainty as well as to the spatial (position and size) uncertainty.
Moreover, recent works have shown that modern neural networks tend to produce unreliable uncertainty information \cite{Niculescu2005,Naeini2015,Kull2017,Guo2018}.
Since the deviation between estimated uncertainty and observed error undergoes a systemic error, it is possible to apply post-hoc calibration methods \cite{Naeini2015,Guo2018}.
Such calibration methods apply a remapping of the estimated uncertainties to more realistic ones that have been observed on a dedicated calibration data set.
We transfer these methods to the task of object detection and address the research question, if correlations exist between semantic (label) and spatial (position and size) uncertainty.
Thus, we derive new calibration methods that are able to capture possible correlations by using further influential factors for uncertainty calibration within the scope object detection.
Our newly proposed methods are evaluated on state-of-the-art detection architectures Faster R-CNN \cite{He2017} and RetinaNet \cite{Lin2017} and the data sets MS COCO \cite{Lin2014} and Cityscapes \cite{Cordts2016}.

\begin{figure}[t!]
    \centering
    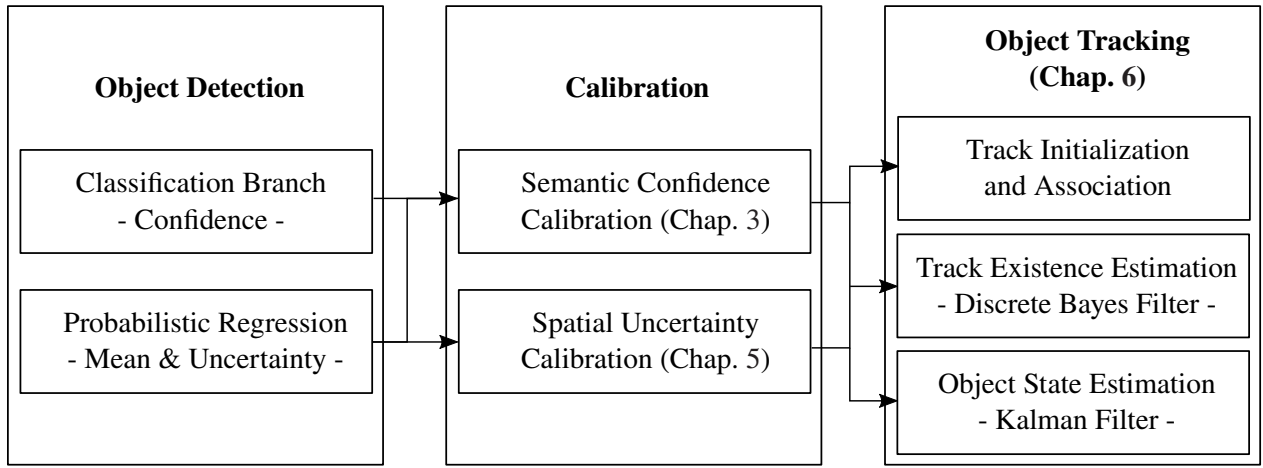%
    \caption[Concept of the environment perception pipeline that is addressed within this work.]{
        Concept of the environment perception pipeline that is addressed within this work.
        An object detection model generates multiple predictions for possible object candidates with a certain semantic and spatial uncertainty.
        We examine the semantic and spatial uncertainty for reliability.
        Furthermore, we provide recalibration methods that can be applied and integrated into the tracking pipeline after the detection process.
        The recalibrated uncertainty information is finally consumed by a tracking framework to track the detected objects over time in a sequence of images.
    }
    \label{fig:introduction:blockimage}
\end{figure}

For environment perception in the scope of driver assistance systems or autonomous driving, the detection of single objects is the first step to implement an object tracking over subsequent frames within a sequence of images.
The object detection is thus the baseline task for object tracking \cite{Bar2004,Van2005}.
Especially for object tracking, reliable uncertainties are crucial, as they are processed by the tracking algorithm and thus have a significant impact on the tracking performance of individual objects.
Therefore, we investigate  if our developed calibration methods have an influence on the task of object tracking.
In this context, we apply our new uncertainty calibration on the output of the baseline object detector and pass the recalibrated uncertainty information to the tracking algorithm.
This process is schematically shown in \figref{fig:introduction:blockimage}.
In this way, we are able to evaluate the influence of the predictive uncertainty in conjunction with our proposed calibration methods on a subsequent process.
Our contributions within this work are summarized in \tabref{tab:introduction:contributions}.
\begin{table}[t!]
    \centering
    \caption[Overview of our contributions within this work.]{
        Overview of our contributions within this work.
    }
    \label{tab:introduction:contributions}
    \begin{tabularx}{\textwidth}{| X | c | c |}
        \hline
        Contribution (short description) & Reference & Section \\ \hline \hline
        Definition of semantic confidence calibration for object detection & \cite{Kueppers2020,Kueppers2022a,Schwaiger2021} & \ref{section:confidence:definition:detection} \\
        Definition of semantic confidence calibration for instance segmentation & \cite{Kueppers2022a} & \ref{section:confidence:definition:instance} \\
        Definition of semantic confidence calibration for semantic segmentation & \cite{Kueppers2022a} & \ref{section:confidence:definition:semantic} \\
        Position-dependent histogram binning calibration & \cite{Kueppers2020,Kueppers2022a} & \ref{section:confidence:methods:binning} \\
        Position-dependent logistic and beta calibration & \cite{Kueppers2020,Kueppers2022a} & \ref{section:confidence:methods:scaling} \\
        Derivation of Bayesian confidence calibration & \cite{Kueppers2021} & \ref{section:bayesian:methods} \\ \hline \hline
        Common mathematical context of multiple definitions for regression calibration & \cite{Kueppers2022b} & \ref{section:regression:definition} \\
        Parametric regression uncertainty calibration using Gaussian processes & \cite{Kueppers2022b} & \ref{section:regression:methods:parametric} \\
        Joint multivariate regression uncertainty calibration using Gaussian processes & \cite{Kueppers2022b} & \ref{section:regression:methods:parametric} \\
        Covariance estimation of regression uncertainty & \cite{Kueppers2022b} & \ref{section:regression:methods:correlations} \\
        Covariance recalibration of regression uncertainty & \cite{Kueppers2022b} & \ref{section:regression:methods:correlations} \\ \hline \hline
        Estimation of object existence over time using the detector's semantic confidence & - & \ref{section:tracking:existence} \\
        Integration of semantic confidence calibration to object existence estimation & - & \ref{section:tracking:existence} \\
        Estimation of object state using the detector's spatial uncertainty & - & \ref{section:tracking:position} \\
        Integration of spatial uncertainty calibration to object estimation & - & \ref{section:tracking:position} \\ \hline
    \end{tabularx}
\end{table}

\section{Structure of this Work}
This work is structured as follows:
First, we introduce the basic architectures for object detection and how uncertainty information can be extracted from existing object detection architectures in \chapref{chapter:basics}.
In \chapref{chapter:confidence}, we review the definitions for semantic confidence calibration and transfer these to the task of object detection, instance segmentation, and semantic segmentation.
We develop metrics to evaluate the uncertainty as well as methods to correct unreliable confidence information. 
Furthermore, we evaluate our proposed metrics and methods using several detection and segmentation models on common public data sets.
Building on top of this, we propose Bayesian confidence calibration in \chapref{chapter:bayesian} that allows to capture additional uncertainty within the uncertainty calibration methods itself.
This adds an additional layer to model uncertainty in the overall perception chain and allows a detection of possible failure modes during inference.
In \chapref{chapter:regression}, we review the definitions for spatial uncertainty evaluation. 
In addition, we extend common metrics as well as methods for uncertainty correction to fit the needs for the task of object detection. 
We also evaluate our proposed methods on different network architectures and different data sets.
The proposed methods for semantic and spatial uncertainty correction are used within \chapref{chapter:tracking} in the context of object tracking to inspect and evaluate their influence on a relevant real-world application.
This demonstrates the effectiveness of the proposed methods.
Finally, we give a conclusion about our evaluation results, our developed methods, and our findings in \chapref{chapter:conclusion}.

\acresetall
\chapter{Object Detection and Uncertainty Modeling}
\label{chapter:basics}

Before we start with our main work, we present the basic principle of modern neural networks as well as the architectures for image-based object detection that have been used throughout this work.
Moreover, we describe the basic types of uncertainties that are analyzed in common literature.
These uncertainties play an important role throughout this work so that we further describe how an object detection model is able to estimate the uncertainty of a prediction.
Finally, we give a brief overview of related work that aims to elaborate possible reasons for an unreliable or misaligned uncertainty estimation of modern neural networks.

\section{Basics of Neural Networks and Image-based Object Detection} 
\label{section:basics:object_detection}

In this work, we focus on the image-based object detection process based on neural network architectures.
Artificial neural networks are models in the field of machine learning.
Machine learning is a concept to enable systems to recognize patterns and characteristics in a known data set.
This can subsequently be used to classify new data on the basis of the previously learned patterns and characteristics.
For example, in image recognition, images are classified into individual categories.
Based on a known data set, the machine learning model is trained to extract the relevant image features to learn a mapping from the input image to the desired category.
The trained model can then be used to classify new images that are not part of the training data set.

In this section, we start by introducing the basic concept of a neural network.
Furthermore, we describe how to train a neural network so that it is able to apply an appropriate mapping from an input image to the desired output.
Finally, we give an overview about the object detection architectures that are based on neural networks and which are used within this work.

\subsection{Fully-Connected Neural Networks}
\label{section:basics:object_detection:fcn}

Basically, a neural network consists of multiple nodes (neurons) that are connected to each other.
Each of these connections has a weight that scales the input signal towards the target neuron.
Commonly, the neurons are organized into multiple consecutive layers, so that a neural network can be interpreted as a sequence of $\numhidden$ consecutive operations (network layers) that take the output of the preceding layer as input to generate a new output.
Given an input image $\inputvariate \in \inputset = [0, 1]^{\Channels \times \Width \times \Height}$ of width $\Width$, height $\Height$, and $\Channels$ channels (e.g., RGB image) with the respective ground-truth label $\groundtruthvariate \in \gtset = \{1, ..., \numclasses\}$ with $\numclasses$ classes, the whole neural network $\model_{\allparameters}$ serves as a mapping from $\inputvariate$ to an overall output $\outputvariate \in \outputset$ (categorical for classification or continuous for regression), so that $\model_{\allparameters}: \inputset \rightarrow \outputset$.
In this scope, $\allparameters \in \parameterset$ denote the network parameters.
We distinguish between the input layer that consumes the image as input, the intermediate or hidden layers, and the output layer that outputs the overall network prediction $\predoutputvariate \in \outputset$ which aims to target the real ground-truth label $\groundtruthvariate \in \outputset$.

\begin{figure}[t]
	\centering
	\begin{overpic}[width=1.0\linewidth, tics=5, trim=0 0 -2em -2em]{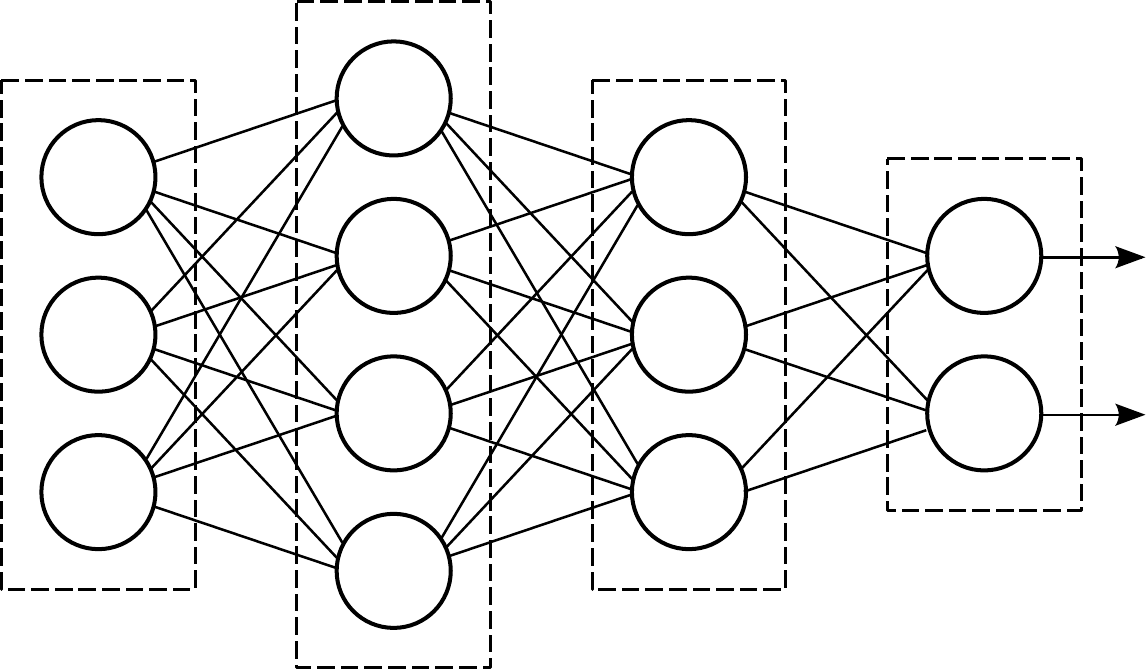}
		\put(0, 56){Input layer ($\indexhidden=1$)}
		\put(23, 56){Hidden layer ($\indexhidden=2$)}
		\put(47.3, 56){Hidden layer ($\indexhidden=3$)}
		\put(71.5, 56){Output layer ($\indexhidden=4$)}
		
		\put(6.9, 40){$\singleinput_{1}$}
		\put(6.9, 27){$\singleinput_{2}$}
		\put(6.9, 14){$\singleinput_{3}$}
		
		\put(96, 33.4){$\predoutput_{1}$}
		\put(96, 20.4){$\predoutput_{2}$}
		
		\put(16, 2){\makecell{Weights\\$\scaleweightmat_{2}$}}
		\put(41, 2){\makecell{Weights\\$\scaleweightmat_{3}$}}
		\put(65.5, 2){\makecell{Weights\\$\scaleweightmat_{4}$}}
	\end{overpic}
	\caption[
	Concept of a fully-connected neural network (without bias terms) with two hidden layers, an input layer, and an output layer to obtain the estimate for $\predoutput_{1}$, $\predoutput_{1}$ based on a given input $\allsingleinput \in \realdigits^{3}$.
	]{
		Concept of a fully-connected neural network (without bias terms) with two hidden layers, an input layer, and an output layer to obtain the estimate for $\predoutput_{1}$, $\predoutput_{2}$ based on a given input $\allsingleinput \in \realdigits^{3}$.
	}
	\label{fig:basics:fullyconnected}
\end{figure}

A basic layer type used in the context of neural networks is the so-called dense or fully-connected layer.
The idea is to use multiple nodes or neurons where each neuron is connected to all neurons in the previous layer.
Each neuron simply computes a weighted sum over the entire input which is given by the output of the preceding neurons.
The fully-connected layer can also be represented by a matrix multiplication between a weight matrix $\scaleweightmat_{\indexhidden} \in \realdigits^{\numneurons_{\indexhidden} \times \numneurons_{\indexhidden-1}}$ and the layer input $\allsingleinput_{\indexhidden} \in \realdigits^{\numneurons_{\indexhidden-1}}$ which is the output of the preceding network layer $\indexhidden - 1$.
In this context, $\numneurons_{\indexhidden-1}$ and $\numneurons_{\indexhidden}$ denote the number of neurons or features of the preceding and the actual layer, respectively.
The matrix $\scaleweightmat_{\indexhidden}$ holds the weights which are used for the connections between the neurons from layer $\indexhidden-1$ to $\indexhidden$.
If the preceding layer is given either by the input image itself or a convolution operation (cf. \secref{section:basics:convolutional_networks}), the (3D) input gets flattened to a single vector, so that $\numneurons_{\indexhidden-1} = \Channels_{\indexhidden-1} \cdot \Width_{\indexhidden-1} \cdot \Height_{\indexhidden-1}$
with $\Width_{\indexhidden-1}$, $\Height_{\indexhidden-1}$, and $\Channels_{\indexhidden-1}$ as the width, height, and number of channels within the preceding layer $\indexhidden-1$.
Let $\layerfunction_{\indexhidden}$ denote the function of layer $\indexhidden$.
A fully-connected layer applies the transformation $\layerfunction_{\indexhidden}(\allsingleinput_{\indexhidden}) = \scaleweightmat_{\indexhidden} \allsingleinput_{\indexhidden} = \allsingleinput_{\indexhidden+1} $ which yields the output vector $\allsingleinput_{\indexhidden+1} \in \realdigits^{\numneurons_{\indexhidden}}$ so that the whole network can be interpreted as a composite function by
\begin{align}
	\label{eq:basics:fullyconnected}
	\model_{\allparameters}(\allsingleinput_\indexsamples) = (\layerfunction_{\numhidden} \circ \ldots \circ \layerfunction_{1})(\allsingleinput_\indexsamples) ,
\end{align}
given a certain sample $\allsingleinput_{\numsamples}$.
A neural network that solely consists of fully-connected layers is denoted a fully-connected neural network.
The concept of a such a network is schematically shown in \figref{fig:basics:fullyconnected}.

\subsubsection{Activation Functions}

\begin{figure}[t]
	\centering
	\begin{overpic}[width=0.8\linewidth, tics=5, trim=-1em 0 -1em -1em]{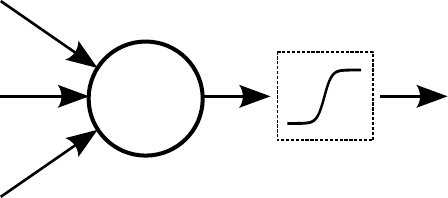}
		\put(1, 42){Input}
		\put(26, 42){Weighted sum}
		\put(58, 42){Activation function}
		\put(92, 42){Output}
		
		\put(3, 38){$\singleinput_{1}$}
		\put(3, 19.5){$\singleinput_{2}$}
		\put(3, 0){$\singleinput_{3}$}
		
		\put(12, 36){$\scaleweight_{1}$}
		\put(12, 21){$\scaleweight_{2}$}
		\put(12, 7){$\scaleweight_{3}$}
		
		\put(95, 19){\large{$o$}}
		\put(31.5, 16){\fontsize{96}{96}\selectfont{\textSigma}}
	\end{overpic}
	\caption[
		Concept of a single neuron with a subsequent non-linear activation function.
	]{
		Concept of a single neuron with a subsequent non-linear activation function to generate an output $o \in \realdigits$.
		The neuron takes either the network input or the preceding layer output and computes a weighted sum over the complete input vector.
		The result is given to the activation function and then passed on to the next network layer.
	}
	\label{fig:basics:neuron}
\end{figure}
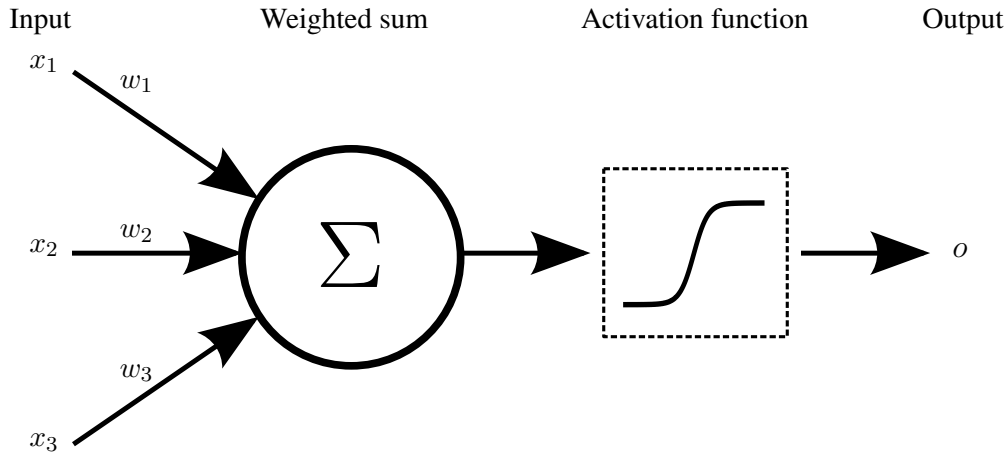

A matrix multiplication is a linear transformations of the incoming data.
Moreover, the concatenation of several linear operations also results in an overall linear transformation.
Therefore, it is not possible to model complex non-linear relationships between input and output space.
For this reason, a so-called activation function is usually used which introduces a non-linearity after the linear transformation and between the individual layers. 
The activation function needs to be differentiable because a neural network is typically trained using the backpropagation algorithm which requires the calculation of the function's derivative.
Given an arbitrary function input $t \in \realdigits$, common choices for the intermediate activation functions are the sigmoid given by
\begin{align}
	\sigmoid(t) = \frac{1}{1 + \exp(-t)} ,
\end{align}
the hyperbolic tangent function (tanh)
\begin{align}
	\tanh(t) = \frac{\exp(2t) - 1}{\exp(2t) + 1} ,
\end{align}
or the \ac{ReLU} given by $\max(0, t)$ with its derivative set to $\frac{\diff \max(0, t)}{\diff t} = 0$ for $t=0$.
Furthermore, an activation function is often used after the output layer.
For example, in binary classification, if output scores in the $[0, 1]$ interval are required, a common choice is the sigmoid function.
In multiclass classification with $\numclasses$ classes and $\mathbf{t} \in \realdigits^{\numclasses}$, the softmax function given by
\begin{align}
	\softmax(\mathbf{t})_{\indexclasses} = \frac{\exp(t_{\indexclasses})}{\sum_{\indexclasses'=1}^{\numclasses} \exp(t_{\indexclasses'})}
\end{align}
is a common choice to obtain predictions whose outputs sum up to $1$, i.e., $\sum_{\indexclasses=1}^{\numclasses} \predoutput_{\indexclasses} = 1$.
In both cases, the output can be interpreted as parameters for a Bernoulli or categorical distribution in binary or multiclass classification, respectively, which allows for a probabilistic interpretation of the network output.
The concept of a single neuron within a fully-connected neural network using a subsequent activation function is schematically shown in \figref{fig:basics:neuron}.

\subsubsection{Training of the Network Weights\protect\footnote{The descriptions for training a neural network are partly adapted from previous works \cite{Maag2021,Colling2022}.}}


According to the principle of \ac{ERM}, it is not possible to know the true real-world distribution of the data (e.g., the distribution of all possible images) so that we can not exactly determine the performance of our algorithm \cite{Vapnik1991}.
Instead, we are restricted to a certain set of training data which is drawn from the true data distribution and on which we can quantify the model performance (the empirical risk).
Let $\dataset = \{(\allsingleinput_\indexsamples, \singlegroundtruth_\indexsamples)\}^{\numsamples}_{\indexsamples=1}$ denote a data set of size $\numsamples$ with images $\allsingleinput_\indexsamples \in \inputset$ and the known target labels $\singlegroundtruth_\indexsamples \in \gtset$.
The training set is generated by a probability distribution $\mathcal{P}$ over $\inputset$.
Furthermore, we assert that a correct labeling function $u^\ast$ exists so that $u^\ast: \inputset \rightarrow \gtset$.
The task for a learning algorithm is to output a prediction rule $\model_{\dataset}: \inputset \rightarrow \outputset$ based on the training set $\dataset$ to classify the images.
The error $\loss_{\mathcal{P}, u^\ast}$ of the prediction rule is given by
\begin{align}
	\loss_{\mathcal{P}, u^\ast}(\model_{\dataset}) = \prob_{\allinputvariates \sampledfrom \mathcal{P}}\big[\model_{\dataset}(\allsingleinput) \neq u^\ast(\allsingleinput)\big] .
\end{align}
Since the distribution $\mathcal{P}$ as well as the true labeling function $u^\ast$ are unknown, the learner needs to find a prediction rule $\model_{\dataset}$ that minimizes the prediction error on the given data set $\dataset$.
It is only possible to observe the true labeling function $u^\ast$ for the samples that are present in the data set $\dataset$ as we know the true mapping from the input $\allsingleinput$ to the desired outcome $\singlegroundtruth$.
Thus, the training error or empirical error of a learner is given by
\begin{align}
	\loss_\dataset(\model_\dataset) = \frac{1}{\numsamples} \sum_{\indexsamples=1}^{\numsamples} \ind(\model_{\dataset}(\allsingleinput_\indexsamples) \neq u^\ast(\allsingleinput_\indexsamples)) ,
\end{align}
where $\ind(\cdot)$ is the indicator function which evaluates to $1$ if the argument is true and $0$ otherwise.

\ac{MLE} is a method to learn the parameters $\allparameters$ that are used to model the underlying probability distribution $\mathcal{P}$ by an approximate distribution $\mathcal{P}_{\allparameters}$ parameterized by $\allparameters$ to construct an appropriate prediction rule $\model_{\allparameters}$ \cite{Aldrich1997}.
The maximum likelihood estimation is a minimization technique for \ac{ERM} using the log loss on $\mathcal{P}_{\allparameters}$ which is given by
\begin{align}
	\label{eq:basics:nll}
	-\log\big(\mathcal{P}_{\allparameters}(\allsingleinput)\big) ,
\end{align}
with parameters $\allparameters$ and input data $\allsingleinput$.
The equation (\ref{eq:basics:nll}) is also known as the \ac{NLL}.
Using this definition, maximizing the log likelihood is equivalent to minimizing the empirical risk, since
\begin{align}
	\label{eq:basics:argmin}
	\argmin_{\allhbparameter} \sum_{\indexsamples=1}^{\numsamples} -\log\big(\mathcal{P}_{\allparameters}(\allsingleinput_\indexsamples)\big) = \argmax_{\allhbparameter} \sum_{\indexsamples=1}^{\numsamples} \log\big(\mathcal{P}_{\allparameters}(\allsingleinput_\indexsamples)\big) ,
\end{align}
and it can be shown that the true risk of parameter $\allparameters$ is given by
\begin{align}
	\expectation_{\allinputvariates \sampledfrom \mathcal{P}}\Big[ -\log\big(\mathcal{P}_{\allparameters}(\allsingleinput)\big) \Big]
	&= \kldivergence\big(\mathcal{P}(\allsingleinput)||\mathcal{P}_{\allparameters}(\allsingleinput)\big) + \entropy\big(\mathcal{P}(\allsingleinput)\big) ,
\end{align}
with $\kldivergence\big(\mathcal{P}(\allsingleinput)||\mathcal{P}_{\allparameters}(\allsingleinput)\big)$ as the Kullback-Leibler divergence \cite{Kullback1951} between true data distribution $\mathcal{P}$ and the estimated one $\mathcal{P}_{\allparameters}$, and the Shannon entropy $\entropy\big(\mathcal{P}(\allsingleinput)\big)$ \cite{Shannon1948} of the true data distribution.

Now, given a neural network $\model$ with $\allparameters$ parameters, the objective of \ac{MLE} is to obtain a $\allestimatedparameters$ that maximizes the likelihood of observing the data.
Thus, during network training, it is required to pass each input image through the network so that the network is requested to output a prediction $\predoutput_\indexsamples = \model_{\allparameters}(\allsingleinput_\indexsamples)$.
Afterwards, the predicted output $\predoutput_\indexsamples$ is evaluated against the ground-truth label $\singlegroundtruth_\indexsamples$.
A cost or loss function $\loss(\allparameters)$ such as the \ac{NLL} in (\ref{eq:basics:nll}) is used to implement the maximum likelihood estimation.
In a binary classification setting ($\gtset = \{0, 1\}$), the likelihood function is a Bernoulli distribution so that the \ac{NLL} is given by
\begin{align}
	\loss(\allparameters) = -\sum_{\indexsamples=1}^{\numsamples} \singlegroundtruth_\indexsamples \log (\predoutput_{\indexsamples}) + (1-\singlegroundtruth_\indexsamples) \log (1 - \predoutput_{\indexsamples}) .
\end{align}

In order to minimize the loss function $\loss(\allparameters)$, the \ac{SGD} algorithm is used to update the network parameters $\allparameters$.
An initial guess for the network parameters $\allparameters$ is used as a starting point to iteratively descend in the direction of the minimum.
The direction of the update is determined by the gradient of the network components w.r.t. the parameters $\allparameters$.
Thus, the update for a weight matrix $\scaleweightmat_{\indexhidden}$ at layer $\indexhidden$ is calculated by
\begin{align}
	\scaleweightmat_{\indexhidden} = \scaleweightmat_{\indexhidden} - \learningrate \frac{\diff \loss}{\diff \scaleweightmat_{\indexhidden}} ,
\end{align}
with $\learningrate \in \realdigitspositive$ as the learning rate which controls the strength of the parameter update.
Commonly, the weight update is not applied using the whole data set $\dataset$.
Instead, the training data is grouped into batches which are small subsets of $\dataset$.
During network training, the loss is repeatedly computed over all batches that are present in the training set.
In this way, the optimization is computationally more efficient and it may help to overcome local minima in the loss function \cite{Bottou2010,Ge2015}.

The gradients, which are used by the \ac{SGD} algorithm, are obtained using backpropagation \cite{Rumelhart1986}.
Since the neural network is grouped into several consecutive layers, the output and thus the loss $\loss(\allparameters)$ are composite functions of multiple consecutive layer operations \big(cf. equation (\ref{eq:basics:fullyconnected})\big).
Thus, for the intermediate network layers, it is required to obtain the gradient using the chain rule
\begin{align}
	\frac{\diff \loss}{\diff \scaleweightmat_{\indexhidden}} = \frac{\diff \loss}{\diff \layerfunction_{\numhidden}} \frac{\diff \layerfunction_{\numhidden}}{\diff \layerfunction_{\numhidden-1}} \ldots \frac{\diff \layerfunction_{\indexhidden+1}}{\diff \layerfunction_{\indexhidden}} \frac{\diff \layerfunction_{\indexhidden}}{\diff \scaleweightmat_{\indexhidden}} ,
\end{align}
to finally yield the gradient for the weight update.
This process is repeated several times to finally approximate the ground-truth data distribution $\pdf_{\groundtruthvariate}(\singlegroundtruth|\allsingleinput)$ by an estimated network distribution $\pdf_{\predoutputvariate}(\predoutput|\allsingleinput)$.

\subsubsection{Regularization}
The minimization of the empirical risk may lead to overfitting, i.e., the estimator $\model_{\dataset}$ works well on the given training data set $\dataset$ but fails to estimate the correct labels on unseen data.
Weight decay \cite{Vapnik1998} is a regularization technique that adds an additional penalty to the loss term which penalizes large layer weights.
Given the loss function w.r.t. a dedicated weight matrix $\scaleweightmat_{\indexhidden}$, the equation for the loss including the weight decay is given by
\begin{align}
	\bar{\loss}(\scaleweightmat_{\indexhidden}) = \loss(\scaleweightmat_{\indexhidden}) + \frac{\learningrate'}{2} || \scaleweightmat_{\indexhidden} ||^2_2 ,
\end{align}
where the parameter $\learningrate' \in \realdigitspositive$ controls the strength of the weight decay.
Using this technique for regularization, the weights of a layer are decreased in each iteration by a constant factor which reduces the model capacity and helps to prevent overfitting.

In addition, dropout has originally been proposed as a regularization technique during network training.
When applying dropout, single network weights are randomly deactivated so that the network is requested to apply a mapping without these specific connections.
This behavior also aims to reduce model capacity and strengthen the remaining network connections.
Moreover, the authors in \cite{Gal2016} showed that dropout can also be used during inference to quantify the uncertainty of the network about a specific prediction.
During inference, the same input is passed multiple times through the network.
In each forward pass, dropout is applied on different randomly chosen network connections.
This yields in multiple predictions for the same input that span a sample distribution as the network output.
The higher the variance of this output distribution, the higher the network's uncertainty about the current prediction.
This process is known as Monte-Carlo dropout \cite{Gal2016}.

Finally, the authors in \cite{Ioffe2015} proposed batch normalization as a technique to normalize the output after each layer $\indexhidden$.
This method aims to improve the network performance by minimizing the distribution shifts (internal covariate shift) after the hidden layer’s activation functions.
Given the layer input $\allsingleinput_{\indexhidden} \in \realdigits^{\numdims}$ with $\numdims$ dimensions, the batch normalization is calculated by
\begin{align}
	\label{eq:basics:batchnorm}
	\singleinput_{\indexhidden, \indexdims}^\ast =  \frac{ \singleinput_{\indexhidden, \indexdims} - \mean_{\mathcal{B}, \indexdims}}{\sqrt{\variance_{\mathcal{B}, \indexdims} + \epsilon}},
\end{align}
where $\mean_{\mathcal{B}, \indexdims}$ and $\variance_{\mathcal{B}, \indexdims}$ are the mean and variance within batch $\mathcal{B}$ and dimension $\indexdims$, respectively, with $\epsilon \in \realdigits$ as a small offset to increase numerical stability.

The authors in \cite{Ioffe2015} note that a normalization of each layer input may affect its representational power.
The authors propose to use additional parameters so that, during network training, it is possible to revert or change the normalization, e.g., to obtain an identity transform.
Therefore, after the batch normalization step in (\ref{eq:basics:batchnorm}), a shifting and rescaling of the input is applied by
\begin{align}
	\singleinput_{\indexhidden, \indexdims}^{\ast\ast} = \scalesub_{\indexdims} \singleinput_{\indexhidden, \indexdims}^\ast + \delta_{\indexdims} ,
\end{align}
with scaling parameter $\scalesub_{\indexdims} \in \realdigitspositive$ and shift parameter $\delta_{\indexdims} \in \realdigits$ for all $\indexdims \in \{1, \ldots \numdims\}$.

\subsection{Convolutional Neural Network}
\label{section:basics:convolutional_networks}

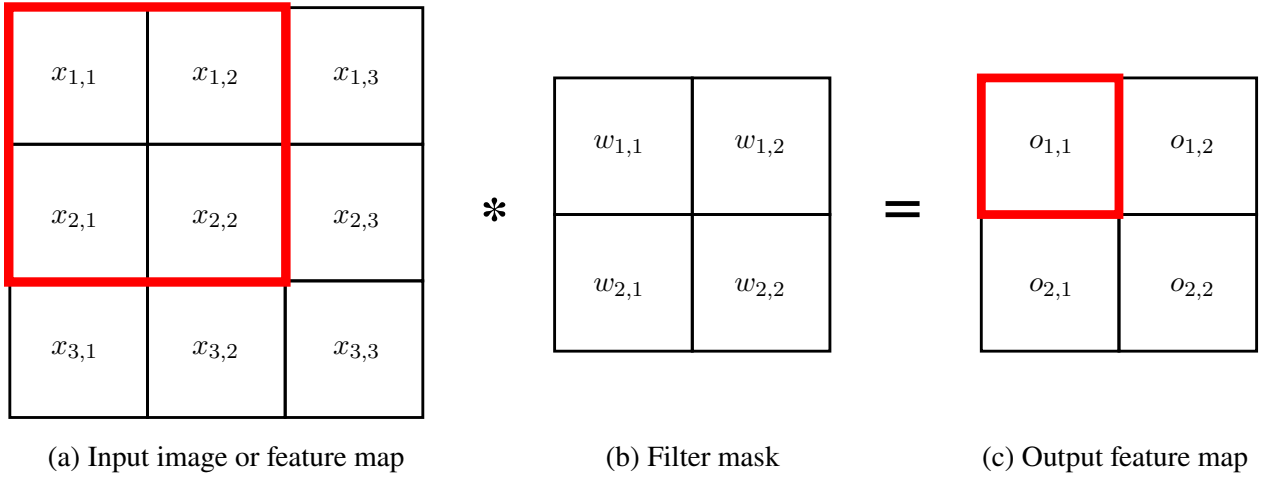
\begin{figure}[t]
	\centering
	\begin{overpic}[width=1.0\linewidth, tics=5, trim=0 -2em 0 0]{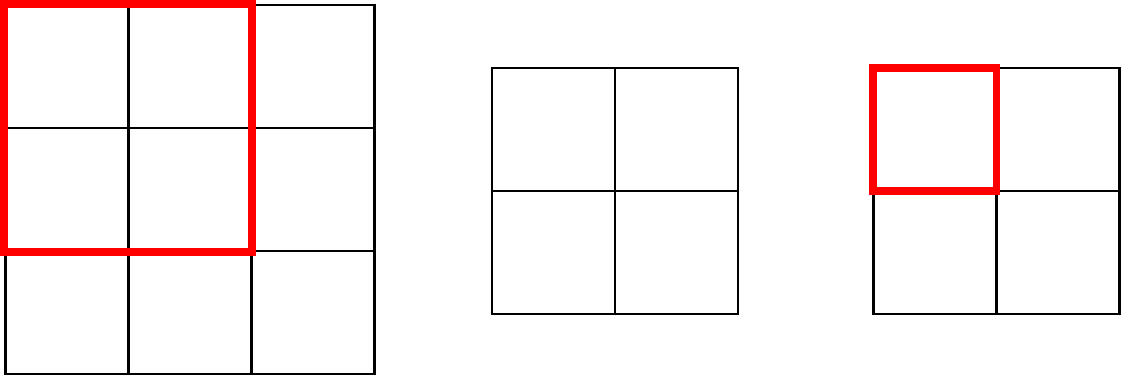}
		\put(3.5, 3){(a) Input image or feature map}
		\put(48, 3){(b) Filter mask}
		\put(78, 3){(c) Output feature map}
		
		\put(38, 21.5){\huge{\textbf{*}}}
		\put(70, 22.5){\huge{\textbf{=}}}
		
		\put(3.75, 34){$\singleinput_{1, 1}$}
		\put(3.75, 22.5){$\singleinput_{2, 1}$}
		\put(3.75, 12){$\singleinput_{3, 1}$}
		\put(15, 34){$\singleinput_{1, 2}$}
		\put(15, 22.5){$\singleinput_{2, 2}$}
		\put(15, 12){$\singleinput_{3, 2}$}
		\put(26.25, 34){$\singleinput_{1, 3}$}
		\put(26.25, 22.5){$\singleinput_{2, 3}$}
		\put(26.25, 12){$\singleinput_{3, 3}$}
		
		\put(47, 28.5){$\scaleweight_{1, 1}$}
		\put(47, 17){$\scaleweight_{2, 1}$}
		\put(58.25, 28.5){$\scaleweight_{1, 2}$}
		\put(58.25, 17){$\scaleweight_{2, 2}$}
		
		\put(81.75, 28.5){$\convolutionoutput_{1, 1}$}
		\put(81.75, 17){$\convolutionoutput_{2, 1}$}
		\put(93, 28.5){$\convolutionoutput_{1, 2}$}
		\put(93, 17){$\convolutionoutput_{2, 2}$}
	\end{overpic}
	\caption[Concept of the convolution operation which is applied on an input image or feature map with a certain filter mask to extract a feature map.]{
		Concept of the convolution operation which is applied on an input image or feature map (a) with a certain filter mask (b) to extract a new feature map (c).
		This result of the convolution can represent certain image features such as edges or other structural information.
	}
	\label{fig:basics:convolution}
\end{figure}

A \ac{CNN} is a network that utilizes at least one convolution operation as an alternative type to the fully-connected layer in order to process the input data \cite{Fukushima1982}.
A convolution is a linear operation on two functions $f: \realdigits \rightarrow \realdigits$ and $g: \realdigits \rightarrow \realdigits$ defined by
\begin{align}
	(f*g)(t) = \int_{-\infty}^{+\infty} f(\tau) g(t - \tau) \diff \tau ,
\end{align}
or, in the discrete case with $f: \intdigits \rightarrow \realdigits$ and $g: \intdigits \rightarrow \realdigits$, by
\begin{align}
	(f*g)(t) = \sum_{\tau \in \intdigits} f(\tau)g(t - \tau) .
\end{align}
In image processing, the convolution describes the process of calculating a weighted sum of neighboring pixels using a filter mask within the original image.
The weights are the coefficients of the filter mask.
The filter mask is moved successively over all image positions to apply the filtering to the complete image and to get feature representations for all image regions.
This process is schematically shown in \figref{fig:basics:convolution}.
More formally, given the layer input $\allsingleinput_{\indexhidden} \in \realdigits^{\Channels_{\indexhidden-1} \times \Width_{\indexhidden-1} \times \Height_{\indexhidden-1}}$ of width $\Width_{\indexhidden-1}$, height $\Height_{\indexhidden-1}$, and $\Channels_{\indexhidden-1}$ channels (obtained by the preceding layer $\indexhidden-1$), the discrete convolution operation is defined by
\begin{align}
	\singleinput_{\indexhidden+1}^{(\channels_{\indexhidden}, i, j)} =  \sum_{\channels_{\indexhidden-1}=1}^{\Channels_{\indexhidden-1}} \sum_{\width=1}^{\convolutionwidth} \sum_{\height=1}^{\convolutionheight} \singleinput_{\indexhidden}^{(\channels_{\indexhidden-1}, i+\width, j+\height)} \cdot \scaleweight_{\indexhidden}^{(\channels_{\indexhidden}, \channels_{\indexhidden-1}, \width, \height)}
\end{align}
given the filter mask $\scaleweightmat_{\indexhidden} \in \realdigits^{\Channels_{\indexhidden} \times \Channels_{\indexhidden-1} \times \convolutionwidth \times \convolutionheight}$ with filter width $\convolutionwidth$, filter height $\convolutionheight$, and $\Channels_{\indexhidden}$ channels.
Note that it is only possible to shift the filter kernel so far until it reaches the image or feature map boundary.
A mitigation of this problem is called padding where missing pixel values are padded by a certain scheme, e.g., by adding zeros to the missing pixel positions (zero padding).
The application of the convolutional operation is a common technique used in computer vision.
The convolution filters are used to obtain feature representations (e.g., detected edges) to further process the input image.
The advantage of applying a convolution is that the layer is equivariant to translation, i.e., the filter mask is applied to all image regions with the same filter weights.
In the context of \ac{CNN}, it is possible to learn the filter mask weights $\scaleweightmat_{\indexhidden}$ by applying stochastic gradient descent using backpropagation algorithm (cf. next section).
Note that, since the convolution is a linear operation similar to the fully-connected layer, a subsequent activation function is also required for a \ac{CNN} after each layer.
In image classification, the output of a \ac{CNN} is commonly a vector that describes the estimated class probabilities given a certain input image.
In order to calculate the output vector, the results after the convolutional layers are commonly passed to a fully-connected layer which generates the final network output.

\subsection{Architectures for Object Detection}
\label{section:basics:object_detection:architectures}

In contrast to simple image recognition, where the label of the whole image is of interest, the task of object detection is the joint task of classification (object type) and regression (object position).
Thus, a neural network architecture for object detection $\model_{\allparameters}$ needs to solve both tasks simultaneously given an input image $\allsingleinput_\indexsamples$.
In the following, we present the basic architectures of modern neural-network based object detectors.

\subsubsection{Feature Extraction and Object Candidates}

Each of the object detector algorithms are based on a backbone network to extract relevant image features which are further processed for object classification and position/shape estimation.
The backbone is usually a \ac{CNN} that processes an image and generates intermediate feature representations of the input using its convolution operations.
This network is commonly adapted from classification and pretrained on a classification data set \cite{Ren2015}.
Thus, passing the input image $\allsingleinput_{\indexsamples}$ through the backbone network $\model_{\allparameters_B}$ results in a feature representation $\model_{\allparameters_B}(\allsingleinput_{\indexsamples}) = \predoutput_B \in \realdigits^{\Channels_B \times \Width_B \times \Height_B}$ of the input image with $\Channels_B$ feature maps of width $\Width_B$ and height $\Height_B$.
At this point, it is necessary to distinguish between a two-stage and a single-stage object detector.
While a single-stage detector directly works with the backbone network, a two-stage detector uses an additional step to generate prior assumptions about possible object candidates.
In a two-stage architecture, the feature maps of the backbone network are passed to a subsequent \ac{RPN} denoted by $\model_{\allparameters_{\text{RPN}}}$ \cite{Ren2015}.
The \ac{RPN} itself is a small \ac{CNN} that outputs a list of $\numproposals$ object proposals by sliding its convolutional network on the feature maps.
These object proposals are initial guesses for possible object candidates.
For the computation of the object proposals, so-called anchor boxes are used for each location of the sliding window.
The anchor boxes can be interpreted as priors for the object proposals, as they are used at each sliding window location of the \ac{RPN} with fixed object locations and fixed aspect ratios.
Subsequently, the \ac{RPN} uses two final fully-connected layer to compute an objectness score as well as the final coordinates for the object proposals.
The objectness score $\predconfidence_\indexproposals$ represents the belief of the \ac{RPN} about the presence or absence of an object within a certain anchor.
This score is used to filter the object proposals so that only proposals with an objectness score above a certain threshold are passed through the network.
Furthermore, the \ac{RPN} outputs the center coordinates $c_{x,\indexproposals}, c_{y,\indexproposals}$ as well as the width and height $\width_\indexproposals, \height_\indexproposals$ of the proposal boxes.
Thus, the output of the \ac{RPN} is given by $(\predconfidence_\indexproposals, c_{x,\indexproposals}, c_{y,\indexproposals}, \width_\indexproposals, \height_\indexproposals)^\T$ for all $\indexproposals \in \{1, \ldots, \numproposals\}$.

\begin{figure}[t!]
	\centering
	\begin{subfigure}{1.0\textwidth}
		\begin{overpic}[width=1.0\linewidth, tics=5]{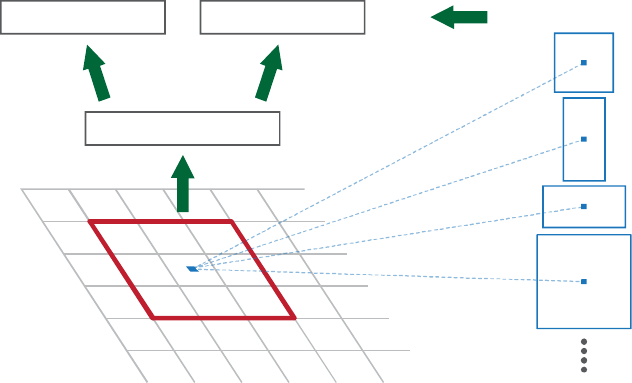}
			\put(4.5, 57){objectness scores}
			\put(37.5, 57){box coordinates}
			\put(86, 57){anchor boxes}
			\put(19.7, 39.5){intermediate features}
			
			\put(1, 53){\textit{cls} layer}
			\put(50, 53){\textit{reg} layer}
			\put(35, 35){intermediate layer}
			
			\put(24, 7){sliding window}
			\put(45, 3){backbone feature map}
		\end{overpic}
		\raggedleft{\scriptsize\ieeecopyright{2015}}
		\vspace{0.25em}
		\caption{
			Concept of the Region Proposal Network (RPN) using a sliding window with different predefined anchor boxes for each position within the output feature maps \cite[p.~3, Fig.~1]{Ren2015}.
		}
		
	\end{subfigure}
	\vspace{1em}

	\begin{subfigure}{0.46\textwidth}
		\begin{overpic}[width=1.0\linewidth, tics=5]{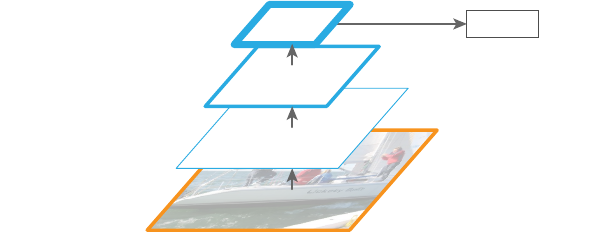}
			\put(77.2, 33.2){\scriptsize{predict}}
		\end{overpic}
		\raggedleft{\scriptsize\ieeecopyright{2017}}
		\vspace{0.25em}
		\caption{
			Basic structure of a Region Proposal Network (RPN) that uses the final output of the backbone network \cite[p.~1, Fig.~1(b)]{Lin2017a}.
		}
	\end{subfigure}%
	\hfill%
	\begin{subfigure}{0.46\textwidth}
		\begin{overpic}[width=1.0\linewidth, tics=5]{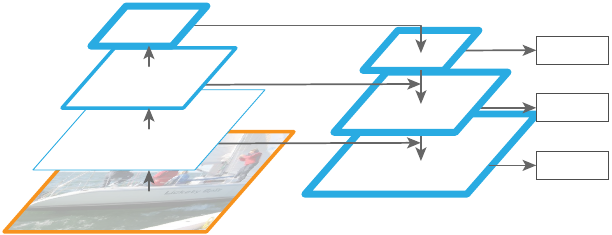}
			\put(88.5, 29){\scriptsize{predict}}
			\put(88.5, 19.5){\scriptsize{predict}}
			\put(88.5, 10){\scriptsize{predict}}
		\end{overpic}
		\raggedleft{\scriptsize\ieeecopyright{2017}}
		\vspace{0.25em}
		\caption{
			Feature Pyramid Network (FPN) that also the intermediate features of the backbone to estimate proposals at different scales \cite[p.~1, Fig.~1(d)]{Lin2017a}.
		}
	\end{subfigure}%
	\caption[Concept of the Region Proposal Network (RPN) and the Feature Pyramid Network (FPN) used for object detection.]{
		Concept of the Region Proposal Network (RPN) and the Feature Pyramid Network (FPN) used for object detection.
		For object detection using neural networks, it is necessary to make an initial assumption about possible object candidates and their locations \cite{Ren2015,Lin2017,Lin2017a}.
		Initially, a RPN has been proposed by \cite{Ren2015} to output object proposals by using a sliding window over the backbone feature maps with predefined anchor boxes (top).
		However, the drawback of this approach is that it only utilizes the features at the output of the backbone network (left bottom) which might only work well at a single scale \cite{Lin2017a}.
		Therefore, the authors in \cite{Lin2017a} proposed a FPN to mitigate this problem by utilizing the intermediate backbone features at different scales.
	}
	\label{fig:basics:proposal_networks}
\end{figure}
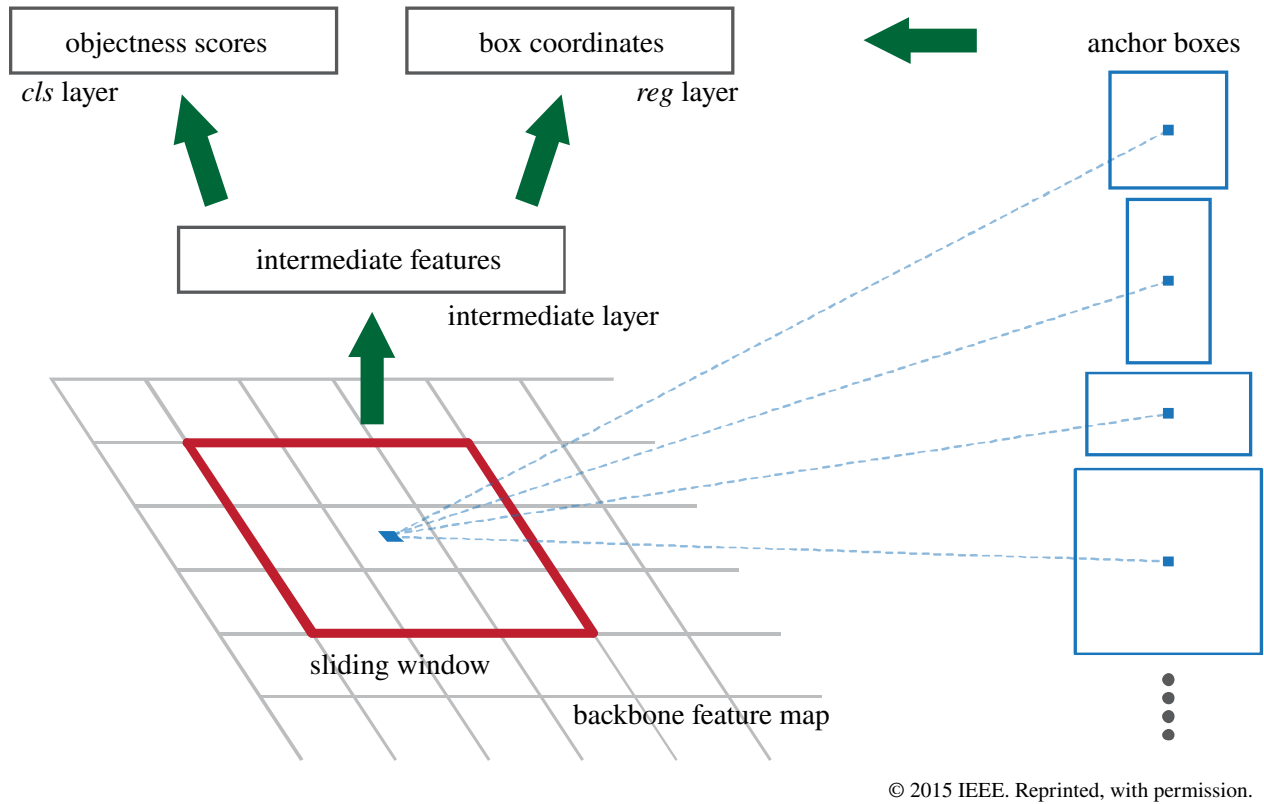
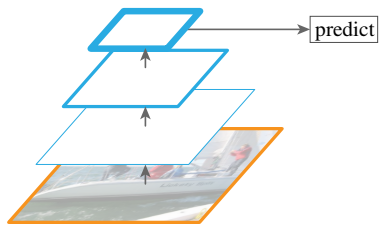
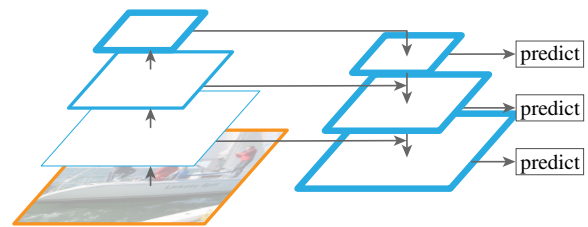

A drawback of the standard \ac{RPN} architecture is that it only consumes the final output of the backbone network which might represent features at a single scale \cite{Lin2017a}.
In common image recognition systems, so-called feature pyramids are used to extract image features at different scales, i.e, features that represent either low-level (local) structures (e.g., single window, arm, leg) or high-level (global) structures (e.g., building, car, human body).
Recently, the authors in \cite{Lin2017a} proposed a \ac{FPN} that is designed to mitigate the limitations of a \ac{RPN} by extracting the feature maps of the backbone network at different scales to construct a feature pyramid.
Furthermore, the different layers/scales of the \ac{FPN} are connected to each other to obtain a more flexible and computationally efficient architecture for object proposal generation.
The concept of the \ac{RPN} as well as the \ac{FPN} architectures are schematically shown in \figref{fig:basics:proposal_networks}.

\subsubsection{Two-Stage Detector: Faster R-CNN}

\begin{figure}[t!]
	\centering
	\includegraphics[width=0.4\textwidth]{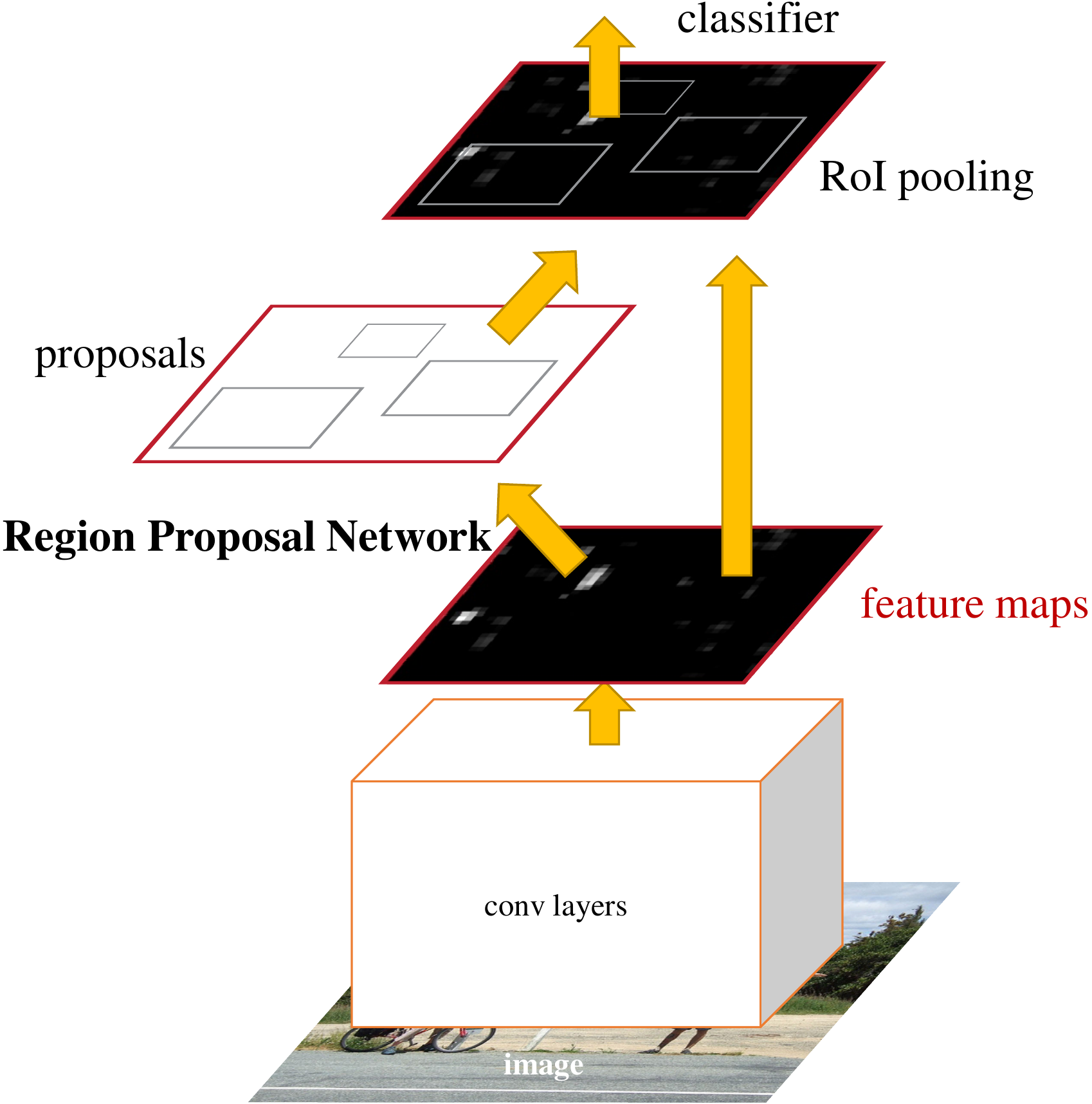}
	
	\vspace{0.75em}
	\raggedleft{\scriptsize\ieeecopyright{2015}}
	
	\caption[Concept of the Faster R-CNN object detection architecture.]{
		Concept of the Faster R-CNN object detection architecture \cite[p.~3, Fig.~2]{Ren2015}.
		A basic backbone network (denotes by ``conv layers'') is used to generate feature representations of the input image.
		Building on top of these features, the region proposal network generates multiple object proposals with possible object candidates.
		In the final step, the final pooling layer (denoted by ``RoI pooling'') discards all proposals that are classified as background.
		Furthermore, the final stage applies a refinement of the bounding boxes that have not been classified as background.
	}
	\label{fig:basics:fasterrcnn}
\end{figure}

Previously, we mentioned the distinction of modern neural-network based object detection algorithms into two-stage architectures such as Faster R-CNN \cite{Ren2015} and single-stage architectures such as Single-Shot detector \cite{Liu2016}, YOLO \cite{Redmon2016}, and RetinaNet \cite{Lin2017}.
A two-stage detector utilizes the previously described object proposal stage using a \ac{RPN} network.
Furthermore, it is also possible to use a preceding \ac{FPN} before the \ac{RPN} for further feature processing to improve the quality of the object proposals \cite{Lin2017a}.
The object proposals as well as the feature maps $\model_{\allparameters_B}(\allsingleinput_{\indexsamples}) = \predoutput_B$ of the backbone network are passed to a fully-connected classification network $\model_{\allparameters_\text{CLS}}$ and to a fully-connected refinement network for the bounding boxes $\model_{\allparameters_\text{BOX}}$ to generate the final object predictions $\predoutput_{\indexsamples}$ with the according bounding box positions $\allpredbboxes_{\indexsamples} \in \bboxset$.
The object proposals are used to crop the relevant features at the proposed object locations from the backbone feature maps $\predoutput_B$.
The classification head $\model_{\allparameters_\text{CLS}}$ takes these features as input to generate the final prediction for the object class which might be one out of $\numclasses$ classes.
The final network output is passed to the softmax activation function to obtain a probabilistic interpretation of the class probabilities for a single object.
For the estimation of the final bounding box of an object, the refinement head $\model_{\allparameters_\text{BOX}}$ works similarly.
The refinement network takes the relevant features from $\predoutput_B$ at the proposed object locations and predicts an offset for the proposed object locations $\Delta{c_{x}}, \Delta{c_{y}}$ and for the proposed width and height $\Delta{\width}, \Delta{\height}$.
Thus, the two-stage detection architecture is a concatenation of the backbone network, the proposal stage, and the detection head, so that the final object predictions are obtained by $(\model_{\allparameters_B} \circ \model_{\allparameters_{\text{RPN}}} \circ \model_{\allparameters_\text{CLS}})(\allsingleinput_\indexsamples)$ for the labels of the objects and $(\model_{\allparameters_B} \circ \model_{\allparameters_{\text{RPN}}} \circ \model_{\allparameters_\text{REG}})(\allsingleinput_\indexsamples)$ for the object positions given an input image $\allsingleinput_\indexsamples$.
During network training, the backbone network, the proposal network as well as the detection head are jointly optimized using the classification and refinement loss of the proposal stage in conjunction with the classification and refinement loss of the detection head.
The concept of the Faster R-CNN detection architecture is schematically shown in \figref{fig:basics:fasterrcnn}.

\subsubsection{Single-Stage Detector: RetinaNet}

\begin{figure}[t!]
    \centering
    \includegraphics[width=\textwidth,]{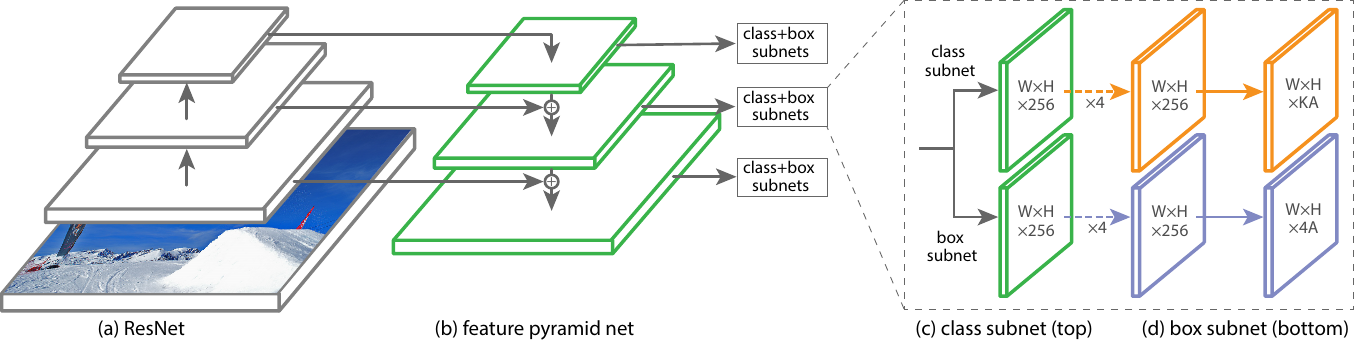}
    
    \vspace{1em}
    \raggedleft{\scriptsize\ieeecopyright{2017}}
    
    \caption[Concept of the RetinaNet object detection architecture.]{
        Concept of the RetinaNet object detection architecture \cite[p.~5, Fig.~3]{Lin2017}.
        Similar to a two-stage detector, the RetinaNet uses a backbone network (left) to generate intermediate feature representations of the input image.
        A feature pyramid network extracts these feature representations at predefined anchor locations and seeks to detect possible correlations between features of different scale.
        Finally, the last refinement stage classifies the anchors based on the intermediate features and applies a bounding box refinement of the anchor boxes to match the estimated objects within an image.
    }
    \label{fig:basics:retinanet}
\end{figure}

A single-stage detection architecture such as a RetinaNet \cite{Lin2017} follows a similar concept of using a pretrained backbone network to generate an intermediate feature representation of the input image.
As opposed to the previously described two-stage object detection approach, a single-stage detector does not utilize an additional proposal stage.
Instead, a single-stage architecture directly works with the anchor boxes that we already know from the \ac{RPN} architecture.
The anchor boxes are placed with fixed object locations and fixed aspect ratios over the whole image output and serve as priors for possible object candidates.
The single-stage object detector seeks to directly learn an appropriate scaling and shifting of these boxes to generate the final object predictions.
Thus, an additional proposal stage is not necessary.
The anchor boxes are connected to certain feature maps of intermediate layers within the backbone network to extract image features at different scales.
Furthermore, the RetinaNet architecture uses a \ac{FPN} before scaling and shifting the anchor boxes to improve the feature extraction process \cite{Lin2017a}.
The output layers $\model_{\allparameters_\text{CLS}}$ and $\model_{\allparameters_\text{BOX}}$ are similar as to the ones used within two-stage detectors as they also estimate the final object class as well as the final object position and size.
The concept of the RetinaNet architecture is schematically shown in \figref{fig:basics:retinanet}.

\section{Different Types of Uncertainty}
\label{section:basics:types_of_uncertainty}

\begin{figure}[b!]
    \begin{subfigure}{0.31\textwidth}
        \includegraphics[width=\linewidth]{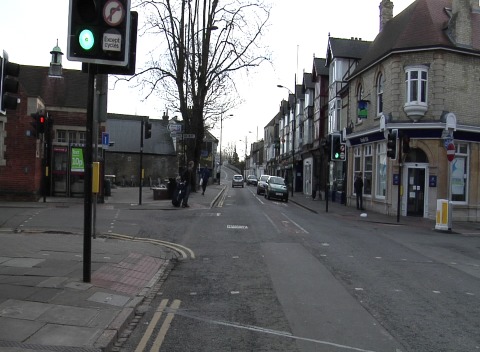}%
        \caption{Input image.}
    \end{subfigure}%
    \hfill%
    \begin{subfigure}{0.31\textwidth}
        \includegraphics[width=\linewidth]{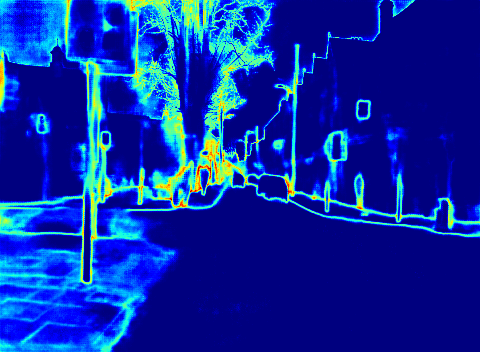}%
        \caption{Aleatoric uncertainty.}
    \end{subfigure}%
    \hfill%
    \begin{subfigure}{0.31\textwidth}
        \includegraphics[width=\linewidth]{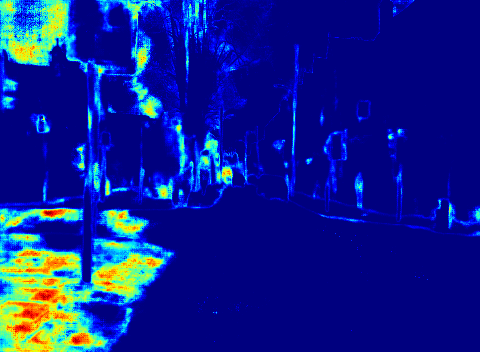}%
        \caption{Epistemic uncertainty.}
    \end{subfigure}
    \vspace{1em}
    
    \raggedleft{\scriptsize\acmcopyright{2017}}
    \caption[Difference between aleatoric and epistemic uncertainty demonstrated at the task of semantic segmentation.]{
        Difference between aleatoric and epistemic uncertainty demonstrated at the task of semantic segmentation \cite[p.~2, Fig.~1]{Kendall2017}.
        We observe that aleatoric data uncertainty occurs on segment boundaries or image regions with blurry contours, e.g., the top of the tree in the image center.
        In contrast, the epistemic uncertainty represents the intrinsic model uncertainty that might occur at different locations that are not well-known by the segmentation model.
    }
    \label{fig:basics:types_of_uncertainty}
\end{figure}

The formerly presented object detection algorithms are deep learning techniques that are able to learn and recognize features and patterns in the given input images.
Although many machine learning and deep learning algorithms are designed to output deterministic estimates, such a process is always subject to certain uncertainties during inference \cite{Kendall2017,Huellermeier2021}.
In this scope, the authors in \cite{Huellermeier2021} identified 3 different reasons why a machine learner is subject to uncertainty:
\begin{enumerate}
    \item The dependency between input space and the respective true outcome (e.g., input image and its associate classification label) may be of stochastic nature \cite{Huellermeier2021}. 
    For example, given two overlapping distributions within the same input space, the translation from input space to the output, which represents the respective distribution label, is not deterministic. 
    This is known as the aleatoric data uncertainty.
    \item Before applied to real-world applications, a machine learning algorithm needs a dedicated training phase to learn the translation from input to output space given a dedicated training set. 
    However, the training set is always a random sample drawn from the data generation process. 
    This leaves a lack of knowledge during application which is referred to as approximation uncertainty by \cite{Huellermeier2021}.
    \item A machine learning or deep learning model might be misspecified \cite{Huellermeier2021}, i.e., it might not be able to correctly capture all dependencies between input and output space. For example, a neural network with low capacity might fail in a classification of input images, whereas a network architecture with high capacity might succeed. The authors in \cite{Huellermeier2021} denote this as model uncertainty \cite{Huellermeier2021}.
\end{enumerate}
These uncertainties are commonly divided into epistemic model and aleatoric data uncertainty \cite{Der2009,Kendall2017,Huellermeier2021}.
The difference between these types of uncertainty is shown in \figref{fig:basics:types_of_uncertainty}.

Epistemic uncertainty represents the lack of knowledge inherent in the model due to a low amount of training data or due to model misspecification \cite{Kendall2017,Huellermeier2021}.
This uncertainty is commonly reducible given more training data \cite{Kendall2017} or by an improved model specification \cite{Huellermeier2021}.
In the context of neural networks, epistemic uncertainty is not directly observable, i.e., we need advanced techniques to access this type of uncertainty.
For this reason, Bayesian neural networks (BNN) have been introduced to mitigate this problem \cite{Mackay1992,Neal2012}.
The idea of a BNN is to place probability distributions over the (commonly deterministic) network weights to yield a probability distribution as the network output.
However, since neural networks exhibit a complex structure and work with intermediate non-linearities, the output distribution is analytically not tractable \cite{Gal2016}.
Therefore, recent works have introduced techniques for an approximation of BNNs.
The authors in \cite{Lakshminarayanan2017} use an ensemble of neural networks during inference with the same architecture to yield a sample distribution as output given a single input.
The authors in \cite{Gal2016} show that dropout, a common regularization technique where single neurons get randomly deactivated within the network training, can also be used as an ensemble approximation during network inference given new data.
Another technique is \ac{SVI} \cite{Hoffman2013} where the network weights are replaced by variational distributions of known parametric form (e.g., Gaussian).
Using \ac{SVI}, it is possible to approximate the variational distribution parameters during network training.
During inference, we can sample from these variational distributions to get multiple parameter sets for the neural network.
However, a known problem of \ac{SVI} is that is does not scale well to large neural networks \cite{Hoffman2013}.
All of these techniques approximate a BNN by yielding a sample distribution for the network output given a certain input.
Recently, the authors in \cite{Postels2019} derived a sampling-free technique to yield epistemic uncertainty by using Gaussian error propagation as an approximation to BNNs.
In this way, it is possible to represent the intermediate as well as the final uncertainty as Gaussian distributions that are propagated through the whole network architecture \cite{Postels2019}.
Using all of these techniques, it is possible to obtain epistemic uncertainty of a neural network.
If transferred to object detection architectures (cf. \secref{section:basics:object_detection}), we face a problem of BNNs during the inference of objects.
A \fasterrcnn{} as well as a \retinanet{} both work with proposal or anchor boxes before the final refinement stage.
If we now apply the sampling techniques as described above, we obtain multiple bounding box estimates for the same image.
Furthermore, the assignment of the proposal/anchor boxes to individual objects might change with each forward pass in the sampling process.
Thus, the authors in \cite{Harakeh2020} and \cite{Feng2021} use clustering techniques to group the predicted objects after sampling.
Consequently, it is also possible to obtain epistemic uncertainty within the detection process.

\begin{figure}[t]
	\centering
	\begin{overpic}[width=0.6\linewidth, tics=5, trim=-1em -1em 0 0]{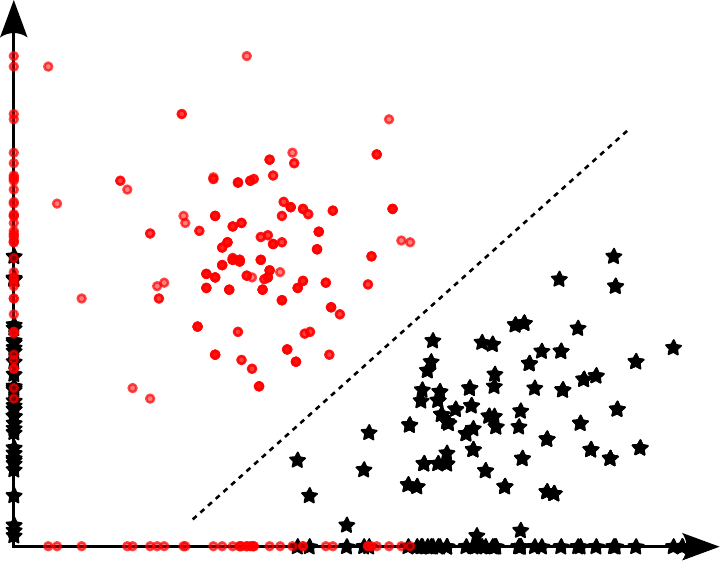}
		\put(42, 0){dimension 1}
		\put(0, 32){\rotatebox{90}{dimension 2}}
	\end{overpic}
	\caption[
		Given two non-overlapping distributions in a 2-D feature space, the distributions are separable and we have no aleatoric uncertainty.
		However, if projected only to a single dimension and provided to a machine learning model, the distributions might overlap which results in regions with aleatoric uncertainty.
	]{
		Given two non-overlapping distributions in a 2-D feature space, the distributions are separable and we have no aleatoric uncertainty.
		However, if projected only to a single dimension and provided to a machine learning model, the distributions might overlap which results in regions with aleatoric uncertainty.
	}
	\label{fig:basics:separable_uncertainty}
\end{figure}

Aleatoric uncertainty represents the uncertainty inherent in the data.
This kind of uncertainty can not be explained away given more data \cite{Kendall2017}.
However, the authors in \cite{Huellermeier2021} argue that aleatoric uncertainty might also result from a misspecification of the problem setting.
For example, given two non-overlapping distributions in a 2-D feature space, the distributions are separable and we have no aleatoric uncertainty.
However, if projected only to a single dimension and provided to a machine learning model, the distributions might overlap in this dimension which results in a region with aleatoric uncertainty.
This is schematically shown in \figref{fig:basics:separable_uncertainty}.
This example shall illustrate that aleatoric uncertainty might be mitigated given enough information to a machine learning algorithm \cite{Huellermeier2021}.
The advantage of working with aleatoric uncertainty is that we can train a neural network to directly output an estimate of this kind of uncertainty.
Since we are working with object detection algorithms, we further need to distinguish between semantic label and spatial position uncertainty.

\subsection{Modeling of Semantic Confidence}
\label{section:introduction:semantic}

Semantic confidence describes the belief of the learning model about the object category of a single object.
A neural network commonly estimates a score $\logitvariate_\indexclasses \in \realdigits$ (logit) for each class $\indexclasses$ in the set of all classes $\outputset = \{1, ..., \numclasses\}$.
Afterwards, the logit of a neural network can be converted to a confidence score $\predconfidencevariate_\indexclasses \in [0, 1]$ for each class $\indexclasses \in \{1, ..., \numclasses\}$ using the softmax function $\softmax(\logitvariate)_\indexclasses$.
The advantage of using the softmax function is that it yields confidence scores which sum up to $1$ and which, in turn, can be used to construct a categorical probability distribution targeting the predicted class for each object.
In contrast, for binary classification with $\outputset = \{0, 1\}$, a neural network only predicts a single logit $\logitvariate$ which is converted to a confidence score using the sigmoid function $\sigmoid(\logitvariate)$.
Similar to the multivariate case, this confidence score is used to construct a Bernoulli distribution which reflects the probability for an outcome belonging either to class $0$ or $1$, respectively.
In both cases, we treat the network output as a probabilistic forecast which allows to derive the uncertainties directly by the network output.

\subsection{Modeling of Spatial Uncertainty}
\label{section:introduction:spatial}

The spatial uncertainty reflects the uncertainty during the prediction of an object position and shape within a single image.
For position and shape estimation, common object detection architectures use anchor or proposal boxes as the basis for inference (cf. \secref{section:basics:object_detection:architectures}).
A neural network learns to rescale and shift these boxes during training to match possible objects within an image.
This architecture has the advantage that a neural network only needs to learn a rescaling and shifting of these boxes which is a simple regression task.
Let $\allinputvariates \in \inputset$ denote the input to an object detector within the input set $\inputset$ and let $\dataset$ denote a training data set of size $\numsamples$ which is used to train the parameters $\allparameters$ of a neural network.
Within the training data set, the variable $\allgroundtruthbboxes \in \bboxset$ represents the ground-truth bounding box information of individual objects of dimension $\numbboxdims$ in the space $\bboxset$ consisting of the position, width, and height information.
A standard regression model interprets the network output $\allpredbboxvariates$ as multiple independent normal distributions for each bounding box dimension with fixed variance, so that $\allpredbboxvariates \sampledfrom \normaldistribution(\meanvec_{\allpredbboxvariates|\allinputvariates}, \cov)$, with $\meanvec_{\allpredbboxvariates|\allinputvariates} \in \bboxset$ as the predicted mean vector and $\cov = \diag(\variance_1, \ldots, \variance_\numbboxdims)$ as the diagonal covariance matrix with variances $\variance_1, \ldots, \variance_\numbboxdims \in \realdigitspositive$.
The variances are commonly treated as fixed constants, so that $\cov = \variance \identity$ with $\variance$ as a common variance (e.g., $1$) and $\identity$ as the identity matrix.
Since the variances are treated as constants, they are neglected within standard regression applications.
A regression model is trained using the \ac{NLL} of the model given the training set, so that the loss is defined by
\begin{align}
    \loss(\allparameters) 	&= -\log\Bigg( \prod_{\indexsamples=1}^{\numsamples} \pdf_{\allpredbboxvariates}(\allgroundtruthbboxes_\indexsamples | \allsingleinput_\indexsamples, \allparameters) \Bigg) \\
    &\propto \sum_{\indexsamples=1}^{\numsamples} \sum_{\indexbboxdims=1}^{\numbboxdims} \Big(\singlegroundtruthbbox_{\indexsamples, \indexbboxdims} - \mean_{\predbboxvariate_{\indexbboxdims}|\allsingleinput_{\indexsamples}} \Big)^2 ,
\end{align}
which is also known as the squared error loss function.

A drawback of this interpretation is that the uncertainty, i.e., the variance, is not considered during model training and inference. 
Thus, no information about the spatial uncertainty is available.
For this reason, recent work has reformulated this loss by also considering the variance as a function of the input data.
Therefore, the covariance matrix $\cov_{\allpredbboxvariates|\allinputvariates} \in \realdigitspositive^{\numbboxdims \times \numbboxdims}$ consisting of the independent variances $\variance_{\predbboxvariate_{\indexbboxdims}|\allinputvariates}$ for each dimension $\indexbboxdims \in \{1, \ldots, \numbboxdims\}$ is also modeled by the regression network for each input $\allsingleinput_\indexsamples$, so that $\cov_{\allpredbboxvariates|\allinputvariates} = \diag\big(\variance_{\predbboxvariate_{1}|\allinputvariates}, \ldots, \variance_{\predbboxvariate_{\indexbboxdims}|\allinputvariates}\big)$ \cite{Kendall2017}.
Thus, the \ac{NLL} extends to
\begin{align}
    \loss(\allparameters) 	&= -\log\Bigg( \prod_{\indexsamples=1}^{\numsamples} \pdf_{\allpredbboxvariates}(\allgroundtruthbboxes_\indexsamples | \allsingleinput_\indexsamples, \allparameters) \Bigg) \\
    &= -\log\Bigg( \prod_{\indexsamples=1}^{\numsamples} \prod_{\indexbboxdims=1}^{\numbboxdims} \frac{1}{\sqrt{2\pi\variance_{\predbboxvariate_{\indexbboxdims}|\allsingleinput_{\indexsamples}}}} \exp\Bigg[\frac{-\big(\singlegroundtruthbbox_{\indexsamples, \indexbboxdims} - \mean_{\predbboxvariate_{\indexbboxdims}|\allsingleinput_{\indexsamples}}\big)^2}{2\variance_{\predbboxvariate_{\indexbboxdims}|\allsingleinput_{\indexsamples}}}\Bigg] \Bigg) \\
    &\propto \sum_{\indexsamples=1}^{\numsamples} \sum_{\indexbboxdims=1}^{\numbboxdims} \frac{1}{2} \stddev_{\predbboxvariate_{\indexbboxdims}|\allsingleinput_{\indexsamples}}^{-2} \Big(\singlegroundtruthbbox_\indexsamples - \mean_{\predbboxvariate_{\indexbboxdims}|\allsingleinput_{\indexsamples}}\Big)^2 + \frac{1}{2}\log \Big(\variance_{\predbboxvariate_{\indexbboxdims}|\allsingleinput_{\indexsamples}} \Big) .
\end{align}
For the joint training of mean and variance, no additional ground-truth information is required.
Instead, the network is able to increase the uncertainty which decreases the loss for uncertain samples during model training.
This type of regression has recently been used in the context of object detection \cite{He2019,Hall2020,Feng2021}.
Therefore, it is possible to obtain semantic as well as spatial uncertainty within the scope of object detection. 

\section{Reasons for Unreliable Uncertainty}
\label{section:basics:unreliable_uncertainty}

Recent work has found that modern neural networks tend to produce unreliable uncertainty information \cite{Niculescu2005,Naeini2015,Guo2018}, i.e., the estimated uncertainty does not match the observed error distribution.
The authors in \cite{Guo2018} found that especially modern architectures with a high model capacity produce too overconfident confidence estimates, whereas simple models with a lower model capacity offer better calibration properties \cite{Guo2018}.
Furthermore, the authors study the effect of additional regularization techniques such as batch nor\-mal\-iza\-tion \cite{Ioffe2015} and weight decay \cite{Vapnik1998} (for a description of these regularization techniques, cf. \secref{section:basics:object_detection:fcn}).
The authors found that the use of batch normalization tends to increase miscalibration of neural networks \cite{Guo2018} but the authors leave the interpretation of this phenomenon open for future work.
Weight decay \cite{Vapnik1998} is an additional penalty added to the loss term which penalizes large layer weights to prevent an overfitting of the network to the training data.
As opposed to batch normalization, the authors in \cite{Guo2018} found that weight decay tends to minimize the model miscalibration.
We support this statement during our experiments as we observed that high network weights tend to produce less calibrated neural networks.

Finally, although the \ac{NLL} is a direct measure for calibration \cite{Gneiting2007,Gneiting2007a,Naeini2015,Kull2015,Guo2018} and used during model training, the authors in \cite{Guo2018} observe that the \ac{NLL} might disconnect from optimizing the calibration properties of the neural network during model training.
This is in agreement with other works as they show that the \ac{NLL} can be decomposed into a classification and a refinement loss \cite{Gneiting2007,DeGroot1983,Kull2015,Song2019}.
As already pointed out in the last section, a neural network predicts a probability distribution targeting the desired outcome, e.g., a Bernoulli distribution within binary classification or a normal distribution within a regression setting.
We further denote the neural network by $\model(\allsingleinput)$ that predicts a probability distribution $\pdf_{\predoutputvariate}(\predoutput | \allsingleinput)$ for a single sample with input $\allsingleinput \in \inputset$ and the according ground-truth $\groundtruthoutput \in \gtset$, where $\gtset = \{0, 1\}$ or $\gtset = \realdigits$ within binary classification or regression, respectively.
Furthermore, let $\allestimatedoutputparameters_{\allsingleinput}$ denote the estimated distribution parameters of $\pdf_{\predoutputvariate}(\predoutput | \allsingleinput)$. 
For example, within binary classification, the network outcome is a Bernoulli distribution $\pdf_{\predoutputvariate}(\predoutput | \allsingleinput) = \bernoullidistribution(\predoutput; \predconfidence_{\allsingleinput})$ with confidence $\predconfidence_{\allsingleinput}$, so that $\allestimatedoutputparameters_{\allsingleinput} = \{\predconfidence_{\allsingleinput}\}$.
In contrast, in the context of probabilistic regression the output is parameterized in terms of a normal distribution $\pdf_{\predoutputvariate}(\predoutput | \allsingleinput) = \normaldistribution(\predoutput; \mean_{\predoutput|\allsingleinput}, \variance_{\predoutput|\allsingleinput})$ with mean $\mean_{\predoutput|\allsingleinput}$ and variance $\variance_{\predoutput|\allsingleinput}$, so that $\allestimatedoutputparameters_{\allsingleinput} = \{\mean_{\predoutput|\allsingleinput}, \variance_{\predoutput|\allsingleinput}\}$.

The predicted probability distribution is a direct estimate of the neural network about its uncertainty within the predicted outcome. 
Depending on the capacity of the used neural network, this estimation might not necessarily follow the real probability distribution about the ground-truth $\pdf_{\groundtruthvariate}(\groundtruthoutput | \allestimatedoutputparameters_{\allsingleinput})$ on the true data generation process given the network output $\allestimatedoutputparameters_{\allsingleinput}$.
The network is trained by the \ac{NLL} whose expectation over the data is denoted by $\expectation_{\allinputvariates, \groundtruthvariate\sampledfrom\pdf_{\allinputvariates,\groundtruthvariate}}\Big[-\log\big( \pdf_{\predoutputvariate}(\groundtruthoutput | \allsingleinput) \big) \Big]$\footnote{For notational brevity, we further use an abbreviated expression to denote the random variable of the expectation, e.g., for $\expectation_{\allinputvariates\sampledfrom\pdf_{\allinputvariates}}[\cdot]$, we write $\expectation_{\allinputvariates}[\cdot]$.}.
In this sense, recent work has derived the decomposition of the expected \ac{NLL} into calibration and refinement loss \cite{Gneiting2007,DeGroot1983,Kull2015,Song2019} given by
\begin{align}
    \expectation_{\allinputvariates, \groundtruthvariate}[\text{NLL}] &= \expectation_{\allinputvariates, \groundtruthvariate} \Big[-\log\big( \pdf_{\predoutputvariate}(\groundtruthoutput | \allsingleinput) \big) \Big] \\ 
    &= \expectation_{\allinputvariates} \bigg[ \expectation_{\groundtruthvariate} \Big[ -\log\big( \pdf_{\predoutputvariate}(\groundtruthoutput | \allsingleinput) \big) + \log\big( \pdf_{\groundtruthvariate}(\groundtruthoutput | \allestimatedoutputparameters_{\allsingleinput}) \big) - \log\big( \pdf_{\groundtruthvariate}(\groundtruthoutput | \allestimatedoutputparameters_{\allsingleinput}) \big) \Big] \bigg] \\
    &= \expectation_{\allinputvariates} \Bigg[ \expectation_{\groundtruthvariate} \bigg[ \log \bigg( \frac{\pdf_{\groundtruthvariate}(\groundtruthoutput | \allestimatedoutputparameters_{\allsingleinput})}{\pdf_{\predoutputvariate}(\groundtruthoutput | \allsingleinput)} \bigg) \bigg] \Bigg] - \expectation_{\allinputvariates, \groundtruthvariate} \Big[ -\log\big( \pdf_{\groundtruthvariate}(\groundtruthoutput | \allestimatedoutputparameters_{\allsingleinput}) \big) \Big] \\
    \label{eq:introduction:reasons:decomposed_nll}
    &= \underbrace{\expectation_{\allinputvariates} \Big[ \kldivergence\big( \pdf_{\groundtruthvariate}(\groundtruthvariate | \allestimatedoutputparameters_{\allsingleinput}) || \pdf_{\predoutputvariate}(\groundtruthvariate | \allsingleinput) \big) \Big]}_{\text{Calibration loss}} + \underbrace{\expectation_{\allinputvariates, \groundtruthvariate} \Big[ -\log \big( \pdf_{\groundtruthvariate}(\groundtruthoutput | \allestimatedoutputparameters_{\allsingleinput}) \big) \Big]}_{\text{Refinement loss}} ,
\end{align}
where 
\begin{align}
    \kldivergence\big( \pdf_{\groundtruthvariate}(\groundtruthvariate | \allestimatedoutputparameters_{\allsingleinput}) || \pdf_{\predoutputvariate}(\groundtruthvariate | \allsingleinput) \big) = \int_{\gtset} \pdf_{\groundtruthvariate}(\groundtruthvariate | \allestimatedoutputparameters_{\allsingleinput}) \log \bigg( \frac{\pdf_{\groundtruthvariate}(\groundtruthvariate | \allestimatedoutputparameters_{\allsingleinput})}{\pdf_{\predoutputvariate}(\singlegroundtruth | \allsingleinput)} \bigg) \diff \singlegroundtruth
\end{align}
is the Kullback-Leibler divergence between the uncertainty distribution $\pdf_{\groundtruthvariate}(\groundtruthvariate | \allestimatedoutputparameters_{\allsingleinput})$ of the true data generation process given the network output $\allestimatedoutputparameters_{\allsingleinput}$ and the estimated uncertainty about the predicted outcome given by $\pdf_{\predoutputvariate}(\groundtruthvariate | \allsingleinput)$.
The refinement loss is responsible to improve the network accuracy on the given ground-truth data set.
The calibration loss serves as a regularizing term to match the predicted network confidence with the observed error.
The authors in \cite{Guo2018} assume that overfitting, which might be a direct cause of a model with too high capacity, might manifest in the calibration error rather than in the refinement loss during model training.
The authors in \cite{Guo2018} argue that once the model is not capable of further improving the network accuracy on the given data, the training process seeks to minimize the calibration loss which results in an overfitting to the training data.

Further related works \cite{Kumar2018,Mukhoti2020} seek to tackle this problem by directly incorporating additional reg\-u\-lar\-iza\-tion during model training.
The authors in \cite{Kumar2018} introduce a regularization technique that is related to the \ac{ECE} known from classification calibration and use this technique to implement a more targeted regularization during model training.
In addition, \cite{Mukhoti2020} study the effect of the focal loss \cite{Lin2017} which is designed to set a higher weight on misclassified samples during model training.
However, none of these approaches address the problem of overfitting during network training directly as they solely incorporate a penalty on high confidences but neglect the baseline generalization performance of a neural network.
For example, in a binary classification setting, a neural network might perfectly predict the right output label for all samples in the training set.
Since the regularization techniques proposed by \cite{Kumar2018} and \cite{Mukhoti2020} are based on the error on the training set, the probability mass is completely set to the correct label in this example and the additional regularization would not have any effect.
This results in a perfect calibration on the training set.
However, this behavior is commonly known as an overfitting on the training set.
In this case, none of the regularization techniques is able to capture the ``true'' probability distribution on real-world data which is out of the training distribution.
This overfitting behavior, which is also a lack of generalization, is a major reason for uncertainty miscalibration \cite{Guo2018,Minderer2021}.

A more recent study on the reasons for miscalibration by \cite{Minderer2021} examines the actual network architectures such as MLP-Mixer (fully-connected classification networks) \cite{Tolstikhin2021}, Transformer clas\-si\-fi\-ca\-tion models \cite{Dosovitskiy2020}, and modern ResNet \cite{He2016} architectures.
In their studies, the authors in \cite{Minderer2021} show that a higher model capacity does not necessarily lead to an increased miscalibration of models as modern architectures tend to improve the generalization performance \cite{Minderer2021}.
Furthermore, their examinations are divided into in-distribution and out-of-distribution data for classification networks on several subsets of the ImageNet \cite{Deng2009} data set.
The authors in \cite{Minderer2021} found that a good calibration performance of a neural network also has a positive influence on the calibration on out-of-distribution data.
Similar to \cite{Guo2018}, their examinations show a correlation between network capacity and miscalibration on in-distribution data which, however, is not as large as within the examinations of \cite{Guo2018}.
Interestingly, \cite{Minderer2021} also found that models with a higher generalization performance tend to have a better calibration performance on out-of-distribution data as these architectures seem to be more robust under domain shift \cite{Minderer2021}.


\acresetall
\chapter{Semantic Confidence Calibration}
\label{chapter:confidence}

In the scope of image-based environment perception, we recently introduced the basic concepts of object detection models that are based on neural networks.
A neural network estimates a score for each prediction that can be interpreted as a probability of correctness of the estimate.
This confidence score reflects the aleatoric uncertainty about the semantic class membership for each prediction \cite{Malinin2018}.
However, modern neural networks in the scope of classification are known to produce overconfident confidence estimates \cite{Naeini2015,Guo2018}, i.e., the predicted confidence does not match the observed accuracy. 
This is a major safety concern as such a bias within the confidence predictions might have a large impact on subsequent processes, e.g., object tracking.
Thus, it is desirable to measure the misalignment between predicted confidence and observed accuracy.
If we assert a systemic error within the objectness confidence, it is further possible to apply post-hoc calibration methods that seek to correct such a misalignment.
For classification tasks, several methods such as Logistic Calibration \cite{Platt1999}, Temperature Scaling \cite{Guo2018}, or Beta Calibration \cite{Kull2017} exist to correct biased confidence scores after model inference. 

\begin{figure}[b!]
    \centering
    \input{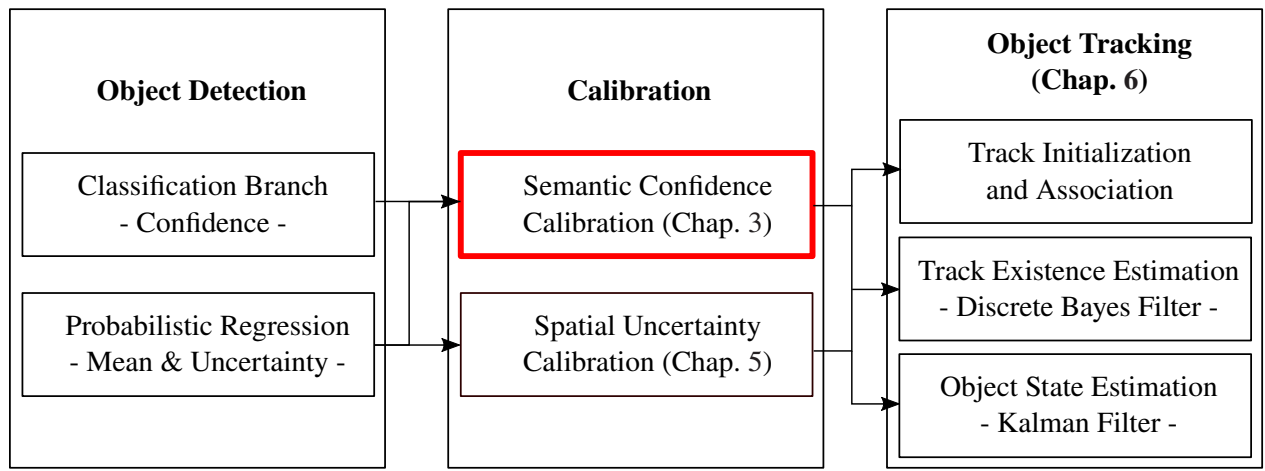}%
    \caption[Concept in this work with focus on semantic confidence calibration in this section.]{
        In this chapter, we focus on the evaluation and calibration of the semantic confidence that is of major importance for the management of tracks within a subsequent object tracking.
        Therefore, reliable and calibrated confidence information are of special interest.
    }
    \label{fig:confidence:blockimage_thesis}
\end{figure}

If we consider the more advanced tasks of object detection and segmentation, we face new challenges as we work with more complex models and within more complex environments.
These tasks are of special interest as they are commonly used within the environment perception process, e.g., for safety-critical applications such as autonomous driving.
For example, a detection model can be used within an object tracking process (cf. \chapref{chapter:tracking}).
In this context, the semantic confidence is used to decide which tracks are kept and which are discarded.
Thus, a reliable self-assessment of the neural network about the semantic uncertainty is mandatory.
However, if we detect a bias in the semantic confidence, calibration methods can be used to correct such a misalignment between predicted confidence and observed model performance.
The concept of confidence calibration within an object tracking pipeline is schematically shown in \figref{fig:confidence:blockimage_thesis}.
In this chapter, we solely focus on the task of object detection and examine the calibration properties of detection and segmentation models.
Furthermore, we investigate novel calibration methods that are designed for detection and segmentation applications.
These models are able to use additional information such as the object location or size for confidence recalibration.
We investigate the influence of semantic confidence calibration to the task of object detection in \chapref{chapter:tracking}.

We present the definition for classification calibration and transfer this to the task of object detection, instance segmentation, and semantic segmentation.
These definitions are given in \secref{section:confidence:definition}.
Object detection is a joint task of classification and regression as a detection model does not only estimate the class membership but also the position of individual objects. 
Thus, we investigate the influence of the model's regression branch to the calibration properties and introduce an additional position-dependence to the definition for confidence calibration. 
Additionally, we extend common calibration metrics (\secref{section:confidence:definition}) as well as common calibration methods (\secref{section:confidence:methods}) to measure miscalibration and to apply post-hoc calibration for detection and segmentation models, respectively. This work has been subject of our previous publications in \cite{Kueppers2020,Schwaiger2021,Kueppers2022a}.
Finally, we evaluate our extended calibration methods as well as the definition for Bayesian confidence calibration in \secref{section:confidence:experiments} for different neural network architectures and for different data sets. We summarize our contributions and findings in \secref{section:confidence:conclusion}.

\textbf{Contributions:} We summarize our contributions in the following:
\begin{itemize}
    \item Definition of confidence calibration for object detection, instance, and semantic segmentation.
    \item Multivariate extension of calibration methods to include additional position information.
    \item Extensive studies on the effect of position-dependent confidence calibration.
\end{itemize}

\section{Related Work in the Context of Confidence Calibration}
\label{section:confidence:related_work}

Recent works have started a discussion about the calibration of the class confidence of probabilistic forecasters \cite{Niculescu2005,Gneiting2007,Naeini2015,Guo2018}. 
The common consent about the calibration of semantic uncertainty is that the estimated confidence of a classifier can be interpreted as a probability mass function (Bernoulli or categorical dis\-tribution for multivariate cases) which should reflect the model's uncertainty about the actual prediction. 
Thus, confidence calibration is defined as the task of matching the predicted confidence with the observed accuracy given a certain confidence level.
It is possible to measure miscalibration using the expected calibration error (ECE) \cite{Naeini2015} which is derived from the definition of classification calibration. 
For ECE computation, all samples of a data set are grouped by their predicted confidence information into several equally sized distinct bins.
Afterwards, it is possible to measure the accuracy within each bin.
This yields multiple accuracy scores that are conditioned on a certain confidence range which is an approximation to the definition of calibration.
Finally, an overall miscalibration score is computed using the weighted sum of the differences between observed accuracy and average confidence within each bin \cite{Naeini2015}.
Besides the ECE, proper scoring rules such as Brier score or negative log likelihood also measure the calibration properties of a classifier inherently \cite{Broecker2009,Ovadia2019}. 
Proper scoring rules are metrics that are minimized if and only if the predicted probability distribution is optimized towards the ground-truth data. 
These scoring rules can be decomposed into a calibration and a refinement part. 
We refer to \cite{DeGroot1983,Broecker2009,Ovadia2019} for a more detailed discussion.

Besides measuring miscalibration, recent work has developed methods for a subsequent (post-hoc) cali\-bration of uncalibrated confidence estimates to calibrated ones.
We distinguish between binning methods and scaling methods.
Similar to the computation of the ECE, the binning methods group all samples by their predicted confidence and use a binning scheme to measure the accuracy conditioned on the confidence score.
This binning scheme is then used to reassign the observed accuracy as the calibrated confidence to grouped samples.
This technique is also referred to as Histogram Binning \cite{Zadrozny2001}.
A more dynamic but related approach is Isotonic Regression \cite{Zadrozny2002} where a strictly monotonically increasing step function is fit to the training data.
This function can be interpreted as a binning scheme with dynamic bin ranges and a dynamic amount of bins.
Further extensions of these methods are \ac{BBQ} \cite{Naeini2015} and \ac{ENIR} \cite{Naeini2015a} which are ensembles of multiple Histogram Binning and multiple Isotonic Regression models, respectively.

In contrast, scaling methods such as Logistic Calibration (or Platt scaling) \cite{Platt1999}, Temperature Scaling \cite{Guo2018}, Beta Calibration \cite{Kull2017}, or Dirichlet calibration \cite{Kull2019} apply a rescaling of the logits before the sigmoid or softmax operation to yield calibrated confidence estimates.
These scaling methods differ in their assumptions on the input data and the resulting recalibration scheme.
Furthermore, Temperature Scaling \cite{Guo2018} and Dirichlet calibration \cite{Kull2019} are designed for the recalibration of multiclass classification tasks. The scaling methods are known to be sample efficient compared to the binning methods \cite{Kumar2019}, i.e., these methods provide a qualitatively good calibration performance given only a small amount of samples, compared to the binning methods. 
However, these methods are also restricted by means of their underlying parametric assumptions \cite{Kumar2019}, e.g., Gaussian distributions for Logistic Calibration \cite{Platt1999,Kull2017}.
For this reason, the authors in \cite{Kumar2019} recently proposed a scaling-binning calibration method that combines the sample-efficiency of the scaling methods with the representational power of a Histogram Binning. 
A similar approach is applied by \cite{Ji2019} who proposed a bin-wise Temperature Scaling for calibration.
Besides the binning and scaling methods, the authors in \cite{Wenger2020} proposed a non-parametric \ac{GP} calibration scheme using a latent GP model with a categorical likelihood to perform recalibration for multiclass classification tasks.
Another approach by \cite{Gupta2021} fits a spline function to the uncalibrated data to achieve confidence calibration.

Some research has also focused on confidence calibration directly during model training by regularization.
The authors in \cite{Pereyra2017} propose a confidence penalty during the training of a neural network model to reduce overconfidence.
Similarly, the authors in \cite{Mukhoti2020} study the effect of the focal loss function \cite{Lin2017} as the training objective to confidence calibration.
They found that focal loss significantly reduces overconfidence but might also lead to underconfident models as well \cite{Schwaiger2021}.
The drawback of these approaches is that they simply apply untargeted penalties to the confidence, regardless of the observed accuracy.
This is addressed by \cite{Kumar2018} who propose the \ac{MMCE} which is a differentiable surrogate for the ECE and which can be used as a regularization term during model training.
A common drawback of calibration methods during model training is that the baseline performance of a forecaster (e.g., accuracy or precision) is commonly way better on the training data compared to new samples during inference.
This has an impact on the regularization and commonly leads to a distortion of the confidences during model training as well.
In this chapter, we focus on post-hoc calibration and especially on the standard calibration methods Histogram Binning \cite{Zadrozny2001}, Logistic Calibration \cite{Platt1999}, and Beta Calibration \cite{Kull2017}, as these methods are most widely used.
Furthermore, it is straightforward to utilize and extend these methods to our task of object detection and segmentation calibration.

Recently, the authors in \cite{Rottmann2020} proposed the concept of meta classification which is quite related to our approach of extended and feature-aware confidence calibration.
Within meta classification, different sources of uncertainty are aggregated which are provided by a forecaster.
Afterwards, a simple classification model (e.g., Logistic Calibration) is used to produce the final outcome.
This concept has been extended to object detection by \cite{Schubert2021} which is denoted as MetaDetect and utilizes further features such as bounding box position or object size.
In the context of this work, we focus on the characteristics of calibration. 
Furthermore, we derive our extended methods directly from existing calibration methods and set them in an overall context. 
In addition, we present a multivariate extension of the calibration methods that also allows for the detection of possible correlations between the individual features.


\section{Definitions and Metrics for Confidence Calibration}
\label{section:confidence:definition}

In this section, we present the definition of semantic confidence calibration for classification and extend it to the tasks of object detection, instance segmentation, and semantic segmentation. 
The derivation of these definitions was the subject of our research in \cite{Kueppers2020}, \cite[p. 228 f.]{Kueppers2022a}, and \cite[p. 230 f.]{Kueppers2022a}.

\subsection{Classification}
\label{section:confidence:definition:classification}

Let $\dataset$ denote a data set of size $\numsamples$ which consists of several images $\allinputvariates \in \inputset$ with height $\Height$, width $\Width$, and number of channels $\Channels$. 
Each image belongs to a certain ground-truth class $\groundtruthvariate \in \gtset = \{1, ..., \numclasses\}$, so that the joint distribution for the input $\allinputvariates$ and the respective ground-truth label $\groundtruthvariate$ is given by $\pdf_{\allinputvariates, \groundtruthvariate}(\allsingleinput, \singlegroundtruth)=\pdf_{\groundtruthvariate}(\singlegroundtruth|\allsingleinput) \pdf_{\allinputvariates}(\allsingleinput)$.
A multiclass classification model seeks to approximate $\pdf_{\groundtruthvariate}(\singlegroundtruth|\allsingleinput)$ by learning its parameters so that it is able to predict a label $\predoutputvariate \in \gtset$ with a certain confidence score $\predconfidencevariate \in \probset = [0, 1]$ that represents the model's belief about the prediction's correctness.
These predictions follow the joint model distribution $\pdf_{\predconfidencevariate, \predoutputvariate}(\predconfidence, \predoutput | \allsingleinput)$.
A classification model is confidence calibrated, if
\begin{align}
	\label{eq:confidence:definition:classification}
	\prob(\predoutputvariate = \groundtruthvariate | \predconfidencevariate = \predconfidence) = \predconfidence
\end{align}
is fullfilled for all $\predconfidence \in \probset$ \cite{Niculescu2005,Naeini2015,Guo2018}. 
This definition implies that the observed accuracy $\prob(\predoutputvariate = \groundtruthvariate | \predconfidencevariate = \predconfidence)$ should match the estimated confidence given a certain confidence level $\predconfidence$.
For example, given $100$ predictions with a confidence score of $0.8$ each, we would expect an accuracy of also $80\%$. 
If we observe a deviation, a model is said to be miscalibrated.
Note that for binary classification with $\gtset = \{0, 1\}$, the network output of a classification model is commonly a sigmoidal function that outputs a score in the $(0, 1)$ interval, indicating the belief for $\predoutputvariate = 1$. 
Thus, our calibration target is the observed \textit{relative frequency} for $\groundtruthvariate = 1$ instead of the accuracy, so that
\begin{align}
	\label{eq:confidence:definition:binary}
	\prob(\groundtruthvariate = 1 | \predconfidencevariate = \predconfidence) = \predconfidence
\end{align}
is required for all $\predconfidence \in \probset$. 

As already stated in the introduction in \secref{section:introduction:semantic}, the predicted probability distribution for $\predoutputvariate$ can be expressed in terms of a Bernoulli distribution with confidence $\predconfidencevariate$ as the probability parameter, so that the estimated probability mass function is defined by
\begin{align}
    \label{eq:confidence:definition:bernoulli}
    \pdf_{\predoutputvariate}(\predoutput | \predconfidence) = \bernoullidistribution(\predoutput; \predconfidence) = \predconfidence^{\predoutput}(1-\predconfidence)^{1-\predoutput} .
\end{align}

The definition for confidence calibration in (\ref{eq:confidence:definition:classification}) can be used to derive the \ac{ECE} \cite{Naeini2015} that is a common metric to measure miscalibration in the scope of classification \cite{Naeini2015,Guo2018}.
We seek to minimize the expectation of the difference between predicted confidence and observed accuracy \cite{Naeini2015} which is denoted by
\begin{align}
	\label{eq:confidence:ece:continuous}
	&\expectation_{\predconfidencevariate \sampledfrom \pdf_{\predconfidencevariate}} \Big[ \big| \prob(\predoutputvariate = \groundtruthvariate | \predconfidencevariate = \predconfidence) - \predconfidence \big| \Big] \\
    &= \int_{0}^{1} \pdf_{\predbboxvariate}(\predconfidence) \cdot \big| \prob(\predoutputvariate = \groundtruthvariate | \predconfidencevariate = \predconfidence) - \predconfidence \big| \diff \predconfidence \\
    \label{eq:confidence:ece:integral}
    &= \int_{0}^{1} \big| \prob(\predoutputvariate = \groundtruthvariate | \predconfidencevariate = \predconfidence) - \predconfidence \big| \diff \cdf_{\predconfidencevariate}(\predconfidence) ,
\end{align}
using the Rieman-Stieltjes integral and with $\cdf_{\predconfidencevariate}(\predconfidence)$ as the \ac{CDF} of the estimated confidence distribution.
However, since $\predconfidencevariate$ is a continuous random variable, we can not get the probability in (\ref{eq:confidence:ece:continuous}) using a finite set of samples \cite{Guo2018}.
Therefore, the \ac{ECE} is approximated by a binning scheme over estimated confidence distribution of $\predconfidencevariate \in \probset$ using $\numbins$ equally sized bins.
Given a finite data set, the \ac{ECE} \cite{Naeini2015}, \cite[p. 11]{Guo2018} is an approximation of (\ref{eq:confidence:ece:integral}) given by
\begin{align}
	\label{eq:confidence:ece:riemann_approx}
	&\ece := \sum^{\numbins}_{\indexbins=1} \prob(\predconfidencevariate \in \bin_\indexbins) \cdot \big| \prob(\predoutputvariate = \groundtruthvariate | \predconfidencevariate = \predconfidence_{\indexbins}) - \predconfidence_{\indexbins} \big| , \\ \nonumber
    &\forall \predconfidence_{\indexbins} \in \bin_\indexbins, \indexbins \in \{1, \ldots \numbins\}.
\end{align}
where $\predconfidence_{\indexbins}$ denote all possible confidences within interval $\bin_\indexbins$.
In other words, the \ac{ECE} is calculated using the weighted sum of the differences between average confidence and observed accuracy/frequency over all bins using a finite set of samples. 
This yields the representation form
\begin{align}
	\ece = \sum^{\numbins}_{\indexbins=1} \frac{\numsamples_\indexbins}{\numsamples} | \acc(\indexbins) - \conf(\indexbins) | ,
\end{align}
where $\numsamples_\indexbins$ denotes the number of samples within bin $\bin_\indexbins$ and
\begin{align}
	\acc(\indexbins) &= \frac{1}{\numsamples_\indexbins} \sum_{\indexsamples_\indexbins=1}^{\numsamples_\indexbins} \ind(\predoutput_{\indexsamples_\indexbins} = \groundtruthoutput_{\indexsamples_\indexbins}) , \\
	\conf(\indexbins) &= \frac{1}{\numsamples_\indexbins} \sum_{\indexsamples_\indexbins=1}^{\numsamples_\indexbins} \predconfidence_{\indexsamples_\indexbins} ,
\end{align}
with indicator function $\ind(\cdot)$ \cite{Naeini2015,Guo2018}.
In this case, $\acc(\indexbins)$ and $\conf(\indexbins)$ denote the average accuracy (or frequency within binary classification) and the average confidence within each bin, respectively.

\subsection{Object Detection}
\label{section:confidence:definition:detection}

Similar to a classifier, an object detection model, which is based on a neural network, also estimates a label $\predoutputvariate \in \outputset$ and an according objectness confidence score $\predconfidencevariate \in \probset$ for each prediction within a single image $\allinputvariates \in \inputset$. 
Let further denote $\allgroundtruthbboxvariates \in \bboxset = [0, 1]^\numbboxdims$ the ground-truth information for the object's position (relative to image size) where $\numbboxdims$ denotes the dimension of the box encoding (commonly center $x$ and $y$ positions $\centerx$, $\centery$ as well as width $\width$ and height $\height$).
Thus, the joint ground-truth data distribution extends to $\pdf_{\allinputvariates, \groundtruthvariate, \allgroundtruthbboxvariates}(\allsingleinput, \singlegroundtruth, \allgroundtruthbboxes) = \pdf_{\groundtruthvariate, \allgroundtruthbboxvariates}(\singlegroundtruth, \allgroundtruthbboxes | \allsingleinput)\pdf_{\allinputvariates}(\allsingleinput)$.
An object detection model thus also needs to infer the object's position to approximate $\pdf_{\groundtruthvariate, \allgroundtruthbboxvariates}(\singlegroundtruth, \allgroundtruthbboxes | \allsingleinput)$. 
In the following, we denote the position predictions as $\allpredbboxvariates \in \bboxset$, so that the overall output distribution of an object detection model is denoted by $\pdf_{\predconfidencevariate, \predoutputvariate, \allpredbboxvariates}(\predconfidence, \predoutput, \allpredbboxes | \allsingleinput)$.
In contrast to the simple case of classification calibration, there are some limitations we need to address within object detection calibration. 
On the one hand, as we know from \secref{section:basics:object_detection}, most object detectors use anchor or proposal boxes to implement the object detection.
These boxes serve as an initial estimate for a possible object location and get scaled and shifted in a subsequent step to match a real object.
However, if only a few amount of objects is present within an image, most anchor/proposal boxes are classified as background.
These predictions are discarded in the final post-processing of a common object detection pipeline.
This approach significantly reduces the model output to the user-relevant predictions.
During the model evaluation, we can denote the amount of missed objects, the so-called \textit{false negatives}.
Since we have no information about what the detection model has classified as background, we can not denote the \textit{true negatives} within a single frame.
However, this information is mandatory for accuracy computation.
For this reason, we further use the precision as our calibration target as it measures the fraction of correctly identified objects.
The precision denotes the fraction of correctly detected and classified objects given all predicted objects by the detection model.
In order to determine the precision, an assignment of the predictions to ground truth objects is necessary. 
This is usually done using the \ac{IoU} score between the predicted and the ground-truth objects. 
Furthermore, this means that the precision and thus the calibration of the detection model depends on the selected \ac{IoU} score. 

On the other hand, as we observed in our examinations in \cite{Kueppers2020} and \cite{Kueppers2022a}, the data distribution of the training set during model training has an influence on the prediction performance of an object detector. 
For example, if most of the training samples have been located within the image center, the predictions near the boundaries might have a reduced performance. 
Therefore, we introduce a position-dependency to the definition of confidence calibration for object detection which is thus given by
\begin{align}
	\label{eq:confidence:definition:detection}
	&\prob(\matchedvariate = 1 | \predconfidencevariate = \predconfidence, \predoutputvariate = \predoutput, \allpredbboxvariates = \allpredbboxes) = \predconfidence \\ \nonumber
	&\forall \predconfidence \in \probset, \predoutput \in \gtset, \allpredbboxes \in \bboxset .
\end{align}
In this case, $\matchedvariate \in \{0, 1\}$ denotes if a predicted object has matched a ground-truth object with a certain \ac{IoU} and thus $\prob(\matchedvariate = 1)$ is a shorthand notation for $\prob(\predoutputvariate = \groundtruthvariate, \allpredbboxvariates = \allgroundtruthbboxvariates)$.

Similar to the standard \ac{ECE} definition in (\ref{eq:confidence:ece:continuous}), we derive the \ac{D-ECE} which is the extension of the \ac{ECE} to object detection. 
Let $\allcollect = (\predconfidence, \predoutput, \allpredbboxes)$ denote a single prediction so that $\allcollect \in \collectset$, where $\collectset$ is the aggregated set of the confidence, label, and bounding box spaces. 
If an object detection model is calibrated, the expected difference between predicted confidence and observed precision conditioned on the model output gets minimal.
The predicted joint output distribution is denoted by $\pdf_{\predconfidencevariate, \predoutputvariate, \allpredbboxvariates}(\predconfidence, \predoutput, \allpredbboxes) = \pdf_{\allcollectvariates}(\allcollect)$, where $\cdf_{\predconfidencevariate,\predoutputvariate,\allpredbboxvariates}(\allcollect)$ is the respective \ac{CDF} function.
Thus, the expectation is denoted by
\begin{align}
	\label{eq:confidence:dece:continuous}
	&\expectation_{\predconfidencevariate,\predoutputvariate,\allpredbboxvariates} \Big[ \big| \prob(\matchedvariate = 1 | \predconfidencevariate = \predconfidence, \predoutputvariate = \predoutput, \allpredbboxvariates = \allpredbboxes) - \predconfidence \big| \Big] \\
    &=\expectation_{\allcollectvariates} \Big[ \big| \prob(\matchedvariate = 1 | \allcollectvariates = \allcollect) - \predconfidence \big| \Big] \\
	&= \int_{\collectset} \pdf_{\allcollectvariates}(\allcollect) \cdot \big| \prob(\matchedvariate = 1 | \allcollectvariates = \allcollect) - \predconfidence \big| \diff \allcollect \\
    \label{eq:confidence:dece:riemann_approx}
    &= \int_{\collectset} \big| \prob(\matchedvariate = 1 | \allcollectvariates = \allcollect) - \predconfidence \big| \diff \cdf_{\allcollectvariates}(\allcollect) .
\end{align}
Similar to (\ref{eq:confidence:ece:riemann_approx}), we can approximate the integral by a multidimensional binning scheme with $\numbins$ distinct bins $\mvbin_\indexbins$ over the joint output space $\collectset$, so that the \ac{D-ECE} is an approximation of the integral in (\ref{eq:confidence:dece:riemann_approx}) by
\begin{align}
	\label{eq:confidence:dece:approximate}
	&\dece := \sum_{\indexbins=1}^{\numbins} \prob(\allcollectvariates \in \mvbin_\indexbins) \cdot \big| \prob(\matchedvariate = 1 | \predconfidencevariate = \predconfidence_\indexbins, \predoutputvariate = \predoutput_\indexbins, \allpredbboxvariates = \allpredbboxes_\indexbins) - \predconfidence_\indexbins \big| \\ \nonumber
    &\forall \predconfidence_\indexbins, \predoutput_\indexbins, \allpredbboxes_\indexbins \in \mvbin_\indexbins, \quad \indexbins \in \{1, \ldots \numbins\} ,
\end{align}
which is finally computed by the weighted sum
\begin{align}
    \dece = \sum_{\indexbins=1}^{\numbins} \frac{\numsamples_\indexbins}{\numsamples} | \precision(\indexbins) - \conf(\indexbins) | ,
\end{align}
with $\numsamples_\indexbins$ as the number of samples within bin $\mvbin_\indexbins$ and $\precision(\indexbins)$ as the precision within bin $\mvbin_\indexbins$.
Thus, the \ac{D-ECE} can be seen as the weighted sum of differences between precision and average confidence over all bins in the output space for a finite set of samples.

\subsection{Instance Segmentation}
\label{section:confidence:definition:instance}

Let $\objectset$ denote the set of all objects over all images in data set $\dataset$, where $\numobjects$ denotes the total amount of objects.
Instance segmentation is the combined task of predicting individual objects and their shape at the pixel-level. 
Thus, for each pixel $\indexpixel \in \pixelset_\indexobjects = \{1, ..., \numpixel_\indexobjects\}$ within the predicted bounding box $\allpredbboxvariates_\indexobjects \in \bboxset$ of an estimated object $\indexobjects \in \objectset$ with predicted label $\predoutputvariate_\indexobjects \in \outputset$, an instance segmentation model also predicts a label $\predoutputvariate_\indexpixel^\ast \in \gtset^\ast = \{0, 1\}$ in conjunction with a confidence score $\predconfidencevariate_\indexpixel^\ast \in \probset$, indicating the estimated membership of each pixel to the object mask. 
We further denote $\groundtruthvariate_\indexpixel^\ast$ as the ground-truth information for the object segmentation masks and $\allbboxvariates_\indexpixel^\ast \in \bboxset^\ast = [0, 1]^{\numbboxdims^\ast}$ as the (normalized) pixel-position within a bounding box $\allpredbboxvariates_\indexobjects$, where $\numbboxdims^\ast$ denotes the size of the used position-encoding for a single pixel.
In contrast to object detection, it is possible to compute the accuracy on pixel-level over all objects and all instance segmentation masks. 
Since the mask inference reduces to a binary classification task, our calibration target for instance segmentation is the pixel-wise relative frequency $\prob(\predoutputvariate_\indexpixel^\ast = 1)$ of each pixel belonging to the object's shape. 
Similar to the definition of object detection calibration, we further introduce a position-dependency. 
For example, pixels close to the shape boundary might have different calibration properties. Therefore, confidence calibration for instance segmentation is defined by
\begin{align}	
	\label{eq:confidence:definition:instance}
	&\prob(\predoutputvariate_\indexpixel^\ast = 1 | \predconfidencevariate_\indexpixel^\ast = \predconfidence^\ast, \allbboxvariates_\indexpixel^\ast = \allpredbboxes^\ast, \predoutputvariate = \predoutput) = \predconfidence^\ast, \\ \nonumber
	&\forall \predconfidence^\ast \in \probset, \allpredbboxes^\ast \in \bboxset^\ast, \predoutput \in \outputset, \indexpixel \in \pixelset_\indexobjects, \indexobjects \in \objectset .
\end{align}
Similar to (\ref{eq:confidence:dece:approximate}), we can use the \ac{D-ECE} to measure the miscalibration of an instance segmentation model. For this purpose, a binning scheme over the joint space for the pixel-wise confidence, the pixel position $\bboxset^\ast$, and all possible labels is used to approximate the \ac{D-ECE}.

\subsection{Semantic Segmentation}
\label{section:confidence:definition:semantic}

In contrast to instance segmentation, a semantic segmentation model does not predict individual objects but rather estimates a label $\predoutputvariate_\indexpixel^\ast \in \gtset = \{1, ..., \numclasses\}$ with corresponding confidence $\predconfidencevariate_\indexpixel^\ast \in \probset$ for each pixel $\indexpixel \in \pixelset$ in an input image $\allinputvariates \in \inputset$, where $\groundtruthvariate_\indexpixel^\ast \in \gtset$ denotes the ground-truth information for pixel $\indexpixel$. 
Therefore, semantic segmentation can be seen as a joint (multiclass) classification task for each pixel in an image. 
Thus, the definition of confidence calibration for semantic segmentation changes to 
\begin{align}
	\label{eq:confidence:definition:semantic}
	&\prob(\predoutputvariate_\indexpixel^\ast = \groundtruthvariate_\indexpixel^\ast | \predconfidencevariate_\indexpixel^\ast = \predconfidence^\ast, \allbboxvariates_\indexpixel^\ast = \allpredbboxes^\ast) = \predconfidence^\ast, \\ \nonumber
	&\forall \predconfidence^\ast \in \probset, \allpredbboxes^\ast \in \bboxset^\ast, \indexpixel \in \pixelset .
\end{align}
In contrast to the definition for instance segmentation calibration in (\ref{eq:confidence:definition:instance}), we have no dependency on individual objects.
Thus, we the definition of calibration for semantic segmentation is related to the calibration definition for classification in (\ref{eq:confidence:definition:classification}) but on pixel-level and with an additional dependency on the pixel position.

\section{Multivariate Confidence Calibration}
\label{section:confidence:methods}

The task of confidence calibration is to remap probability estimates so that they reflect the observed accuracy, frequency, or precision. 
These calibration methods are applied as post-hoc calibration that is applied after model inference.
In a first step, we review existing calibration techniques and group these methods into binning and scaling methods.
Scaling calibration methods perform post-hoc recalibration by rescaling the output of a forecaster to achieve well-calibrated confidence estimates \cite{Platt1999,Guo2018}.
In contrast, binning methods divide the probability space into several distinct bins (similar to the calculation of the \ac{ECE}) and use this binning scheme to derive a calibrated confidence \cite{Zadrozny2001,Naeini2015}.

\begin{figure}[t!]
    \centering
    \def\stackalignment{r}
    \stackunder{%
        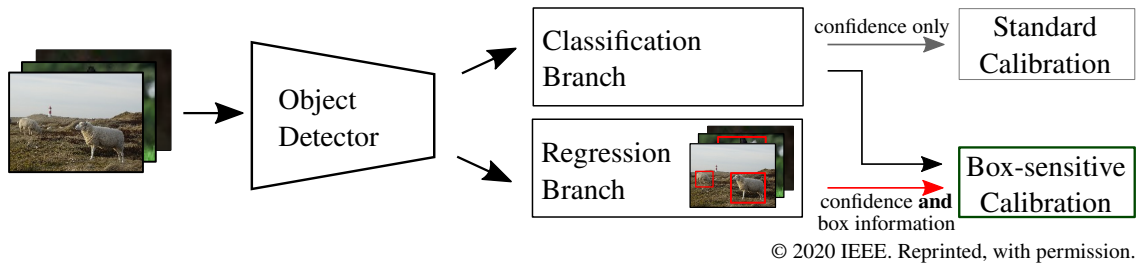%
    }{\scriptsize\ieeecopyright{2020}}
    \caption[Concept of box-sensitive confidence calibration w.r.t. the regression branch of an object detector.]{
        Concept of box-sensitive confidence calibration w.r.t. the regression branch of an object detector \cite[p. 2, Fig. 2]{Kueppers2020}.
        A detection model predicts several objects within an image with a certain position and a certain class.
        Standard calibration methods (top row) only use the classification branch of a detection model for confidence calibration.
        In contrast, our position-dependent calibration methods also utilize the additional regression branch to keep track of possible correlations between box position and miscalibration.
    }
    \label{fig:confidence:blockimage}
\end{figure}

\begin{figure}[t!]
    \centering
    \def\stackalignment{r}
    \stackunder{%
\begin{tikzpicture}

\tikzstyle{every node}=[font=\scriptsize]
\pgfplotsset{every y tick label/.append style={font=\tiny, xshift=0.5ex}}
\pgfplotsset{ every non boxed x axis/.append style={x axis line style=-}, every non boxed y axis/.append style={y axis line style=-}}

\begin{axis}[
width=0.35\textwidth,
colorbar horizontal,
colormap={mymap}{[1pt]
	rgb(0pt)=(0,0,0.5);
	rgb(22pt)=(0,0,1);
	rgb(25pt)=(0,0,1);
	rgb(68pt)=(0,0.86,1);
	rgb(70pt)=(0,0.9,0.967741935483871);
	rgb(75pt)=(0.0806451612903226,1,0.887096774193548);
	rgb(128pt)=(0.935483870967742,1,0.0322580645161291);
	rgb(130pt)=(0.967741935483871,0.962962962962963,0);
	rgb(132pt)=(1,0.925925925925926,0);
	rgb(178pt)=(1,0.0740740740740741,0);
	rgb(182pt)=(0.909090909090909,0,0);
	rgb(200pt)=(0.5,0,0)
},
colorbar style={
	title style={at={(0.5,0)},anchor=north,yshift=3em},
	title=(a) Uncalibrated,
	xticklabels={0, 0\%, 5\%, 10\%, 15\%, 20\%, 25\%},
	x tick scale label style={yshift=0.5cm},
	at={(1,1.03)},anchor=south east,
	width=1.0*\pgfkeysvalueof{/pgfplots/parent axis width},
	xticklabel pos=upper,
},
point meta max=0.1,
point meta min=0,
tick align=outside,
tick pos=left,
x grid style={white!69.0196078431373!black},
xlabel={$c_x$ position},
xmin=0, xmax=14,
xtick style={color=black},
xtick={0,3.5,7,10.5,14},
xticklabels={0.0,0.25,0.5,0.75,1.0},
y grid style={white!69.0196078431373!black},
ylabel={$c_y$ position},
ymin=0, ymax=14,
ytick style={color=black},
ytick={0,3.5,7,10.5,14},
yticklabels={0.0,0.25,0.5,0.75,1.0}
]
\addplot graphics [includegraphics cmd=\pgfimage,xmin=-0.5, xmax=14.5, ymin=-0.5, ymax=14.5] {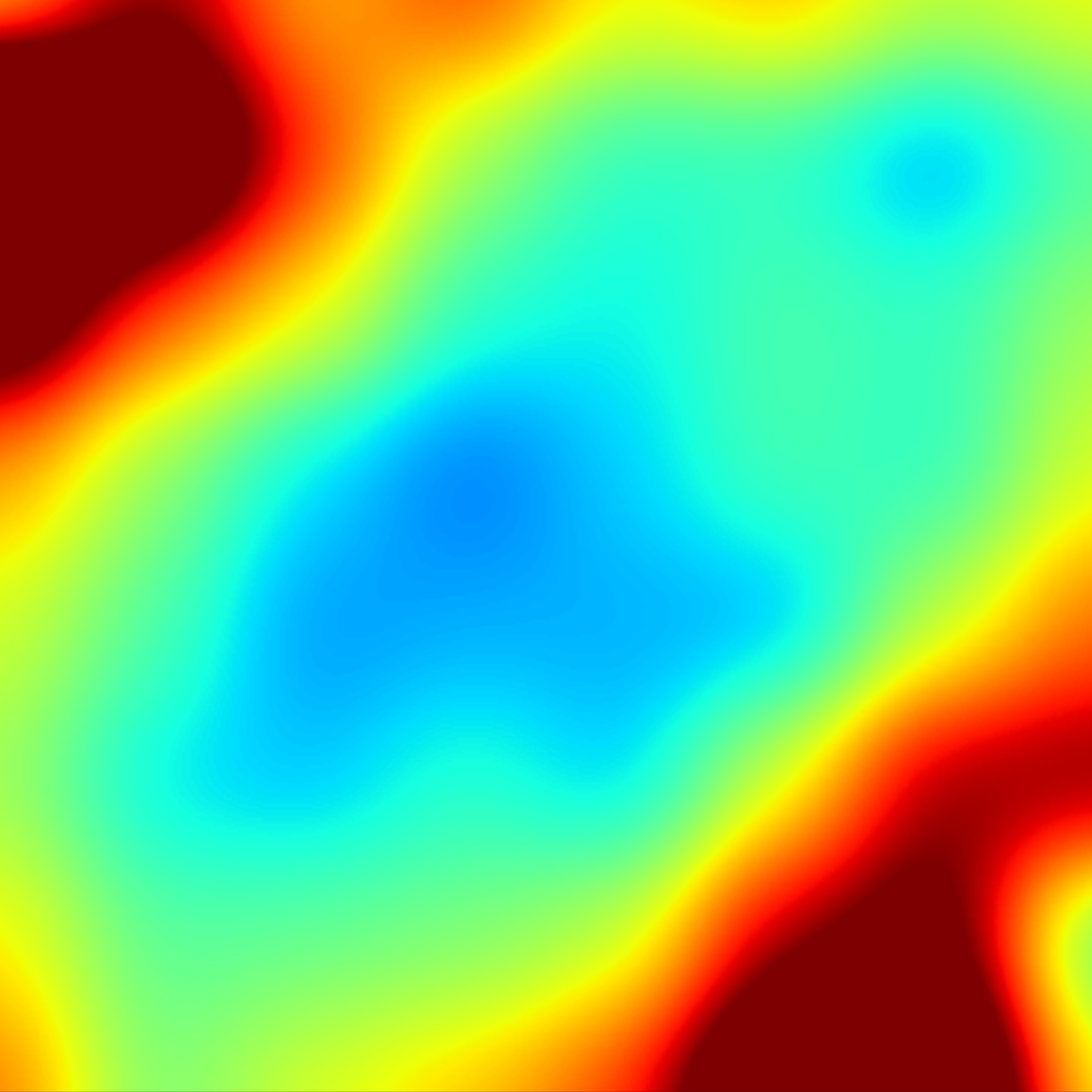};
\end{axis}

\end{tikzpicture}%
\begin{tikzpicture}

\tikzstyle{every node}=[font=\scriptsize]
\pgfplotsset{every y tick label/.append style={font=\tiny, xshift=0.5ex}}
\pgfplotsset{ every non boxed x axis/.append style={x axis line style=-}, every non boxed y axis/.append style={y axis line style=-}}

\begin{axis}[
width=0.35\textwidth,
colorbar horizontal,
colormap={mymap}{[1pt]
	rgb(0pt)=(0,0,0.5);
	rgb(22pt)=(0,0,1);
	rgb(25pt)=(0,0,1);
	rgb(68pt)=(0,0.86,1);
	rgb(70pt)=(0,0.9,0.967741935483871);
	rgb(75pt)=(0.0806451612903226,1,0.887096774193548);
	rgb(128pt)=(0.935483870967742,1,0.0322580645161291);
	rgb(130pt)=(0.967741935483871,0.962962962962963,0);
	rgb(132pt)=(1,0.925925925925926,0);
	rgb(178pt)=(1,0.0740740740740741,0);
	rgb(182pt)=(0.909090909090909,0,0);
	rgb(200pt)=(0.5,0,0)
},
colorbar style={
	title style={at={(0.5,0)},anchor=north,yshift=3em},
	title=(b) After standard calibration,
	xticklabels={0, 0\%, 5\%, 10\%, 15\%, 20\%, 25\%},
	x tick scale label style={yshift=0.5cm},
	at={(1,1.03)},anchor=south east,
	width=1.0*\pgfkeysvalueof{/pgfplots/parent axis width},
	xticklabel pos=upper,
},
point meta max=0.1,
point meta min=0,
tick align=outside,
tick pos=left,
x grid style={white!69.0196078431373!black},
xlabel={$c_x$ position},
xmin=0, xmax=14,
xtick style={color=black},
xtick={0,3.5,7,10.5,14},
xticklabels={0.0,0.25,0.5,0.75,1.0},
y grid style={white!69.0196078431373!black},
ymin=0, ymax=14,
ytick style={color=black},
ytick={},
yticklabels={},
ytick style={draw=none},
]
\addplot graphics [includegraphics cmd=\pgfimage,xmin=-0.5, xmax=14.5, ymin=-0.5, ymax=14.5] {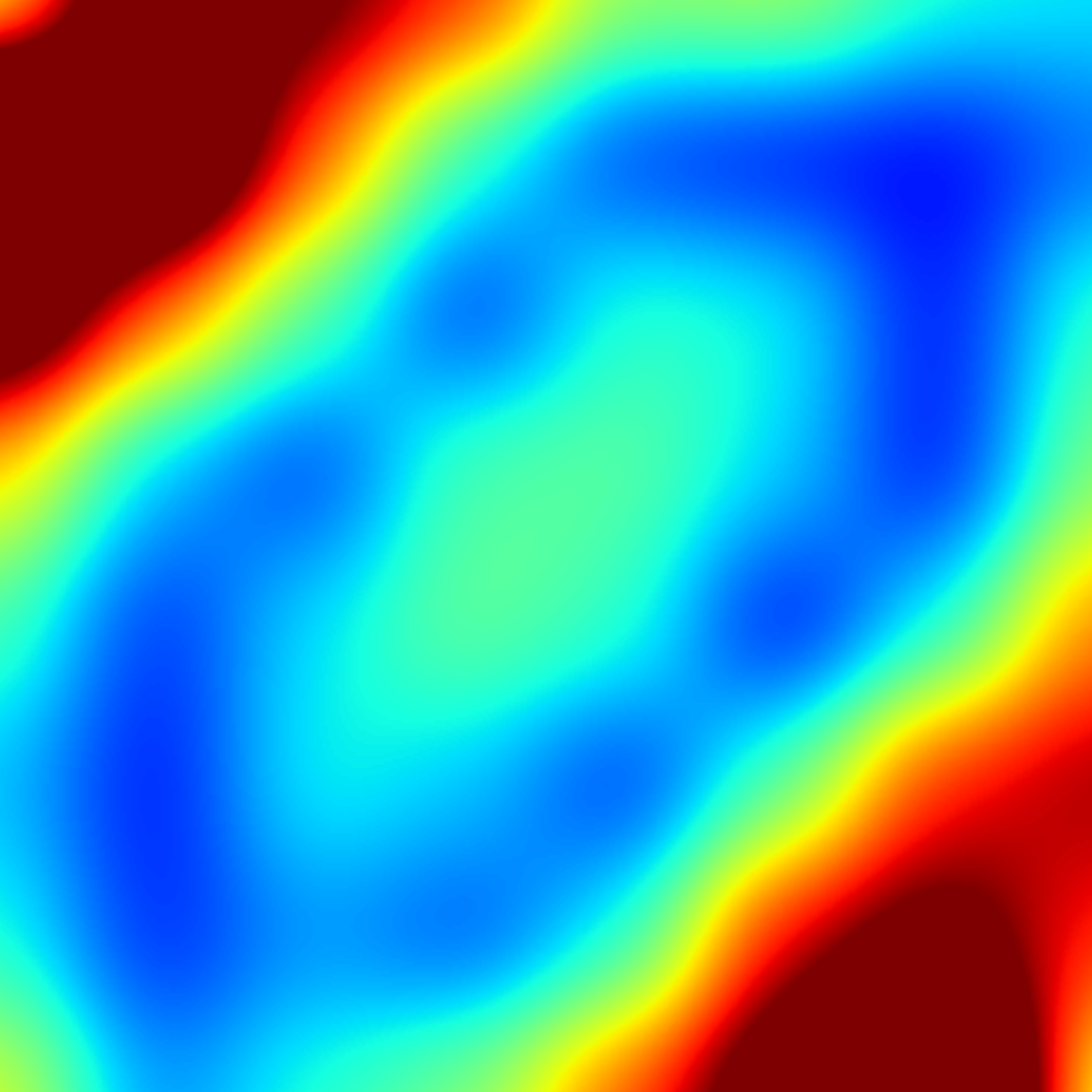};
\end{axis}

\end{tikzpicture}%
\begin{tikzpicture}

\tikzstyle{every node}=[font=\scriptsize]
\pgfplotsset{every y tick label/.append style={font=\tiny, xshift=0.5ex}}
\pgfplotsset{ every non boxed x axis/.append style={x axis line style=-}, every non boxed y axis/.append style={y axis line style=-}}

\begin{axis}[
width=0.35\textwidth,
colorbar horizontal,
colormap={mymap}{[1pt]
	rgb(0pt)=(0,0,0.5);
	rgb(22pt)=(0,0,1);
	rgb(25pt)=(0,0,1);
	rgb(68pt)=(0,0.86,1);
	rgb(70pt)=(0,0.9,0.967741935483871);
	rgb(75pt)=(0.0806451612903226,1,0.887096774193548);
	rgb(128pt)=(0.935483870967742,1,0.0322580645161291);
	rgb(130pt)=(0.967741935483871,0.962962962962963,0);
	rgb(132pt)=(1,0.925925925925926,0);
	rgb(178pt)=(1,0.0740740740740741,0);
	rgb(182pt)=(0.909090909090909,0,0);
	rgb(200pt)=(0.5,0,0)
},
colorbar style={
	title style={at={(0.5,0)},anchor=north,yshift=3em},
	title=(c) After position-dependent calibration,
	xticklabels={0, 0\%, 5\%, 10\%, 15\%, 20\%, 25\%},
	x tick scale label style={yshift=0.5cm},
	at={(1,1.03)},anchor=south east,
	width=1.0*\pgfkeysvalueof{/pgfplots/parent axis width},
	xticklabel pos=upper,
},
point meta max=0.1,
point meta min=0,
tick align=outside,
tick pos=left,
x grid style={white!69.0196078431373!black},
xlabel={$c_x$ position},
xmin=0, xmax=14,
xtick style={color=black},
xtick={0,3.5,7,10.5,14},
xticklabels={0.0,0.25,0.5,0.75,1.0},
y grid style={white!69.0196078431373!black},
ymin=0, ymax=14,
ytick style={color=black},
ytick={},
yticklabels={},
ytick style={draw=none},
]
\addplot graphics [includegraphics cmd=\pgfimage,xmin=-0.5, xmax=14.5, ymin=-0.5, ymax=14.5] {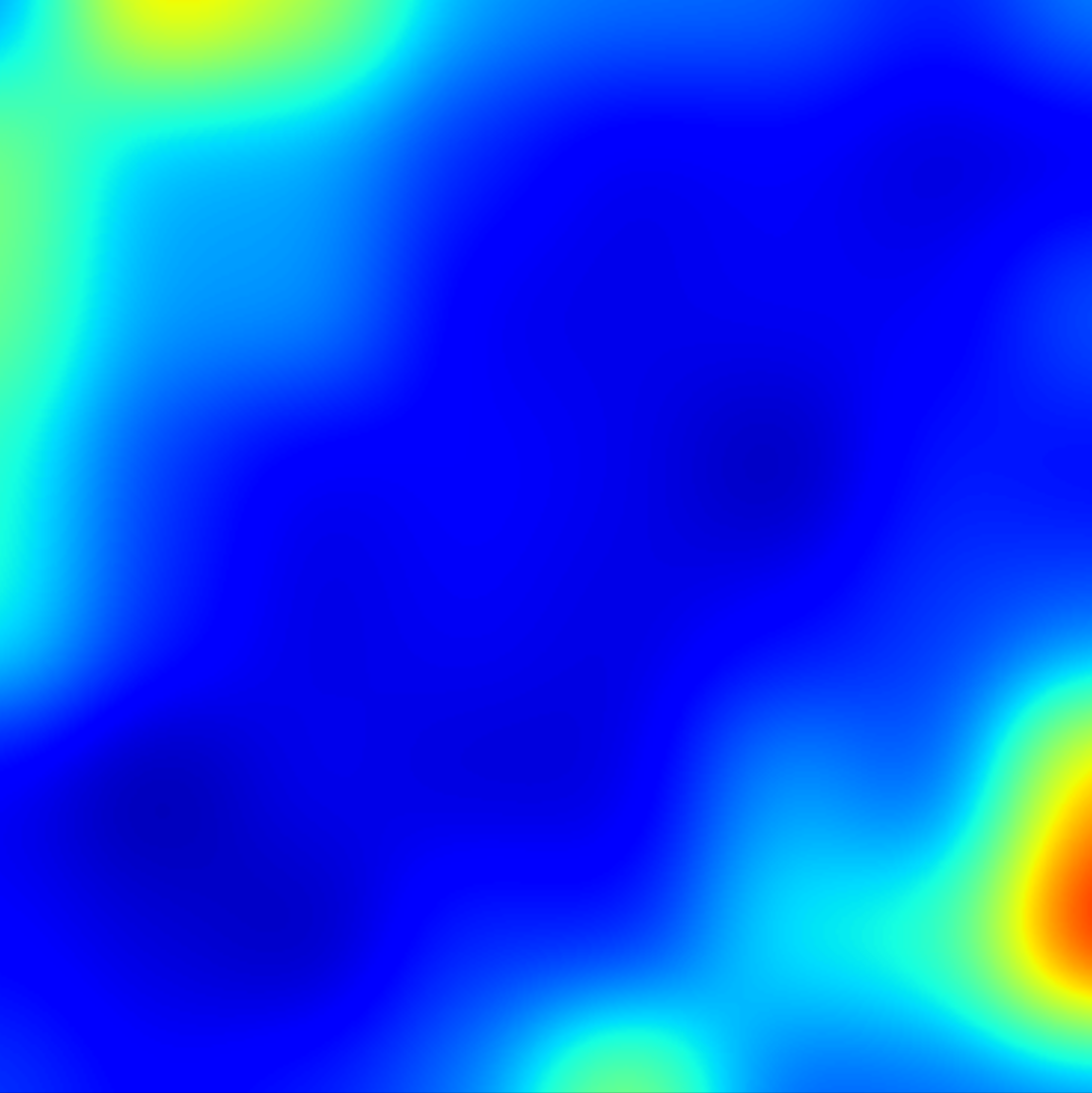};
\end{axis}

\end{tikzpicture}%
    }{\scriptsize\springercopyright{2022}}
    \caption[Qualitative example of multivariate confidence calibration on an artificially created data set.]{
        Qualitative example of multivariate confidence calibration on an artificially created data set \cite[p. 5, Fig. 3]{Kueppers2020}, \cite[p. 232, Fig. 1]{Kueppers2022a}.
        The samples are drawn with a certain confidence score $\predconfidencevariate$ and a matched flag $\matchedvariate \in \{0, 1\}$.
        Similar to the computation of the \ac{D-ECE}, the position space is divided into several bins.
        We measure the gap between average confidence and observed precision within each bin and visualize the position-dependent miscalibration in terms of a heatmap.
        (a) The deviation between average confidence and average precision follows a correlated bivariate normal distribution and increases towards the distribution boundary.
        (b) Common calibration methods are only able to rescale to average confidence in this scenario. This yields new confidences which are calibrated on average but still show a position-dependent calibration error.
        (c) In contrast, our multivariate calibration methods are able to successfully recalibrate the confidence information by means of the position information.
    }
    \label{fig:confidence:qualitative}
\end{figure}

Moreover, we refer to the definition for position-dependent calibration from the previous section and extend these methods to perform confidence calibration by means of the additional position information which is provided by an object detection or segmentation model.
This concept is schematically shown in \figref{fig:confidence:blockimage}.
The advantage of these extended calibration methods is that they are sensitive to possible correlations between miscalibration and object location/shape.
If there is a position dependence in the confidence miscalibration, the common calibration methods will not be able to perform a proper recalibration.
The effect of position-dependent calibration is demonstrated in an artificial example in \figref{fig:confidence:qualitative}.

The extended calibration methods yield a calibrated confidence $\calibratedvariate \in [0, 1]$ which reflects the probability of a predicted object to match an existing one given the uncalibrated confidence $\predconfidencevariate$, the object category $\predoutputvariate$, and the position information $\allpredbboxvariates$ in the joint space $\collectset$, so that $\calmodel: \collectset \rightarrow \probset$.
Therefore, the predicted probability distribution for the categorical random variable $\matchedvariate$ (object detection) after calibration is also a Bernoulli distribution whose probability parameter $\calibratedvariate$ can be expressed as a function of the uncalibrated confidence and the predicted position information, so that the \ac{PMF} of $\matchedvariate$ is given by
\begin{align}
    \label{eq:confidence:calibrated}
    \prob\big(\matched | \calmodel(\predconfidence, \predoutput, \allpredbboxes)\big) = \bernoullidistribution\big(\matched; \calmodel(\predconfidence, \predoutput, \allpredbboxes)\big) .
\end{align}

In the context of the multivariate scaling methods, we further distinguish between conditionally independent and dependent variants.
Both variants are able to model the influence of additional position information to the calibration output.
However, the conditionally independent methods assume independent random variables as input which simplifies the calibration computation but also leads to a reduced representational power.
In contrast, the conditional dependent methods model the input distribution as a joint multivariate probability distribution so that it is possible to capture possible correlations between the input quantities.

The multivariate extensions of the calibration methods was subject of our work in \cite{Kueppers2020}.
In the following, we derive the position-dependent calibration techniques for the binning and scaling methods, respectively.

\subsection{Histogram Binning}
\label{section:confidence:methods:binning}

Similar to the approximate computation of the \ac{ECE} (cf. \secref{section:confidence:definition}), the Histogram Binning method uses a binning scheme with $\numbins$ equally sized bins over the confidence space $\probset$ to group all samples in $\dataset$ by their confidence \cite{Zadrozny2001}. Afterwards, it is possible to measure the accuracy, frequency, or precision w.r.t. the confidence. This allows for a remapping of uncalibrated confidences to calibrated ones \cite{Zadrozny2001}.
More formally, let $\numsamples$ denote the amount of predictions obtained by a neural network with a certain label and confidence.
Additionally, let $\numbins$ denote the number of bins with interval boundaries $0 = \boundary_1 < \boundary_2 < ... < \boundary_{\numbins+1} = 1$ as well as the according calibration parameters $\allhbparameter = \big\{\hbparameter_{\indexbins} | \indexbins \in \{1, ..., \numbins\} \big\}$, which reflect the observed accuracy, frequency, or precision within each bin. The objective of Histogram Binning is the minimization of
\begin{align}
	\allpredhbparameter = \argmin_{\allhbparameter} \sum_{\indexsamples=1}^{\numsamples} \sum_{\indexbins=1}^{\numbins} \ind(\predconfidence_\indexsamples \in \bin_\indexbins) (\hbparameter_{\indexbins} - \groundtruthoutput_\indexsamples)^2 ,
\end{align}
to infer the optimal recalibration parameters $\allpredhbparameter$, where $\ind(\predconfidence_\indexsamples \in \bin_\indexbins)$ is the indicator function that a predicted sample falls into bin $\bin_\indexbins$ \cite{Zadrozny2001}.
By the strong law of large numbers, each parameter $\hbparameter_{\indexbins}$ converges to the fraction of positive samples within each bin \cite{Zadrozny2001,Guo2018}.
In contrast to isotonic regression \cite{Zadrozny2002}, Histogram Binning does not guarantee a monotonically increasing mapping from uncalibrated to calibrated confidence scores as there is no restriction for neighboring bins to yield smaller or larger recalibration parameters $\hbparameter_{\indexbins}$.
Therefore, Histogram Binning might also affect the order of the predicted samples and thus might have an influence on the computation of the average precision score.

Similar to the multivariate extension of the \ac{ECE} in (\ref{eq:confidence:dece:approximate}), we can further use a multivariate binning scheme over the joint space $\collectset$ which consists of the probability space $\probset$, all possible labels $\gtset$, and the spatial space $\bboxset$. 
In this way, all samples in $\dataset$ are grouped in $\numbins$ distinct bins $\mvbin_\indexbins$ by their confidence, label, and position information to construct a recalibration mapping, so that the objective function slightly changes to
\begin{align}
	\allpredhbparameter = \argmin_{\allhbparameter} \sum_{\indexsamples=1}^{\numsamples} \sum_{\indexbins=1}^{\numbins} \ind(\allcollect_\indexsamples \in \mvbin_\indexbins) (\hbparameter_{\indexbins} - \groundtruthoutput_\indexsamples)^2 ,
\end{align}
where $\ind(\allcollect_\indexsamples \in \mvbin_\indexbins)$ is the indicator function that sample $\indexsamples$ falls into bin $\mvbin_\indexbins$ with a certain label in $\gtset$ and a certain set of boundaries for the confidence in $\probset$ and the position information in $\bboxset$.

\subsection{Scaling Methods}
\label{section:confidence:methods:scaling}

In contrast to binning methods such as Histogram Binning, scaling methods perform a rescaling of the network output to yield calibrated confidences.
For this purpose, the methods Logistic Calibration \cite{Platt1999} and Beta Calibration \cite{Kull2017} either rescale the logit before applying a sigmoid (or softmax for multiclass problems) or directly rescale the confidence output after the sigmoid or softmax function, respectively.
We further denote the logit by $\logitvariate \in \realdigits$ which is related to the same distribution as the predicted confidence $\predconfidencevariate$, so that the predicted joint model distribution is given by $\pdf_{\logitvariate, \predoutputvariate, \allpredbboxvariates}(\logit, \predoutput, \allpredbboxes)$.

The advantage of scaling methods is that they require significantly less parameters compared to binning methods and thus are able to achieve a good calibration results given less data. 
We further inspect the rescaling of the confidence output in more detail.
For binary classification, the distribution of the confidence $\predconfidencevariate \in \probset$ can be represented using the probability density functions $\pdf_{\predconfidencevariate}(\predconfidence | \groundtruthvariate = 1)$ and $\pdf_{\predconfidencevariate}(\predconfidence | \groundtruthvariate = 0)$ for the positive ``$\positive$'' ($\groundtruthvariate = 1$) and the negative classes ``$\negative$'' ($\groundtruthvariate = 0$), respectively.
If we further treat these density functions as the likelihood for $\groundtruthvariate$ given $\predconfidencevariate$, we can use this representation to derive a recalibration scheme by the likelihood ratio 
between the likelihood for $\groundtruthvariate=1$ and $\groundtruthvariate=0$ \cite{Kull2017}. Using the logarithm of this ratio, we further refer to the log likelihood ratio as
\begin{align}
	\loglikelihoodratio(\predconfidence) = \log \frac{\pdf_{\predconfidencevariate}(\predconfidence | \groundtruthvariate = 1)}{\pdf_{\predconfidencevariate}(\predconfidence | \groundtruthvariate = 0)} .
\end{align}
The calibrated probability for $\prob(\groundtruthvariate = 1 | \predconfidencevariate = \predconfidence)$ can be derived by the ratio
\begin{align}
    & \frac{\prob(\groundtruthvariate = 1 | \predconfidencevariate = \predconfidence)}{\prob(\groundtruthvariate = 0 | \predconfidencevariate = \predconfidence)} = \frac{\pdf_{\predconfidencevariate}(\predconfidence | \groundtruthvariate = 1)}{\pdf_{\predconfidencevariate}(\predconfidence | \groundtruthvariate = 0)} \frac{\prob(\groundtruthvariate = 1)}{\prob(\groundtruthvariate = 0)} .
\end{align}
If we assume a uniform prior for the positive and negative classes so that $\prob(\groundtruthvariate = 1) = \prob(\groundtruthvariate = 0)$, the ratio $\frac{\prob(\groundtruthvariate = 1)}{\prob(\groundtruthvariate = 0)}$ evaluates to $1$ and can be neglected \cite{Kull2017}. 
In addition, if $\prob(\groundtruthvariate = 0 | \predconfidencevariate = \predconfidence) = 1 - \prob(\groundtruthvariate = 1 | \predconfidencevariate = \predconfidence)$ within binary classification, the likelihood ratio equals the posterior odds \cite{Kull2017}, \cite[p. 279]{Barber2012}, and we can derive a calibrated probability by
\begin{alignat}{2}
    & \frac{\prob(\groundtruthvariate = 1 | \predconfidencevariate = \predconfidence)}{\prob(\groundtruthvariate = 0 | \predconfidencevariate = \predconfidence)} &&= \frac{\pdf_{\predconfidencevariate}(\predconfidence | \groundtruthvariate = 1)}{\pdf_{\predconfidencevariate}(\predconfidence | \groundtruthvariate = 0)} \\
    \Leftrightarrow \quad & \prob(\groundtruthvariate = 1 | \predconfidencevariate = \predconfidence) &&= \big(1 - \prob(\groundtruthvariate = 1 | \predconfidencevariate = \predconfidence)\big) \exp\big(\loglikelihoodratio(\predconfidence)\big) \\
    \Leftrightarrow \quad & \prob(\groundtruthvariate = 1 | \predconfidencevariate = \predconfidence) &&= \exp\big(\loglikelihoodratio(\predconfidence)\big) - \prob(\groundtruthvariate = 1 | \predconfidencevariate = \predconfidence) \exp\big(\loglikelihoodratio(\predconfidence)\big) \\
    \Leftrightarrow \quad & \frac{\exp\big(\loglikelihoodratio(\predconfidence)\big)}{\prob(\groundtruthvariate = 1 | \predconfidencevariate = \predconfidence)} &&= 1 + \exp\big(\loglikelihoodratio(\predconfidence)\big) \\
    \Leftrightarrow \quad & \frac{\prob(\groundtruthvariate = 1 | \predconfidencevariate = \predconfidence)}{\exp\big(\loglikelihoodratio(\predconfidence)\big)} &&= \frac{1}{1 + \exp\big(\loglikelihoodratio(\predconfidence)\big)}  \\
    \label{eq:confidence:scaling:derivation}
    \Leftrightarrow \quad & \prob(\groundtruthvariate = 1 | \predconfidencevariate = \predconfidence) &&= \frac{\exp\big(\loglikelihoodratio(\predconfidence)\big)}{1 + \exp\big(\loglikelihoodratio(\predconfidence)\big)} = \sigmoid\big(\loglikelihoodratio(\predconfidence)\big) ,
\end{alignat}
which recovers the logistic (sigmoid) function $\sigmoid\big(\loglikelihoodratio(\predconfidence)\big)$ \cite{Kull2017}.
This derivation also holds if we consider the logits $\logitvariate$ instead of the confidence $\predconfidencevariate$, i.e., for the derivation of the Logistic Calibration function.
In this case, we utilize the log likelihood ratio for the logits by
\begin{align}
    \loglikelihoodratio(\logit) = \log \frac{\pdf_\logitvariate(\logit | \groundtruthvariate=1)}{\pdf_\logitvariate(\logit | \groundtruthvariate=0)} ,
\end{align}
which actually yields the probability for $\prob(\groundtruthvariate = 1 | \logitvariate = \logit)$.
However, common literature treats this formu\-lation equivalently to $\prob(\groundtruthvariate = 1 | \predconfidencevariate = \predconfidence)$ since we are interested in the calibration properties by means of an interpretable confidence score.
This equivalent interpretation for calibration holds as the logit $\logitvariate$ is commonly remapped to a confidence $\predconfidencevariate = \sigmoid(\logitvariate)$ using the bijective sigmoid function.
Since the recalibration mapping is also a logistic and thus monotonically increasing function, the standard scaling methods do not affect the order of the samples provided by a forecaster.
Therefore, the calibration does not affect the computation of the average precision score.

According to the definition of confidence calibration for object detection models in \secref{section:confidence:definition}, we further want to include additional information such as position and shape of the predicted objects into a calibration mapping for detection models. 
For this reason, we can also use the derivation of a calibrated probability in (\ref{eq:confidence:scaling:derivation}) for multivariate probability distributions given the joint distribution $\pdf_{\predconfidencevariate, \predoutputvariate, \allpredbboxvariates}(\predconfidence, \predoutput, \allpredbboxes | \matchedvariate)$.
We further use the shorthand notation $\pdf_{\allcollectvariates_{\predconfidencevariate}}(\allcollect_{\predconfidencevariate} | \matchedvariate)$ for $\allcollectvariates_{\predconfidencevariate} = (\predconfidencevariate, \predoutputvariate, \allpredbboxvariates) \in \collectset_{\predconfidencevariate}$.
The equivalent joint distribution for the logits is further denoted by $\pdf_{\allcollectvariates_\logitvariate}(\allcollect_\logitvariate | \matchedvariate)$ for $\allcollectvariates_\logitvariate = (\logitvariate, \predoutputvariate, \allpredbboxvariates) \in \collectset_{\logitvariate}$.
To derive a multivariate calibration mapping with $\numdims$ dimensions, we can assume conditional independence between all quantities in $\collectset_{\predconfidencevariate}$ or $\collectset_\logitvariate$ so that the log likelihood ratio can be rewritten to 
\begin{align}
	\label{eq:confidence:multivariate:independent}
	\loglikelihoodratio(\allcollect) = \sum_{\indexdims=1}^{\numdims} \log \frac{\pdf_\allcollectvariates(\collect_\indexdims | \matchedvariate = 1)}{\pdf_\allcollectvariates(\collect_\indexdims | \matchedvariate = 0)} ,
\end{align}
for any $\allcollectvariates$ either in $\collectset_{\predconfidencevariate}$ or $\collectset_\logitvariate$. 
In contrast, it is also possible to derive a calibration mapping assuming dependencies between all quantities in $\collectset_{\predconfidencevariate}$ or $\collectset_\logitvariate$ using the log likelihood ratio for multivariate distributions.
In the following, we will derive the multivariate extension for the calibration methods Logistic Calibration \cite{Platt1999} and Beta Calibration \cite{Kull2017}.

\subsubsection{Logistic Calibration}
A popular recalibration method for binary classification is Logistic Calibration (or Platt scaling) \cite{Platt1999} which assumes normally distributed logit scores with mean values $ \mean^{\positive},  \mean^{\negative} \in \realdigits$ for the positive and negative classes, respectively, and equal variance $\variance \in \realdigitspositive$. Therefore, the probability density functions are defined as $\pdf_\logitvariate(\logit | \groundtruthvariate = 1) = \normaldistribution(\logit; \mean^{\positive}, \variance)$ and $\pdf_\logitvariate(\logit | \groundtruthvariate = 0) = \normaldistribution(\logit; \mean^{\negative}, \variance)$ \cite{Platt1999,Kull2017}. The log likelihood ratio $\loglikelihoodratio(\logit)$ is thus given by
\begin{align}
	\loglikelihoodratio(\logit) &= \log \frac{\pdf_\logitvariate(\logit | \groundtruthvariate = 1)}{\pdf_\logitvariate(\logit | \groundtruthvariate = 0)} = \frac{1}{2\variance} \Big[ (\logit - \mean^{\negative})^2 - (\logit - \mean^{\positive})^2 \Big] \\
	&= \frac{1}{2\variance} \Big[ 2\logit (\mean^{\positive} - \mean^{\negative}) - (\mean_{\positive}^2 - \mean_{\negative}^2) \Big] \\
	&= \frac{\mean^{\positive} - \mean^{\negative}}{2\variance} \Big[ \logit - (\mean^{\positive} - \mean^{\negative}) \Big] \\
	&= \scaleweight (\logit - \meansub) ,
\end{align}
where $\scaleweight = \frac{1}{2\variance}(\mean^{\positive} - \mean^{\negative})$ and $\meansub = \mean^{\positive} - \mean^{\negative}$ \cite{Kull2017}. In practice, Logistic Calibration utilizes a disentangled representation by the scale weight $\scaleweight \in \realdigitspositive$ and a bias $\scalebias \in \realdigits$ which can be interpreted as $\scalebias = -\meansub\scaleweight$. These parameters are obtained by \ac{MLE} using the \ac{NLL} loss and are used to rescale the logits $\logitvariate$, so that a calibrated confidence estimate is derived by
\begin{align}
	\prob\big(\groundtruthvariate = 1 | \predconfidencevariate = \sigmoid(\logit)\big) = \sigmoid(\scaleweight \cdot \logit + \scalebias) ,
\end{align}
\cite{Platt1999}. Similarly, we can derive the multivariate extension of the log likelihood ratio assuming conditional independence between all quantities, so that the likelihood ratio is derived in the sense of (\ref{eq:confidence:multivariate:independent}) using normal distributions for each quantity $\indexdims$ which yields the log likelihood ratio
\begin{align}
	\label{eq:confidence:scaling:logistic:independent}
	\loglikelihoodratio(\allcollect_\logitvariate) = \allcollect^\T_\logitvariate \scalevec + \scalebias ,
\end{align}
where $\scalevec \in \realdigits^{\numdims}$. 
In contrast, if we assume conditional dependence between all quantities, the log likelihood ratio $\loglikelihoodratio(\allcollect)$ is represented as the fraction of two Gaussians with mean vectors $\meanvec^\positive, \meanvec^\negative \in \realdigits^{\numdims}$ for the positive and negative classes, respectively, and the positive semidefinit covariance matrices $\cov^\positive, \cov^\negative \in \realdigits^{\numdims \times \numdims}$. This yields the log likelihood ratio
\begin{align}
	\label{eq:confidence:scaling:logistic:dependent}
	\loglikelihoodratio(\allcollect_\logitvariate) = \frac{1}{2} \Big[ (\allcollect^\T_{\negative} \cov^{-1}_{\negative} \allcollect^{\negative}) - (\allcollect^\T_{\positive} \cov^{-1}_{\positive} \allcollect^{\positive})\Big] + \scalebias ,
\end{align}
where $\allcollect^\positive = \allcollect_\logitvariate - \meanvec^\positive$, $\allcollect^\negative = \allcollect_\logitvariate - \meanvec^\negative$, and $\scalebias = \log \frac{|\cov^{\negative}|}{|\cov^{\positive}|}$. During parameter optimization, we use $\cov^{-1} = (\decomposed^\T\decomposed)^{-1} = \decomposed^{-1}(\decomposed^{-1})^\T$ and directly infer $\decomposed^{-1}$ to guarantee symmetric and positive semidefinit covariance matrices.

\subsubsection{Beta Calibration}
Recently, the authors in \cite{Kull2017} introduced the Beta Calibration method which takes advantage of the fact that the confidence is defined in $[0, 1]$.
Therefore, it is possible to directly rescale the confidence using beta distributions for $\pdf_{\predconfidencevariate}(\predconfidence | \groundtruthvariate)$.
The authors in \cite{Kull2017} use the derivation for the calibrated confidence in (\ref{eq:confidence:scaling:derivation}) using the uncalibrated confidence $\predconfidence \in \probset$ as function input and derive the log likelihood ratio
\begin{align}
	\loglikelihoodratio(\predconfidence) &= \log \Bigg[ \frac{\betafunc(\betafunca^\negative, \betafuncb^\negative)}{\betafunc(\betafunca^\positive, \betafuncb^\positive)} \frac{\predconfidence^{\betafunca^\positive-1} (1-\predconfidence)^{\betafuncb^\positive-1}}{\predconfidence^{\betafunca^\negative-1} (1-\predconfidence)^{\betafuncb^\negative-1}} \Bigg] \\
    \label{eq:confidence:scaling:betacal:standard}
	&= \betaparama \cdot \log(\predconfidence) - \betaparamb \cdot \log(1-\predconfidence) + \betaparamc ,
\end{align}
between two beta distributions for $\groundtruthvariate = 1$ and $\groundtruthvariate = 0$, respectively, with the parameters $\betaparama = \betafunca^\positive - \betafunca^\negative$, $\betaparamb = \betafuncb^\negative - \betafuncb^\positive$, and $\betaparamc = \log \frac{\betafunc(\betafunca^\negative, \betafuncb^\negative)}{\betafunc(\betafunca^\positive, \betafuncb^\positive)}$.
Furthermore, $\betafunc(\betafunca, \betafuncb)$ is the beta function.
In practice, the parameters $\betaparama, \betaparamb \in \realdigitspositive$ and $\betaparamc \in \realdigits$ are estimated by \ac{MLE} using the \ac{NLL} loss \cite{Kull2017}.

Similar to the multivariate extension of the Logistic Calibration method for object detection calibration, we can derive a multivariate and conditional independent calibration mapping according to (\ref{eq:confidence:multivariate:independent}) for the Beta Calibration method as well.
Since Beta Calibration aims to directly rescale the confidences, we further use $\allcollectvariates \in \collectset_{\predconfidencevariate}$ as the surrogate for $\allcollectvariates_{\predconfidencevariate}$ for notational simplicity.
Thus, the log likelihood ratio for the multivariate (independent) Beta Calibration is given by
\begin{align}
	\label{eq:confidence:scaling:betacal:independent}
	\loglikelihoodratio(\allcollect) = \scalebias + \sum_{\indexdims=1}^{\numdims} \betaparama_\indexdims \log(\collect_\indexdims) - \betaparamb_\indexdims \log(1-\collect_\indexdims) ,
\end{align}
where $\betaparama_\indexdims = \betafunca_\indexdims^\positive - \betafunca_\indexdims^\negative$, $\betaparamb_\indexdims = \betafuncb_\indexdims^\negative - \betafuncb_\indexdims^\positive$, and $\scalebias = \sum_{\indexdims=1}^{\numdims} \log \frac{\betafunc(\betafunca_\indexdims^\negative, \betafuncb_\indexdims^\negative)}{\betafunc(\betafunca_\indexdims^\positive, \betafuncb_\indexdims^\positive)}$ with the multivariate beta function $\betafunc(\allbetafunca)$.
Similar to the univariate case, we optimize $\allbetaparama, \allbetaparamb \in \realdigitspositive^{\numdims}$ and $\betaparamc \in \realdigits$ in practice. 
The multivariate extension of the Beta Calibration method under the assumption of conditional dependence is not that straight forward since the natural multivariate extension of a beta distribution is a Dirichlet distribution $\dirichletdistribution(\allcollect; \allbetafunca)$ with shape parameters $\betafunca \in \realdigitspositive^{\numdims}$. 
The Dirichlet distribution is defined for $\sum_{\indexdims=1}^{\numdims} \collect_\indexdims = 1$.
However, this is not suitable in our case as we do not work with a multivariate probability vector as the input to the calibration function.
The input $\allcollect \in \collectset$ is not restricted to a sum of $1$, so that we propose to utilize a multivariate beta distribution that has been defined by the authors in \cite{Libby1982} and is given by
\begin{align}
	\pdf_\allcollectvariates(\allcollect | \matchedvariate) = \betafunc(\allbetafunca)^{-1} \cdot \frac{\prod_{\indexdims=1}^{\numdims} \Big[ \betaratio_\indexdims^{\betafunca_\indexdims} (\collect_\indexdims^\ast)^{\betafunca_\indexdims+1} \collect_\indexdims^{-2} \Big] }{ \Big[ 1 + \sum_{\indexdims=1}^{\numdims} \betaratio_\indexdims \collect_\indexdims^\ast \Big]^{\sum_{\indexdims=0}^{\numdims} \betafunca_\indexdims}} ,
\end{align}
with shape parameters $\betafunca_\indexdims, \betafuncb_\indexdims \in \realdigitspositive$ for all $\indexdims \in \{0, ..., \numdims\}$ and the abbreviations $\betaratio_\indexdims = \frac{\betafuncb_\indexdims}{\betafuncb_0}$ and $\allcollect^\ast = \frac{\allcollect}{1-\allcollect}$. This allows for a derivation of a log likelihood ratio defined by
\begin{align}
	\label{eq:confidence:scaling:betacal:dependent}
	\loglikelihoodratio(\allcollect) = &\sum_{\indexdims=1}^{\numdims} \Big[ 
		\betafunca_\indexdims^\positive \log(\betaratio_\indexdims^\positive) -
		\betafunca_\indexdims^\negative \log(\betaratio_\indexdims^\negative) +
		(\betafunca_\indexdims^\positive - \betafunca_\indexdims^\negative) \log(\collect_\indexdims^\ast)
	\Big] + \\ \nonumber
	&\sum_{\indexdims=0}^{\numdims} \Bigg[ 
		\betafunca_\indexdims^\negative \log \Bigg( \sum_{\indexdimsalt=1}^{\numdims} \betaratio_\indexdimsalt^\negative \collect_\indexdimsalt^\ast \Bigg) - 
		\betafunca_\indexdims^\positive \log \Bigg( \sum_{\indexdimsalt=1}^{\numdims} \betaratio_\indexdimsalt^\positive \collect_\indexdimsalt^\ast \Bigg)
	\Bigg] + \scalebias ,
\end{align}
where $\betafunca^\positive, \betafunca^\negative$ and $\betaratio^\positive, \betaratio^\negative$ denote the shape parameters for the positive and negative classes, respectively, and $\betaparamc = \log \frac{\betafunc(\allbetaparama^\negative)}{\betafunc(\allbetaparama^\positive)}$.



\section{Experiments for Semantic Confidence Calibration} 
\label{section:confidence:experiments}

In this section, we evaluate our calibration methods using different detection and segmentation archi\-tectures that are based on neural networks. 
We describe our experimental setup and show the calibration results for object detection, instance segmentation, and semantic segmentation. 
For each of these tasks, we use the MS COCO \cite{Lin2014} and the Cityscapes \cite{Cordts2016} validation data sets.
The respective data sets are divided into two equally sized parts for calibration training and evaluation, respectively.
The respective training sets are used for the training of the forecaster itself, whereas no ground-truth label information are available for the respective test sets, so that our experiments for calibration evaluation are limited to the validation sets.
For Cityscapes, we use the Munster \& Lindau images for calibration training and the Frankfurt images for evaluation, whereas the MS COCO validation set is split randomly.
The experiments for multivariate confidence calibration evaluation have been subject of our publications in \cite{Kueppers2020} and \cite[p. 235 ff.]{Kueppers2022a}.

Note that the MetaDetect framework by \cite{Schubert2021} (mentioned in \secref{section:confidence:related_work}) also applies an extended uncertainty evaluation of detection methods, which is related to our multivariate (conditional independent) calibration methods presented in \secref{section:confidence:methods:scaling}.
Similarly, the authors in \cite{Maag2020} propose an equivalent approach for the uncertainty quantification within semantic segmentation.
We leave a comparison of the works by \cite{Schubert2021} and \cite{Maag2020} with our methods subject of future works.

\subsection{Object Detection}
\label{section:confidence:experiments:detection}

We follow the experimental setup in \cite[p. 235 f.]{Kueppers2022a} and evaluate the Histogram Binning \cite{Zadrozny2001}, Logistic Calibration \cite{Platt1999}, and Beta Calibration \cite{Kull2017} methods either using the confidence information only or by using all available information such as confidence, position, and shape of each detected object. 
We evaluate these methods on the MS COCO \cite{Lin2014} and the Cityscapes \cite{Cordts2016} validation data sets that consist of $5.000$ and $500$ annotated images, respectively. 
For MS COCO, we use a pretrained \fasterrcnnx[101]{} \cite{Ren2015} and a pretrained \retinanet[101]{} \cite{Lin2017} model provided by \cite{Wu2019} for inference. 
For the experiments on the Cityscapes data set, we use the bounding box information provided by a pretrained \maskrcnn[50]{} \cite{He2017}.
Samples with an uncalibrated confidence below $0.3$ are neglected to keep the focus only on relevant detections.
The experiments are performed for the classes \textit{person}, \textit{rider}, \textit{car}, \textit{truck}, \textit{bus}, \textit{train}, \textit{motorcycle}, and \textit{bicycle}, as these object categories are present in both data sets, which allows for a better comparison of the results.
We further utilize the \ac{D-ECE}, Brier score, and \ac{NLL} as metrics for calibration evaluation as well as the \ac{AUPRC} to measure the model's detection performance. 
We divide our examinations into standard calibration evaluation where only the confidence information is used and into the multivariate calibration where the confidence in conjunction with all available bounding box information is used for calibration. 
For \ac{D-ECE} calculation in the confidence-only case, a binning scheme with $\numbins = 20$ equally sized bins is used over the confidence space. 
In contrast, the \ac{D-ECE} for the multivariate case uses $\numbins_\indexdims = 5$ bins for each dimension $\indexdims \in \{1, \ldots, \numdims\}$ which yields a total amount of $3,125$ bins. 
Uninformative bins with less than $8$ samples are neglected during \ac{D-ECE} computation.
For the computation of the \ac{D-ECE}, it is necessary to define the features that are used for the underlying binning scheme.
Thus, comparing \ac{D-ECE} scores obtained by different subsets of features is not applicable since the conditional probability distributions of $\predconfidencevariate$ differ from each other. 
For example, a \ac{D-ECE} score obtained by using the confidence, $\centerx$, and $\centery$ position should not be compared to a \ac{D-ECE} score that is based on the confidence, width, and height, as the conditional distributions for the confidence, and thus the approximate binning schemes, differ from each other.

\begin{figure}[t!]
    \centering
    \def\stackalignment{r}
    \begin{subfigure}{\textwidth}
        \stackunder{%
            \includegraphics[width=0.3475\textwidth,]{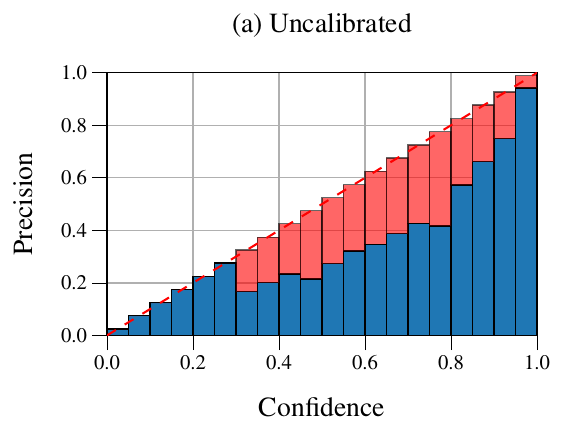}%
            \includegraphics[width=0.3175\textwidth,]{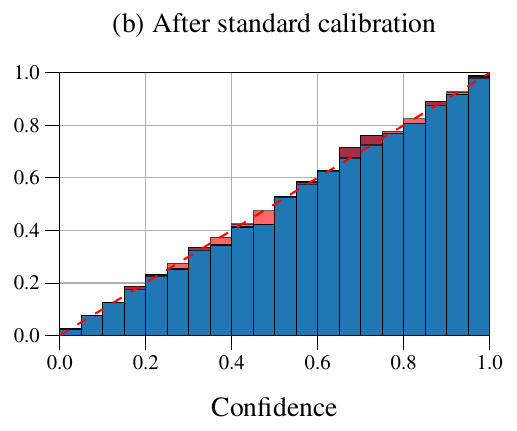}%
            \includegraphics[width=0.3175\textwidth,]{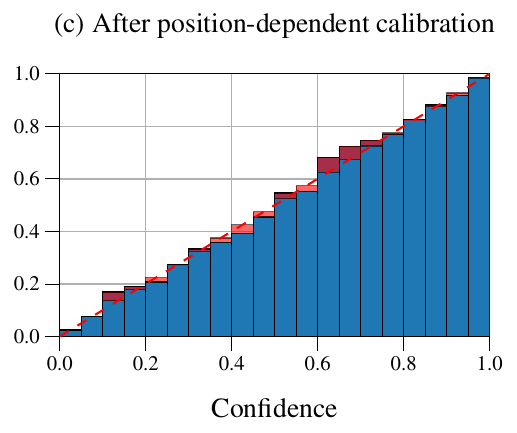}%
        }{\scriptsize\springercopyright{2022}}
        \caption{Reliability diagrams w.r.t. the confidence only \cite[p. 237, Fig. 3]{Kueppers2022a}.}
        \label{fig:confidence:detection:reliability:0d}
    \end{subfigure}
    \vspace{0.5cm}
    \begin{subfigure}{\textwidth}
        \stackunder{%
            \includegraphics[width=0.3385\textwidth,]{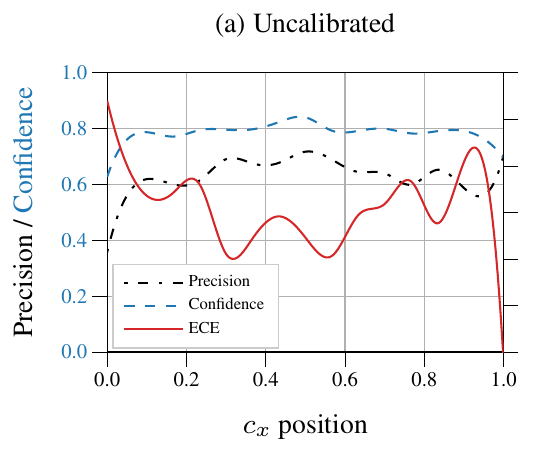}%
            \includegraphics[width=0.2925\textwidth,]{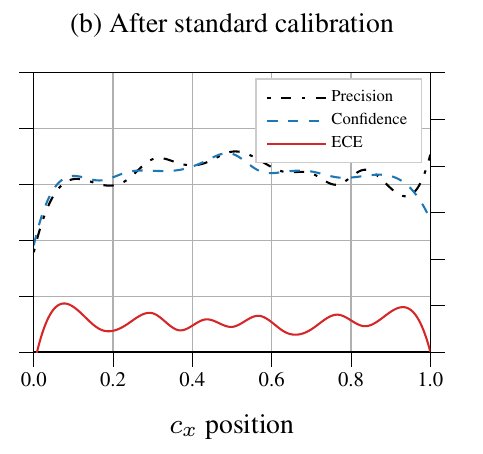}%
            \includegraphics[width=0.3495\textwidth,]{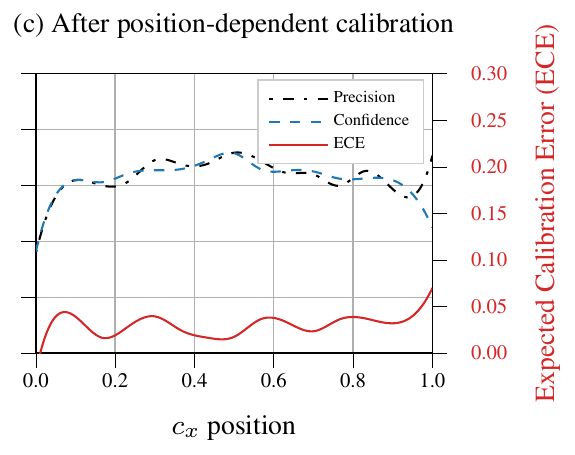}%
        }{\scriptsize\springercopyright{2022}}
        \caption{Reliability diagrams w.r.t. the $\centerx$ position of the predicted objects (1d) \cite[p. 240, Fig. 4]{Kueppers2022a}.}
        \label{fig:confidence:detection:reliability:1d}
    \end{subfigure}
    \vspace{0.5cm}
    \begin{subfigure}{\textwidth}
        \stackunder{%
            \includegraphics[width=0.362\textwidth,]{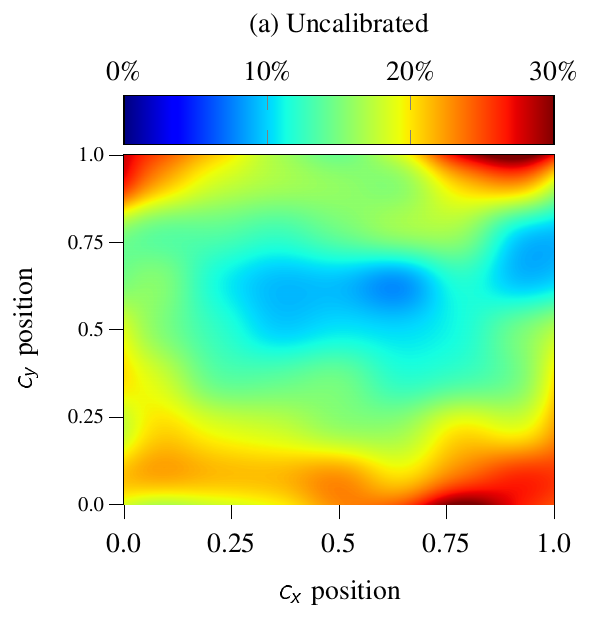}%
            \includegraphics[width=0.3025\textwidth,]{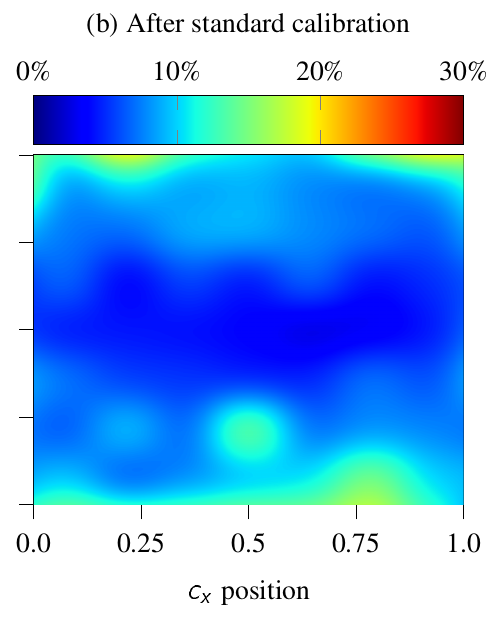}%
            \includegraphics[width=0.3025\textwidth,]{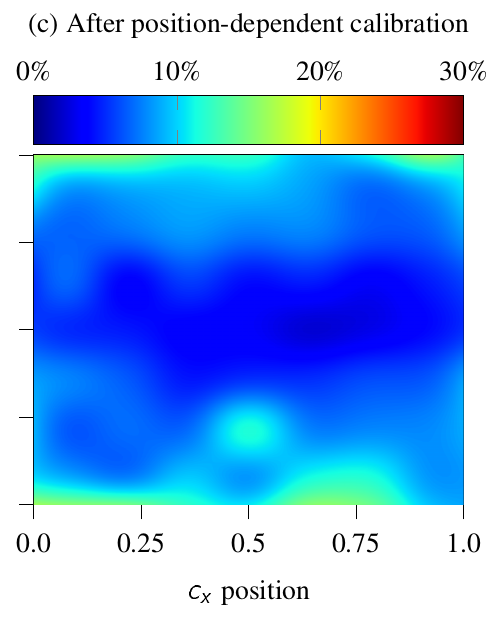}%
        }{\scriptsize\springercopyright{2022}}
        \caption{Reliability diagrams w.r.t. the $\centerx$ and $\centerx$ position of the predicted objects (2d) \cite[p. 240, Fig. 5]{Kueppers2022a}.}
        \label{fig:confidence:detection:reliability:2d}
    \end{subfigure}
    \caption[Reliability diagrams (object detection) for a \fasterrcnn{} on the MS COCO calibration validation set for class \textit{pedestrian}.]{
        Reliability diagrams (object detection) for a \fasterrcnn{} on the MS COCO calibration validation set for class \textit{pedestrian} \cite[p. 237 ff., Fig. 3-5]{Kueppers2022a} with uncalibrated confidences $\predconfidencevariate \geq 0.3$. 
        The uncalibrated baseline model is consistently overconfident for all confidence levels with increasing miscalibration towards the image boundaries.
        In this example, standard calibration shows good performance which is slightly improved using position-dependent calibration.
    }
    \label{fig:confidence:detection:reliability}
\end{figure}

According to our definition for object detection calibration in (\ref{eq:confidence:definition:detection}), the calibration target is the precision, which depends on the \ac{IoU} score used to distinguish between correctly predicted objects and false negatives. 
Thus, we run our evaluations using an IoU threshold of $0.50$ and $0.75$. 
The calibration results for the confidence-only case as well as for the multivariate calibration case are given in \tabref{tab:confidence:detection:evaluation:confidence} and \tabref{tab:confidence:detection:evaluation:position}, respectively. For further insights, we show the reliability diagrams for all calibration cases in \figref{fig:confidence:detection:reliability}.
\begin{table}[t!]
    \centering
    \caption[Calibration results for object detection where only the predicted confidence is used for calibration and evaluation.]{
        Calibration results for object detection where only the predicted confidence $\predconfidencevariate$ is used for calibration and D-ECE evaluation. 
        We observe that the scaling methods Logistic Calibration and Beta Calibration consistently achieve the best calibration results for \ac{D-ECE}, Brier score, and \ac{NLL}. 
        In contrast to Histogram Binning, the scaling methods apply a monotonically increasing mapping of uncalibrated confidence estimates to calibrated ones which does not affect the \ac{AUPRC} \cite[p. 238, Tab. 1]{Kueppers2022a}.
    }
    \begin{tabular}{c|c|l|cccc}
    \hline
    Network & IoU & Calibration method & D-ECE &        Brier &          NLL &          AUPRC \\ \hline \hline
    \multirow{8}{*}{\rotatebox[origin=c]{90}{\makecell{\fasterrcnn{}\\(trained on MS COCO)}}} & \multirow{4}{*}{0.50} & Uncalibrated  & 0.153 & 0.176 & 0.536 & 0.920 \\
    & & Histogram Binning & 0.026 & 0.146 & 0.470 & 0.878 \\
    & & Logistic Calibration & 0.021 & \textbf{0.142} & \textbf{0.433} & 0.920 \\
    & & Beta Calibration & \textbf{0.020} & \textbf{0.142} & \textbf{0.433} & 0.920 \\ \cline{2-7}
    &  \multirow{4}{*}{0.75} & Uncalibrated & 0.294 & 0.257 & 0.829 & 0.866 \\
    & & Histogram Binning & \textbf{0.026} & 0.155 & 0.510 & 0.770 \\
    & & Logistic Calibration & 0.027 & \textbf{0.144} & \textbf{0.448} & 0.866 \\
    & & Beta Calibration & 0.030 & \textbf{0.144} & 0.449 & 0.866 \\ \hline \hline
    \multirow{8}{*}{\rotatebox[origin=c]{90}{\makecell{\retinanet{}\\(trained on MS COCO)}}} & \multirow{4}{*}{0.50} & Uncalibrated & 0.083 & 0.157 & 0.478 & 0.907 \\
    & & Histogram Binning & 0.025 & 0.152 & 0.482 & 0.889 \\
    & & Logistic Calibration & 0.024 & \textbf{0.150} & \textbf{0.451} & 0.907 \\
    & & Beta Calibration & \textbf{0.022} & \textbf{0.150} & \textbf{0.451} & 0.907 \\ \cline{2-7}
    &  \multirow{4}{*}{0.75} &  Uncalibrated & 0.151 & 0.172 & 0.518 & 0.855 \\
    & & Histogram Binning & 0.030 & 0.142 & 0.457 & 0.832 \\
    & & Logistic Calibration & 0.032 & \textbf{0.140} & 0.439 & 0.855 \\
    & & Beta Calibration & \textbf{0.026} & \textbf{0.140} & \textbf{0.437} & 0.855 \\ \hline \hline
    \multirow{8}{*}{\rotatebox[origin=c]{90}{\makecell{\maskrcnn{}\\(trained on Cityscapes)}}} & \multirow{4}{*}{0.50} & Uncalibrated & 0.108 & 0.145 & 0.496 & 0.952 \\
    & & Histogram Binning & 0.033 & 0.133 & 0.493 & 0.902 \\
    & & Logistic Calibration & \textbf{0.029} & \textbf{0.124} & \textbf{0.378} & 0.952 \\
    & & Beta Calibration & \textbf{0.029} & 0.125 & 0.379 & 0.952 \\ \cline{2-7}
    &  \multirow{4}{*}{0.75} & Uncalibrated & 0.296 & 0.269 & 1.055 & 0.896 \\
    & & Histogram Binning & \textbf{0.036} & 0.160 & 0.547 & 0.757 \\
    & & Logistic Calibration & 0.042 & \textbf{0.135} & \textbf{0.421} & 0.896 \\
    & & Beta Calibration & 0.044 & \textbf{0.135} & 0.422 & 0.896 \\ \hline
\end{tabular}%
    \label{tab:confidence:detection:evaluation:confidence}
\end{table}
\begin{table}[t!]
    \centering
    \caption[Calibration results for object detection where all information are used for confidence calibration and calibration evaluation.]{
        Calibration results for object detection where all information $\allcollectvariates = (\predconfidencevariate, \predoutputvariate, \allpredbboxvariates)^\T$ are used for confidence calibration and D-ECE evaluation.
        In this case, $\allpredbboxvariates$ denotes the bounding box encoding that consists of the center $x$ and $y$ positions $\centerx$, $\centery$ as well as of the width $w$ and height $h$. 
        Similar to the results for the confidence-only case in Tab. \ref{tab:confidence:detection:evaluation:confidence}, the scaling methods consistently achieve the best results. 
        However, the multivariate confidence calibration is not a monotonically increasing function any more and thus has an influence on the \ac{AUPRC}, which is marginal in this case \cite[p. 239, Tab. 2]{Kueppers2022a}.
    }
     \begin{tabular}{c|c|l|cccc}
    \hline
    Network & IoU & Calibration method & D-ECE &        Brier &          NLL &          AUPRC \\ \hline \hline
    \multirow{12}{*}{\rotatebox[origin=c]{90}{\makecell{\fasterrcnn{}\\(trained on MS COCO)}}} & \multirow{6}{*}{0.50} & Uncalibrated & 0.119 & 0.176 & 0.536 & \textbf{0.920} \\
    & & Histogram Binning & 0.052 & 0.174 & 0.712 & 0.829 \\
    & & Logistic Calibration (independent) & \textbf{0.041} & \textbf{0.143} & \textbf{0.436} & 0.919 \\
    & & Logistic Calibration (dependent) & 0.043 & 0.146 & 0.456 & 0.915 \\
    & & Beta Calibration (independent) & 0.042 & 0.145 & 0.442 & 0.916 \\
    & & Beta Calibration (dependent) & 0.046 & 0.146 & 0.444 & 0.914 \\ \cline{2-7}
    &  \multirow{6}{*}{0.75} & Uncalibrated & 0.227 & 0.257 & 0.829 & \textbf{0.866} \\
    & & Histogram Binning & 0.059 & 0.186 & 0.689 & 0.723 \\
    & & Logistic Calibration (independent) & \textbf{0.044} & \textbf{0.145} & \textbf{0.452} & 0.864 \\
    & & Logistic Calibration (dependent) & 0.047 & 0.149 & 0.469 & 0.856 \\
    & & Beta Calibration (independent) & 0.047 & 0.146 & 0.454 & 0.862 \\
    & & Beta Calibration (dependent) & 0.047 & 0.147 & 0.456 & 0.861 \\ \hline \hline
    \multirow{12}{*}{\rotatebox[origin=c]{90}{\makecell{\retinanet{}\\(trained on MS COCO)}}} & \multirow{6}{*}{0.50} & Uncalibrated & 0.072 & 0.157 & 0.478 & 0.907 \\
    & & Histogram Binning & 0.046 & 0.175 & 0.739 & 0.842 \\
    & & Logistic Calibration (independent) & \textbf{0.045} & \textbf{0.149} & \textbf{0.450} & \textbf{0.908} \\
    & & Logistic Calibration (dependent) & 0.049 & 0.153 & 0.474 & 0.903 \\
    & & Beta Calibration (independent) & 0.046 & 0.150 & 0.458 & 0.906 \\
    & & Beta Calibration (dependent) & 0.053 & 0.155 & 0.467 & 0.901 \\ \cline{2-7}
    &  \multirow{6}{*}{0.75} & Uncalibrated & 0.110 & 0.172 & 0.518 & \textbf{0.855} \\
    & & Histogram Binning & 0.049 & 0.162 & 0.668 & 0.756 \\
    & & Logistic Calibration (independent) & 0.048 & 0.140 & \textbf{0.439} & \textbf{0.855} \\
    & & Logistic Calibration (dependent) & 0.048 & 0.142 & 0.463 & 0.848 \\
    & & Beta Calibration (independent) & \textbf{0.047} & \textbf{0.139} & \textbf{0.439} & 0.853 \\
    & & Beta Calibration (dependent) & 0.053 & 0.144 & 0.450 & 0.844 \\ \hline \hline
    \multirow{12}{*}{\rotatebox[origin=c]{90}{\makecell{\maskrcnn{}\\(trained on Cityscapes)}}} & \multirow{6}{*}{0.50} & Uncalibrated & 0.102 & 0.145 & 0.496 & \textbf{0.952} \\
    & & Histogram Binning & 0.053 & 0.150 & 0.536 & 0.857 \\
    & & Logistic Calibration (independent) & 0.038 & \textbf{0.125} & \textbf{0.381} & 0.950 \\
    & & Logistic Calibration (dependent) & 0.045 & 0.133 & 0.437 & 0.948 \\
    & & Beta Calibration (independent) & \textbf{0.036} & 0.127 & 0.404 & 0.950 \\
    & & Beta Calibration (dependent) & 0.063 & 0.134 & 0.413 & 0.925 \\ \cline{2-7}
    &  \multirow{6}{*}{0.75} & Uncalibrated & 0.281 & 0.269 & 1.055 & 0.896 \\
    & & Histogram Binning & 0.080 & 0.194 & 0.606 & 0.685 \\
    & & Logistic Calibration (independent) & 0.056 & \textbf{0.135} & \textbf{0.424} & \textbf{0.901} \\
    & & Logistic Calibration (dependent) & 0.064 & 0.139 & 0.462 & 0.896 \\
    & & Beta Calibration (independent) & \textbf{0.052} & 0.137 & 0.503 & 0.895 \\
    & & Beta Calibration (dependent) & 0.096 & 0.160 & 0.491 & 0.832 \\ \hline
\end{tabular}%
    \label{tab:confidence:detection:evaluation:position}
\end{table}

For each inspected object detection model, we observe a high deviation between predicted confidence and observed precision which is indicated by high scores, especially for \ac{D-ECE} but also for Brier score and \ac{NLL}. 
This holds for the confidence-only case as well as for the position-dependent case. 
The reliability diagrams in \figref{fig:confidence:detection:reliability:0d} reveal an overconfidence of the \fasterrcnn{} on the MS COCO data set. 
This is in agreement with the current state of research which evaluates common neural network architectures as overconfident \cite{Guo2018,Minderer2021}. 
In contrast, the \retinanet{} architecture is trained using the focal loss \cite{Lin2017} which is known of resulting in low confidence estimates and thus in underconfidence \cite{Mukhoti2020,Schwaiger2021}. 
In both cases, the object detection models are miscalibrated which is alleviated using confidence calibration. 
In our experiments, we found that especially the scaling methods Logistic Calibration and Beta Calibration are able to successfully recalibrate the detection models. 
Furthermore, these methods provide a monotonically increasing mapping of uncalibrated confidences to calibrated one in the standard (confidence only) case and therefore do not affect the \ac{AUPRC} scores.
In contrast, Histogram Binning has no restrictions and thus leads to a degradation of the prediction performance.

Similar to the standard (confidence only) calibration, our position-dependent calibration and especially the scaling methods are also able to provide a meaningful recalibration mapping. 
The reliability diagrams show minor improvements in the position-dependent calibration of the models especially in the 2d case. 
In contrast to confidence-only calibration, the position-dependent calibration is not a monotonically increasing mapping and thus does affect the \ac{AUPRC} scores. 
However, the effect on the \ac{AUPRC} scores is marginal for the scaling methods.
By comparing the conditional independent calibration methods with their conditional dependent counterparts, we observe no improvements in calibration and only minor differences in the results.
We could find a low connection between position and miscalibration which, however, is not as strong as initially suggested.
This might be the reason why the conditional dependent methods do not yield further improvements in calibration in our experiments.
Therefore, we conclude that the standard scaling methods already provide sufficient confidence calibration, which is further improved by our conditionally independent position-dependent calibration methods. 
This is a valuable extension especially for safety-critical applications and subsequent processes which will be further investigated in \chapref{chapter:tracking}.

\subsection{Instance Segmentation}
The experiments shown here are part of our previous work in \cite[p. 239 ff.]{Kueppers2022a}.
For the evaluation of calibration within the task of instance segmentation, we apply a pretrained \maskrcnn{} \cite{He2017} as well as a pretrained \pointrend{} \cite{Kirillov2020} on the MS COCO and Cityscapes validation data sets. 
Since the target is to perform recalibration for the instance segmentation masks, we can use each pixel as an own input to the calibration functions. 
This leads to a large training data set (e.g., approx. 45 million samples for the class \textit{pedestrian} within the Cityscapes data set), resulting in a data set that is too large to train the scaling methods on common hardware in a reasonable time.
Furthermore, it has recently been shown that binning methods such as Histogram Binning yield a more robust calibration mapping compared to scaling methods given a large amount of data \cite{Kumar2019}.
In contrast, a large data set is available for the task of instance segmentation calibration.
Therefore, we only use the Histogram Binning as calibration method for instance segmentation calibration.

\begin{table}[t!]
    \centering
    \caption[Calibration results for instance segmentation where only the predicted pixel confidence is used for calibration and evaluation.]{
        Calibration results for instance segmentation where only the predicted pixel confidence $\predconfidencevariate_\indexpixel$ is used for calibration and D-ECE evaluation \cite[p. 242, Tab. 3-4]{Kueppers2022a}.
        The Histogram Binning is able to reduce miscalibration of the mask scores while preserving the mask quality in all cases.
    }
    \def\stackalignment{r}
    \stackunder{%
        \begin{tabular}{c|c|c|l|cccc}
    \hline
    Network & Data set & IoU & Calibration method & D-ECE & Brier & NLL & AUPRC \\ \hline \hline
    \multirow{8}{*}{\maskrcnn{}} & \multirow{4}{*}{Cityscapes} & \multirow{2}{*}{0.50} & Uncalibrated & 0.071 & 0.110 & 0.432 & \textbf{0.724} \\
    & & & Histogram Binning & \textbf{0.057} & \textbf{0.099} & \textbf{0.320} & 0.723 \\ \cline{3-8}
    & & \multirow{2}{*}{0.75} & Uncalibrated & 0.129 & 0.147 & 0.622 & \textbf{0.375} \\
    & & & Histogram Binning & \textbf{0.059} & \textbf{0.108} & \textbf{0.340} & \textbf{0.375} \\ \cline{2-8}
    & \multirow{4}{*}{MS COCO} & \multirow{2}{*}{0.50} & Uncalibrated & 0.220 & 0.222 & 0.940 & \textbf{0.663}  \\
    & & & Histogram Binning & \textbf{0.064} & \textbf{0.150} & \textbf{0.442} & 0.662 \\ \cline{3-8}
    & & \multirow{2}{*}{0.75} & Uncalibrated & 0.266 & 0.250 & 1.070 & \textbf{0.237} \\
    & & & Histogram Binning & \textbf{0.060} & \textbf{0.144} & \textbf{0.423} & 0.235 \\ \hline \hline
    \multirow{8}{*}{\pointrend{}} & \multirow{4}{*}{Cityscapes} & \multirow{2}{*}{0.50} & Uncalibrated & 0.129 & 0.160 & 0.785 & \textbf{0.709} \\
    & & & Histogram Binning & \textbf{0.027} & \textbf{0.105} & \textbf{0.326} & 0.698 \\ \cline{3-8}
    & & \multirow{2}{*}{0.75} & Uncalibrated & 0.187 & 0.192 & 0.929 & \textbf{0.347} \\
    & & & Histogram Binning & \textbf{0.039} & \textbf{0.115} & \textbf{0.349} & 0.344 \\ \cline{2-8}
    & \multirow{4}{*}{MS COCO} & \multirow{2}{*}{0.50} & Uncalibrated & 0.223 & 0.222 & 0.946 & \textbf{0.672} \\
    & & & Histogram Binning & \textbf{0.063} & \textbf{0.144} & \textbf{0.428} & 0.664 \\ \cline{3-8}
    & & \multirow{2}{*}{0.75} & Uncalibrated & 0.266 & 0.248 & 1.060 & \textbf{0.258} \\
    & & & Histogram Binning & \textbf{0.067} & \textbf{0.138} & \textbf{0.411} & 0.238 \\ \hline
\end{tabular}%

    }{\small\springercopyright{2022}}
    \label{tab:confidence:instance:evaluation:confidence}
\end{table}%
\begin{table}[h!]
    \centering
    \caption[Calibration results for instance segmentation where all information are used for confidence calibration and calibration evaluation.]{
        Calibration results for instance segmentation where all information $\allcollectvariates = (\predconfidencevariate_\indexpixel, \predoutputvariate, \allpredbboxvariates_\indexpixel)^\T$ are used for confidence calibration and D-ECE evaluation \cite[p. 242, Tab. 3-4]{Kueppers2022a}.
        The multivariate Histogram Binning also reduces miscalibration. Furthermore, it significantly improves the mask quality.
    }
    \def\stackalignment{r}
    \stackunder{%
        \begin{tabular}{c|c|c|l|cccc}
    \hline
    Network & Data set & IoU & Calibration method & D-ECE & Brier & NLL & AUPRC \\ \hline \hline
    \multirow{8}{*}{\maskrcnn{}} & \multirow{4}{*}{Cityscapes} & \multirow{2}{*}{0.50} & Uncalibrated & 0.112 & \textbf{0.110} & \textbf{0.432} & 0.724 \\
    & & & Histogram Binning & \textbf{0.101} & 0.117 & 0.530 & \textbf{0.787} \\ \cline{3-8}
    & & \multirow{2}{*}{0.75} & Uncalibrated & 0.145 & 0.147 & 0.622 & 0.375 \\
    & & & Histogram Binning & \textbf{0.103} & \textbf{0.120} & \textbf{0.523} & \textbf{0.479} \\ \cline{2-8}
    & \multirow{4}{*}{MS COCO} & \multirow{2}{*}{0.50} & Uncalibrated & 0.234 & 0.222 & 0.940 & 0.663  \\
    & & & Histogram Binning & \textbf{0.136} & \textbf{0.171} & \textbf{0.776} & \textbf{0.760} \\ \cline{3-8}
    & & \multirow{2}{*}{0.75} & Uncalibrated & 0.272 & 0.250 & 1.070 & 0.237 \\
    & & & Histogram Binning & \textbf{0.129} & \textbf{0.165} & \textbf{0.720} & \textbf{0.425} \\ \hline \hline
    \multirow{8}{*}{\pointrend{}} & \multirow{4}{*}{Cityscapes} & \multirow{2}{*}{0.50} & Uncalibrated & 0.209 & \textbf{0.160} & \textbf{0.785} & 0.709 \\
    & & & Histogram Binning & \textbf{0.190} & 0.190 & 1.299 & \textbf{0.758} \\ \cline{3-8}
    & & \multirow{2}{*}{0.75} & Uncalibrated & 0.254 & 0.192 & \textbf{0.929} & 0.347 \\
    & & & Histogram Binning & \textbf{0.184} & \textbf{0.191} & 1.247 & \textbf{0.486} \\ \cline{2-8}
    & \multirow{4}{*}{MS COCO} & \multirow{2}{*}{0.50} & Uncalibrated & 0.240 & 0.222 & \textbf{0.946} & 0.672 \\
    & & & Histogram Binning & \textbf{0.161} & \textbf{0.180} & 1.005 & \textbf{0.751} \\ \cline{3-8}
    & & \multirow{2}{*}{0.75} & Uncalibrated & 0.274 & 0.248 & 1.060 & 0.258 \\
    & & & Histogram Binning & \textbf{0.153} & \textbf{0.173} & \textbf{0.936} & \textbf{0.388} \\ \hline
\end{tabular}%

    }{\small\springercopyright{2022}}
    \label{tab:confidence:instance:evaluation:position}
\end{table}

\begin{figure}[t!]
    \centering
    \def\stackalignment{r}
    \begin{subfigure}{\textwidth}
        \stackunder{%
            \includegraphics[width=0.3495\textwidth,]{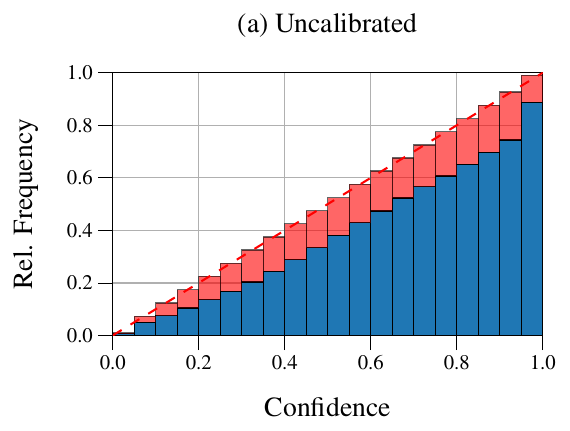}%
            \includegraphics[width=0.3175\textwidth,]{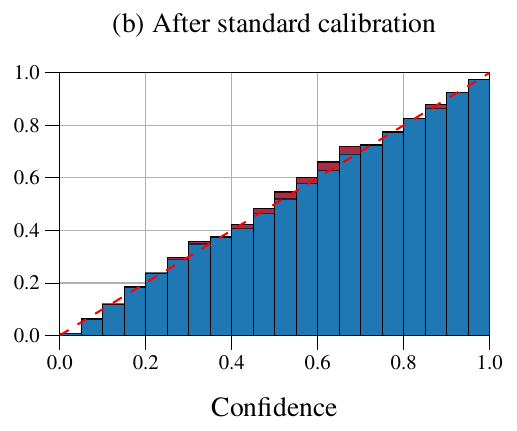}%
            \includegraphics[width=0.3175\textwidth,]{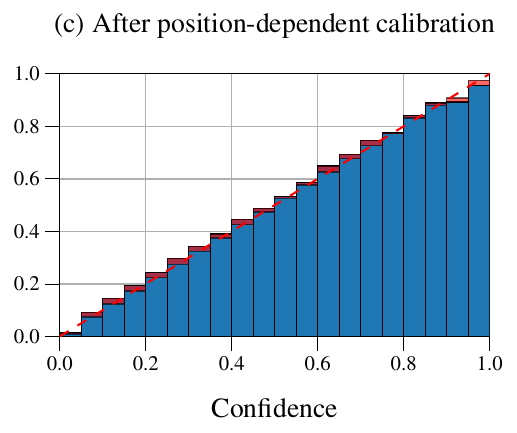}%
        }{\scriptsize\springercopyright{2022}}
        \caption{Reliability diagrams w.r.t. the confidence only \cite[p. 243, Fig. 6]{Kueppers2022a}.}
        \label{fig:confidence:instance:reliability:0d}
    \end{subfigure}
    \vspace{0.5cm}
    \begin{subfigure}{\textwidth}
        \stackunder{%
            \includegraphics[width=0.34\textwidth,]{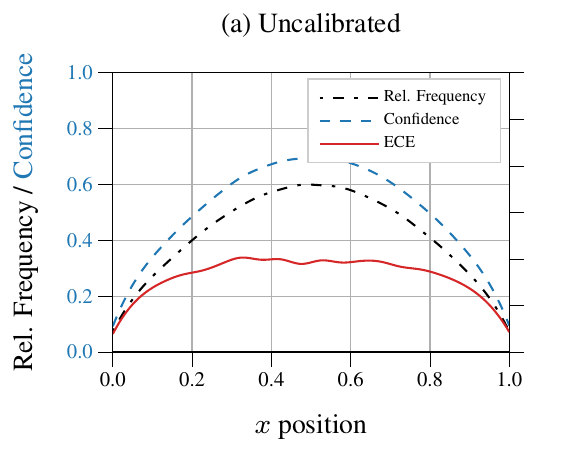}%
            \includegraphics[width=0.2925\textwidth,]{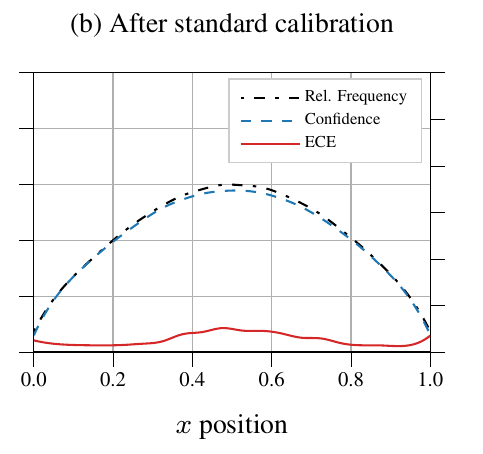}%
            \includegraphics[width=0.3495\textwidth,]{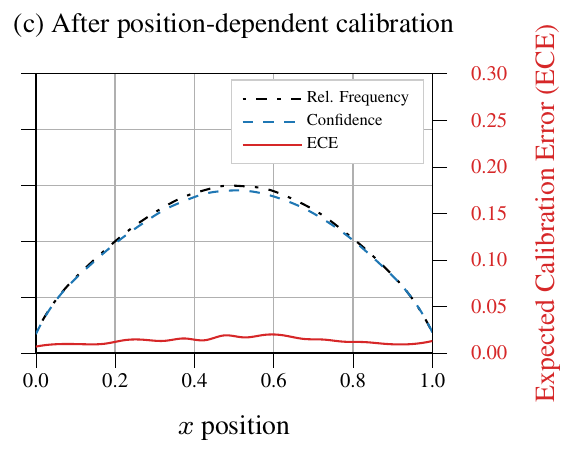}%
        }{\scriptsize\springercopyright{2022}}
        \caption{Reliability diagrams w.r.t. the relative $x$ position of each mask pixel (1d) \cite[p. 243, Fig. 7]{Kueppers2022a}.}
        \label{fig:confidence:instance:reliability:1d}
    \end{subfigure}
	\vspace{0.5cm}
    \begin{subfigure}{\textwidth}
        \stackunder{%
            \includegraphics[width=0.36\textwidth,]{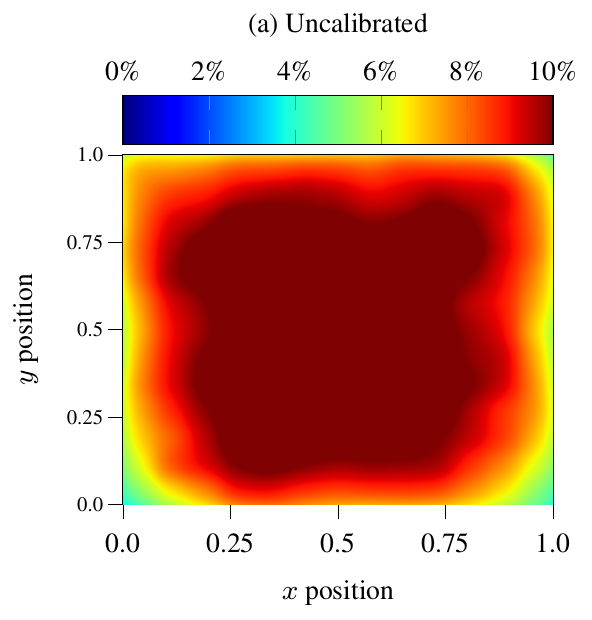}%
            \includegraphics[width=0.30675\textwidth,]{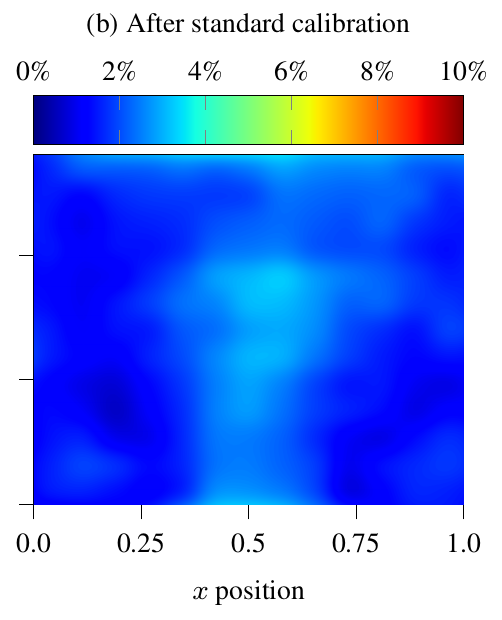}%
            \includegraphics[width=0.30675\textwidth,]{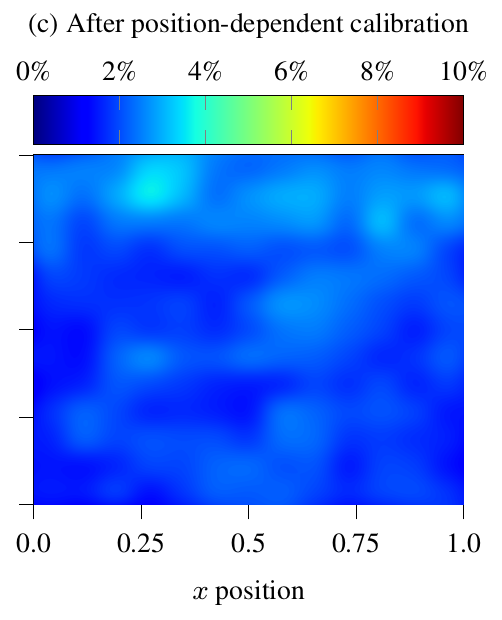}%
        }{\scriptsize\springercopyright{2022}}
        \caption{Reliability diagrams w.r.t. the relative $x$ and $y$ position of each mask pixel (2d) \cite[p. 243, Fig. 8]{Kueppers2022a}.}
        \label{fig:confidence:instance:reliability:2d}
    \end{subfigure}
    \caption[Reliability diagrams (instance segmentation) for a \maskrcnn{} on the MS COCO calibration validation set for class \textit{pedestrian}.]{
        Reliability diagrams (instance segmentation) for a \maskrcnn{} on the MS COCO calibration validation set for class \textit{pedestrian} \cite[p. 243, Fig. 6-8]{Kueppers2022a}.
        The uncalibrated pixel confidences are consistently too overconfident for all confidence levels.
        Furthermore, the gap between predicted pixel confidence and observed frequency increases towards the mask's center.
        This is mitigated by standard calibration as well as by our extended methods.
    }
    \label{fig:confidence:instance:reliability}
    
\end{figure}

Instance segmentation is a joint task of object detection and semantic segmentation within a predicted bounding box. Similar to object detection, it is necessary to distinguish between true and false positives which requires a certain \ac{IoU} threshold. Thus, we run our evaluations using an IoU threshold of $0.50$ and $0.75$, respectively.
Similar to our experiments for object detection, we further investigate the effect of standard (confidence-only) calibration as well as of position-dependent calibration. We use the pixel $x$ and $y$ position as position information which are normalized to the bounding box size to get $x, y \in [0, 1]$ for calibration. Furthermore, we suspect a correlation between miscalibration and a pixel's distance to the next segment boundary. Thus, we also use the (normalized) distance as an additional feature for calibration.
For calibration evaluation, we use the \ac{D-ECE} with $\numbins_\indexdims = 15$ bins for each dimension $\indexdims \in \{1, \ldots, \numdims\}$. Furthermore, the Brier and \ac{NLL} scores are used as complementary evaluation metrics. Finally, we report the \ac{mIoU} which is the mean \ac{IoU} score over all classes and which is an indicator for the quality of the predicted instance segmentation masks. 
The evaluation results for the confidence-only case as well as for the multivariate calibration case are given in Tab. \ref{tab:confidence:instance:evaluation:confidence} and Tab. \ref{tab:confidence:instance:evaluation:position}, respectively. For further insights, we show the reliability diagrams for the confidence only as well as for including position information in \figref{fig:confidence:instance:reliability}.

As visualized in \figref{fig:confidence:instance:reliability}, the instance segmentation models are consistently overconfident in their predictions for the pixel confidence. 
This miscalibration is reduced by the standard Histogram Binning as well as by the position-dependent Histogram Binning. 
In some cases, the position-dependent Histogram Binning does not lead to an improvement in the complementary Brier and \ac{NLL} scores.
On the one hand, we can observe a strong connection between position information and miscalibration. 
In this case, both calibration schemes lead to an improvement in calibration, whereas the position-dependent variant results in a more uniform calibration over the $x$ and $y$ space compared to its confidence-only counterpart. 
On the other hand, our experiments show that position-dependent calibration is able to significantly improve the quality of the segmentation masks which is indicated by the gain in the \ac{mIoU} scores for all evaluated models. 
For standard calibration, the mask scores are only rescaled by their confidence which might lead to a better calibration but sometimes also to unwanted losses of mask segments (especially small objects in the background). 
In contrast, position-dependent calibration is able to apply a recalibration that is also aware of possible correlations between pixel confidence and object size. 
We assume that this leads to improved estimates of the mask confidences even for smaller objects.
Therefore, we conclude that especially the position-dependent calibration is a valuable contribution towards reliable confidence information and improved seg\-mentation masks for the task of instance segmentation.

\subsection{Semantic Segmentation}
As opposed to instance segmentation, it is not necessary to identify single objects within semantic segmen\-tation but to determine the membership of each image pixel to a general class. 
Therefore, the experiments for semantic segmentation calibration do not rely on a preceding detection stage and we can use each image pixel as an input for calibration training and evaluation using the Histogram Binning.
We use a pretrained \deeplabp{} \cite{Chen2018a} and a pretrained \deeplab[2]{} \cite{Chen2018} on the Cityscapes and MS COCO validation data sets, respectively, as well as a pretrained \hrnet{} \cite{Sun2019,Wang2020,Yuan2020}.
We further use the same additional features for confidence calibration such as relative $x$, $y$ position and the pixel's distance to the next segment boundary. 
The calibration results are shown in \tabref{tab:confidence:semantic:evaluation:confidence} for the confidence-only calibration case as well as in \tabref{tab:confidence:semantic:evaluation:position} for the position-dependent case.
Furthermore, we show the respective reliability diagrams in Fig.\ref{fig:confidence:semantic:reliability} to gain further insights in the calibration properties of the examined models.
\begin{table}[t!]
    \centering
    \caption[Calibration results for semantic segmentation where only the predicted pixel confidence is used for calibration and evaluation.]{
        Calibration results for semantic segmentation where only the predicted pixel confidence $\predconfidencevariate_\indexpixel$ is used for calibration and D-ECE evaluation \cite[p. 245, Tab. 5]{Kueppers2022a}.
        The uncalibrated segmentation models are already well-calibrated. The Histogram Binning does not affect the mask quality and only leads to minor improvements in confidence calibration.
    }
    \def\stackalignment{r}
    \stackunder{%
        \begin{tabular}{c|c|l|cccc}
    \hline
    Network & Data set & Calibration Method & D-ECE & Brier & NLL & mIoU \\ \hline \hline
    \multirow{2}{*}{\deeplabp{}} & \multirow{2}{*}{Cityscapes} & Uncalibrated & 0.0016 & 0.060 & \textbf{0.139} & \textbf{0.623} \\
    & & Histogram Binning & \textbf{0.0008} & 0.060 & 0.170 & 0.619 \\ \hline \hline
    \multirow{2}{*}{\deeplab[2]{}} & \multirow{2}{*}{MS COCO} & Uncalibrated & 0.0009 & 0.458 & \textbf{1.173} & \textbf{0.933} \\
    & & Histogram Binning & \textbf{0.0006} & \textbf{0.456} & 1.515 & \textbf{0.933} \\ \hline \hline
    \multirow{4}{*}{\hrnet{}} & \multirow{2}{*}{Cityscapes} & Uncalibrated & \textbf{0.0007} & 0.057 & \textbf{0.115} & \textbf{0.629} \\
    & & Histogram Binning & 0.0008 & 0.057 & 0.148 & 0.628 \\ \cline{2-7}
    & \multirow{2}{*}{MS COCO} & Uncalibrated & 0.0046 & 0.779 & 5.812 & \textbf{0.939} \\
    & & Histogram Binning & \textbf{0.0006} & \textbf{0.563} & \textbf{2.261} & \textbf{0.939} \\ \hline
\end{tabular}%
    }{\scriptsize\springercopyright{2022}}
    \label{tab:confidence:semantic:evaluation:confidence}
\end{table}
\begin{table}[t!]
    \centering
    \caption[Calibration results for semantic segmentation where all information are used for confidence calibration and calibration evaluation.]{
        Calibration results for semantic segmentation where all information $\allcollectvariates = (\predconfidencevariate_\indexpixel, \allpredbboxvariates_\indexpixel)^\T$ are used for confidence calibration and D-ECE evaluation \cite[p. 245, Tab. 5]{Kueppers2022a}.
        In contrast to the case in \tabref{tab:confidence:semantic:evaluation:confidence}, the position-dependent calibration is not able to improve calibration. Furthermore, it leads to a minor degradation of the mask quality as indicated by the \ac{mIoU}.
    }
    \def\stackalignment{r}
    \stackunder{%
         \begin{tabular}{c|c|l|cccc}
    \hline
    Network & Data set & Calibration Method & D-ECE & Brier & NLL & mIoU \\ \hline \hline
    \multirow{2}{*}{\deeplabp{}} & \multirow{2}{*}{Cityscapes} & Uncalibrated & 0.0019 & \textbf{0.060} & \textbf{0.139} & \textbf{0.623} \\
    & & Histogram Binning & 0.0019 & 0.062 & 0.189 & 0.589 \\ \hline \hline
    \multirow{2}{*}{\deeplab[2]{}} & \multirow{2}{*}{MS COCO} & Uncalibrated & 0.0015 & \textbf{0.458} & \textbf{1.173} & \textbf{0.933} \\
    & & Histogram Binning & 0.0015 & 0.485 & 1.790 & 0.913 \\ \hline \hline
    \multirow{4}{*}{\hrnet{}} & \multirow{2}{*}{Cityscapes} & Uncalibrated & \textbf{0.0015} & \textbf{0.057} & \textbf{0.115} & \textbf{0.629} \\
    & & Histogram Binning & 0.0019 & 0.060 & 0.171 & 0.582 \\ \cline{2-7}
    & \multirow{2}{*}{MS COCO} & Uncalibrated & 0.0046 & 0.779 & 0.812 & \textbf{0.939} \\
    & & Histogram Binning & \textbf{0.0014} & \textbf{0.571} & \textbf{0.372} & 0.931 \\ \hline
\end{tabular}%
    }{\scriptsize\springercopyright{2022}}
    \label{tab:confidence:semantic:evaluation:position}
\end{table}
\begin{figure}[t!]
    \centering
    \def\stackalignment{r}
    \begin{subfigure}{\textwidth}
        \stackunder{%
            \includegraphics[width=0.3495\textwidth,]{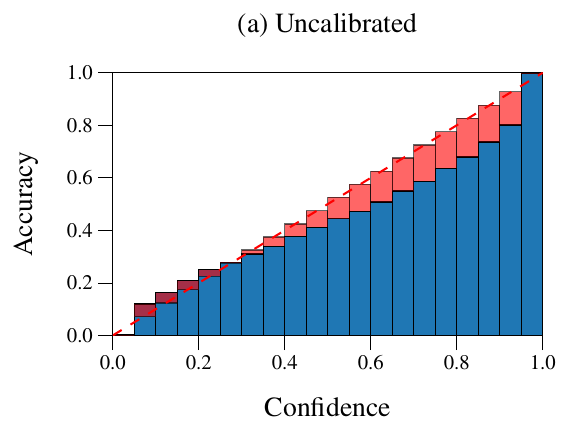}%
            \includegraphics[width=0.3175\textwidth,]{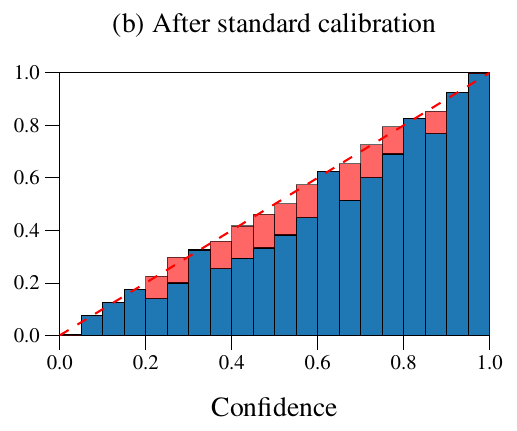}%
            \includegraphics[width=0.3175\textwidth,]{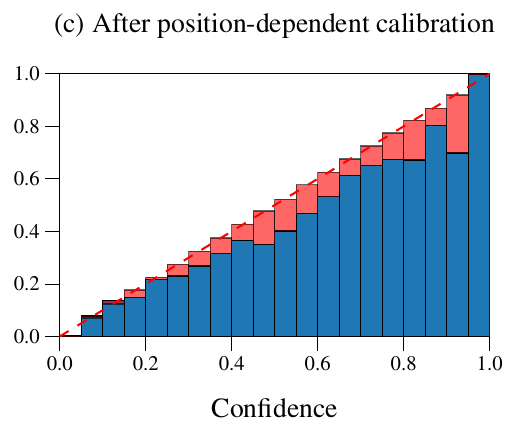}%
        }{\scriptsize\springercopyright{2022}}
        \caption{Reliability diagrams w.r.t. the confidence only \cite[p. 246, Fig. 11]{Kueppers2022a}.}
        \label{fig:confidence:semantic:reliability:0d}
    \end{subfigure}
	\vspace{0.5cm}
    \begin{subfigure}{\textwidth}
        \stackunder{%
            \includegraphics[width=0.34\textwidth,]{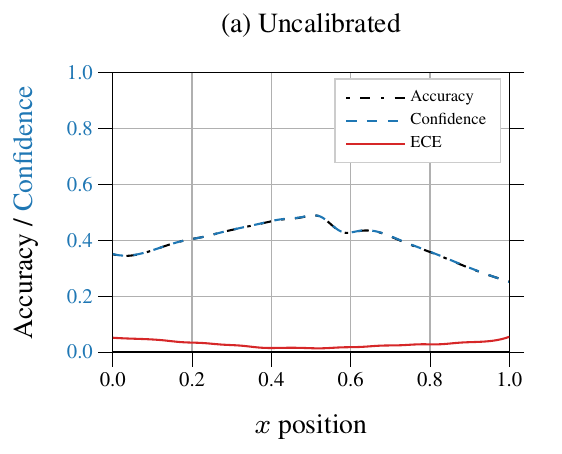}
            \includegraphics[width=0.294\textwidth,]{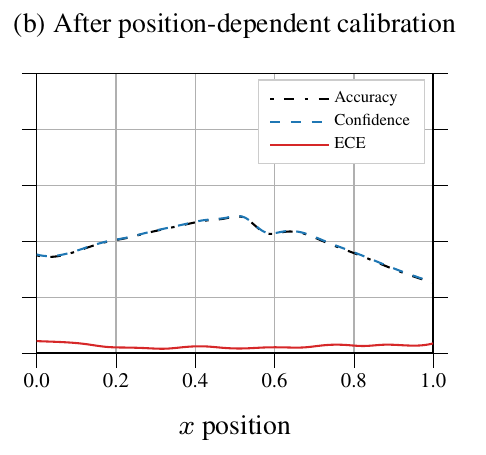}%
            \includegraphics[width=0.348\textwidth,]{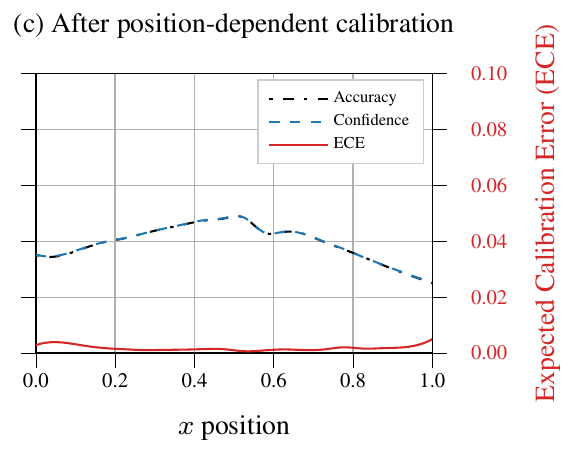}%
        }{\scriptsize\springercopyright{2022}}
        \caption{Reliability diagrams w.r.t. the relative $x$ position of each mask pixel (1d) \cite[p. 247, Fig. 12]{Kueppers2022a}.}
        \label{fig:confidence:semantic:reliability:1d}
    \end{subfigure}
	\vspace{0.5cm}
    \begin{subfigure}{\textwidth}
        \stackunder{%
            \includegraphics[width=0.361\textwidth,]{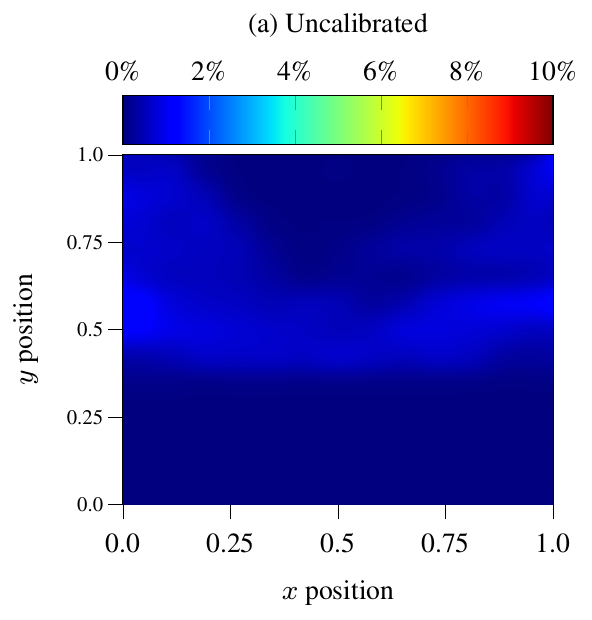}%
            \includegraphics[width=0.30675\textwidth,]{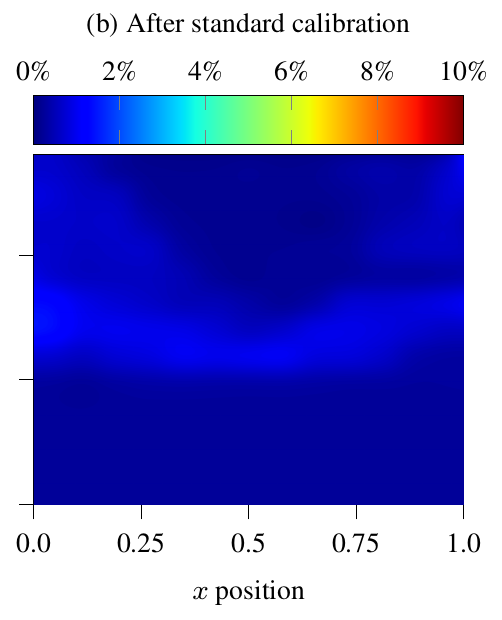}%
            \includegraphics[width=0.30675\textwidth,]{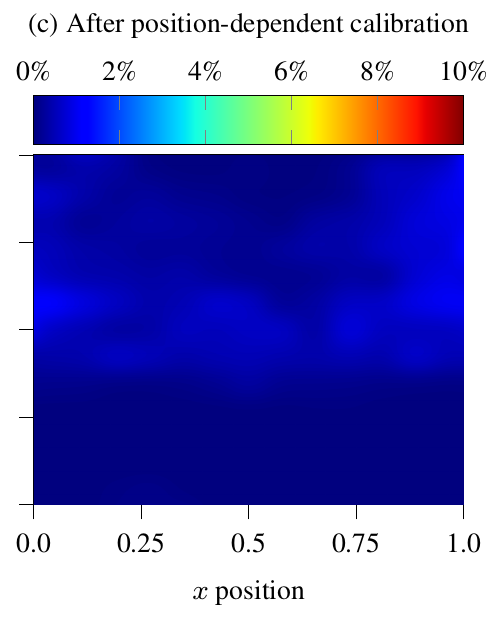}%
        }{\scriptsize\springercopyright{2022}}
        \caption{Reliability diagrams w.r.t. the relative $x$ and $y$ position of each mask pixel (2d) \cite[p. 247, Fig. 13]{Kueppers2022a}.}
        \label{fig:confidence:semantic:reliability:2d}
    \end{subfigure}
    \caption[Reliability diagrams (semantic segmentation) for a \deeplabp{} on the Cityscapes calibration validation set for class \textit{pedestrian}.]{
        Reliability diagrams (semantic segmentation) for a \deeplabp{} on the Cityscapes calibration validation set for class \textit{pedestrian} \cite[p. 246 f., Fig. 11-13]{Kueppers2022a}.
        The uncalibrated segmentation model already provides well-calibrated confidences (most samples are located at confidence levels of $0$ or $1$).
        This can be also seen in the 1d and 2d reliability diagrams.
        Thus, neither standard nor position-dependent calibration lead to significant changes in calibration.
    }
    \label{fig:confidence:semantic:reliability}
\end{figure}
In contrast to instance segmentation, the examined semantic segmentation models provide already well-calibrated confidence estimates on pixel-level. 
In this case, neither standard Histogram Binning nor our position-dependent calibration are able to further improve the calibration. 
However, position-dependent cali\-bration leads to a slight degradation of the mask quality as stated by the \ac{mIoU} scores. 
Although the reliability diagrams in \figref{fig:confidence:semantic:reliability:0d} show an overconfidence for the confidence range $\probvariate \in [0.35, 0.95]$, most pixels have a low confidence of $\probvariate < 0.1$, which results in an overall low miscalibration score (\ac{D-ECE}).
We suspect that the major difference between instance and semantic segmentation in calibration in our experiments arises from the different approaches for model training. 
As already pointed out, instance segmentation is a joint task of object detection and segmentation. 
Thus, the quality of the predicted bounding boxes directly affects the performance of the segmentation head. 
It is a more challenging task to identify single objects in an image and to jointly optimize the network for object detection and segmentation.
Furthermore, a semantic segmentation model uses the whole image for model training and thus has more pixels available, whereas an instance segmentation model is restricted to the pixels within a detected bounding box. 
Therefore, we conclude that the semantic segmentation models, that have been investigated in our experiments, already provide well-calibrated confidence estimates and thus do not require an additional post-hoc calibration step.

\section{Conclusion for Semantic Confidence Calibration}
\label{section:confidence:conclusion}

Object detection is the joint task of predicting the position, shape, and class of individual objects.
For each predicted object, a confidence score is estimated for the predicted class that can be interpreted as a probability of the correctness of the predicted class label.
In this chapter, we examined this semantic confidence score for consistency, i.e., if the estimated uncertainty corresponds to the observed error.
We started by deriving the definitions of semantic confidence calibration for the tasks of object detection, instance segmentation, and semantic segmentation.
These definitions allow to investigate if the position information has an influence on the semantic confidence as well.
This also holds for instance and semantic segmentation where the position of each pixel might have an influence to the confidence.
Thus, these definitions allow for an extension of the common Expected Calibration Error (ECE) metric to the task of object detection and segmentation.
We used the new Detection ECE (D-ECE) to measure the semantic miscalibration of detection and segmentation models.

We extended the common calibration methods Histogram Binning \cite{Zadrozny2001}, Logistic Calibration \cite{Platt1999}, and Beta Calibration \cite{Kull2017} to include additional information such as position and shape into a calibration and to capture possible correlations between the given information.
These multivariate calibration methods can be used to apply a post-hoc calibration of the confidence scores provided by a detection or segmentation model.
The multivariate scaling methods Logistic Calibration and Beta Calibration calibration are further divided into conditionally independent and dependent variants.
Both variants use the confidence and the position of the detected objects for confidence recalibration.
Furthermore, both variants are able to model an influence of the additional position information to the calibration result.
However, the conditionally independent scaling methods assume that the input random variables follow independent probability distributions.
In contrast, the conditionally dependent methods are able to capture possible dependencies in the input.

The proposed multivariate calibration methods have been evaluated using different neural network archi\-tectures and different data sets.
Our investigations show that in our experiments, the semantic segmentation models are already well calibrated so that our calibration methods have not been able to gain further improvements.
In contrast, the multivariate confidence calibration has a positive effect on the calibration properties of object detection and especially on instance segmentation models.
Although we could only find a minor connection between position and miscalibration, our extended methods show a qualitatively good calibration performance.
In the case of instance segmentation, the extended multivariate calibration is able to not only improve the calibration properties of the model but also to enhance the model's prediction performance.
We suspect that the major difference between instance and semantic segmentation is a result of the different training procedures that are used for model training.

Therefore, we conclude that our multivariate extension of the calibration methods provide a powerful framework for the recalibration of semantic uncertainty.
Since object detection is a part of the image-based environment perception process (cf. \chapref{chapter:introduction}), we further investigate the effect of semantic confidence calibration on object detection within a subsequent object tracking in \chapref{chapter:tracking}.

\acresetall
\chapter{Bayesian Confidence Calibration}
\label{chapter:bayesian}

\begin{figure}[b!]
    \centering
    \def\stackalignment{r}
    \stackunder{%
        \begin{overpic}[width=1.0\linewidth]{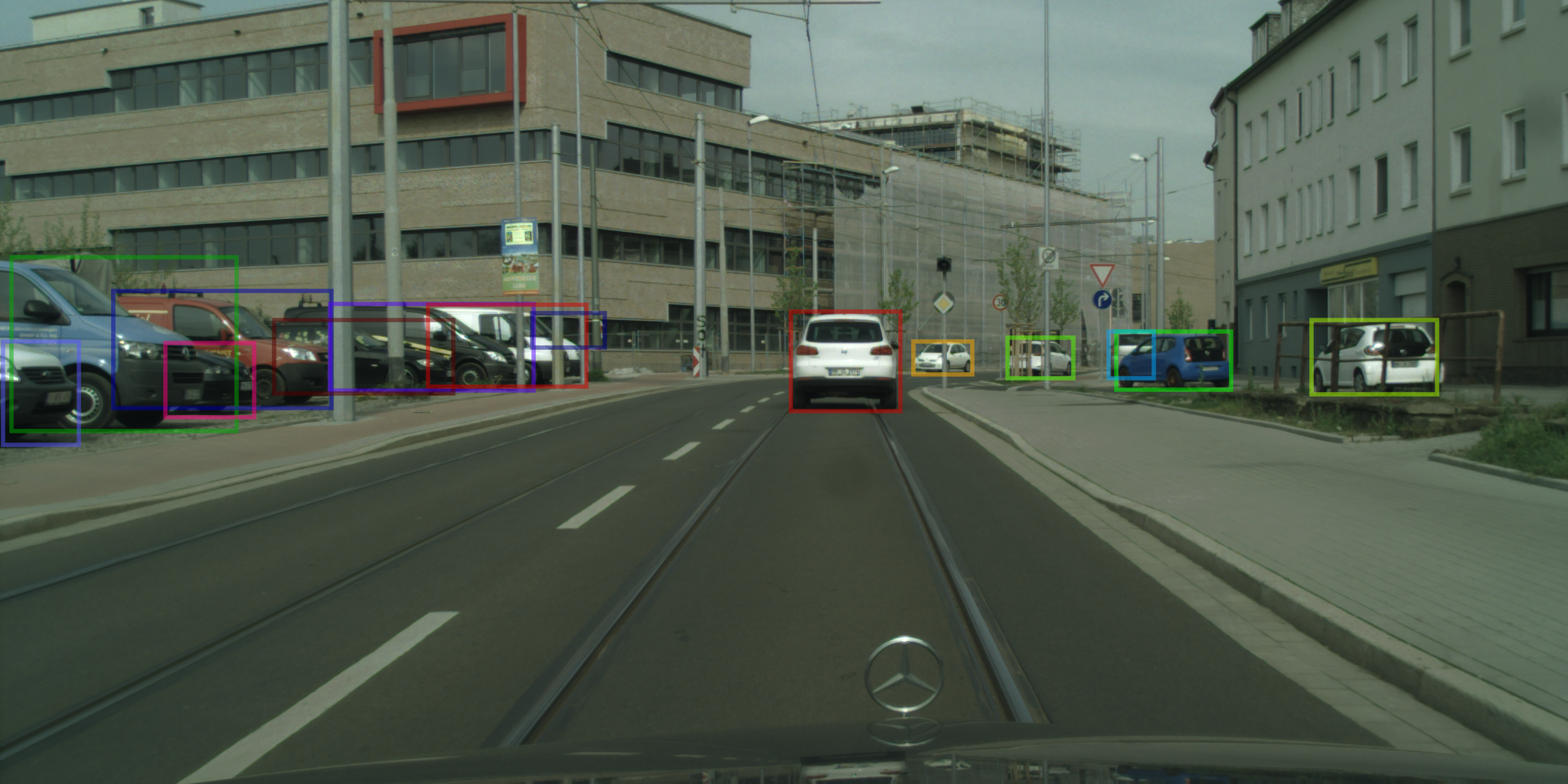}
            \put(52,31.5){\textbf{\scriptsize\color{red}\circled{1}}}
            \put(85,31){\textbf{\scriptsize\color{cyan}\circled{2}}}
            \put(64.5,30){\textbf{\scriptsize\color{orange}\circled{3}}}
            \put(5, 10){
                \colorbox{white}{
                    \begin{minipage}[l]{9.2em}
                        \setlength{\tabcolsep}{1pt}
                        \begin{tabular}{crrl}
                            {\normalsize \color{red}\circled{1}} & \normalsize $100\% \rightarrow$ & \normalsize $98\%$ & \normalsize {\raisebox{0.5ex}{\scriptsize$\substack{+1.2\% \\ -1.3\%}$}}\\
                            {\normalsize \color{cyan}\circled{2}} & \normalsize $100\% \rightarrow$ & \normalsize $96\%$ & \normalsize {\raisebox{0.5ex}{\scriptsize$\substack{+3.8\% \\ -2.7\%}$}}\\
                            {\normalsize \color{orange}\circled{3}} & \normalsize $99\% \rightarrow$ & \normalsize $85\%$ & \normalsize {\raisebox{0.5ex}{\scriptsize$\substack{+15.0\% \\ -14.8\%}$}}
                        \end{tabular}
                    \end{minipage}
                }
            }
        \end{overpic}
    }{\scriptsize\ieeecopyright{2021}}
    \caption[Qualitative example of Bayesian confidence calibration for object detection.]{
        Qualitative example of Bayesian confidence calibration for object detection \cite[p.~1, Fig.~1]{Kueppers2021}.
        For each detection, a calibration method reassigns a new confidence score.
        If we apply Bayesian confidence calibration, it is possible to yield a prediction interval for the calibrated confidence to express the epistemic uncertainty in calibration.
    }
    \label{fig:bayesian:qualitative}
\end{figure}
In the previous chapter \ref{chapter:confidence}, we derived the definitions for semantic confidence calibration from simple classification to the more complex detection, instance, and semantic segmentation tasks.
Furthermore, we proposed methods for a position-dependent confidence calibration.
But what if the calibration method itself is uncertain in some cases, e.g., for an input sample which falls into a region with a sparsely populated training set?
For example, if a calibration method is embedded into a safety-relevant context, it might still lead to a false sense of safety if applied to situations which are unknown either for the baseline model or for the calibration mapping.
In this case, it is advantageous to get a self-assessment about the epistemic calibration model uncertainty.
In contrast to the examinations for the aleatoric uncertainty given by the detector confidence, we seek for the epistemic uncertainty which is inherent in a calibration function itself.

Therefore, we introduce the term of Bayesian confidence calibration which adapts the idea of probabilistic modeling similar to Bayesian neural networks.
Within a Bayesian neural network, the weights in the network layers are treated as probability distributions that express the uncertainty in the weights \cite{Gal2016,Gal2016a}.
In this way, it is possible to sample multiple realizations of the network weights from these distributions.
Each of these weight combinations is then used to generate a sample distribution for each output.
In practice, it is possible to approximate such a Bayesian neural network using dropout during the inference which is also known as Monte-Carlo dropout \cite{Gal2016}.

We adapt this idea of modeling the epistemic uncertainty and seek to transfer this to our calibration functions.
Similarly, a probability distribution is placed over each calibration weight.
However, in contrast to Bayesian neural networks, we can not use Monte-Carlo dropout for calibration methods as these functions do not use any dropout or only use a few weights for recalibration, e.g., Temperature Scaling \cite{Guo2018}.
Therefore, we approximate the weight distributions using \ac{SVI} which allows for training variational distributions for each calibration weight.
In this way, it is possible to obtain multiple calibrated confidence estimates for a single input sample which, in turn, describe a sample distribution and thus reflect the epistemic uncertainty of the calibration mapping itself.
The concept of Bayesian confidence calibration is qualitatively shown in \figref{fig:bayesian:qualitative}.
The work presented in this chapter was subject of the publication in \cite{Kueppers2021}.

In \secref{section:bayesian:methods}, we derive the Bayesian confidence calibration and describe how to obtain the calibrated confidence estimates in conjunction with its epistemic uncertainty.
The Bayesian calibration framework is afterwards used for the evaluation of the calibration performance and of the quality of the provided epistemic uncertainty in \secref{section:bayesian:experiments}.
In our studies, we focus on the task of object detection since the Bayesian confidence calibration framework requires a sampling during inference which has been computational too expensive for the application to instance or semantic segmentation.
We give a conclusion about the methods and our findings in \secref{section:bayesian:conclusion}.

As opposed to \chapref{chapter:confidence} and \chapref{chapter:regression}, we do not aim to integrate this Bayesian framework into a subsequent process such as object tracking in this work.
Our target is to show a possibility to capture additional model uncertainties to increase the awareness of possible failure modes which is important for safety-critical applications.
Nevertheless, the integration of Bayesian confidence calibration, e.g., into an object tracking process, is an interesting use case which we let open for future work.

\textbf{Contributions:} Our contributions are summarized by:
\begin{itemize}
    \item Definition of Bayesian confidence calibration.
    \item Methods for Bayesian confidence calibration.
\end{itemize}

\section{Epistemic Uncertainty Modeling of Confidence Calibration}
\label{section:bayesian:methods}

We use the preceding notation and follow the definition of confidence calibration for classifi\-cation in (\ref{eq:confidence:definition:classification}) as well as for object detection in (\ref{eq:confidence:definition:detection}). 
In this scope, we work with an arbitrary classification or detection model that estimates a label $\predoutputvariate \in \outputset$ as well as a confidence $\predconfidencevariate \in \probset$ for each prediction where the confidence expresses the model's belief about the correctness of the actual prediction.
We denote this belief as $\matchedvariate \sampledfrom \bernoullidistribution(\predconfidencevariate)$.
An object detection model also outputs a position and shape estimate $\allpredbboxvariates \in \bboxset$ for each predicted objects, so that the joint conditioned model distribution is denoted by  $\pdf_{\predconfidencevariate, \predoutputvariate, \allpredbboxvariates}(\predconfidence, \predoutput, \allpredbboxes | \allsingleinput)$.
We further denote $\allcollectvariates = (\predconfidencevariate, \predoutputvariate, \allpredbboxvariates)^\T \in \collectset$\footnote{In the preceding chapter, we distinguished between $\collectset_\probvariate$ as the aggregated output space using the confidence $\predconfidencevariate$, and $\collectset_\logitvariate$ as the output space using the raw network logits $\logitvariate$. Since the derivation for Bayesian confidence calibration holds for both cases, we use the shorthand notation of $\collectset$ to represent either $\collectset_\probvariate$ or $\collectset_\logitvariate$.} as the aggregated variable for the model output. 
These predictions aim to target the true label $\groundtruthvariate \in \outputset$ and the true position $\allgroundtruthbboxvariates \in \bboxset$ which follow the joint ground-truth data distribution $\pdf_{\allinputvariates, \groundtruthvariate, \allgroundtruthbboxvariates}(\allsingleinput, \singlegroundtruth, \allgroundtruthbboxes) = \pdf_{\groundtruthvariate, \allgroundtruthbboxvariates}(\singlegroundtruth, \allgroundtruthbboxes | \allsingleinput)\pdf_{\allinputvariates}(\allsingleinput)$.
For confidence calibration, we adapt the scaling methods presented in \secref{section:confidence:methods:scaling} whose calibration parameters $\allparameters \in \parameterset$ are commonly obtained by \ac{MLE} using a data set $\dataset = \{(\allcollect_\indexsamples, \matched_\indexsamples)\}^{\numsamples}_{\indexsamples=1}$ with $\numsamples$ samples, where $\parameterset$ denotes the set of all possible parameters.
A position-dependent calibration function $\calmodel_{\allparameters}: \collectset \rightarrow \probset$ serves as a mapping from the uncalibrated confidence, the estimated label and the predicted position to a calibrated confidence estimate $\calibratedvariate \in [0, 1]$, so that $\calibratedvariate = \calmodel_{\allparameters}(\allcollectvariates)$.
The calibrated confidence is then used as the Bernoulli parameter for $\matchedvariate$, so that $\matchedvariate \sampledfrom \bernoullidistribution(\calibratedvariate)$.

Similar to neural networks, a calibration mapping may also exhibit epistemic uncertainty, e.g., due to an insufficient amount of training data (cf. \secref{section:basics:unreliable_uncertainty}).
Thus, we are interested in modeling the epistemic uncertainty of the calibrated confidence $\calibratedvariate$ for a new sample $\allcollect^\ast \in \collectset$ during inference given the calibration parameters $\allparameters$ and the calibration training data set $\dataset$.
Therefore, we do not interpret the calibrated confidence as a deterministic distribution parameter but rather as a random variable that follows a certain probability distribution $\calibratedvariate|\allcollect^\ast, \dataset \sampledfrom \pdf_{\calibratedvariate}$.
In Bayesian statistics, we can infer the conditional distribution of $\calibratedvariate$ using the posterior predictive distribution that denotes the probability of $\calibratedvariate$ weighted by the posterior for the calibration parameters $\pdf_{\allparameters}(\allparameters | \dataset)$.
The posterior for the calibration parameters is marginalized out so that the posterior predictive distribution is given by
\begin{align}
    \label{eq:bayesian:predictive}
    \pdf_{\calibratedvariate}(\calibrated^\ast | \allcollect^\ast, \dataset) &= \int_{\parameterset} \pdf_{\calibratedvariate}(\calibrated^\ast | \allcollect^\ast, \allparameters) \pdf_{\allparameters}(\allparameters |  \dataset) \diff \allparameters ,
\end{align}
where $\pdf_{\calibratedvariate}(\calibrated^\ast | \allcollect^\ast, \allparameters)$ is obtained using the calibration function $\calmodel_{\allparameters}(\allcollect^\ast)$ given the parameters $\allparameters$.

For posterior inference, we place a prior distribution $\pdf_{\allparameters}(\allparameters)$ over the parameters $\allparameters$ so that it is possible to obtain a posterior distribution $\pdf_{\allparameters}(\allparameters | \dataset)$ given the training data set $\dataset$.
Thus, the posterior is defined by
\begin{align}
    \label{eq:bayesian:posterior}
    \pdf_{\allparameters}(\allparameters | \dataset) = \pdf_{\allparameters}(\allparameters |  \mathbf{s}, \mathbf{\matched}) = \frac{\pdf_{\calibratedvariate}(\mathbf{\matched} | \mathbf{s}, \allparameters) \pdf_{\allparameters}(\allparameters)}{\int_{\parameterset} \pdf_{\calibratedvariate}(\mathbf{\matched} | \mathbf{s}, \allparameters^\ast) \pdf_{\allparameters}(\allparameters^\ast) \diff \allparameters^\ast} ,
\end{align}
using $\mathbf{\matched} = (\matched_1, \ldots, \matched_\numsamples)^\T$ and $\mathbf{s} = (\allcollect_1, \ldots, \allcollect_\numsamples)^\T$, where $\pdf_{\calibratedvariate}(\mathbf{\matched} | \mathbf{s}, \allparameters)$ is the model likelihood of $\calmodel_{\allparameters}$ given the data set $\dataset$.
Thus, the posterior reflects the probability distribution for the model parameters $\allparameters$ given the training data $\dataset$. 
However, since the integral over the complete parameter set $\parameterset$ is intractable, it is not possible to analytically obtain the posterior for $\allparameters$ \cite{Jordan1999,Hoffman2013}. 
Therefore, we adapt \ac{SVI} as an approximation method to infer the posterior \cite{Jordan1999,Hoffman2013,Gal2016a} where a variational distribution $\variationaldist_{\allparameters}(\allparameters | \variationalparameter)$ of known functional form is used with distribution parameters $\variationalparameter$ to approximate the real posterior, so that $\variationaldist_{\allparameters}(\allparameters | \variationalparameter) \approx \pdf_{\allparameters}(\allparameters | \dataset)$.
Note that using variational distributions, possible correlations between the calibration parameters are neglected.
Furthermore, using \ac{MCMC} for approximating the posterior is known to be asymptotically exact \cite{Blei2017}, whereas \ac{SVI} methods are limited by the use of variational distributions.
However, \ac{SVI} comes with computationally low costs and scales well to large data sets compared to \ac{MCMC} methods.
Thus, we further use \ac{SVI} for approximating the posterior using a Gaussian to implement the variational distribution as it has a known functional form and is easy to evaluate.

\begin{figure}[t]
    \centering
    \begin{overpic}[width=1.0\linewidth]{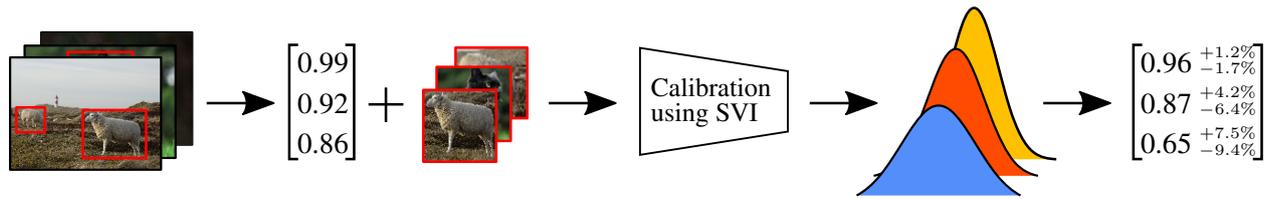}
        \put(23, 10.65){0.99}
        \put(23, 7.4){0.92}
        \put(23, 4.15){0.86}
        \put(51.2, 8.7){\small Calibration}
        \put(51.2, 6.7){\small using SVI}
        \put(90.25, 10.65){0.96 {\raisebox{0.5ex}{\scriptsize$\substack{+1.2\% \\ -1.7\%}$}}}
        \put(90.25, 7.4){0.87 {\raisebox{0.5ex}{\scriptsize$\substack{+4.2\% \\ -6.4\%}$}}}
        \put(90.25, 4.15){0.65 {\raisebox{0.5ex}{\scriptsize$\substack{+7.5\% \\ -9.4\%}$}}}
    \end{overpic}
    \raggedleft{\scriptsize\ieeecopyright{2021}}
    \caption[Concept of Bayesian confidence calibration.]{
        Concept of Bayesian confidence calibration \cite[p.~2, Fig.~2]{Kueppers2021}.
        A detector estimates several objects with a certain confidence and a certain position within an image.
        In Bayesian confidence calibration, the parameters of the calibration function are replaced by variational (Gaussian) distribution and learned by Stochastic Variational Inference (SVI).
        When the calibration method is applied to the detector output, we sample from the variational distributions to obtain multiple parameter sets.
        These parameter sets are then used to generate a sample distribution for the calibration output that represents the uncertainty within the recalibrated confidence.
    }
    \label{fig:bayesian:blockimage}
\end{figure}

The distribution parameters $\variationalparameter$ of $\variationaldist_{\allparameters}(\allparameters | \variationalparameter)$ are obtained using the \ac{ELBO} loss \cite{Jordan1999,Gal2016a}.
The basic idea is that the evidence $\pdf(\dataset)$ is the target distribution to maximize the model likelihood given a certain parameter set $\allparameters$. 
The \ac{ELBO} loss treats the evidence $\pdf(\dataset)$ as an upper bound during the maximization of the model likelihood and is derived by
\begin{align}
	\log\big(\pdf(\dataset)\big) 
	&= \log\Bigg( \int_{\parameterset} \pdf(\dataset, \allparameters) \diff \allparameters \Bigg) \\
	&= \log\Bigg( \int_{\parameterset} \pdf(\dataset, \allparameters) \frac{\variationaldist_{\allparameters}(\allparameters | \variationalparameter)}{\variationaldist_{\allparameters}(\allparameters | \variationalparameter)} \diff \allparameters \Bigg) \\
	&= \log \Bigg( \expectation_{\variationaldist_{\allparameters}} \bigg[ \frac{\pdf_{\allparameters}(\dataset, \allparameters)}{\variationaldist_{\allparameters}(\allparameters | \variationalparameter)} \bigg] \Bigg) \\
	&\geq \expectation_{\variationaldist_{\allparameters}} \Big[ \log\big(\pdf_{\allparameters}(\dataset, \allparameters)\big)\Big] - \expectation_{\variationaldist_{\allparameters}} \Big[\log\big(\variationaldist_{\allparameters}(\allparameters | \variationalparameter)\big)\Big] ,
\end{align}
using Jensen's inequality on the log probability of observations.
Note that the \ac{ELBO} is related to the Kullback-Leibler divergence between the variational distribution $\variationaldist_{\allparameters}(\allparameters | \variationalparameter)$ and the posterior $\pdf_{\allparameters}(\allparameters | \dataset)$ by
\begin{align}
	\label{eq:bayesian:elbo:kldivergence}
	\kldivergence\big(\variationaldist_{\allparameters}(\allparameters | \variationalparameter) || \pdf_{\allparameters}(\allparameters | \dataset) \big) 
	&= \expectation_{\variationaldist_{\allparameters}} \Bigg[\log \bigg( \frac{\variationaldist_{\allparameters}(\allparameters | \variationalparameter)}{\pdf_{\allparameters}(\allparameters | \dataset)}\bigg) \Bigg] \\
	&= \expectation_{\variationaldist_{\allparameters}} \Big[ \log \big( \variationaldist_{\allparameters}(\allparameters | \variationalparameter) \big) \Big] - \expectation_{\variationaldist_{\allparameters}} \Big[\log\big(\pdf_{\allparameters}(\allparameters | \dataset\big) \Big] \\
	&= \expectation_{\variationaldist_{\allparameters}} \Big[ \log \big( \variationaldist_{\allparameters}(\allparameters | \variationalparameter) \big) \Big] - \expectation_{\variationaldist_{\allparameters}} \Big[\log\big(\pdf_{\allparameters}(\dataset, \allparameters\big) \Big] + \log\big(\pdf(\dataset)\big) \\
	&= -\bigg[ \expectation_{\variationaldist_{\allparameters}} \Big[\log\big(\pdf_{\allparameters}(\dataset, \allparameters\big) \Big] - \expectation_{\variationaldist_{\allparameters}} \Big[ \log \big( \variationaldist_{\allparameters}(\allparameters | \variationalparameter) \big) \Big] \bigg] + \log\big(\pdf(\dataset)\big) ,
\end{align}
which is the negative \ac{ELBO} and the log evidence $\pdf(\dataset)$.
Thus, maximizing the \ac{ELBO} leads to a minimization of the Kullback-Leibler divergence in (\ref{eq:bayesian:elbo:kldivergence}).
Similar to the training of neural networks, \ac{SVI} utilizes a gradient descent approach using backpropagation to optimize the variational distribution parameters $\variationalparameter$.
Once the approximate distribution for the posterior has been learned, we can plug in the variational distribution into the posterior predictive in (\ref{eq:bayesian:predictive}) by
\begin{align}
    \label{eq:bayesian:predictive:variational}
    \pdf_{\calibratedvariate}(\calibrated^\ast | \allcollect^\ast, \dataset) &= \int_{\parameterset} \pdf_{\calibratedvariate}(\calibrated^\ast | \allcollect^\ast, \allparameters) \pdf_{\allparameters}(\allparameters | \dataset) \diff \allparameters \\
    &\approx \int_{\parameterset} \pdf_{\calibratedvariate}(\calibrated^\ast | \allcollect^\ast, \allparameters) \variationaldist_{\allparameters}(\allparameters | \variationalparameter) \diff \allparameters ,
\end{align}
which denotes the probability distribution of $\calibratedvariate^\ast$ given the new sample $\allcollect^\ast$ and the training set $\dataset$. 
The distribution for $\calibratedvariate^\ast$ should express the uncertainty for the calibrated probability.
However, for subsequent applications such as object detection (cf. \chapref{chapter:tracking}), it is mandatory to provide a scalar for the calibrated probability to finally construct the Bernoulli $\matchedvariate^\ast$.
Thus, the expectation $\expectation_{\calibratedvariate^\ast}[\calibratedvariate^\ast | \allcollect^\ast, \dataset]$ can be used to assess a mean estimate for the calibrated confidence.
In application, we sample $\numstochastic$ parameter combinations $\allestimatedparameters \in \parameterset$ from the posterior distribution $\allestimatedparameters \sampledfrom \variationaldist_{\allparameters}(\allparameters | \variationalparameter)$ and perform calibration with each parameter set to approximate the expectation and the variance.
We denote the mean by $\mean_{\calibratedvariate}$ that is an approximation of the expectation to obtain the calibrated confidence by
\begin{align}
    \label{eq:bayesian:bernoulli}
    \expectation_{\calibratedvariate^\ast}[\calibratedvariate^\ast | \allcollect^\ast, \dataset] \approx \frac{1}{\numstochastic} \sum^\numstochastic_{\indexstochastic=1} \pdf_{\calibratedvariate}(\calibrated^\ast | \allcollect^\ast, \allestimatedparameters_{\indexstochastic}) = \mean_{\calibratedvariate^\ast} ,
\end{align}
which is used to construct the Bernoulli distribution for $\matchedvariate^\ast$, so that
\begin{align}
    \pdf_{\matchedvariate^\ast}(\matched^\ast | \mean_{\calibratedvariate^\ast}) = \mean_{\calibratedvariate^\ast}^{\matched^\ast} (1-\mean_{\calibratedvariate^\ast})^{1-\matched^\ast} ,
\end{align}
yields the final calibrated confidence and probability distribution for $\matchedvariate^\ast$.
The concept of Bayesian confidence calibration is schematically shown in \figref{fig:bayesian:blockimage}.
Note that this kind of calibration does not necessarily lead to a monotonically increasing calibration mapping any more which might affect the baseline average precision. 
This influence is thus investigated in our experiments for Bayesian confidence calibration.
In general, it is also possible to obtain a variance by sampling from the posterior in the same way.
However, we face some challenges for the uncertainty quantification.
On the one hand, the final probability for $\matchedvariate^\ast$ is expressed in terms of a Bernoulli distribution. 
Furthermore, we do not have a direct ground-truth for $\calibratedvariate^\ast$ available to evaluate the epistemic uncertainty.
We mitigate this problem by using the same approach to quantify the ground-truth information for $\calibratedvariate^\ast$ as within the \ac{D-ECE} calculation.
We apply a binning scheme over the uncalibrated confidence estimates for all samples within a data set $\dataset$ and measure the precision within each bin.
This yields the ground-truth information $\groundtruthconfidencevariate \in [0, 1]$ for all samples in $\dataset$ in each bin which allows for an evaluation of the calibrated uncertainty.
Similarly to the standard \ac{ECE} or \ac{D-ECE} calculation, using a binning scheme leads to a modeling error which, however, we tolerate in our experiments for practical reasons.

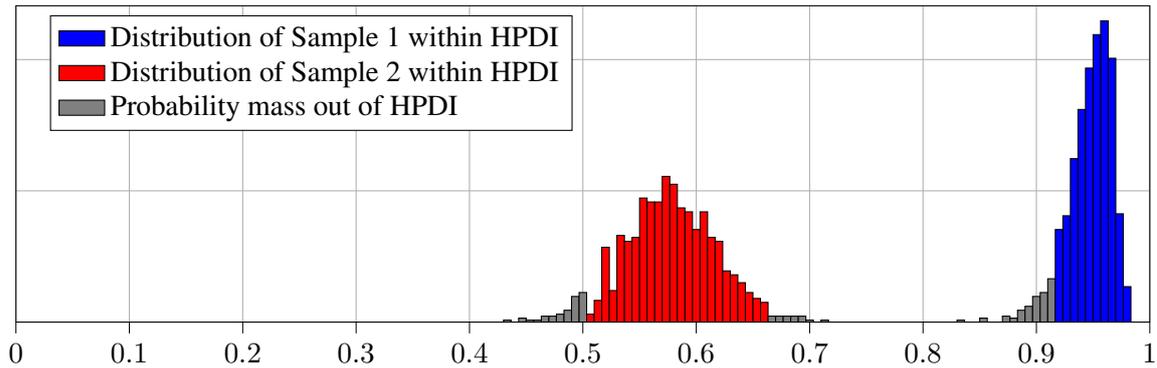
\begin{figure}[t!]
    \centering
\begin{tikzpicture}

\pgfplotsset{
    legend image with text/.style={
        legend image code/.code={%
            \node[anchor=center] at (0.3cm,0cm) {#1};
        }
    },
}

\begin{axis}[
width=\textwidth,
height=0.35\textwidth,
tick align=outside,
tick pos=left,
x grid style={white!69.0196078431373!black},
xmajorgrids,
xmin=0, xmax=1,
xtick style={color=black},
y grid style={white!69.0196078431373!black},
ymajorgrids,
ymin=0, ymax=24.0975229811888,
ytick style={color=black},
ytick={},
yticklabels={},
ytick style={draw=none},
legend style={
    legend cell align=left,
},
legend pos=north west,
]
\addlegendimage{area legend, black, fill=blue}
\addlegendentry{Distribution of Sample 1 within HPDI}

\addlegendimage{area legend, black, fill=red}
\addlegendentry{Distribution of Sample 2 within HPDI}

\addlegendimage{area legend, black, fill=white!50.1960784313725!black}
\addlegendentry{Probability mass out of HPDI}

\draw[draw=black,fill=white!50.1960784313725!black] (axis cs:0.829999983310699,0) rectangle (axis cs:0.836666643619537,0.150000143051284);
\draw[draw=black,fill=white!50.1960784313725!black] (axis cs:0.850000023841858,0) rectangle (axis cs:0.856666684150696,0.300000286102568);
\draw[draw=black,fill=white!50.1960784313725!black] (axis cs:0.870000004768372,0) rectangle (axis cs:0.876666665077209,0.450000429153852);
\draw[draw=black,fill=white!50.1960784313725!black] (axis cs:0.876666665077209,0) rectangle (axis cs:0.883333325386047,0.300000286102568);
\draw[draw=black,fill=white!50.1960784313725!black] (axis cs:0.883333325386047,0) rectangle (axis cs:0.889999985694885,0.900000858307703);
\draw[draw=black,fill=white!50.1960784313725!black] (axis cs:0.889999985694885,0) rectangle (axis cs:0.896666646003723,1.20000114441027);
\draw[draw=black,fill=white!50.1960784313725!black] (axis cs:0.896666646003723,0) rectangle (axis cs:0.903333365917206,1.94998442543071);
\draw[draw=black,fill=white!50.1960784313725!black] (axis cs:0.903333365917206,0) rectangle (axis cs:0.910000026226044,2.25000214576926);
\draw[draw=black,fill=white!50.1960784313725!black] (axis cs:0.910000026226044,0) rectangle (axis cs:0.916666686534882,3.30000314712825);
\draw[draw=black,fill=blue] (axis cs:0.916666686534882,0) rectangle (axis cs:0.923333346843719,7.05000672341034);
\draw[draw=black,fill=blue] (axis cs:0.923333346843719,0) rectangle (axis cs:0.930000007152557,8.10000772476933);
\draw[draw=black,fill=blue] (axis cs:0.930000007152557,0) rectangle (axis cs:0.936666667461395,12.4500118732566);
\draw[draw=black,fill=blue] (axis cs:0.936666667461395,0) rectangle (axis cs:0.943333327770233,16.2000154495387);
\draw[draw=black,fill=blue] (axis cs:0.943333327770233,0) rectangle (axis cs:0.949999988079071,19.3500184536156);
\draw[draw=black,fill=blue] (axis cs:0.949999988079071,0) rectangle (axis cs:0.956666648387909,21.9000208854874);
\draw[draw=black,fill=blue] (axis cs:0.956666648387909,0) rectangle (axis cs:0.963333308696747,22.9500218868464);
\draw[draw=black,fill=blue] (axis cs:0.963333249092102,0) rectangle (axis cs:0.969999969005585,20.099839462132);
\draw[draw=black,fill=blue] (axis cs:0.970000028610229,0) rectangle (axis cs:0.976666688919067,8.25000786782061);
\draw[draw=black,fill=blue] (axis cs:0.976666688919067,0) rectangle (axis cs:0.983333349227905,2.70000257492311);
\draw[draw=black,fill=white!50.1960784313725!black] (axis cs:0.430000007152557,0) rectangle (axis cs:0.436666667461395,0.150000143051284);
\draw[draw=black,fill=white!50.1960784313725!black] (axis cs:0.443333327770233,0) rectangle (axis cs:0.450000017881393,0.299998945001498);
\draw[draw=black,fill=white!50.1960784313725!black] (axis cs:0.450000017881393,0) rectangle (axis cs:0.456666678190231,0.150000143051284);
\draw[draw=black,fill=white!50.1960784313725!black] (axis cs:0.456666678190231,0) rectangle (axis cs:0.463333338499069,0.150000143051284);
\draw[draw=black,fill=white!50.1960784313725!black] (axis cs:0.463333338499069,0) rectangle (axis cs:0.469999998807907,0.450000429153852);
\draw[draw=black,fill=white!50.1960784313725!black] (axis cs:0.469999998807907,0) rectangle (axis cs:0.476666659116745,0.450000429153852);
\draw[draw=black,fill=white!50.1960784313725!black] (axis cs:0.476666629314423,0) rectangle (axis cs:0.483333319425583,0.599997890002995);
\draw[draw=black,fill=white!50.1960784313725!black] (axis cs:0.483333349227905,0) rectangle (axis cs:0.490000009536743,0.900000858307703);
\draw[draw=black,fill=white!50.1960784313725!black] (axis cs:0.490000009536743,0) rectangle (axis cs:0.496666669845581,1.95000185966669);
\draw[draw=black,fill=white!50.1960784313725!black] (axis cs:0.496666669845581,0) rectangle (axis cs:0.503333330154419,2.25000214576926);
\draw[draw=black,fill=red] (axis cs:0.503333330154419,0) rectangle (axis cs:0.509999990463257,0.600000572205136);
\draw[draw=black,fill=red] (axis cs:0.509999990463257,0) rectangle (axis cs:0.516666650772095,1.65000157356412);
\draw[draw=black,fill=red] (axis cs:0.516666650772095,0) rectangle (axis cs:0.523333311080933,5.70000543594879);
\draw[draw=black,fill=red] (axis cs:0.523333311080933,0) rectangle (axis cs:0.530000030994415,2.39998083129934);
\draw[draw=black,fill=red] (axis cs:0.530000030994415,0) rectangle (axis cs:0.536666691303253,6.60000629425649);
\draw[draw=black,fill=red] (axis cs:0.536666691303253,0) rectangle (axis cs:0.543333351612091,6.15000586510264);
\draw[draw=black,fill=red] (axis cs:0.543333351612091,0) rectangle (axis cs:0.550000011920929,6.45000615120521);
\draw[draw=black,fill=red] (axis cs:0.550000011920929,0) rectangle (axis cs:0.556666672229767,9.45000901223088);
\draw[draw=black,fill=red] (axis cs:0.556666672229767,0) rectangle (axis cs:0.563333332538605,9.15000872612832);
\draw[draw=black,fill=red] (axis cs:0.563333332538605,0) rectangle (axis cs:0.569999992847443,9.15000872612832);
\draw[draw=black,fill=red] (axis cs:0.569999992847443,0) rectangle (axis cs:0.576666653156281,11.100010585795);
\draw[draw=black,fill=red] (axis cs:0.576666653156281,0) rectangle (axis cs:0.583333313465118,10.5000100135899);
\draw[draw=black,fill=red] (axis cs:0.583333313465118,0) rectangle (axis cs:0.589999973773956,8.70000829697446);
\draw[draw=black,fill=red] (axis cs:0.589999914169312,0) rectangle (axis cs:0.596666634082794,8.39993290954769);
\draw[draw=black,fill=red] (axis cs:0.596666693687439,0) rectangle (axis cs:0.603333353996277,7.05000672341034);
\draw[draw=black,fill=red] (axis cs:0.603333353996277,0) rectangle (axis cs:0.610000014305115,8.4000080108719);
\draw[draw=black,fill=red] (axis cs:0.610000014305115,0) rectangle (axis cs:0.616666674613953,6.45000615120521);
\draw[draw=black,fill=red] (axis cs:0.616666674613953,0) rectangle (axis cs:0.623333334922791,6.15000586510264);
\draw[draw=black,fill=red] (axis cs:0.623333334922791,0) rectangle (axis cs:0.629999995231628,3.90000371933338);
\draw[draw=black,fill=red] (axis cs:0.629999995231628,0) rectangle (axis cs:0.636666655540466,3.60000343323081);
\draw[draw=black,fill=red] (axis cs:0.636666655540466,0) rectangle (axis cs:0.643333315849304,3.00000286102568);
\draw[draw=black,fill=red] (axis cs:0.643333315849304,0) rectangle (axis cs:0.649999976158142,2.25000214576926);
\draw[draw=black,fill=red] (axis cs:0.649999976158142,0) rectangle (axis cs:0.656666696071625,1.79998562347451);
\draw[draw=black,fill=red] (axis cs:0.656666696071625,0) rectangle (axis cs:0.663333356380463,1.50000143051284);
\draw[draw=black,fill=white!50.1960784313725!black] (axis cs:0.663333356380463,0) rectangle (axis cs:0.670000016689301,0.450000429153852);
\draw[draw=black,fill=white!50.1960784313725!black] (axis cs:0.670000016689301,0) rectangle (axis cs:0.676666676998138,0.450000429153852);
\draw[draw=black,fill=white!50.1960784313725!black] (axis cs:0.676666676998138,0) rectangle (axis cs:0.683333337306976,0.450000429153852);
\draw[draw=black,fill=white!50.1960784313725!black] (axis cs:0.683333337306976,0) rectangle (axis cs:0.689999997615814,0.450000429153852);
\draw[draw=black,fill=white!50.1960784313725!black] (axis cs:0.689999997615814,0) rectangle (axis cs:0.696666657924652,0.450000429153852);
\draw[draw=black,fill=white!50.1960784313725!black] (axis cs:0.696666657924652,0) rectangle (axis cs:0.70333331823349,0.150000143051284);
\draw[draw=black,fill=white!50.1960784313725!black] (axis cs:0.709999918937683,0) rectangle (axis cs:0.716666638851166,0.149998801956209);
\end{axis}

\end{tikzpicture}
    \vspace{0.25em}
    
    \raggedleft{\scriptsize\ieeecopyright{2021}}
    \caption[Qualitative example of two samples after Bayesian confidence calibration with their respective probability distributions for the calibrated confidence.]{
        Qualitative example of two samples after Bayesian confidence calibration with their respective probability distributions for the calibrated confidence \cite[p. 4, Fig. 3]{Kueppers2021}.
        The probability distributions do not necessarily follow a Gaussian distribution.
        Therefore, we seek for the highest posterior density interval (HPDI) to express the epistemic uncertainty of a calibration mapping.
    }
    \label{fig:bayesian:qualitative:distribution}
\end{figure}

On the other hand, the probability distribution for the calibrated confidences might not necessarily follow a distribution of a known parametric form, e.g., a Gaussian.
This is demonstrated in Fig. \ref{fig:bayesian:qualitative:distribution}.
Thus, a sampled variance might not properly reflect the epistemic uncertainty of a Bayesian confidence mapping.
Therefore, we seek to express the epistemic calibration uncertainty in terms of a prediction interval.
A prediction interval is represented by interval boundaries $\boundary_{\quantile}, \boundarysec_{\quantile} \in [0, 1]$ which enclose a certain probability mass $\quantile \in [0, 1]$, so that
\begin{align}
    \int_{\boundary_{\quantile}}^{\boundarysec_{\quantile}} \pdf_{\calibratedvariate^\ast}(\calibrated^\ast | \allcollect^\ast, \dataset) \diff \calibrated^\ast = \quantile .
\end{align}
We further denote the prediction interval as $\credibleinterval_\quantile^\ast = [\hat{\boundary}_{\quantile}^\ast, \hat{\boundarysec}_{\quantile}^\ast]$.
The advantage of using prediction intervals is that these intervals reflect the probability for a new sample located within the interval boundaries.
However, since prediction intervals are not uniquely defined, we consider the \ac{HPDI} to express epistemic uncertainty, which reflects the narrowest credible interval given a certain confidence level $\quantile$.
Using the \ac{HPDI}, we can evaluate the consistency of the predicted uncertainty by comparing the prediction interval with the observed interval coverage.
This is a requirement by the definition for uncertainty calibration \cite{Kuleshov2018} and will be discussed in \chapref{chapter:regression} in more detail.
For this reason, we adapt the \ac{PICP} \cite{Pearce2018} as a metric that measures the fraction of samples whose ground-truth score fall into the estimated prediction interval for a certain confidence level, so that
\begin{align}
    \label{eq:bayesian:picp}
    \picp(\quantile) := \frac{1}{\numsamples} \sum^\numsamples_{\indexsamples=1} \ind(\groundtruthconfidencevariate_\indexsamples \in \credibleinterval_{\quantile, \indexsamples}) .
\end{align}
Furthermore, we use the \ac{MPIW} \cite{Pearce2018} as a complementary metric to evaluate the sharpness of the predicted distribution.
Thus, we finally considered all necessary aspects to determine and evaluate the epistemic uncertainty of a Bayesian calibration method. 


\section{Experiments for Bayesian Confidence Calibration}
\label{section:bayesian:experiments}

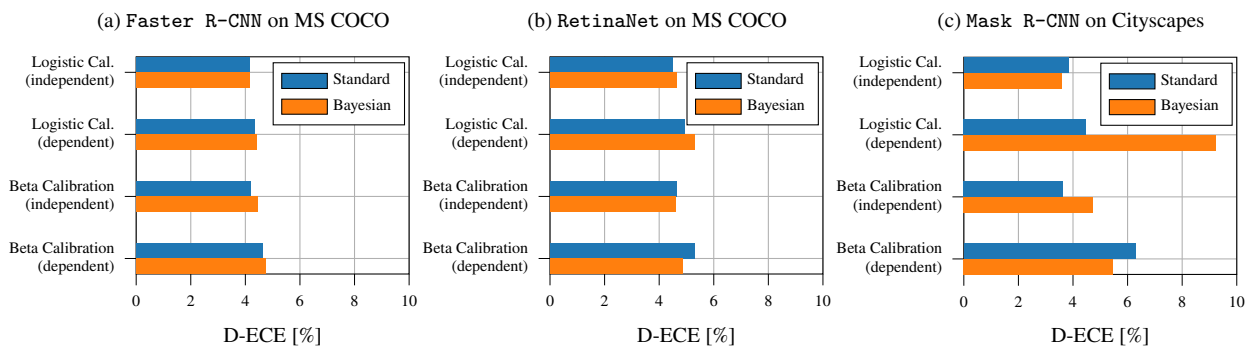
\begin{figure}[b!]
    \centering
    \begin{subfigure}{0.33\textwidth}
\begin{tikzpicture}

\definecolor{color0}{rgb}{0.12156862745098,0.466666666666667,0.705882352941177}
\definecolor{color1}{rgb}{1,0.498039215686275,0.0549019607843137}

\pgfplotsset{every axis/.append style={label style={font=\scriptsize}, tick label style={font=\tiny}}}

\begin{axis}[
width=0.95\linewidth,
legend style={
    legend cell align=left,
    font=\tiny,
    at={(0.5, 0.975)},
    anchor=north west,
},
title={(a) \fasterrcnn{} on MS COCO},
title style={font=\scriptsize, xshift=-1em},
tick align=outside,
tick pos=left,
x grid style={white!69.0196078431373!black},
xmajorgrids,
xmin=0, xmax=0.1,
xtick style={color=black},
xtick={0, 0.02, 0.04, 0.06, 0.08, 0.1},
xticklabels={0, 2, 4, 6, 8, 10},
xlabel={D-ECE [\%]},
y grid style={white!69.0196078431373!black},
ymajorgrids,
ymin=0, ymax=3.5,
ytick style={color=black},
ytick={0.25, 1.25, 2.25, 3.25},
yticklabels={{Beta Calibration\\(dependent)}, {Beta Calibration\\(independent)},  {Logistic Cal.\\(dependent)},  {Logistic Cal.\\(independent)}},
yticklabel style={align=right},
scaled x ticks=false,
]
\addlegendimage{area legend, black, fill=color0}
\addlegendentry{Standard}
\addlegendimage{area legend, black, fill=color1}
\addlegendentry{Bayesian}

\draw[draw=none,fill=color0] (axis cs:0,3.25) rectangle (axis cs:0.04142180201908559,3.5);
\draw[draw=none,fill=color0] (axis cs:0,2.25) rectangle (axis cs:0.04330975297589995,2.5);
\draw[draw=none,fill=color0] (axis cs:0,1.25) rectangle (axis cs:0.04182126686425273,1.5);
\draw[draw=none,fill=color0] (axis cs:0,0.25) rectangle (axis cs:0.04641451496133137,0.5);
\draw[draw=none,fill=color1] (axis cs:0,3) rectangle (axis cs:0.04167867341179454,3.25);
\draw[draw=none,fill=color1] (axis cs:0,2) rectangle (axis cs:0.044239327161890246,2.25);
\draw[draw=none,fill=color1] (axis cs:0,1) rectangle (axis cs:0.04433888818110838,1.25);
\draw[draw=none,fill=color1] (axis cs:0,0) rectangle (axis cs:0.047346005955575685,0.25);

\end{axis}

\end{tikzpicture}
    \end{subfigure}%
    \begin{subfigure}{0.33\textwidth}
\begin{tikzpicture}

\definecolor{color0}{rgb}{0.12156862745098,0.466666666666667,0.705882352941177}
\definecolor{color1}{rgb}{1,0.498039215686275,0.0549019607843137}

\pgfplotsset{every axis/.append style={label style={font=\scriptsize}, tick label style={font=\tiny}}}

\begin{axis}[
width=0.95\linewidth,
legend style={
    legend cell align=left,
    font=\tiny,
    at={(0.5, 0.975)},
    anchor=north west,
},
title={(b) \retinanet{} on MS COCO},
title style={font=\scriptsize, xshift=-1em},
tick align=outside,
tick pos=left,
x grid style={white!69.0196078431373!black},
xmajorgrids,
xmin=0, xmax=0.1,
xtick style={color=black},
xtick={0, 0.02, 0.04, 0.06, 0.08, 0.1},
xticklabels={0, 2, 4, 6, 8, 10},
xlabel={D-ECE [\%]},
y grid style={white!69.0196078431373!black},
ymajorgrids,
ymin=0, ymax=3.5,
ytick style={color=black},
ytick={0.25, 1.25, 2.25, 3.25},
yticklabels={{Beta Calibration\\(dependent)}, {Beta Calibration\\(independent)},  {Logistic Cal.\\(dependent)},  {Logistic Cal.\\(independent)}},
yticklabel style={align=right},
scaled x ticks=false,
]
\addlegendimage{area legend, black, fill=color0}
\addlegendentry{Standard}
\addlegendimage{area legend, black, fill=color1}
\addlegendentry{Bayesian}

\draw[draw=none,fill=color0] (axis cs:0,3.25) rectangle (axis cs:0.045014683272571854,3.5);
\draw[draw=none,fill=color0] (axis cs:0,2.25) rectangle (axis cs:0.04932524882622342,2.5);
\draw[draw=none,fill=color0] (axis cs:0,1.25) rectangle (axis cs:0.04636613143261246,1.5);
\draw[draw=none,fill=color0] (axis cs:0,0.25) rectangle (axis cs:0.0528165711813638,0.5);
\draw[draw=none,fill=color1] (axis cs:0,3) rectangle (axis cs:0.04645845826838001,3.25);
\draw[draw=none,fill=color1] (axis cs:0,2) rectangle (axis cs:0.052809982092132224,2.25);
\draw[draw=none,fill=color1] (axis cs:0,1) rectangle (axis cs:0.04608726347542448,1.25);
\draw[draw=none,fill=color1] (axis cs:0,0) rectangle (axis cs:0.048687983224270086,0.25);

\end{axis}

\end{tikzpicture}
    \end{subfigure}%
    \begin{subfigure}{0.33\textwidth}
\begin{tikzpicture}

\definecolor{color0}{rgb}{0.12156862745098,0.466666666666667,0.705882352941177}
\definecolor{color1}{rgb}{1,0.498039215686275,0.0549019607843137}

\pgfplotsset{every axis/.append style={label style={font=\scriptsize}, tick label style={font=\tiny}}}

\begin{axis}[
width=0.95\linewidth,
legend style={
    legend cell align=left,
    font=\tiny,
    at={(0.5, 0.975)},
    anchor=north west,
},
title={(c) \maskrcnn{} on Cityscapes},
title style={font=\scriptsize, xshift=-1em},
tick align=outside,
tick pos=left,
x grid style={white!69.0196078431373!black},
xmajorgrids,
xmin=0, xmax=0.1,
xtick style={color=black},
xtick={0, 0.02, 0.04, 0.06, 0.08, 0.1},
xticklabels={0, 2, 4, 6, 8, 10},
xlabel={D-ECE [\%]},
y grid style={white!69.0196078431373!black},
ymajorgrids,
ymin=0, ymax=3.5,
ytick style={color=black},
ytick={0.25, 1.25, 2.25, 3.25},
yticklabels={{Beta Calibration\\(dependent)}, {Beta Calibration\\(independent)},  {Logistic Cal.\\(dependent)},  {Logistic Cal.\\(independent)}},
yticklabel style={align=right},
scaled x ticks=false,
]
\addlegendimage{area legend, black, fill=color0}
\addlegendentry{Standard}
\addlegendimage{area legend, black, fill=color1}
\addlegendentry{Bayesian}

\draw[draw=none,fill=color0] (axis cs:0,3.25) rectangle (axis cs:0.03834349838375852,3.5);
\draw[draw=none,fill=color0] (axis cs:0,2.25) rectangle (axis cs:0.04474331977749641,2.5);
\draw[draw=none,fill=color0] (axis cs:0,1.25) rectangle (axis cs:0.036332783330396475,1.5);
\draw[draw=none,fill=color0] (axis cs:0,0.25) rectangle (axis cs:0.06298837930102794,0.5);
\draw[draw=none,fill=color1] (axis cs:0,3) rectangle (axis cs:0.03583091392239447,3.25);
\draw[draw=none,fill=color1] (axis cs:0,2) rectangle (axis cs:0.09234384971931944,2.25);
\draw[draw=none,fill=color1] (axis cs:0,1) rectangle (axis cs:0.04731268902431987,1.25);
\draw[draw=none,fill=color1] (axis cs:0,0) rectangle (axis cs:0.05459545216745486,0.25);

\end{axis}

\end{tikzpicture}
    \end{subfigure}%
    \caption[Comparison of the calibration performance between standard calibration and Bayesian confidence calibration.]{
        Comparison of the calibration performance between standard calibration and Bayesian confidence calibration for the different scaling methods and on different networks and data sets.
        Apart from one exception, we observe a consistent calibration performance of the Bayesian methods compared to their standard counterparts which are built using maximum likelihood estimation.
    }
    \label{fig:bayesian:comparison}
\end{figure}

For the experiments of the Bayesian confidence calibration methods, we use the same setup as within our experiments for object detection evaluation in \secref{section:confidence:experiments:detection} and use a \fasterrcnn{}, \retinanet{}, and \maskrcnn{} on the MS COCO and Cityscapes data sets, respectively.
We further use the \ac{D-ECE}, Brier score, and \ac{NLL} to evaluate the calibration performance of the Bayesian methods.
Furthermore, we also use the \ac{PICP} and \ac{MPIW} \cite{Pearce2018} metrics for a fixed prediction interval of $\quantile = 0.95$ to evaluate the uncertainty (for a detailed description of these metrics, see \secref{section:bayesian:methods}).
For further uncertainty evaluation, we perform inference of each network on a different data set. 
For example, a \fasterrcnn{}, which has been trained for MS COCO, is also used for inference on the Cityscapes validation set.
We reuse the respective calibration methods for each network configuration and do not perform a retraining of the methods on the new data sets.
In this way, we are able to study the effect of a possible covariate shift on the uncertainty of the calibrated confidence.
The evaluation results for the confidence-only case are shown in \tabref{tab:bayesian:evaluation:confidence}, as well as for the position-dependent calibration cases in \tabref{tab:bayesian:evaluation:position:independent} and \tabref{tab:bayesian:evaluation:position:dependent} for conditional independent and conditional dependent calibration, respectively.
Additionally, we compare the Bayesian calibration methods with their deterministic counterparts (cf. \tabref{tab:confidence:detection:evaluation:confidence} and \tabref{tab:confidence:detection:evaluation:confidence}) that is shown in \figref{fig:bayesian:comparison}.

\begin{table}[t!]
    \centering
    \caption[Calibration results for Bayesian confidence calibration where only the predicted confidence is used for calibration and evaluation.]{
        Calibration results for Bayesian confidence calibration where only the predicted confidence $\predconfidencevariate$ is used for calibration and evaluation.
        Moreover, the networks as well as the trained calibration methods are used on a different data set to evaluate the effect of a possible covariate shift on the Bayesian confidence calibration.
        The best scores are highlighted in bold.
        In this case, logistic and beta calibration show equal calibration performance.
        In addition, both calibration methods consistently indicate a higher uncertainty on the data sets for which they have not been trained for.
    }
     \begin{tabular}{l|c|c|l|cccccc}
     \hline
     Network & IoU & \makecell{Evaluation\\data set} & \makecell{Calibration\\method} & D-ECE &        Brier &          NLL &          AUPRC &           PICP &          MPIW \\ \hline \hline
     \multirow{12}{*}{\rotatebox[origin=c]{90}{\makecell{\fasterrcnn{}\\(trained on MS COCO)}}} & \multirow{6}{*}{0.50} & \multirow{3}{*}{MS COCO} & Uncalibrated &                0.153 &                0.176 &              0.536 &             0.920 &               - &               - \\
     & & & Logistic Cal. &                \textbf{0.021} &                \textbf{0.142} &              \textbf{0.433} &             0.920 &               \textbf{0.900} &               0.085 \\
     & & & Beta Cal. &                0.022 &                \textbf{0.142} &              \textbf{0.433} &             0.920 &               0.872 &               0.092 \\ \cline{3-10}
     & & \multirow{3}{*}{Cityscapes} & Uncalibrated &                \textbf{0.093} &                0.168 &              0.552 &             0.887 &               - &               - \\
     & & & Logistic Cal. &                0.115 &                \textbf{0.166} &              \textbf{0.505} &             0.887 &               0.971 &               0.116 \\
     & & & Beta Cal. &                0.117 &                0.167 &              0.509 &             0.887 &               \textbf{0.954} &               0.113 \\ \cline{2-10}
     & \multirow{6}{*}{0.75} & \multirow{3}{*}{MS COCO} & Uncalibrated &                0.294 &                0.257 &              0.829 &             0.866 &               - &               - \\
     & & & Logistic Cal. &                \textbf{0.027} &                \textbf{0.144} &              0.449 &             0.866 &               \textbf{0.922} &               0.085 \\
     & & & Beta Cal. &                \textbf{0.027} &                \textbf{0.144} &              \textbf{0.448} &             0.866 &               \textbf{0.978} &               0.093 \\ \cline{3-10}
     & & \multirow{3}{*}{Cityscapes} & Uncalibrated &                0.296 &                0.275 &              0.917 &             0.814 &               - &               - \\
     & & & Logistic Cal. &                \textbf{0.064} &                0.163 &              0.499 &             0.814 &               \textbf{0.969} &               0.121 \\
     & & & Beta Cal. &                0.066 &                \textbf{0.162} &              \textbf{0.498} &             0.814 &               0.995 &               0.123 \\ \hline \hline
     \multirow{12}{*}{\rotatebox[origin=c]{90}{\makecell{\retinanet{}\\(trained on MS COCO)}}} & \multirow{6}{*}{0.50} & \multirow{3}{*}{MS COCO} & Uncalibrated &                0.083 &                0.157 &              0.478 &             0.907 &               - &               - \\
     & & & Logistic Cal. &                \textbf{0.024} &                \textbf{0.149} &              \textbf{0.451} &             0.907 &               \textbf{0.894} &               0.076 \\
     & & & Beta Cal. &                0.026 &                0.150 &              0.452 &             0.907 &               0.852 &               0.095 \\ \cline{3-10}
     & & \multirow{3}{*}{Cityscapes} & Uncalibrated &                0.131 &                0.174 &              0.522 &             0.879 &               - &               - \\
     & & & Logistic Cal. &                \textbf{0.115} &                \textbf{0.171} &              0.515 &             0.879 &               \textbf{0.941} &               0.094 \\
     & & & Beta Cal. &                0.116 &                \textbf{0.171} &              \textbf{0.513} &             0.879 &               0.887 &               0.116 \\ \cline{2-10}
     & \multirow{6}{*}{0.75} & \multirow{3}{*}{MS COCO} & Uncalibrated &                0.151 &                0.172 &              0.518 &             0.855 &               - &               - \\
     & & & Logistic Cal. &                0.033 &                \textbf{0.140} &              0.439 &             0.855 &               \textbf{0.950} &               0.078 \\
     & & & Beta Cal. &                \textbf{0.027} &                \textbf{0.140} &              \textbf{0.437} &             0.855 &               0.985 &               0.095 \\ \cline{3-10}
     & & \multirow{3}{*}{Cityscapes} & Uncalibrated &                0.149 &                0.190 &              0.565 &             0.789 &               - &               - \\
     & & & Logistic Cal. &                0.063 &                0.161 &              0.495 &             0.789 &               \textbf{0.966} &               0.097 \\
     & & & Beta Cal. &                \textbf{0.058} &                \textbf{0.160} &              \textbf{0.492} &             0.789 &               0.986 &               0.113 \\ \hline \hline
     \multirow{12}{*}{\rotatebox[origin=c]{90}{\makecell{\maskrcnn{}\\(trained on Cityscapes)}}} & \multirow{6}{*}{0.50} & \multirow{3}{*}{MS COCO} & Uncalibrated &                0.108 &                0.145 &              0.496 &             0.952 &               - &               - \\
     & & & Logistic Cal. &                \textbf{0.030} &                \textbf{0.124} &              \textbf{0.378} &             0.952 &               1.000 &               0.166 \\
     & & & Beta Cal. &                0.042 &                0.126 &              0.382 &             0.952 &               0.783 &               0.098 \\ \cline{3-10}
     & & \multirow{3}{*}{MS COCO} & Uncalibrated &                0.274 &                0.283 &              1.084 &             0.799 &               - &               - \\
     & & & Logistic Cal. &                0.137 &                \textbf{0.215} &              \textbf{0.641} &             0.799 &               0.999 &               0.190 \\
     & & & Beta Cal. &                \textbf{0.136} &                0.216 &              0.683 &             0.799 &               \textbf{0.902} &               0.132 \\ \cline{2-10}
     & \multirow{6}{*}{0.75} & \multirow{3}{*}{Cityscapes} & Uncalibrated &                0.296 &                0.269 &              1.055 &             0.896 &               - &               - \\
     & & & Logistic Cal. &                \textbf{0.042} &                \textbf{0.134} &              \textbf{0.420} &             0.896 &               0.997 &               0.203 \\
     & & & Beta Cal. &                0.055 &                0.137 &              0.429 &             0.896 &               \textbf{0.958} &               0.112 \\ \cline{3-10}
     & & \multirow{3}{*}{MS COCO} & Uncalibrated &                0.439 &                0.397 &              1.632 &             0.689 &               - &               - \\
     & & & Logistic Cal. &                \textbf{0.086} &                \textbf{0.184} &              \textbf{0.555} &             0.689 &               1.000 &               0.205 \\
     & & & Beta Cal. &                0.116 &                0.192 &              0.598 &             0.689 &               \textbf{0.986} &               0.133 \\ \hline
 \end{tabular}%
    \label{tab:bayesian:evaluation:confidence}
\end{table}
\begin{table}[t!]
    \centering
    \caption[Calibration results for Bayesian confidence calibration using the conditional \textbf{independent} scaling methods where all available information are used for calibration and evaluation.]{
        Calibration results for Bayesian confidence calibration using the conditional \textbf{independent} scaling methods where all information $\allcollectvariates = (\predconfidencevariate, \predoutputvariate, \allpredbboxvariates)^\T$ are used for calibration and evaluation.
        The underlined scores are the best calibration results compared to their conditional dependent counterparts in \tabref{tab:bayesian:evaluation:position:dependent}.
        For the position-dependent case, we observe an equal performance of both calibration methods.
        Although the calibration methods indicate a higher uncertainty on foreign data, the calibration uncertainty is consistently below the desired confidence level.
    }
     \begin{tabular}{l|c|c|l|cccccc}
    \hline
    Network & IoU & \makecell{Evaluation\\data set} & \makecell{Calibration\\method} & D-ECE &        Brier &          NLL &          AUPRC &           PICP &          MPIW \\ \hline \hline
    \multirow{12}{*}{\rotatebox[origin=c]{90}{\makecell{\fasterrcnn{}\\(trained on MS COCO)}}} & \multirow{6}{*}{0.50} & \multirow{3}{*}{MS COCO} & Uncalibrated &                0.119 &                0.176 &              0.536 &             \textbf{0.920} &               - &               - \\
    & & & Logistic Cal. &                \underline{\textbf{0.042}} &                \underline{\textbf{0.143}} &              \underline{\textbf{0.434}} &             \underline{\textbf{0.920}} &               0.627 &               0.108 \\
    & & & Beta Cal. &                0.044 &                0.144 &              0.439 &             0.918 &               \textbf{0.738} &               0.175 \\ \cline{3-10}
    & & \multirow{3}{*}{Cityscapes} & Uncalibrated &                \textbf{0.086} &                0.168 &              0.552 &             \textbf{0.887} &               - &               - \\
    & & & Logistic Cal. &                0.107 &                0.167 &              0.506 &             \underline{\textbf{0.887}} &               0.724 &               0.134 \\
    & & & Beta Cal. &                0.090 &                \underline{\textbf{0.163}} &              \underline{\textbf{0.499}} &             0.886 &               \textbf{0.800} &               0.231 \\ \cline{2-10}
    & \multirow{6}{*}{0.75} & \multirow{3}{*}{MS COCO} & Uncalibrated &                0.227 &                0.257 &              0.829 &             \textbf{0.866} &               - &               - \\
    & & & Logistic Cal. &                \underline{\textbf{0.045}} &                \underline{\textbf{0.145}} &              \underline{\textbf{0.450}} &             \underline{0.865} &               0.611 &               0.106 \\
    & & & Beta Cal. &                0.048 &                0.146 &              0.453 &             0.863 &               \textbf{0.805} &               0.163 \\ \cline{3-10}
    & & \multirow{3}{*}{Cityscapes} & Uncalibrated &                0.274 &                0.275 &              0.917 &             0.814 &               - &               - \\
    & & & Logistic Cal. &                \underline{\textbf{0.063}} &                \underline{\textbf{0.163}} &              \underline{\textbf{0.499}} &             \underline{0.814} &               0.748 &               0.151 \\
    & & & Beta Cal. &                0.081 &                0.164 &              0.502 &             \underline{0.814} &               \textbf{0.881} &               0.232 \\ \hline \hline
    \multirow{12}{*}{\rotatebox[origin=c]{90}{\makecell{\retinanet{}\\(trained on MS COCO)}}} & \multirow{6}{*}{0.50} & \multirow{3}{*}{MS COCO} & Uncalibrated &                0.072 &                0.157 &              0.478 &             0.907 &               - &               - \\
    & & & Logistic Cal. &                \underline{\textbf{0.046}} &                \underline{\textbf{0.149}} &              \underline{\textbf{0.449}} &             \underline{\textbf{0.909}} &               0.612 &               0.112 \\
    & & & Beta Cal. &                \underline{\textbf{0.046}} &                0.150 &              0.452 &             0.908 &               \textbf{0.710} &               0.182 \\ \cline{3-10}
    & & \multirow{3}{*}{Cityscapes} & Uncalibrated &                0.123 &                0.174 &              0.522 &             0.879 &               - &               - \\
    & & & Logistic Cal. &                \textbf{0.109} &                \textbf{0.172} &              \textbf{0.519} &             0.880 &               0.657 &               0.122 \\
    & & & Beta Cal. &                0.114 &                0.179 &              0.546 &             \underline{\textbf{0.881}} &               \textbf{0.792} &               0.252 \\ \cline{2-10}
    & \multirow{6}{*}{0.75} & \multirow{3}{*}{MS COCO} & Uncalibrated &                0.110 &                0.172 &              0.518 &             0.855 &               - &               - \\
    & & & Logistic Cal. &                \underline{\textbf{0.046}} &                \underline{\textbf{0.139}} &              \underline{\textbf{0.437}} &             \underline{\textbf{0.856}} &               0.660 &               0.101 \\
    & & & Beta Cal. &                \underline{\textbf{0.046}} &                \underline{\textbf{0.139}} &              0.439 &             0.855 &               \textbf{0.810} &               0.165 \\ \cline{3-10}
    & & \multirow{3}{*}{Cityscapes} & Uncalibrated &                0.136 &                0.190 &              0.565 &             0.789 &               - &               - \\
    & & & Logistic Cal. &                \textbf{0.067} &                \underline{\textbf{0.161}} &              \textbf{0.498} &             0.790 &               0.703 &               0.121 \\
    & & & Beta Cal. &                0.086 &                0.164 &              0.509 &             \textbf{0.791} &               \textbf{0.870} &               0.232 \\ \hline \hline
    \multirow{12}{*}{\rotatebox[origin=c]{90}{\makecell{\maskrcnn{}\\(trained on Cityscapes)}}} & \multirow{6}{*}{0.50} & \multirow{3}{*}{Cityscapes} & Uncalibrated &                0.102 &                0.145 &              0.496 &             0.952 &               - &               - \\
    & & & Logistic Cal. &                \underline{\textbf{0.036}} &                \underline{\textbf{0.124}} &              \underline{\textbf{0.378}} &             \underline{0.952} &               \underline{\textbf{0.910}} &               0.185 \\
    & & & Beta Cal. &                0.047 &                \underline{\textbf{0.124}} &              0.380 &             \underline{0.952} &               0.766 &               0.265 \\ \cline{3-10}
    & & \multirow{3}{*}{MS COCO} & Uncalibrated &                0.234 &                0.283 &              1.084 &             \textbf{0.799} &               - &               - \\
    & & & Logistic Cal. &                \underline{\textbf{0.142}} &                \underline{\textbf{0.218}} &              \underline{\textbf{0.651}} &             \underline{0.784} &               \textbf{0.926} &               0.278 \\
    & & & Beta Cal. &                0.162 &                0.243 &              0.767 &             0.760 &               0.812 &               0.354 \\ \cline{2-10}
    & \multirow{6}{*}{0.75} & \multirow{3}{*}{Cityscapes} & Uncalibrated &                0.281 &                0.269 &              1.055 &             0.896 &               - &               - \\
    & & & Logistic Cal. &                \underline{\textbf{0.056}} &                0.136 &              \underline{\textbf{0.423}} &             0.898 &               \textbf{0.969} &               0.227 \\
    & & & Beta Cal. &                0.061 &                \underline{\textbf{0.135}} &              0.427 &             \underline{\textbf{0.899}} &               0.881 &               0.253 \\ \cline{3-10}
    & & \multirow{3}{*}{MS COCO} & Uncalibrated &                0.361 &                0.397 &              1.632 &             0.689 &               - &               - \\
    & & & Logistic Cal. &                \textbf{0.118} &                \underline{\textbf{0.190}} &              \underline{\textbf{0.572}} &             \underline{\textbf{0.703}} &               \textbf{0.983} &               0.299 \\
    & & & Beta Cal. &                0.134 &                0.209 &              0.658 &             0.666 &               0.888 &               0.336 \\ \hline
\end{tabular}%
    \label{tab:bayesian:evaluation:position:independent}
\end{table} 
\begin{table}[t!]
    \centering
    \caption[Calibration results for Bayesian confidence calibration using the conditional \textbf{dependent} calibration methods where all available information are used for calibration and evaluation.]{
        Calibration results for Bayesian confidence calibration using the conditional \textbf{dependent} calibration methods where all information $\allcollectvariates = (\predconfidencevariate, \predoutputvariate, \allpredbboxvariates)^\T$ are used for calibration and evaluation.
        The underlined scores are the best calibration results of compared to their conditional independent counterparts in \tabref{tab:bayesian:evaluation:position:independent}.
        The extended methods show similar calibration performance and further provide a meaningful uncertainty with a prediction interval coverage probability which is consistently close to the desired confidence level.
    }
    \begin{tabular}{l|c|c|l|cccccc}
    \hline
    Network & IoU & \makecell{Evaluation\\data set} & \makecell{Calibration\\method} & D-ECE &        Brier &          NLL &          AUPRC &           PICP &          MPIW \\ \hline \hline
    \multirow{12}{*}{\rotatebox[origin=c]{90}{\makecell{\fasterrcnn{}\\(trained on MS COCO)}}} & \multirow{6}{*}{0.50} & \multirow{3}{*}{MS COCO} & Uncalibrated & 0.119 & 0.176 & 0.536 & \textbf{0.920} & - & - \\
    & & & Logistic Cal. & \textbf{0.044} & \textbf{0.144} & \textbf{0.439} & 0.919 &0 \underline{\textbf{0.885}} & 0.284 \\
    & & & Beta Cal. & 0.047 & 0.145 & 0.445 & 0.916 & 0.837 & 0.263 \\ \cline{3-10}
    & & \multirow{3}{*}{Cityscapes} & Uncalibrated  & \textbf{0.086} & 0.168 & 0.552 & \textbf{0.887} & - & - \\
    & & & Logistic Cal. & 0.117 & 0.170 & 0.528 & 0.885 & 0.927 & 0.400 \\
    & & & Beta Cal. & \underline{0.105} & \textbf{0.167} & \textbf{0.519} & 0.885 & \underline{\textbf{0.928}} & 0.369 \\ \cline{2-10}
    & \multirow{6}{*}{0.75} & \multirow{3}{*}{MS COCO} & Uncalibrated & 0.227 & 0.257 & 0.829 & \textbf{0.866} & - & - \\
    & & & Logistic Cal. & \textbf{0.047} & \textbf{0.146} & \textbf{0.457} & 0.862 & 0.990 & 0.300 \\
    & & & Beta Cal. & \textbf{0.047} & 0.147 & 0.458 & 0.862 & \underline{\textbf{0.942}} & 0.262 \\ \cline{3-10}
    & & \multirow{3}{*}{Cityscapes} & Uncalibrated  & 0.274 & 0.275 & 0.917 & \underline{\textbf{0.814}} & - & - \\ 
    & & & Logistic Cal. & 0.077 & \textbf{0.165} & 0.508 & \textbf{0.814} & 0.991 & 0.464 \\
    & & & Beta Cal. & \textbf{0.068} & \textbf{0.165} & \textbf{0.507} & 0.806 & \textbf{0.986} & 0.415 \\ \hline \hline
    \multirow{12}{*}{\rotatebox[origin=c]{90}{\makecell{\retinanet{}\\(trained on MS COCO)}}} & \multirow{6}{*}{0.50} & \multirow{3}{*}{MS COCO} & Uncalibrated & 0.072 & 0.157 & 0.478 & 0.907 & - & - \\
    & & & Logistic Cal. & 0.053 & \textbf{0.151} & \textbf{0.456} & \textbf{0.908} & \underline{\textbf{0.878}} & 0.333 \\
    & & & Beta Cal. & \textbf{0.049} & \textbf{0.151} & \textbf{0.456} & 0.906 & 0.802 & 0.292 \\ \cline{3-10}
    & & \multirow{3}{*}{Cityscapes} & Uncalibrated  & 0.123 & 0.174 & 0.522 & 0.879 & - & - \\
    & & & Logistic Cal. & 0.118 & 0.171 & 0.527 & 0.877 & \underline{\textbf{0.962}} & 0.463 \\
    & & & Beta Cal. & \underline{\textbf{0.088}} & \underline{\textbf{0.167}} & \underline{\textbf{0.516}} & \underline{\textbf{0.881}} & 0.824 & 0.336 \\ \cline{2-10}
    & \multirow{6}{*}{0.75} & \multirow{3}{*}{MS COCO} & Uncalibrated & 0.110 & 0.172 & 0.518 & 0.855 & - & - \\
    & & & Logistic Cal. & 0.048 & \textbf{0.140} & \textbf{0.439} & \underline{\textbf{0.856}} & \underline{\textbf{0.978}} & 0.318 \\
    & & & Beta Cal. & \textbf{0.047} & 0.141 & 0.441 & 0.853 & 0.906 & 0.265 \\ \cline{3-10}
    & & \multirow{3}{*}{Cityscapes} & Uncalibrated  & 0.136 & 0.190 & 0.565 & 0.789 & - & - \\
    & & & Logistic Cal. & \underline{\textbf{0.066}} & \underline{\textbf{0.161}} & \underline{\textbf{0.497}} & 0.784 & 0.995 & 0.477 \\
    & & & Beta Cal. & 0.080 & 0.163 & 0.506 & \underline{\textbf{0.792}} & \underline{\textbf{0.968}} & 0.361 \\ \hline \hline
    \multirow{12}{*}{\rotatebox[origin=c]{90}{\makecell{\maskrcnn{}\\(trained on Cityscapes)}}} & \multirow{6}{*}{0.50} & \multirow{3}{*}{Cityscapes} & Uncalibrated & 0.102 & 0.145 & 0.496 &\textbf{0.952} &- &- \\
    & & & Logistic Cal. & 0.092 & 0.144 & 0.472 &0.935 &0.655 &0.365 \\
    & & & Beta Cal. & \textbf{0.055} & \textbf{0.131} & \textbf{0.395} &0.949 &\textbf{0.776} &0.475 \\ \cline{3-10}
    & & \multirow{3}{*}{MS COCO} & Uncalibrated  & 0.234 & 0.283 & 1.084 & \textbf{0.799} & - & - \\
    & & & Logistic Cal. & \textbf{0.155} & \textbf{0.241} & 0.815 & 0.751 & 0.862 & 0.476 \\
    & & & Beta Cal. & 0.177 & 0.242 & \textbf{0.743} & 0.773 & \textbf{0.885} & 0.588 \\ \cline{2-10}
    & \multirow{6}{*}{0.75} & \multirow{3}{*}{Cityscapes} & Uncalibrated  & 0.281 & 0.269 & 1.055 &\textbf{0.896} &- &- \\
    & & & Logistic Cal. & 0.081 & 0.150 & 0.480 &0.889 &0.809 &0.574 \\
    & & & Beta Cal. & \textbf{0.067} & \textbf{0.140} & \textbf{0.436} &0.895 &\underline{\textbf{0.945}} &0.536 \\ \cline{3-10}
    & & \multirow{3}{*}{MS COCO} & Uncalibrated & 0.361 & 0.397 & 1.632 & 0.689 & - & - \\
    & & & Logistic Cal. & \underline{\textbf{0.101}} & \textbf{0.193} & \textbf{0.598} & \textbf{0.685} & \underline{\textbf{0.947}} & 0.665 \\
    & & & Beta Cal. & 0.140 & 0.213 & 0.648 & \textbf{0.675} & 0.969 & 0.556 \\ \hline
\end{tabular}%

    \label{tab:bayesian:evaluation:position:dependent}
\end{table}

\begin{figure}[t!]
    \centering
    \begin{subfigure}[b]{0.46\textwidth}
\begin{tikzpicture}

\pgfplotsset{every axis/.append style={label style={font=\scriptsize}}}

\begin{axis}[
width=\linewidth,
height=0.8\linewidth,
tick align=outside,
tick pos=left,
x grid style={white!69.0196078431373!black},
xlabel={relative cx},
xmin=0, xmax=1,
xtick style={color=black},
y grid style={white!69.0196078431373!black},
ylabel={relative cy},
ymin=0, ymax=1,
ticks=none,
axis line style={draw=none}
]
\addplot graphics [includegraphics cmd=\pgfimage,xmin=0, xmax=1, ymin=0, ymax=1] {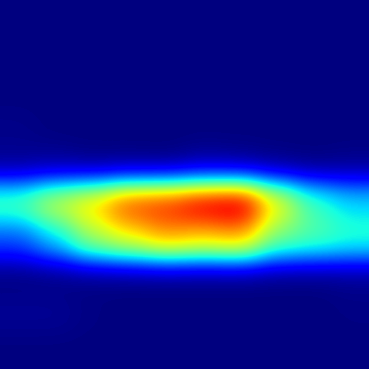};
\end{axis}

\end{tikzpicture}
        \subcaption{Data distribution for the $\centerx$ and $\centery$ position of objects for Cityscapes.}
    \end{subfigure}%
    \hspace{1em}%
    \begin{subfigure}[b]{0.46\textwidth}
        \input{images/bayesian/covariate_shift/lr_dependent_cls-00_iou0.50_2d_cx_cy-00.tex}
        \subcaption{Calibration error w.r.t. the epistemic uncertainty of single predictions for Cityscapes.}
    \end{subfigure}
    \vspace{1em}
    \hrule
    \vspace{1em}
    
    \begin{subfigure}[b]{0.46\textwidth}
\begin{tikzpicture}

\pgfplotsset{every axis/.append style={label style={font=\scriptsize}}}

\begin{axis}[
width=\linewidth,
height=0.8\linewidth,
tick align=outside,
tick pos=left,
x grid style={white!69.0196078431373!black},
xlabel={relative cx},
xmin=0, xmax=1,
xtick style={color=black},
y grid style={white!69.0196078431373!black},
ylabel={relative cy},
ymin=0, ymax=1,
ticks=none,
axis line style={draw=none}
]
\addplot graphics [includegraphics cmd=\pgfimage,xmin=0, xmax=1, ymin=0, ymax=1] {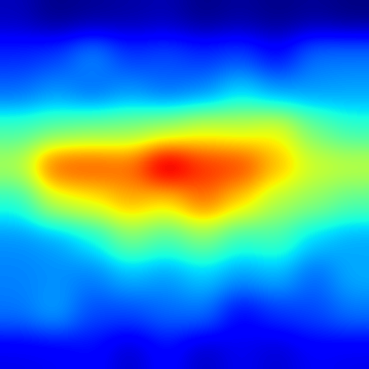};
\end{axis}

\end{tikzpicture}
        \subcaption{Data distribution for the $\centerx$ and $\centery$ position of objects for MS COCO.}
    \end{subfigure}%
    \hspace{1em}%
    \begin{subfigure}[b]{0.46\textwidth}
        \input{images/bayesian/covariate_shift/lr_dependent_cls-00_iou0.50_2d_cx_cy-01.tex}
        \subcaption{Calibration error w.r.t. the epistemic uncertainty of single predictions for MS COCO.}
    \end{subfigure}
    \vspace{0.25em}
    
    \raggedleft{\scriptsize\ieeecopyright{2021}}
    %
    \caption[Data distribution of the $\centerx$ and $\centery$ and the correlation between epistemic calibration uncertainty and calibration error of a logistic calibration model for a \maskrcnn{} on different data sets.]{
        Data distribution of the $\centerx$ and $\centery$ position and the correlation between epistemic calibration uncertainty and calibration error of a logistic calibration model for a \maskrcnn{} on different data sets. \cite[p. 5, Fig. 4]{Kueppers2021}.
        The \maskrcnn{} as well as the confidence calibration have been trained on the Cityscapes data set.
        In the top row (b), we evaluate the correlation between epistemic uncertainty obtained by Bayesian calibration and the calibration error itself for the Cityscapes data set.
        The orange-dotted lines denote the \{25, 50, 75\} percentiles of the samples.
        We repeat this for the MS COCO data set (d) to evaluate a possible change in uncertainty.
        We can inspect that the calibration error increases for a large epistemic uncertainty in both cases.
        In addition, the average epistemic uncertainty increases for the MS COCO data set as indicated by the percentiles.
    }
    \label{fig:bayesian:distribution}
\end{figure}

As already stated, Bayesian confidence calibration does not guarantee a monotonically increasing calibration mapping and thus might affect the baseline average precision.
However, we only find a marginal effect of the Bayesian confidence calibration on the average precision which, thus, does not degenerate baseline performance.
The Bayesian calibration methods (confidence-only and position-dependent) consistently re\-duce miscalibration and show a similar calibration performance compared to the standard methods which are trained by simple maximum likelihood estimation.
This also holds for the miscalibration on the foreign data sets in most cases.
Moreover, we observe a higher epistemic uncertainty when calibration is applied on the data sets for which the calibration methods have not been trained for (see MPIW scores).
In our experiments, the confidence-only methods provide consistent uncertainty quantifications whose prediction interval coverage probability (PICP) is close to the desired confidence level of $\quantile=0.95$.
We further inspect the position-dependent calibration and compare the conditionally independent methods with their conditionally dependent counterparts.
Although the conditionally independent calibration methods provide a better calibration mapping in general, they underestimate the epistemic uncertainty in most cases.
In contrast, the conditionally dependent calibration methods consistently provide good uncertainty estimates which are close to the desired confidence level.
This also holds for calibration on the foreign data sets.
For further uncertainty evaluation, a visual example is given in \figref{fig:bayesian:distribution} which demonstrates the effect of Bayesian confidence calibration on foreign data sets.
In this example, the Bayesian calibration method was trained on the Cityscapes data set where most objects are located in the image center.
If we compare the epistemic uncertainty with the miscalibration for each sample individually, we can observe an increasing miscalibration for increasing epistemic uncertainty.
This is the desired effect for our Bayesian confidence calibration methods.
In the next step, this trained calibration mapping is used for calibration on the MS COCO data set.
We observe a wider distribution for the location of objects within the data set compared to Cityscapes.
On the one hand, we also observe an increasing miscalibration for increasing epistemic uncertainty.
On the other hand, the epistemic uncertainty is higher on average compared to the calibration results for the Cityscapes data set.
This is visualized by the orange-dotted lines which indicate the \{25, 50, 75\} percentiles of the samples.
In both scenarios, we observe an increasing calibration error for samples with a prediction interval width above $0.5$.
This indicates that a reliable prediction for the calibrated confidence of such samples is a challenging task.
One way to handle this phenomenon is to place a threshold for the prediction interval width to detect critical samples during the inference.
The high variation of the calibration error for uncertain samples might also be reduced by using more training data.
Therefore, we conclude that the epistemic uncertainty is a valuable indicator for a possibly higher miscalibration as well as a sufficient criterion for a possible covariate shift of the data during inference.

\section{Conclusion for Bayesian Confidence Calibration} 
\label{section:bayesian:conclusion}

A reliable uncertainty assessment is crucial especially for safety-critical applications.
A step towards more safety is the usage of appropriate calibration methods to get a more interpretable assessment for the probability of being correct.
However, confidence calibration might also be misleading in situations that are unknown either to the baseline object detector or to the calibration mapping itself.
Thus, we are interested in the epistemic model uncertainty of such a calibration mapping to be able to indicate a possibly high model uncertainty in critical situations.
For this reason, we derive the term of Bayesian confidence calibration that transfers the idea of epistemic uncertainty modeling in Bayesian statistics to the task of confidence calibration.
In this context, we interpret the calibrated confidence as a random variable whose probability distribution is determined by probabilistic calibration methods.
A prior distribution is placed over the calibration weights which allows for a probabilistic computation of the calibrated confidences.
In this way, it is possible to quantify the (epistemic) model uncertainty of a calibration mapping itself which might be used to assess the reliability in calibrated confidence estimates.

The investigations for the Bayesian confidence calibration models show that it is possible to yield a quali\-tatively good calibration mapping by using stochastic variational inference for calibration training.
On the one hand, the Bayesian calibration methods provide a meaningful uncertainty quantification for the epistemic uncertainty inherent in the calibration mapping itself.
Especially the conditionally dependent calibration methods provide good uncertainty estimates for the epistemic calibration uncertainty.
On the other hand, we find a connection between epistemic uncertainty and miscalibration which might be used as a sufficient criterion to detect a possible covariate shift within the calibration mapping.
This allows for an indication of possibly unknown out-of-distribution samples during inference.
However, for samples with a higher epistemic uncertainty, we also observe an increasing calibration error above a certain threshold.
Thus, a reliable estimation of the confidence is a challenging task for these samples which needs further investigations.
Nevertheless, we conclude that Bayesian confidence calibration is a valuable contribution for safety-critical applications as it allows for an additional indication of epistemic uncertainty without losing calibration performance.
Finally, it might be used as an additional component for a possible out-of-distribution recognition.
The assessment of the epistemic uncertainty for the object existence might also be used within an object tracking framework using a particle filter for the object confidence estimation.
This is subject of future work.

\acresetall
\chapter{Spatial Uncertainty Calibration}
\label{chapter:regression}

\begin{figure}[b!]
    \centering
\begin{tikzpicture}

\begin{axis}[
tick pos=left,
title={(a) Uncalibrated},
height=0.4\textwidth,
width=0.4\textwidth,
xmin=-0.5, xmax=611.5,
y dir=reverse,
ymin=-0.5, ymax=611.5,
axis lines=none,
]
\addplot graphics [includegraphics cmd=\pgfimage,xmin=-0.5, xmax=611.5, ymin=611.5, ymax=-0.5] {};
\addplot graphics [includegraphics cmd=\pgfimage,xmin=-0.5, xmax=611.5, ymin=611.5, ymax=-0.5] {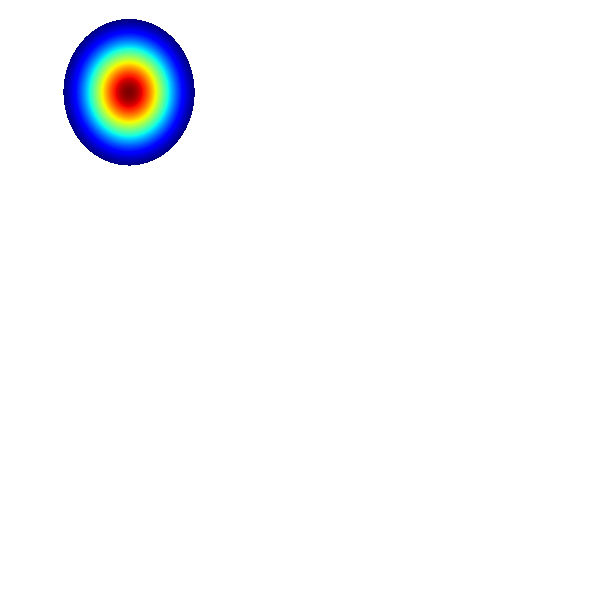};
\addplot graphics [includegraphics cmd=\pgfimage,xmin=-0.5, xmax=611.5, ymin=611.5, ymax=-0.5] {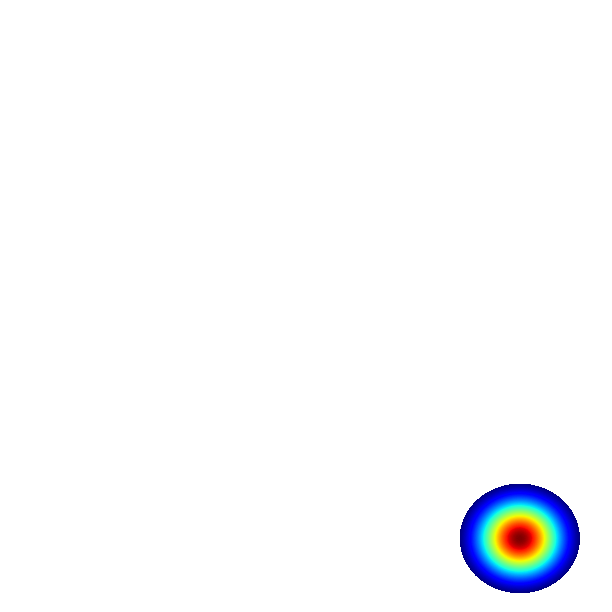};
\end{axis}

\end{tikzpicture}
\begin{tikzpicture}

\begin{axis}[
tick pos=left,
title={(b) GP-Beta},
height=0.4\textwidth,
width=0.4\textwidth,
xmin=-0.5, xmax=611.5,
y dir=reverse,
ymin=-0.5, ymax=611.5,
axis lines=none,
]
\addplot graphics [includegraphics cmd=\pgfimage,xmin=-0.5, xmax=611.5, ymin=611.5, ymax=-0.5] {};
\addplot graphics [includegraphics cmd=\pgfimage,xmin=-0.5, xmax=611.5, ymin=611.5, ymax=-0.5] {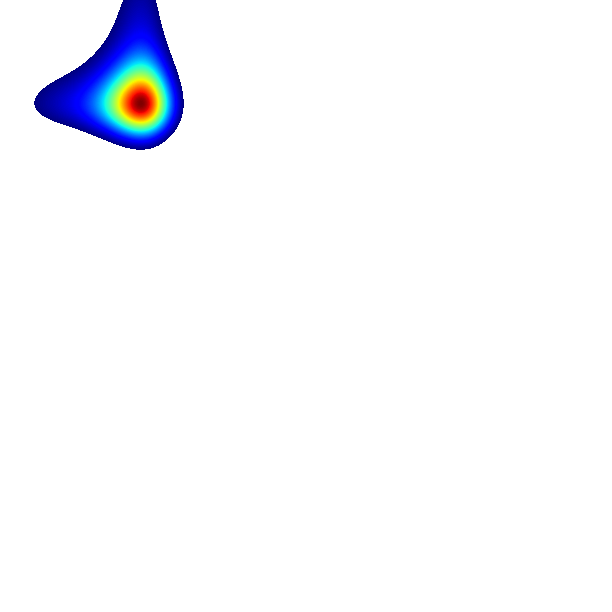};
\addplot graphics [includegraphics cmd=\pgfimage,xmin=-0.5, xmax=611.5, ymin=611.5, ymax=-0.5] {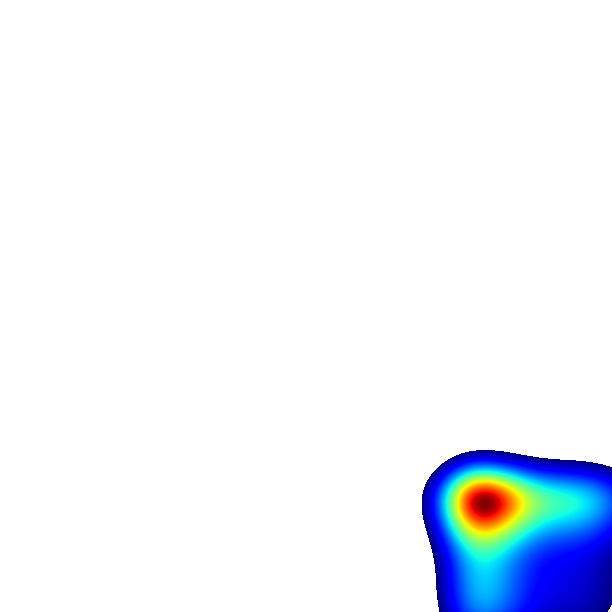};
\end{axis}

\end{tikzpicture}
\begin{tikzpicture}

\begin{axis}[
tick pos=left,
title={(c) GP-Normal (mv.)},
height=0.4\textwidth,
width=0.4\textwidth,
xmin=-0.5, xmax=611.5,
y dir=reverse,
ymin=-0.5, ymax=611.5,
axis lines=none,
]
\addplot graphics [includegraphics cmd=\pgfimage,xmin=-0.5, xmax=611.5, ymin=611.5, ymax=-0.5] {};
\addplot graphics [includegraphics cmd=\pgfimage,xmin=-0.5, xmax=611.5, ymin=611.5, ymax=-0.5] {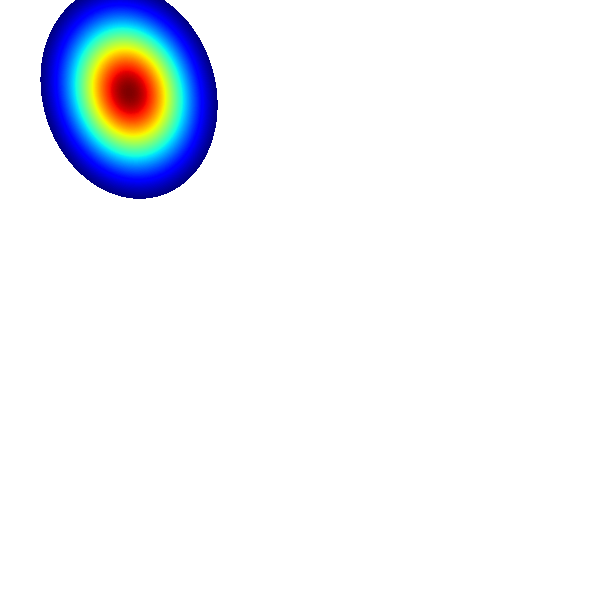};
\addplot graphics [includegraphics cmd=\pgfimage,xmin=-0.5, xmax=611.5, ymin=611.5, ymax=-0.5] {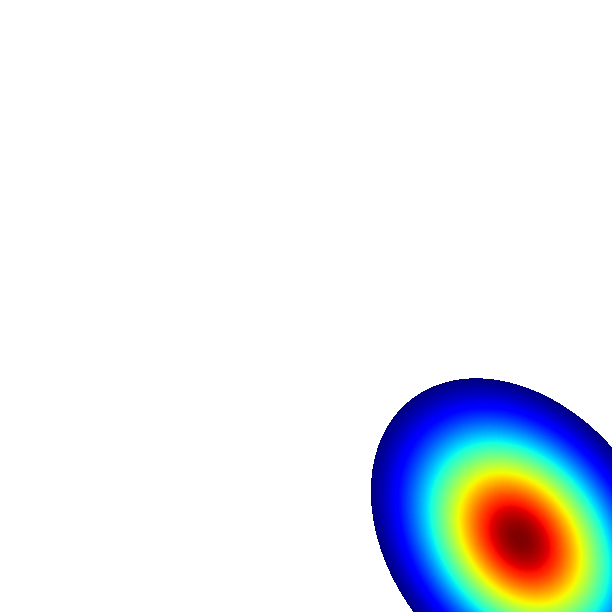};
\end{axis}

\end{tikzpicture}
    \caption[Qualitative example for spatial uncertainty calibration on predictions of a probabilistic \retinanet{} on the MS COCO data set.]{
        Qualitative example for spatial uncertainty calibration on predictions of a probabilistic \retinanet{} (cf. \secref{section:tracking:experiments}) on the MS COCO data set \cite[p.~2, Fig.~1]{Kueppers2022b}.
        (a) The probabilistic detection model outputs normal distribution for the position, width, and height information that are modeled as independent Gaussians.
        (b) The GP-Beta \cite{Song2019} is a non-parametric calibration method that is able to estimate a calibrated probability distribution of arbitrary shape.
        (c) In contrast, our multivariate GP-Normal (cf. \secref{section:regression:methods:correlations}) is a parametric method yielding a multivariate Gaussian as calibration output.
        Thus, this method is able to represent possible correlations between the dimensions.
    }
    \label{fig:regression:qualitative}
\end{figure}
For environment perception, it is necessary to identify not only the semantic label of an object but also its position.
Thus, in the scope of object detection, a neural network needs to infer the class as well as the position and shape of individual objects within an image.
In this chapter, we focus on the regression branch of an object detection model.
While the classification head commonly outputs a score indicating the model's belief about the predicted class (cf. \chapref{chapter:confidence}), the regression branch is usually designed to output the position/shape information without any uncertainty information.
However, a detection model can also be trained so that it outputs probabilistic estimates for the position information using Gaussian distributions \cite{He2019,Hall2020,Harakeh2020,Feng2021}.
This has already been described in \secref{section:introduction:spatial}.
On the one hand, this additional uncertainty can be interpreted as a self-assessment of the model's belief about the predicted position/shape information.
On the other hand, the uncertainty can also be used for subsequent processes such as Kalman filtering for object tracking.
Similar to semantic confidence calibration, the predicted variance can be interpreted as an estimate of the aleatoric uncertainty which is inherent in the model input \cite{Kendall2017}.
We can evaluate if the predicted uncertainty matches the observed prediction error.
This is of major importance especially if such a detection model is used in the context safety-critical applications such as autonomous driving.
Similar to the methods for semantic confidence calibration, it is possible to examine the probabilistic forecaster regarding a systemic bias in its uncertainty estimations for the bounding box position.
If we detect such a deviation, we can apply post-hoc calibration methods to perform a recalibration of the spatial uncertainty.
The concept of uncertainty calibration is qualitatively shown in \figref{fig:regression:qualitative}.

\begin{figure}[t!]
    \centering
    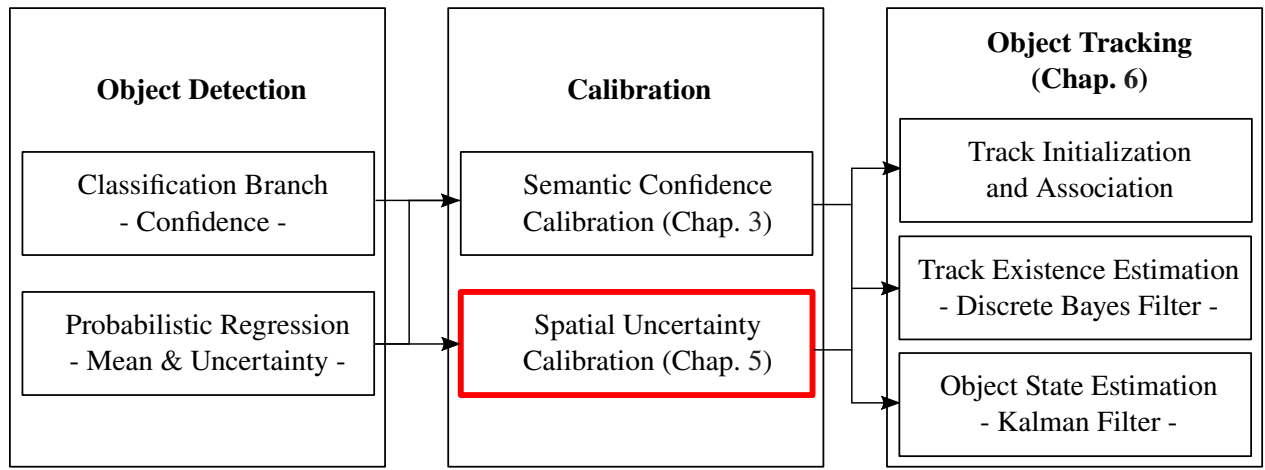%
    \caption[In this chapter, we focus on the evaluation and calibration of spatial uncertainty which is a crucial part of the image-based environment perception process.]{
        In this chapter, we focus on the evaluation and calibration of spatial uncertainty which is a crucial part of the image-based environment perception process.
        The recalibrated spatial uncertainty might be used for subsequent applications, e.g., object tracking (cf. \chapref{chapter:tracking}).
        Therefore, a reliable assessment of spatial uncertainty is mandatory.
    }
    \label{fig:regression:blockimage_thesis}
\end{figure}

Reliable uncertainty evaluation is crucial especially if subsequent processes depend on these estimates, e.g., object tracking.
Thus, we interpret the assessment as well as the recalibration of spatial uncertainty as part of a superordinate process chain that could be used e.g. for image-based environment perception in a vehicle.
The concept of uncertainty calibration as a part of this process chain is schematically shown in \figref{fig:regression:blockimage_thesis}.
We start by reviewing the state-of-the-art for regression uncertainty calibration in \secref{section:regression:related_work}.
Furthermore, we review the existing definitions for regression uncertainty calibration and set them into a common mathematical context to each other.
On the basis of these definitions, we describe the respective calibration metrics and derive new metrics to measure multivariate miscalibration in \secref{section:regression:definition}.
In addition to these metrics, recent work has developed several post-hoc methods
for the calibration of regression uncertainty.
These techniques can be distinguished into parametric and non-parametric calibration methods.
We review these methods and provide detailed mathematical descriptions in \secref{section:regression:methods}.
Furthermore, we extend these methods in \secref{section:regression:methods:parametric:gp} to construct a new calibration method GP-Normal for a flexible and parametric recalibration scheme.
These methods are evaluated on different data sets and for different object detection models in \secref{section:regression:experiments}.
Finally, we give a conclusion about our findings in \secref{section:regression:conclusion}.

\textbf{Contributions:} In summary, the following contributions can be found in this section::
\begin{itemize}
    \item Common mathematical context for the definitions of regression uncertainty calibration.
    \item Derivation of the M-QCE and C-QCE metrics to measure multivariate regression calibration.
    \item New methods GP-Normal and GP-Cauchy for parametric recalibration using Gaussian processes.
    \item Joint multivariate calibration of multiple dimensions.
    \item Covariance estimation \& recalibration for multivariate regression tasks.
    \item Extensive studies on the effect of uncertainty recalibration for probabilistic object detectors.
\end{itemize}

\section{Related Work in the Context of Spatial Uncertainty Calibration} 
\label{section:regression:related_work}

In the scope of regression uncertainty, recent work has proposed several definitions to define the term of calibration.
The authors in \cite{Kuleshov2018} define the term of \quantilecalibration{} which requires that the predicted quantiles for a certain quantile level $\quantile \in [0, 1]$ should cover $100\quantile\%$ of the ground-truth scores given a finite data set.
Furthermore, the authors adapt the Isotonic Regression calibration method known from semantic confidence calibration \cite{Zadrozny2002} and use this method to rescale the predicted cumulative distribution to achieve \quantilecalibration{} \cite{Kuleshov2018}.
Thus, the calibration is applied in a post-hoc step after model training.
In contrast, the authors in \cite{Song2019} argue that the definition of \quantilecalibration{} only faces the marginal coverage probability which does not consider the actually predicted probability distribution.
The authors argue that this is in contrast to the common understanding of uncertainty calibration known from semantic confidence calibration which is conditioned on the actually predicted confidence \cite{Song2019} (cf. definition (\ref{eq:confidence:definition:classification}) in \secref{section:confidence:definition:classification}).
Therefore, the authors propose the term of \distributioncalibration{} which requires that a predicted distribution should match the observed error distribution \textit{given a certain probability distribution}.
As calibration method, the authors further propose the GP-Beta method \cite{Song2019} which adapts the Beta calibration method from semantic confidence calibration \cite{Kull2017} to perform a rescaling of the predicted cumulative dis\-tribution (similar to Isotonic Regression).
To achieve \distributioncalibration{}, the authors adapt a Gaussian process to obtain the recalibration parameters for each distribution, individually.
Independently, the authors in \cite{Levi2019} and \cite{Laves2020} propose the term of \variancecalibration{} which is designed for parametric normal distributions and requires that the predicted variance should match the observed mean squared error (which is equivalent to the observed variance) \textit{for a certain variance level}.
For variance calibration, the authors adapt the simple Temperature Scaling from semantic confidence calibration \cite{Guo2018} to rescale the predicted variance by a single scalar \cite{Levi2019,Laves2020}.
We use these definitions as well as the respective calibration methods for our examinations and set them in a common mathematical context in \secref{section:regression:definition} and \secref{section:regression:methods}, respectively.

Recent work has developed a parametric approach to construct a probabilistic object detection model \cite{He2019,Hall2020,Harakeh2020,Feng2021}. 
A different approach of yielding aleatoric uncertainty is quantile regression \cite{Fasiolo2020,Chung2021} where a model does not predict a certain score but directly infers the quantile boundaries.
However, quantile regression has not been used for object detection so far but is definitely an interesting approach to investigate for future work.
Besides the previously mentioned post-hoc calibration methods, recent work has also proposed techniques to achieve intrinsically calibrated probabilistic models during model training.
The authors in \cite{Feng2019} propose a calibration loss which adds a second regularization term that aims to minimize the difference between predicted variance and observed squared error.
Another approach provided by \cite{Cui2020} adapts maximum mean discrepancy (MMD) to perform a distribution matching between predicted and observed distribution during model training.
Similarly, the authors in \cite{Bhatt2021} propose the method $f$-Cal which is also used for a distribution matching during model training.
The advantage of these calibration techniques is that they do not require a dedicated held-out calibration training set.
However, as already mentioned in \secref{section:confidence:related_work}, the drawback of calibration during model training is that they aim to calibrate against the model's performance on the training set which is commonly much better compared to unseen data during inference.
This may lead to a distortion after calibration.
Therefore, we focus on post-hoc methods for regression calibration.



\section{Definitions and Metrics for Uncertainty Calibration}
\label{section:regression:definition}

In this section, we review the definitions of \quantilecalibration{} \cite{Kuleshov2018}, \distributioncalibration{} \cite{Song2019}, and \variancecalibration{} \cite{Levi2019,Laves2020} for regression uncertainty calibration and place them in a common mathematical context.
For this purpose, we follow the mathematical notation of the previous \chapref{chapter:confidence} and consider an object detection model that predicts individual objects with a certain label $\predoutputvariate \in \gtset = \{1, ..., \numclasses\}$ with an according confidence score $\predconfidencevariate \in \probset$ given the input samples $\allinputvariates \in \inputset$, where $\numclasses$ denotes the available number of classes.
These predictions aim to target the real objects with ground-truth information for the class $\groundtruthvariate \in \gtset$.
Furthermore, let $\bboxset$ denote the set of all possible bounding boxes with $\numbboxdims$ as the size of the used box encoding.
Commonly an encoding with the position, width, and height is used.

In contrast to the label prediction, an object detection model does not assess the uncertainty of the predicted object location by default.
Therefore, we consider a probabilistic object detector \cite{He2019,Hall2020,Feng2021} (cf. \secref{section:introduction:spatial}) that interprets the regression output $\allpredbboxvariates$ as a Gaussian so that $\allpredbboxvariates|\allinputvariates \sampledfrom \normaldistribution(\meanvec_{\allpredbboxvariates|\allinputvariates}, \cov_{\allpredbboxvariates|\allinputvariates})$
with mean vector $\meanvec_{\allpredbboxvariates|\allinputvariates} \in \bboxset$ and the variances $\variance_{1}, \ldots, \variance_{\numbboxdims} \in \realdigitspositive^{\numbboxdims}$ for each bounding box quantity, so that $\cov_{\allpredbboxvariates|\allinputvariates} = \diag(\variance_{1}, \ldots, \variance_{\numbboxdims})$.
Thus, the network output for the bounding box regression can be interpreted as a random variable with a \ac{PDF} denoted by $\pdf_{\allpredbboxvariates|\allinputvariates}(\allpredbboxes) = \normaldistribution(\allpredbboxes; \meanvec_{\allpredbboxvariates|\allinputvariates}, \cov_{\allpredbboxvariates|\allinputvariates})$ which targets the true object location $\allgroundtruthbboxvariates \in \bboxset$.
Let further denote $\cdf_{\allpredbboxvariates|\allinputvariates}(\allpredbboxes)$ as the respective \ac{CDF} where $\cdf: \bboxset \rightarrow [0, 1]$.
In the following, we also need the quantile function that returns the quantile boundaries for each bounding box quantity given a certain quantile $\quantile \in [0, 1]$.
However, the quantile function is only defined for the univariate case.
Given the univariate \ac{CDF} $\cdf_{\predbboxvariate|\allinputvariates}(\allpredbboxes)$ for a single bounding box quantity $\predbboxvariate$, the respective quantile function is denoted by $\ppf_{\predbboxvariate|\allinputvariates}(\quantile)$ so that $\ppf_{\allpredbboxvariates|\allinputvariates}: [0, 1] \rightarrow \realdigits$.
Note that using independent quantile functions for each bounding box quantity leads to a negligence of possible correlations.

\begin{figure}[ht!]
    \centering
    \begin{subfigure}{0.8\textwidth}
        \begin{overpic}[width=1.0\linewidth]{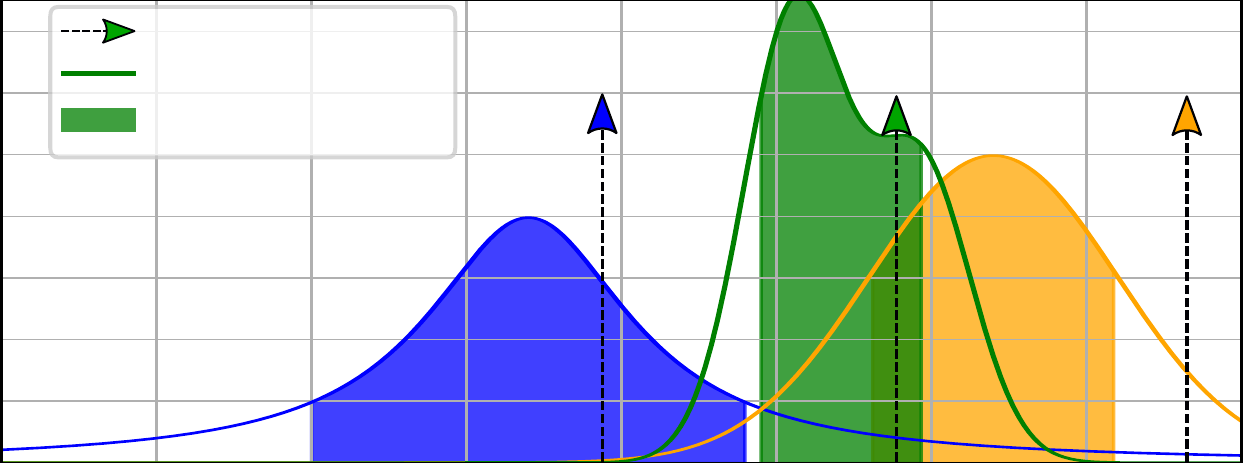}
            \put(12, 34.1){\scriptsize Ground-truth samples}
            \put(12, 30.8){\scriptsize Predicted distribution}
            \put(12, 27){\scriptsize $67\%$ prediction interval}
        \end{overpic}
        \caption{
            Principle of \quantilecalibration{} \cite{Kuleshov2018} demonstrated by 3 predicted distributions with a prediction interval of $67\%$. 
            The goal within \quantilecalibration{} is to cover approx. $67\%$ of the ground-truth samples within the predicted quantiles.
        }
        \label{fig:regression:definitions:quantile}
    \end{subfigure}
    \vspace{1em}
    
    \begin{subfigure}{0.8\textwidth}
        \begin{overpic}[width=1.0\linewidth]{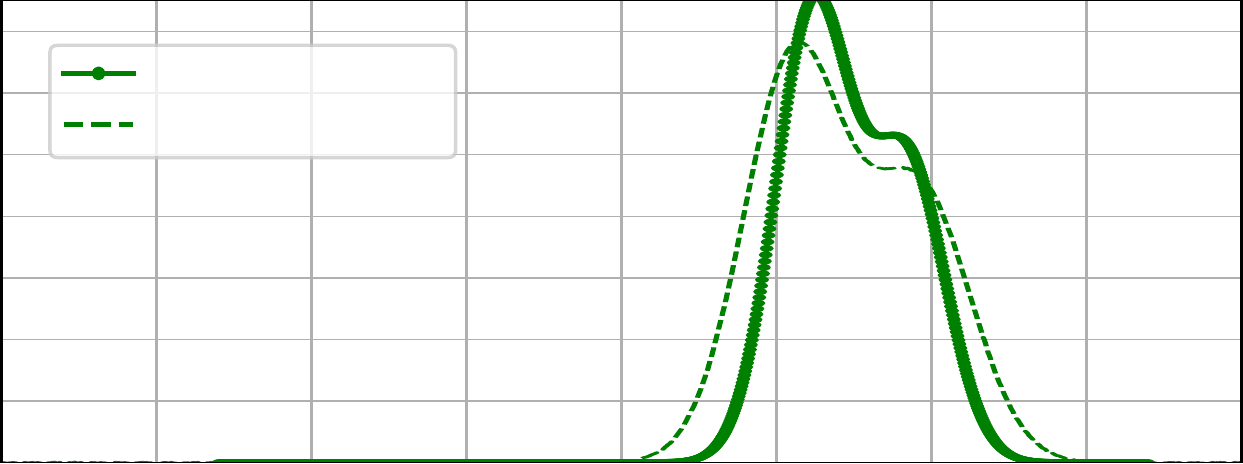}
            \put(12, 30.8){\scriptsize Predicted distribution}
            \put(12, 26.5){\scriptsize Observed error distribution}
        \end{overpic}
        \caption{
            Principle of \distributioncalibration{} \cite{Song2019}.
            The definition for \distributioncalibration{} requires that the observed error distribution matches the predicted distribution given all probability distributions with the same shape, e.g., as shown above.
        }
        \label{fig:regression:definitions:distribution}
    \end{subfigure}
    \vspace{1em}
    
    \begin{subfigure}{0.8\textwidth}
        \begin{overpic}[width=1.0\linewidth]{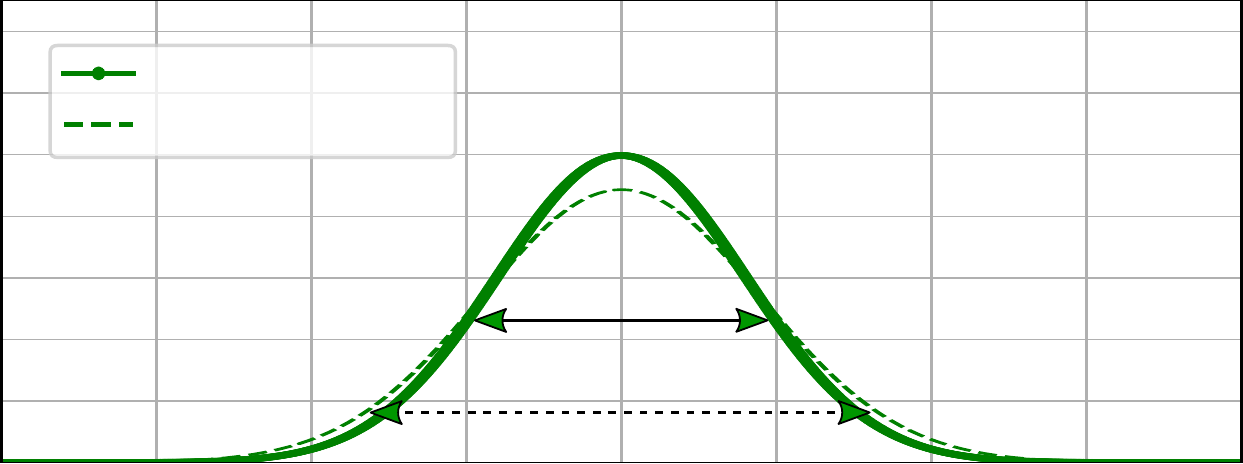}
            \put(12, 30.8){\scriptsize Predicted distribution}
            \put(12, 26.5){\scriptsize Observed error distribution}
            \put(45, 12){\scriptsize Predicted $\variance$}
            \put(43.5, 5){\scriptsize Observed MSE}
        \end{overpic}
        \caption{
            Principle of \variancecalibration{} \cite{Levi2019,Laves2020}.
            For all predicted normal distributions with the same variance, the term of \variancecalibration{} requires that the observed variance, i.e., the observed Mean Squared Error (MSE) matches the predicted variance.
        }
        \label{fig:regression:definitions:variance}
    \end{subfigure}
    
    \caption[Overview over the different definitions for regression uncertainty calibration.]{
        Overview over the different definitions for regression uncertainty calibration.
        We distinguish between (a) \quantilecalibration{}, (b) \distributioncalibration{}, and (c) \variancecalibration{}.
    }
    \label{fig:regression:definitions}
\end{figure}

The target of probabilistic object detection is to predict the mean as precisely as possible to the ground-truth objects, but also to indicate a high uncertainty if the model fails to reliably predict the object position or shape.
Thus, it is required to evaluate the quality of the predicted uncertainties and to examine for calibration.
In the following, we review the different definitions of uncertainty calibration for the task of regression.
We give an overview over the different types of regression calibration in \figref{fig:regression:definitions}.
Note that the probabilistic object detector does not predict any covariances between the bounding box quantities which yields multiple independent normal distributions for each quantity.
Since most of the existing definitions for regression calibration face only univariate probability distributions, we further revisit the definitions for single bounding box quantities which are denoted by $\predbboxvariate \sampledfrom \normaldistribution(\mean_{\predbboxvariate|\allinputvariates}, \variance_{\predbboxvariate|\allinputvariates})$ for notational simplicity.

\subsection{Quantile Calibration}
\label{section:regression:definition:quantile}

The first definition for uncertainty calibration is \quantilecalibration{} that has initially been proposed by the authors in \cite{Kuleshov2018}.
A probabilistic forecaster is \quantilecalibrated{} if the predicted quantiles for a quantile level $\quantile$ cover $100\quantile\%$ of the ground-truth samples.
For example, consider 100 samples where a probabilistic forecaster predicts a mean and variance that targets the ground-truth value.
If we consider a quantile level of $0.8$, we would expect that $80$ of the ground-truth scores are covered by the predicted intervals.
This must hold for all quantile levels $\quantile \in [0, 1]$.
Therefore, a probabilistic object detector is \quantilecalibrated{} if
\begin{align}
    \label{eq:regression:definition:one_sided:univariate}
    \prob\big(\groundtruthbboxvariate \leq \ppf_{\predbboxvariate|\allinputvariates}(\quantile)\big) = \quantile , \quad \forall \quantile \in [0, 1] ,
\end{align}
holds for each bounding box dimension \cite{Kuleshov2018}.
This definition also holds for two sided quantiles $\quantile = \quantile_2 - \quantile_1$ where $\quantile_1 < \quantile_2$ \cite{Kuleshov2018}, so that
\begin{align}
    \label{eq:regression:definition:two_sided:univariate}
    \prob\big(\ppf_{\predbboxvariate|\allinputvariates}(\quantile_1) \leq \groundtruthbboxvariate \leq \ppf_{\predbboxvariate|\allinputvariates}(\quantile_2) \big) = \quantile_2 - \quantile_1 , \quad \forall \quantile_1, \quantile_2 \in [0, 1] .
\end{align}
The principle of \quantilecalibration{} is schematically shown in \figref{fig:regression:definitions:quantile}.
A probabilistic forecaster is commonly evaluated for \quantilecalibration{} using the Pinball loss $\loss_\text{Pin}$ \cite{Steinwart2011} for a certain $\quantile$ that is defined by
\begin{align}
    \label{eq:regression:metrics:pinball}
    \loss_\text{Pin}(\quantile) := 
    \begin{cases}
        \big(\groundtruthbboxvariate - \ppf_{\predbboxvariate|\allinputvariates}(\quantile)\big) \quantile \quad &\text{ if } \groundtruthbboxvariate \geq \ppf_{\predbboxvariate|\allinputvariates}(\quantile) \\
        \big(\ppf_{\predbboxvariate|\allinputvariates}(\quantile) - \groundtruthbboxvariate\big)(1-\quantile) \quad &\text{ if } \groundtruthbboxvariate < \ppf_{\predbboxvariate|\allinputvariates}(\quantile)
    \end{cases} .
\end{align}
A mean $\xoverline[0.95]{\loss}_\text{Pin} := \expectation_\quantile[\loss_\text{Pin}(\quantile)]$ can also be used to denote the calibration properties for several quantile levels.
Further metrics are the \ac{PICP} (cf. (\ref{eq:bayesian:picp}) in \secref{section:bayesian:methods}) \cite{Pearce2018} and the \ac{MPIW} \cite{Pearce2018} for a certain $\quantile$ (both have been introduced in \secref{section:bayesian:methods}).

A major difficulty for measuring \quantilecalibration{} is that prediction intervals for certain quantile levels are not uniquely defined.
Therefore, we seek for the \ac{HPDI} which denotes the narrowest prediction interval given a certain quantile level $\quantile$.
The computation of the \ac{HPDI} for non-parametric probability distributions requires a numerical approach for approximation.
In contrast, it is considerably easier to determine the \ac{HPDI} and thus the prediction interval coverage of a Gaussian distribution with a known parametric form.
In addition, we can also determine the \ac{HPDR} which is the multivariate counterpart of the \ac{HPDI} and reflects a certain region with probability mass $\quantile$ for a certain quantile $\quantile$.
Similar to the univariate case, we would expect that approx. $100\quantile\%$ of the ground-truth samples are covered by this prediction region.
To determine the prediction region coverage of the true value vector $\allgroundtruthbboxvariates$ given a predicted Gaussian distribution with mean $\meanvec_{\allpredbboxvariates|\allinputvariates}$ and covariance matrix $\cov_{\allpredbboxvariates|\allinputvariates}$, we adapt the \ac{NEES} from Kalman filter consistency evaluation \cite[pp. 232]{Bar2004} \cite[pp. 292]{Van2005} that is defined by
\begin{align}
    \label{eq:regression:metrics:nees}
    \nees_{\allpredbboxvariates|\allinputvariates} := \big( \allgroundtruthbboxvariates - \meanvec_{\allpredbboxvariates|\allinputvariates} \big)^\T \cov_{\allpredbboxvariates|\allinputvariates}^{-1} \big( \allgroundtruthbboxvariates - \meanvec_{\allpredbboxvariates|\allinputvariates} \big) ,
\end{align}
which is also known as the squared Mahalanobis distance between the predicted distribution and the ground-truth vector \cite{Mahalanobis1936}.
Furthermore, we assume that the estimator has no bias in its predictions.
Thus, the prediction interval coverage for the ground-truth vector $\allgroundtruthbboxvariates$ can be determined using the $\chi^2$-test.
It is well known, see e.g. \cite{Van2005}, that the \ac{NEES} can be interpreted as the sum of $\numbboxdims$ independent squared random variables with zero mean and unit variance \cite[p. 295]{Van2005}.
This sum is also represented by a $\chi^2_\numbboxdims$ distribution with $\numbboxdims$ degrees of freedom \cite[p. 295]{Van2005}.
The \ac{NEES} $\nees_{\allpredbboxvariates|\allinputvariates}$ is highly connected to the \ac{HPDR} for Gaussian distributions as its equation in (\ref{eq:regression:metrics:nees}) can also be interpreted as an equation for an ellipsoid which describes the contours/isolines for certain quantile levels.
To test for prediction interval coverage, it is only required to determine if the radius of the ellipsoid for the ground-truth vector $\allgroundtruthbboxvariates$ is below the radius for the target quantile which is given by $\chi^2_\numbboxdims(\quantile)$\footnote{For notational simplicity, we further refer to $\chi^2_{\numbboxdims}(\quantile)$ as the percent point function (inverse \ac{CDF}) of a $\chi^2$-distributed random variable with $\numbboxdims$ degrees of freedom and quantile $\quantile \in [0, 1]$.}.
Therefore, for a certain sample $\allsingleinput \in \inputset$, the ground-truth vector is covered by the estimated prediction interval with a certain quantile level $\quantile$, if $\nees_{\allpredbboxvariates|\allsingleinput} \leq \chi^2_\numbboxdims(\quantile)$ is fulfilled.
More formally, we can denote the gap between the desired quantile level $\quantile$ and the observed quantile coverage by
\begin{align}
    \label{eq:regression:metrics:mqce:continuous}
    \Big| \prob\big(\nees_{\allpredbboxvariates|\allinputvariates} \leq \chi^2_\numbboxdims(\quantile)\big) - \quantile \Big| .
\end{align}

We use this derivation to construct the \ac{M-QCE} that is a new metric to measure for \quantilecalibration{} given multivariate normal distributions.
The advantage of the \ac{M-QCE} is that it tests for \quantilecalibration{} given multivariate normal distributions that might also represent possible correlations between the random variables.
The \ac{M-QCE} is related to the \ac{PICP} but determines the absolute difference between prediction interval coverage probability and the target quantile.
Therefore, the \ac{M-QCE} can directly be interpreted as an error metric reflecting the deviation between expected and observed quantile coverage.
On a finite data set $\dataset = \big\{(\allgroundtruthbboxes_{\indexsamples}, \meanvec_{\allpredbboxvariates|\allsingleinput_{\indexsamples}}, \cov_{\allpredbboxvariates|\allsingleinput_{\indexsamples}})\big\}^{\numsamples}_{\indexsamples=1}$ with $\numsamples$ samples, the \ac{M-QCE} is designed to approximate (\ref{eq:regression:metrics:mqce:continuous}) by
\begin{align}
    \label{eq:regression:metrics:mqce:discrete}
    \mqce(\quantile) := \Bigg|\frac{1}{\numsamples} \sum^\numsamples_{\indexsamples=1} \ind\big(\nees_{\allpredbboxvariates|\allsingleinput_{\indexsamples}} \leq \chi^2_\numbboxdims(\quantile)\big) - \quantile \Bigg| .
\end{align}
It is also possible to denote a mean $\xoverline[0.95]{\mqce} := \expectation_\quantile[\mqce(\quantile)]$ for several quantile levels to measure the overall properties for \quantilecalibration{} of a probabilistic forecaster.

\subsection{Distribution Calibration}
\label{section:regression:definition:distribution}

If we review the definition for semantic confidence calibration (cf. (\ref{eq:confidence:definition:classification}) in \secref{section:confidence:definition:classification}), we can see that the probability for the observed accuracy is conditioned on the predicted confidence.
This ensures that a forecaster not only predicts globally (marginally) calibrated confidences but also must provide informative confidence estimates for each confidence level separately.
In contrast, the definition for \quantilecalibration{} only considers the marginal probability over all predicted probability distributions. 
This allows a forecaster to have putatively good calibration properties if the predicted quantiles match the observed quantile coverage on average, even if it is poorly calibrated at a local level (e.g., for subsets with neighboring samples).
Therefore, the authors in \cite{Song2019} recently introduced the definition of \distributioncalibration{} for regression uncertainty calibration.
Let $\distset$ denote the set of all possible probability distributions, so that $\distvariate_{\bboxvariate} \in \distset$.
A probabilistic forecaster is \distributioncalibrated{} if the predicted probability distribution $\pdf_{\predbboxvariate}$ matches the observed (error) distribution $\distvariate_{\bboxvariate}$ given a certain probability distribution $\dist$ \cite{Song2019}, so that
\begin{align}
    \label{eq:regression:definition:distribution}
    \pdf_{\predbboxvariate}(\predbbox | \distvariate_{\bboxvariate} = \dist) = \dist(\predbbox) ,
\end{align}
must hold for all $\dist \in \distset$ and for all $\predbbox \in \bboxset$ \cite{Song2019}.
For example, consider a probabilistic forecaster that predicts several samples with equal distributions (e.g., with same mean and variance using Gaussian distributions).
With these predictions, it is possible to construct an error distribution for the corresponding ground-truth samples $\groundtruthbboxvariate$ as well.
If the predicted distributions match the observed ones, a probabilistic forecaster is \distributioncalibrated{} \cite{Song2019}.
This principle is schematically shown in \figref{fig:regression:definitions:distribution}.
The authors in \cite{Song2019} also point out that a \distributioncalibrated{} forecaster is also \quantilecalibrated{} \cite{Song2019}.
In the context of object detection, false positive predictions may occur which, however, do not have a corresponding ground-truth label.
Thus, we can not compute error statistics for these predictions so that they are discarded in this process.

The authors in \cite{Song2019} use \ac{NLL} for measuring \distributioncalibration{}. 
The authors decompose the \ac{NLL} into a calibration and a refinement loss and show the benefits of \distributioncalibration{} to the former calibration loss.
Although the definition for \distributioncalibration{} is more restrictive and constructed in the sense of the well-known definition for semantic confidence calibration, the set of possible probability distributions $\distset$ can be very large and intractable if it is not restricted to a parametric distribution family.
This leads us to the next definition for regression uncertainty calibration.

\subsection{Variance Calibration}
\label{section:regression:definition:variance}

Recently, the authors in \cite{Levi2019} and \cite{Laves2020} independently introduced the definition for \variancecalibration{} which is designed to evaluate the calibration properties of Gaussian distributions.
Given a joint ground-truth data distribution $\pdf_{\allinputvariates,\groundtruthbboxvariate}(\allsingleinput,\singlegroundtruthbbox)$ with input images $\allinputvariates \in \inputset$ and ground-truth bounding box positions $\groundtruthbboxvariate \in \bboxset$,
a probabilistic forecaster is \variancecalibrated{} if the predicted variance matches the observed one given a certain variance level \cite{Levi2019,Laves2020}, so that
\begin{align}
    \expectation_{\allinputvariates,\groundtruthbboxvariate}\big[(\singlegroundtruthbbox - \mean_{\predbboxvariate|\allinputvariates})^2 | \variance_{\predbboxvariate|\allinputvariates} = \variance \big] = \variance,
\end{align}
must hold for all $\variance \in \realdigitspositive$ \cite{Levi2019,Laves2020}.
We further refer to $\normaldistribution(\mean_{\groundtruthbboxvariate}, \variance_{\groundtruthbboxvariate})$ as the observed (error) distribution with mean $\mean_{\groundtruthbboxvariate} \in \realdigits$ and observed variance $\variance_{\groundtruthbboxvariate} \in \realdigitspositive$.
A \variancecalibrated{} forecaster is also \quantilecalibrated{} if the observed error distribution is a Gaussian and the forecaster is not biased in its predictions.
The error distribution is then equal to the predicted normal distribution $\normaldistribution(\mean_{\predbboxvariate|\allinputvariates}, \variance_{\predbboxvariate|\allinputvariates})$ which results in equal quantile functions $\ppf_{\predbboxvariate|\allinputvariates}(\quantile) = \ppf_{\groundtruthbboxvariate}(\quantile)$ so that the predicted quantiles match the observed ones.
In application, the observed variance is equal to the \ac{MSE} for a certain predicted variance level.
Furthermore, if the observed variance does not depend on the predicted mean, i.e., $\distcovariance(\mean_{\predbboxvariate|\allinputvariates}, \variance_{\groundtruthbboxvariate}) = 0$, then a \variancecalibrated{} forecaster is also \distributioncalibrated{} as the set of probability distributions $\distset$ is restricted to normal distributions and the mean $\mean_{\groundtruthbboxvariate}$ is no influential factor.
Therefore, \variancecalibration{} can be interpreted as a variant of \distributioncalibration{} for normal distributions with additional requirements on the forecaster and the ground-truth data.
This principle is schematically shown in \figref{fig:regression:definitions:variance}.

To test for \variancecalibration{}, the authors in \cite{Levi2019} propose the \ac{ENCE} which measures the unweighted and normalized difference between predicted and observed standard deviation for a certain standard deviation level.
Similar to the \ac{ECE} from semantic confidence calibration, the \ac{ENCE} utilizes a binning scheme over the predicted standard deviation with $\numbins$ distinct bins to measure calibration by means of the predicted standard deviation, so that the \ac{ENCE} is defined by
\begin{align}
    \ence := \frac{1}{\numbins} \sum^\numbins_{\indexbins=1} \frac{|\text{RMSE}(\indexbins) - \text{RMV}(\indexbins)|}{\text{RMV}(\indexbins)} ,
\end{align}
where $\text{RMSE}(\indexbins)$ and $\text{RMV}(\indexbins)$ denote the root mean squared error and the root mean variance within bin $\indexbins$, respectively.
In practice, we choose the maximum of the estimated standard deviation $\stddev_\text{max}$ and divide the interval $[0, \stddev_\text{max}]$ into $\numbins$ equally sized bins.
Similarly, the authors in \cite{Laves2020} propose the \ac{UCE} which measures the weighted and unnormalized difference between predicted and observed variance for a certain variance level.
Thus, the \ac{UCE} is defined by
\begin{align}
    \uce := \sum^\numbins_{\indexbins=1} \frac{\numsamples_\indexbins}{\numsamples} |\text{MSE}(\indexbins) - \text{MV}(\indexbins) | ,
\end{align}
where $\text{MSE}(\indexbins)$ and $\text{MV}(\indexbins)$ denote the mean squared error and the mean variance within bin $\indexbins$, respectively \cite{Laves2020}.
The advantage of the \ac{ENCE} is that the miscalibration can be quantified relative to the predicted uncertainty.
Therefore, the ENCE is independent of the size of the investigation space $\bboxset$.

A drawback of both metrics is that they can not capture the properties of multivariate normal distributions with possible correlations.
Using these formulations, it is not straightforward to measure calibration by means of predicted covariances.
Furthermore, our recently derived \ac{M-QCE} metric only measures the marginal calibration error which is not sensitive to calibration by means of different variances.
Therefore, we first consider the \ac{SGV} as a property of a mul\-ti\-vari\-ate normal distribution which is defined by $\variance_\text{SG} = \det(\cov_{\allpredbboxvariates|\allinputvariates})^{\frac{1}{\numbboxdims}}$ \cite{Sengupta1987,Sengupta2004}.
The \ac{SGV} reflects the dispersion of a distribution over all dimensions $\numbboxdims$.
This allows distributions with similar dispersion to be grouped and compared to each other.
The distribution $\pdf_{\variance_\text{SG}}$ of the \ac{SGV} is directly connected to the output of the object detector as it is derived by the random variable $\allpredbboxvariates$.

In (\ref{eq:regression:metrics:mqce:continuous}), we already proposed the \ac{M-QCE} to evaluate the \ac{HPDR} of multivariate Gaussian distributions.
However, the \ac{M-QCE} evaluates the calibration properties of a forecaster over all predictions.
Instead, we seek to evaluate a forecaster conditioned on its predictions, similar to the \ac{ECE} known from semantic confidence calibration evaluation (cf. \secref{section:confidence:definition:classification}).
The \ac{ECE} is conditioned on the predicted confidence to measure miscalibration by means of the model output.
Similarly, we can measure the quantile coverage conditioned on the \ac{SGV} as the model output by
\begin{align}
    \label{eq:regression:metrics:cqce:continuous}
    \expectation_{\variance_\text{SG} \sampledfrom \pdf_{\variance_\text{SG}}}\Big[\big| \prob\big(\nees_{\allpredbboxvariates|\allinputvariates} \leq \chi^2_{\numbboxdims}(\quantile) | \variance_\text{SG}\big) - \quantile \big|\Big].
\end{align}

We can now reformulate the \ac{M-QCE} to the \ac{C-QCE} which measures the error between predicted quantile and observed quantile coverage as a function of the model output given by the \ac{SGV}.
In practice, we further use the square root of the \ac{SGV} to achieve a better data distribution of the dispersion for binning.
Similar to the \ac{UCE} and \ac{ENCE}, a binning scheme with $\numbins$ bins over the square root of the \ac{SGV} is applied, so that the \ac{C-QCE} is an approximation of (\ref{eq:regression:metrics:cqce:continuous}) on a finite data set $\dataset$ given by
\begin{align}
    \label{eq:regression:metrics:cqce:discrete}
    \cqce(\quantile) := \sum^\numbins_{\indexbins=1} \frac{\numsamples_\indexbins}{\numsamples} | \text{freq}(\indexbins) - \quantile | , 
\end{align}
where 
\begin{align}
    \text{freq}(\indexbins) = \frac{1}{\numsamples_\indexbins} \sum_{\indexsamples \in \indexset_\indexbins} \ind\big(\nees_{\allpredbboxvariates|\allsingleinput_\indexsamples} \leq \chi^2_{\numbboxdims}(\quantile)\big) ,
\end{align}
denotes the prediction interval coverage frequency within bin $\indexbins$, where $\indexset_\indexbins$ is the set of sample indices with all samples falling into bin $\indexbins$.
In the following, these metrics are used for uncertainty calibration evaluation.

\section{Review of Methods for Regression Uncertainty Calibration}
\label{section:regression:methods}

In this section, we present state-of-the-art calibration methods such as Isotonic Regression \cite{Kuleshov2018}, Variance Scaling \cite{Levi2019,Laves2020}, and GP-Beta \cite{Song2019}.
These techniques perform calibration w.r.t. one of the previously introduced definitions for regression uncertainty calibration. The effects of calibration by means of individual calibration targets are qualitatively shown in \figref{fig:regression:methods:artificial}.
\begin{figure}[t!]
    \centering
    \begin{subfigure}[t]{0.99\linewidth}
        \input{images/regression/artificial/artificial_0.tikz}
        \subcaption{Artificial data (green) and uncalibrated estimator with according predicted uncertainty (blue).}
    \end{subfigure}
    \vspace{1em}
    
    \begin{subfigure}[t]{0.5\textwidth}
        \input{images/regression/artificial/artificial_1.tikz}
        \subcaption{Isotonic Regression \cite{Kuleshov2018}.}
    \end{subfigure}%
    \begin{subfigure}[t]{0.5\textwidth}
        \input{images/regression/artificial/artificial_2.tikz}
        \subcaption{Variance Scaling \cite{Levi2019,Laves2020}.}
    \end{subfigure}%
    \vspace{1em}
    
    \begin{subfigure}[t]{0.5\textwidth}
        \input{images/regression/artificial/artificial_3.tikz}
        \subcaption{GP-Beta \cite{Song2019}.}
    \end{subfigure}%
    \begin{subfigure}[t]{0.5\textwidth}
        \input{images/regression/artificial/artificial_4.tikz}
        \subcaption{GP-Normal \cite{Kueppers2022b}.}
    \end{subfigure}%

    \caption[Qualitative example on artificial data to demonstrate the differences between methods for \quantilecalibration{}, \variancecalibration{}, and \distributioncalibration{}.]{
        Qualitative example on artificial data to demonstrate the differences between methods for \quantilecalibration{} (ii), \variancecalibration{} (iii), and \distributioncalibration{} (iv \& v) \cite[p.~7, Fig.~2]{Kueppers2022b}.
        The ground-truth data is obtained by sampling from a cosine with aleatoric Gaussian noise (green points).
        The noise amplitude is proportional to the y-value of the cosine function (with a small offset), thus, the aleatoric uncertainty is correlated with the function value.
        Furthermore, we assume an unbiased estimator (blue) with randomly sampled variance for each point which, however, is equally sampled for the whole function.
        In this example, we can see that the methods for \quantilecalibration{} (ii) as well as for \variancecalibration{} (iii) can not to capture the dependency between aleatoric uncertainty and the function value.
        In contrast, the methods for \distributioncalibration{} (iv \& v) are able to recalibrate the predicted uncertainty scores.
    }
    \label{fig:regression:methods:artificial}
\end{figure}
We can divide these calibration techniques into parametric and non-parametric methods.
The former methods yield a parametric probability distribution (e.g., Gaussian) as calibration output, whereas the latter ones yield probability distributions of arbitrary shape.
While the non-parametric distributions are more flexible in representing any data distributions, it might be necessary to utilize a parametric distributions after calibration, e.g., for subsequent applications such as Kalman filtering for object tracking (cf. \chapref{chapter:tracking}).
Therefore, we propose an extension to the existing calibration methods that is able to perform parametric uncertainty calibration in the sense of \distributioncalibration{}.
We further derive a calibration scheme for a joint multivariate uncertainty recalibration of multiple dimensions.
Finally, we adapt the Gaussian process scheme for a covariance estimation and recalibration.
This allows for a post-hoc introduction of correlations between independently inferred probability distributions.
Note that we assert normal distributions as the input to the calibration methods as we work with the output of probabilistic detection models that have been introduced in \secref{section:basics:object_detection}.
We start by reviewing the state-of-the-art methods for regression uncertainty calibration and present our extensions for parametric and joint uncertainty calibration that are subject of our publication in \cite{Kueppers2022b}.

\subsection{Non-Parametric Calibration}
\label{section:regression:methods:nonparametric}

A non-parametric calibration method takes an uncalibrated probability distribution and outputs a calibrated distribution that has no analytical form, i.e., the \ac{PDF} is not fixed to any arbitrary shape.
The advantage of this distribution representation is that calibration methods are not restricted to any assumptions and thus can flexibly represent any data distribution.
The existing non-parametric calibration methods are designed to output a non-parametric density function given any probability distribution as input to fit the calibrated distribution to the observed (error) distribution.
Commonly, the calibration techniques are applied on the univariate \ac{CDF} of the input distributions (for each dimension independently) which denotes the cumulative probability mass of a certain point.
Since the total probability mass of any distribution is $1$, the \ac{CDF} also always outputs scores only within the $[0, 1]$ interval.
This allows for the application of calibration techniques known from semantic confidence calibration, since these methods are bound to the $[0, 1]$ interval as well.
Specifically, the Isotonic Regression \cite{Zadrozny2002,Kuleshov2018} as well as the Beta Calibration \cite{Kull2017,Song2019} methods have recently been adapted for regression calibration which are presented in the following.

\subsubsection{Isotonic Regression}
\label{section:regression:methods:nonparametric:isotonic}

As already described in \secref{section:regression:definition:quantile}, the same authors in \cite{Kuleshov2018} propose a recalibration framework to achieve a \quantilecalibrated{} forecaster.
We further denote this as marginal calibration as the target is to construct a recalibration method that leads to a \quantilecalibrated{} model where only the marginal calibration properties over all given samples are of interest.
Let $\pdf_{\predbboxvariate}(\predbbox)$ and $\cdf_{\predbboxvariate|\allinputvariates}(\predbbox)$ denote the \ac{PDF} and \ac{CDF} of the predicted object locations in a single dimension, respectively.
For any quantile level $\quantile \in [0, 1]$, the authors in \cite{Kuleshov2018} seek to estimate the true probability of $\prob\big(\groundtruthbboxvariate \leq \ppf_{\predbboxvariate|\allinputvariates}(\quantile)\big)$ \big(cf. (\ref{eq:regression:definition:one_sided:univariate})\big) given a finite data set.
If we observe a deviation between the estimated prediction intervals and observed interval coverage (using the true target score), a calibration function $\calmodel$ will be necessary to recalibrate the estimated quantile boundaries.
As the recalibration target, the authors utilize the \ac{ECDF} which can be interpreted as an estimator for the probability distribution of the true underlying data generation process.
The \ac{ECDF} is a step function that denotes, for each sample $\indexsamples$, the fraction of observations whose values are less than or equal to the actual sample.
According to the Glivenko–Cantelli theorem \cite{Glivenko1933,Tucker1959}, the \ac{ECDF} converges to the underlying distribution as the number of observations grows.
In the case of \quantilecalibration{}, we are interested in the probability that an observation falls into the estimated prediction interval.
The authors in \cite{Kuleshov2018} utilize the \ac{ECDF} to represent the according probability distribution.
Given $\numsamples$ observations with known (univariate) ground-truth bounding box position $\singlegroundtruthbbox_\indexsamples$ as well as the estimated \ac{CDF} $\cdf_{\predbboxvariate|\allsingleinput_{\indexsamples}}$ for each sample (obtained by a probabilistic object detector), the \ac{ECDF} can be constructed by
\begin{align}
	\cdf_{\text{emp}}(\singlegroundtruthbbox_\indexsamples) = \frac{1}{\numsamples} \sum^{\numsamples}_{\indexsamples^\ast=1} \ind\Big(\cdf_{\predbboxvariate|\allsingleinput_{\indexsamples^\ast}}(\singlegroundtruthbbox_{\indexsamples^\ast}) \leq \cdf_{\predbboxvariate|\allsingleinput_{\indexsamples}}(\singlegroundtruthbbox_{\indexsamples}) \Big) .
\end{align}
By constructing the \ac{ECDF} for the probability distribution of the prediction interval coverage, the inverse empirical quantile function reflects the desired estimate for $\prob\big(\groundtruthbboxvariate \leq \ppf_{\predbboxvariate|\allinputvariates}(\quantile)\big)$ which is the final target for \quantilecalibration{}.
Thus, matching the predicted \ac{CDF} scores with the according \ac{ECDF} counterpart finally yields in \quantilecalibrated{} estimates.

For calibration, the function $\calmodel(\cdot)$ serves as a mapping from the uncalibrated \ac{CDF} to a calibrated one, so that $\calmodel: [0, 1] \rightarrow [0, 1]$.
The authors in \cite{Kuleshov2018} propose to use the Isotonic Regression calibration method known from semantic confidence calibration \cite{Zadrozny2002}.
As originally proposed by the authors in \cite{Zadrozny2002}, Isotonic Regression fits a piece-wise constant and monotonically increasing function as recalibration function to map uncalibrated confidence estimates to calibrated ones.
Thus, Isotonic Regression can be seen as a variant of Histogram Binning \cite{Zadrozny2001}, but with flexible bin sizes and a flexible amount of bins.
A training data set for the Isotonic Regression method can be constructed using the predicted \ac{CDF} $\cdf_{\predbboxvariate|\allsingleinput_{\indexsamples}}(\singlegroundtruthbbox_{\indexsamples})$ as the input to the regression function and the according \ac{ECDF} score $\cdf_{\text{emp}}(\singlegroundtruthbbox_\indexsamples)$ as the regression target for each sample $\indexsamples$ with ground-truth $\singlegroundtruthbbox_{\indexsamples}$, so that $\dataset = \{\cdf_{\predbboxvariate|\allsingleinput_{\indexsamples}}(\singlegroundtruthbbox_{\indexsamples}), \cdf_{\text{emp}}(\singlegroundtruthbbox_\indexsamples)\}^{\numsamples}_{\indexsamples=1}$.
We can use the training set $\dataset$ to construct the regression function which serves as a mapping from uncalibrated quantiles to calibrated ones \cite{Kuleshov2018}.

During inference, the mapping function is used to transform the \ac{CDF} of each input data individually.
For each input sample, we draw a number of $\numstochastic$ points denoted by $\dataset_{\indexsamples}^\ast = \big\{\big(\bbox_{\indexstochastic}, \cdf_{\predbboxvariate|\allsingleinput_{\indexsamples}}(\bbox_{\indexstochastic})\big)\big\}^{\numstochastic}_{\indexstochastic=1}$ to represent the predicted \ac{CDF} of the input data with index $\indexsamples \in \{1, \ldots, \numsamples\}$.
Afterwards, we can pass each of these points through the calibration mapping to finally yield a non-parametric representation of the calibrated cumulative.
Thus, the calibrated \ac{CDF} $\cdfcalibrated_{\predbboxvariate|\allinputvariates}(\predbbox)$ for each point in $\dataset_{\indexsamples}^\ast$ is given by
\begin{align}
    \label{eq:regression:isotonic:calibrated:cdf}
    \cdfcalibrated_{\predbboxvariate|\allsingleinput_{\indexsamples}}(\bbox_{\indexstochastic, \indexsamples}) = \calmodel\big(\cdf_{\predbboxvariate|\allsingleinput_{\indexsamples}}(\bbox_{\indexstochastic, \indexsamples})\big), \quad \forall \indexstochastic \in \{1, \ldots, \numstochastic\} .
\end{align}
An approximation for the respective calibrated density scores $\pdfcalibrated_{\predbboxvariate|\allsingleinput_{\indexsamples}}(\bbox_{\indexstochastic, \indexsamples})$ at the point locations $\bbox_{\indexstochastic, \indexsamples}$ can be obtained by differentiation for each sample with index $\indexsamples$.

\subsubsection{Beta Calibration with Gaussian Process Parameter Estimation}
\label{section:regression:methods:nonparametric:gp}

Similarly to Isotonic Regression, the authors in \cite{Song2019} adapted the Beta Calibration method \cite{Kull2017} from the scope of confidence calibration and applied it to regression uncertainty calibration (cf. equation (\ref{eq:confidence:scaling:betacal:standard} in \secref{section:confidence:methods:scaling} for a detailed description of Beta Calibration).
Let $\calmodel_\beta$ denote the Beta Calibration function which transforms the quantiles $\quantile \in [0, 1]$ (obtained by the \ac{CDF}) to calibrated ones.
For regression calibration, the Beta Calibration parameters $\betaparama, \betaparamb \in \realdigitspositive$ and $\betaparamc \in \realdigits$ are used to rescale the cumulative of the input distribution \cite{Song2019}, so that the calibrated \ac{CDF} is given by
\begin{align}
    \label{eq:regression:methods:gpbeta:cdf}
    \cdfcalibrated_{\predbboxvariate|\allinputvariates}(\predbbox) 
    &= \sigmoid\Big(\betaparama \cdot \log\big(\cdf_{\predbboxvariate|\allinputvariates}(\predbbox)\big) - \betaparamb \cdot \log\big(1-\cdf_{\predbboxvariate|\allinputvariates}(\predbbox)\big) + \betaparamc\Big) \\
    &= \sigmoid\Big( \logit_\beta\big(\cdf_{\predbboxvariate|\allinputvariates}(\predbbox)\big) \Big) \\
    &= \calmodel_\beta\Big(\logit_\beta\big(\cdf_{\predbboxvariate|\allinputvariates}(\predbbox)\big)\Big),
\end{align}
with $\sigmoid(\cdot)$ as the sigmoid function, where
\begin{align}
    \logit_\beta\big(\cdf_{\predbboxvariate|\allinputvariates}(\predbbox)\big) = \betaparama \cdot \log\big(\cdf_{\predbboxvariate|\allinputvariates}(\predbbox)\big) - \betaparamb \cdot \log\big(1-\cdf_{\predbboxvariate|\allinputvariates}(\predbbox)\big) + \betaparamc .
\end{align}
Similar to the previous Isotonic Regression \cite{Kuleshov2018}, the Beta Calibration function seeks to serve as a transformation of uncalibrated quantiles to calibrated ones.
Since the \ac{PDF} can also be interpreted as the derivative of the \ac{CDF}, it is possible to derive the \ac{PDF} by differentiation \cite{Song2019}, so that the calibrated \ac{PDF} $\pdfcalibrated_{\predbboxvariate|\allinputvariates}$ is given by
\begin{align}
    \label{eq:regression:methods:gpbeta:pdf}
    \pdfcalibrated_{\predbboxvariate|\allinputvariates}(\predbbox) 
    &= \frac{\diff \calmodel_\beta\Big(\logit_\beta\big(\cdf_{\predbboxvariate|\allinputvariates}(\predbbox)\big)\Big)}{\diff \predbbox} \\
    &= \frac{\diff \calmodel_\beta\Big(\logit_\beta\big(\cdf_{\predbboxvariate|\allinputvariates}(\predbbox)\big)\Big)}{\diff \logit_\beta\big(\cdf_{\predbboxvariate|\allinputvariates}(\predbbox)\big)}
    \frac{\diff \logit_\beta\big(\cdf_{\predbboxvariate|\allinputvariates}(\predbbox)\big)}{\diff \cdf_{\predbboxvariate|\allinputvariates}(\predbbox)}
    \frac{\diff \cdf_{\predbboxvariate|\allinputvariates}(\predbbox)}{\diff \predbbox} \\
    & = \linkfunction_\beta\big(\cdf_{\predbboxvariate|\allinputvariates}(\predbbox)\big) \cdot \pdf_{\predbboxvariate|\allinputvariates}(\predbbox) ,
\end{align}
where $\frac{\diff \cdf_{\predbboxvariate|\allinputvariates}(\predbbox)}{\diff \predbbox}$ reduces to the uncalibrated input \ac{PDF} and
\begin{align}
	\linkfunction_\beta\big(\cdf_{\predbboxvariate|\allinputvariates}(\predbbox)\big) = \frac{\diff \calmodel_\beta\Big(\logit_\beta\big(\cdf_{\predbboxvariate|\allinputvariates}(\predbbox)\big)\Big)}{\diff \logit_\beta\big(\cdf_{\predbboxvariate|\allinputvariates}(\predbbox)\big)}
	\frac{\diff \logit_\beta\big(\cdf_{\predbboxvariate|\allinputvariates}(\predbbox)\big)}{\diff \cdf_{\predbboxvariate|\allinputvariates}(\predbbox)} ,
\end{align}
is the derivative of $\calmodel_\beta$ w.r.t. $\predbbox$ \cite{Song2019}.
The derivative of the sigmoid function can be expressed by $\sigmoid(\logit_\beta)' = \sigmoid(\logit_\beta) \big(1 - \sigmoid(\logit_\beta)\big)$, so that the function $\linkfunction_\beta$ can be derived by
\begin{align}
    \linkfunction_\beta\big(\cdf_{\predbboxvariate|\allinputvariates}(\predbbox)\big) 
    &= \frac{\diff \calmodel_\beta\Big( \logit_\beta\big(\cdf_{\predbboxvariate|\allinputvariates}(\predbbox)\big) \Big)}{\diff \cdf_{\predbboxvariate|\allinputvariates}(\predbbox)} 
    = \frac{\diff \calmodel_\beta\Big( \logit_\beta\big(\cdf_{\predbboxvariate|\allinputvariates}(\predbbox)\big) \Big)}{\diff \logit_\beta\big(\cdf_{\predbboxvariate|\allinputvariates}(\predbbox)\big)} \frac{\diff \logit_\beta\big(\cdf_{\predbboxvariate|\allinputvariates}(\predbbox)\big)}{\diff \cdf_{\predbboxvariate|\allinputvariates}(\predbbox)}  \\
    &= \sigmoid\Big( \logit_\beta\big(\cdf_{\predbboxvariate|\allinputvariates}(\predbbox)\big) \Big) \bigg[1 - \sigmoid\Big( \logit_\beta\big(\cdf_{\predbboxvariate|\allinputvariates}(\predbbox)\big) \Big)\bigg] \bigg[\frac{\betaparama}{\cdf_{\predbboxvariate|\allinputvariates}(\predbbox)} + \frac{\betaparamb}{1-\cdf_{\predbboxvariate|\allinputvariates}(\predbbox)} \bigg] \\
    &= \cdfcalibrated_{\predbboxvariate|\allinputvariates}(\predbbox) \Big(1 - \cdfcalibrated_{\predbboxvariate|\allinputvariates}(\predbbox)\Big) \bigg[\frac{\betaparama}{\cdf_{\predbboxvariate|\allinputvariates}(\predbbox)} + \frac{\betaparamb}{1-\cdf_{\predbboxvariate|\allinputvariates}(\predbbox)} \bigg] .
\end{align}

If the calibration parameters were trained by standard \ac{MLE} using the \ac{NLL} as the loss function, this calibration method would also perform marginal calibration yielding \quantilecalibrated{} distributions.
This corresponds to the training procedure used within marginal recalibration which we already introduced in the last section.
However, the authors in \cite{Song2019} propose the term of \distributioncalibration{} and thus seek to derive a calibration function that applies uncertainty calibration with a specific parameter set for each input distribution, individually.
As opposed to the training of the Isotonic Regression method \cite{Kuleshov2018}, we do not compute an empirical \ac{CDF} over the training data set $\dataset$ as this would result in marginal recalibration.
Instead, the authors in \cite{Song2019} use a \ac{GP} for inferring the calibration parameters $\betaparama(\cdot), \betaparamb(\cdot)$ and $\betaparamc(\cdot)$ as a function of the provided sample.
The authors propose to use three latent functions $\scaleweight_\betaparama(\cdot)$, $\scaleweight_\betaparamb(\cdot)$, and $\scaleweight_\betaparamc(\cdot)$ which are directly mapped to the calibration parameters $\betaparama(\cdot), \betaparamb(\cdot)$ and $\betaparamc(\cdot)$ by
\begin{align}
	\nonumber\betaparama(\cdot) &= \exp(\gpscale_\betaparama \scaleweight_\betaparama(\cdot) + \gpbias_\betaparama) , \\
	\label{eq:regression:methods:gpbeta:parameters}
	\betaparamb(\cdot) &= \exp(\gpscale_\betaparamb \scaleweight_\betaparamb(\cdot) + \gpbias_\betaparamb) , \\ \nonumber
	\betaparamc(\cdot) &= \gpscale_\betaparamc \scaleweight_\betaparamc(\cdot) + \gpbias_\betaparamc ,
\end{align}
where the exponential functions guarantee that $\betaparama, \betaparamb \in \realdigitspositive$.
The advantage of using the intermediate functions $\scaleweight_\betaparama$, $\scaleweight_\betaparamb$, and $\scaleweight_\betaparamc$ is that we have no restrictions on the \ac{GP}, so that the latent functions can directly be drawn from the \ac{GP}.
Furthermore, each additional scaling factor $\gpscale_\betaparama, \gpscale_\betaparamb, \gpscale_\betaparamc \in \realdigits$ and bias $\gpbias_\betaparama, \gpbias_\betaparamb, \gpbias_\betaparamc \in \realdigits$ are used to prevent possible distortion during the \ac{GP} initialization phase \cite{Song2019}.

The idea of using a \ac{GP} for \distributioncalibration{} is that neighboring samples in the input space (with similar mean and variance) are likely to produce similar outputs.
The concept of \distributioncalibration{} using a \ac{GP} model is to construct an error distribution for each sample individually based on the local neighborhood.
These error distributions can be compared to the ones that have been estimated by the object detector, so that the \ac{GP} can finally be used for a recalibration to achieve \distributioncalibration{}.
For this reason, the authors in \cite{Song2019} introduce a multi-output \ac{GP} \cite{Alvarez2008,Moreno2018,Skolidis2011} to infer the latent functions $\scaleweight_\betaparama(\cdot)$, $\scaleweight_\betaparamb(\cdot)$, and $\scaleweight_\betaparamc(\cdot)$ by means of the uncalibrated input.
We further refer to the function weight vector as $\scalevec = (\scaleweight_\betaparama, \scaleweight_\betaparamb, \scaleweight_\betaparamc)^\T$.
A \ac{GP} is a stochastic process such that a finite collection of random variables follows a joint multivariate normal distribution.
The \ac{GP} is parameterized by a mean and a positive semidefinit kernel or covariance function.
Especially the kernel function $\kernel(\cdot, \cdot)$ is important as it is used to construct the covariance matrix of the \ac{GP} which reflects the correlations between the random variables.
Common kernel functions for vector-valued inputs are the squared exponential kernel, the rational quadratic kernel, or the periodic kernel.

However, in our case, the input to the kernel function are the (uncalibrated) predictions of a probabilistic object detector that consist of the estimated distribution parameters for a Gaussian distribution.
Thus, the authors in \cite{Song2019} utilize a univariate Gaussian embedding using the \ac{RBF} kernel \cite{Song2008} which is given by
\begin{align}
	\label{eq:regression:methods:gpbeta:kernel:univariate}
	\kernel\big((\mean_i, \variance_i), (\mean_j, \variance_j)\big) = \frac{\parameter}{|\variance_{ij}|^{\frac{1}{2}}} \exp\bigg(-\frac{1}{2\variance_{ij}} (\mean_i - \mean_j)^2\bigg) ,
\end{align}
with length scale parameter $\parameter \in \realdigitspositive$, where $\variance_{ij} = \variance_i + \variance_j + \parameter^2$ \cite{Song2019}.
Unfortunately, the authors in \cite{Song2008} do not provide a mathematical proof that this kernel function yields a positive semidefinit covariance matrix.
However, we conducted extensive studies so that we have been able to empirically verify that the resulting covariance matrices are valid.

Let $\dataset = \big\{\big(\singlegroundtruthbbox_{\indexsamples}, \mean_{\predbboxvariate|\allsingleinput_\indexsamples}, \variance_{\predbboxvariate|\allsingleinput_\indexsamples}\big)\big\}^{\numsamples}_{\indexsamples=1}$ denote the training set with $\numsamples$ samples consisting of the ground-truth $\singlegroundtruthbbox_{\indexsamples}$ as well as of the predicted mean $\mean_{\predbboxvariate|\allsingleinput_\indexsamples}$ and variance $\variance_{\predbboxvariate|\allsingleinput_\indexsamples}$.
The predicted mean and variance are both obtained by a probabilistic object detector.
Note that we are working with the univariate (independent) bounding box quantities.
Furthermore, we denote the mean vector by $\boldsymbol{\mean}_\dataset = (\mean_{\predbboxvariate|\allsingleinput_1}, \ldots, \mean_{\predbboxvariate|\allsingleinput_\numsamples})^\T$ and the variance vector by $\boldsymbol{\stddev}_\dataset^2 = (\variance_{\predbboxvariate|\allsingleinput_1}, \ldots, \variance_{\predbboxvariate|\allsingleinput_\numsamples})^\T$ which both hold the predictions of the whole data set $\dataset$.
Using the kernel function in (\ref{eq:regression:methods:gpbeta:kernel:univariate}), we denote the covariance matrix $\kernelmatrix_\dataset \in \realdigits^{\numsamples \times \numsamples}$ which is given by $\kernelmatrix_\dataset = \kernel\big((\boldsymbol{\mean}_\dataset, \boldsymbol{\stddev}_\dataset^2), (\boldsymbol{\mean}_\dataset, \boldsymbol{\stddev}_\dataset^2)' \big)$.
Moreover, we denote the respective function weight matrix by ${\scaleweightmat_\dataset = \big(\scalevec_1^\T, \ldots, \scalevec_\numsamples^\T\big)^\T}$ that contains the latent functions for all samples in $\dataset$, where $\scaleweightmat_\dataset \in \scaleweightset$.
The matrix $\scaleweightmat_\dataset$ is also a function of the input samples, which we will omit in the following for notation simplicity.

In our case, when modeling $\scaleweightmat_\dataset$ using a \ac{GP} with zero mean and covariance matrix $\kernelmatrix_\dataset$, the latent functions follow a joint multivariate normal distribution, so that
\begin{align}
	\pdf(\scaleweightmat_\dataset | \dataset ) = \normaldistribution( \scaleweightmat_\dataset | 0, \kernelmatrix_\dataset \kroneckerprod \coregion_\beta ) ,
\end{align}
where $\kroneckerprod$ is the Kronecker product and $\coregion_\beta \in \realdigits^{3 \times 3}$ is the coregionalization matrix that captures the dependencies between all latent functions.
The target is to obtain a vector-valued output for each sample (the weights $\scaleweight_\betaparama(\cdot)$, $\scaleweight_\betaparamb(\cdot)$, and $\scaleweight_\betaparamc(\cdot)$). 
Such a multi-output \ac{GP} is also called an intrinsic coregionalization model (ICM) \cite{Goovaerts1997} which models the dependencies between the outputs as a linear combination of multiple latent variables that share the same covariance kernel.
Thus, for a single sample, the weights are drawn by the \ac{GP} model so that
\begin{align}
	\scaleweight_\betaparama, \scaleweight_\betaparamb, \scaleweight_\betaparamc \sampledfrom \gp(0, \kernel(\cdot, \cdot), \coregion_\beta) ,
\end{align}
where $\gp(0, \kernel(\cdot, \cdot), \coregion_\beta)$ denotes the \ac{GP} model with zero mean, kernel function $\kernel$ and coregionalization matrix $\coregion_\beta$.
In summary, the \ac{GP} model uses the scaling weights $\gpscale_\betaparama, \gpscale_\betaparamb, \gpscale_\betaparamc$, the scaling biases $\gpbias_\betaparama, \gpbias_\betaparamb, \gpbias_\betaparamc$, the coregionalization matrix $\coregion_\beta$ and the length scale parameter $\parameter$ as trainable parameters.

In practice, the authors in \cite{Song2019} use an approximate variational \ac{GP} for regression uncertainty calibration.
We can not use the standard log marginal likelihood for the training of the \ac{GP} parameters as we use intermediate latent functions that are passed through a non-linear exponential function and that do not have a ground-truth \cite{Song2019}.
Instead, it is necessary to utilize the posterior predictive distribution from Bayesian statistics to obtain an appropriate likelihood function.
Given the observations $\mathbf{\singlegroundtruthbbox}_\dataset = (\singlegroundtruthbbox_{1}, \ldots, \singlegroundtruthbbox_{\numsamples})^\T$ for the ground-truth bounding box positions in the data set $\dataset$ (for a single dimension), the model likelihood is given by
\begin{align}
	\pdf(\mathbf{\singlegroundtruthbbox}_\dataset|\dataset) 
	&= \int_{\scaleweightset} \pdf(\mathbf{\singlegroundtruthbbox}_\dataset | \scaleweightmat_\dataset, \dataset) \pdf(\scaleweightmat_\dataset | \dataset) \diff \scaleweightmat_\dataset \\
	\label{eq:regression:gpbeta:likelihood}
	&= \int_{\scaleweightset} \prod_{\indexsamples=1}^{\numsamples} \Big[ \pdf\big(\singlegroundtruthbbox_{\indexsamples} | \scalevec_\indexsamples, \mean_{\predbboxvariate|\allsingleinput_\indexsamples}, \variance_{\predbboxvariate|\allsingleinput_\indexsamples}\big) \Big] \pdf\big(\scaleweightmat_\dataset | \mean_{\predbboxvariate|\allsingleinput_\indexsamples}, \variance_{\predbboxvariate|\allsingleinput_\indexsamples}\big) \diff \scaleweightmat_\dataset ,
\end{align}
where
\begin{align}
    \pdf\big(\singlegroundtruthbbox_{\indexsamples} | \scalevec_\indexsamples, \mean_{\predbboxvariate|\allsingleinput_\indexsamples}, \variance_{\predbboxvariate|\allsingleinput_\indexsamples}\big) = \pdf_{\predbboxvariate|\allsingleinput_{\indexsamples}}(\singlegroundtruthbbox_{\indexsamples}) \cdot \linkfunction_\beta\big( \cdf_{\predbboxvariate|\allsingleinput_{\indexsamples}}(\singlegroundtruthbbox_{\indexsamples}) \big) ,
\end{align}
is the likelihood obtained using the link function $\linkfunction_\beta(\cdot)$ with weights $\scalevec_{\indexsamples}$ \cite{Song2019}.
Furthermore, the term $\pdf\big(\scaleweightmat_\dataset | \mean_{\predbboxvariate|\allsingleinput_\indexsamples}, \variance_{\predbboxvariate|\allsingleinput_\indexsamples}\big)$ is the Gaussian likelihood of the \ac{GP} model itself.

The \ac{GP} is approximate because the kernel matrix, which is constructed during the computation of the \ac{GP}, scales quadratically with the amount of training samples which yields the complexity $\complexity(\numsamples^2)$.
Therefore, the authors in \cite{Song2019} use a scalable inference scheme \cite{Snelson2006,Hensman2015} where a set of $\numinducing$ inducing points is learned to represent the data set so that each inducing point has an own mean $\mean_\indexinducing$ and variance $\variance_\indexinducing$ that needs to be learned. We refer to \cite{Snelson2006}, \cite{Hensman2015}, and \cite{Song2019} for a detailed discussion about approximate \ac{GP} models.

With this method, it is finally possible to perform \distributioncalibration{} by mapping uncalibrated Gaussians to calibrated non-parametric probability distributions.
Similar to the previously described Isotonic Regression method \cite{Kuleshov2018}, we start by generating a set of $\numstochastic$ points that describe the uncalibrated \ac{PDF} and \ac{CDF} denoted by $\dataset_{\indexsamples}^\ast = \big\{\big(\bbox_{\indexstochastic}, \pdf_{\predbboxvariate|\allsingleinput_{\indexsamples}}(\bbox_{\indexstochastic}),  \cdf_{\predbboxvariate|\allsingleinput_{\indexsamples}}(\bbox_{\indexstochastic})\big)\big\}^{\numstochastic}_{\indexstochastic=1}$.
Furthermore, we draw a set of weights from the trained \ac{GP} model given the new inference data.
The calibrated \ac{PDF} and \ac{CDF} are approximated for each point in $\dataset_{\indexsamples}^\ast$ by computing the average of the rescaled \ac{PDF} and \ac{CDF} estimates using $\linkfunction_\beta$ and $\calmodel_\beta$ in (\ref{eq:regression:methods:gpbeta:pdf}) and (\ref{eq:regression:methods:gpbeta:cdf}), respectively, given the sampled set of weights.
In this way, we can finally construct the recalibrated and non-parametric probability distribution for each sample during inference.

\subsection{Parametric Calibration}
\label{section:regression:methods:parametric}

In contrast to the previously introduced non-parametric calibration techniques, parametric methods seek for a recalibration using a known analytical representation of the probability distribution.
In this context, we present the Variance Scaling which is a parametric calibration method for normal distributions.

\subsubsection{Variance Scaling}
\label{section:regression:methods:parametric:variance}

The Variance Scaling method for \variancecalibration{} has independently been introduced by \cite{Levi2019} and \cite{Laves2020} and can be interpreted as a kind of temperature scaling \cite{Guo2018} from classification calibration for the uncalibrated variances of normal distributions.
In this way, a single scaling parameter $\scaleweight_{\stddev} \in \realdigitspositive$ is learned to yield a calibrated normal distribution by
\begin{align}
    \label{eq:regression:methods:varscaling:pdf}
    \pdfcalibrated_{\predbboxvariate|\allinputvariates}(\predbbox) = \normaldistribution\Big(\predbbox; \mean_{\predbboxvariate|\allinputvariates}, (\scaleweight_{\stddev} \cdot \stddev_{\predbboxvariate|\allinputvariates})^2\Big) ,
\end{align}
where the scaling parameter is a fixed constant after calibration training \cite{Levi2019,Laves2020}.
The parameter is trained using \ac{MLE} with the \ac{NLL} as the optimization objective by
\begin{align}
    \loss(\scaleweight_{\stddev}) 
    &= -\sum^\numsamples_{\indexsamples=1} \log \Bigg[ \frac{1}{\sqrt{2\pi} (\scaleweight_{\stddev} \cdot \stddev_{\predbboxvariate|\allsingleinput_\indexsamples})} \exp\Big(-\frac{1}{2}(\scaleweight_{\stddev} \cdot \stddev_{\predbboxvariate|\allsingleinput_\indexsamples})^{-2} (\singlegroundtruthbbox_{\indexsamples} - \mean_{\predbboxvariate|\allsingleinput_\indexsamples})^2\Big) \Bigg] \\
    &\propto -\numsamples \log(\scaleweight_{\stddev}) - \frac{1}{2\scaleweight_{\stddev}^2} \sum^\numsamples_{\indexsamples=1} \stddev_{\predbboxvariate|\allsingleinput_\indexsamples}^{-2} (\singlegroundtruthbbox_{\indexsamples} - \mean_{\predbboxvariate|\allsingleinput_\indexsamples})^2 ,
\end{align}
which is to be minimized.
Thus, we seek to get the minimum of the optimization objective which can analytically be determined using its derivative $\quad\frac{\diff \loss(\scaleweight_{\stddev})}{\diff \scaleweight_{\stddev}} = 0$ \cite{Laves2020}, so that
\begin{align}
    &-\frac{\numsamples}{\scaleweight_{\stddev}} + \frac{1}{\scaleweight_{\stddev}^3} \sum^\numsamples_{\indexsamples=1} \stddev_{\predbboxvariate|\allsingleinput_\indexsamples}^{-2} (\singlegroundtruthbbox_{\indexsamples} - \mean_{\predbboxvariate|\allsingleinput_\indexsamples})^2 = 0 \\
    &\Leftrightarrow\quad-\numsamples\scaleweight_{\stddev}^2 \sum^\numsamples_{\indexsamples=1} \stddev_{\predbboxvariate|\allsingleinput_\indexsamples}^{-2} (\singlegroundtruthbbox_{\indexsamples} - \mean_{\predbboxvariate|\allsingleinput_\indexsamples})^2 = 0 \\
    &\Leftrightarrow\quad \scaleweight_{\stddev} = \pm \sqrt{\frac{1}{\numsamples} \sum^\numsamples_{\indexsamples=1} \stddev_{\predbboxvariate|\allsingleinput_\indexsamples}^{-2} (\singlegroundtruthbbox_{\indexsamples} - \mean_{\predbboxvariate|\allsingleinput_\indexsamples})^2}
\end{align}
gives us the analytical solution for $\scaleweight_{\stddev}$ given a certain set $\dataset$ of training samples \cite{Laves2020}, where the data set $\dataset = \big\{\big(\singlegroundtruthbbox_{\indexsamples}, \mean_{\predbboxvariate|\allsingleinput_\indexsamples}, \variance_{\predbboxvariate|\allsingleinput_\indexsamples}\big)\big\}^{\numsamples}_{\indexsamples=1}$ with $\numsamples$ samples consists of the ground-truth $\singlegroundtruthbbox_{\indexsamples}$ as well as of the predicted mean $\mean_{\predbboxvariate|\allsingleinput_\indexsamples}$ and variance $\variance_{\predbboxvariate|\allsingleinput_\indexsamples}$.

\section{Parametric Calibration using Gaussian Processes}
\label{section:regression:methods:parametric:gp}

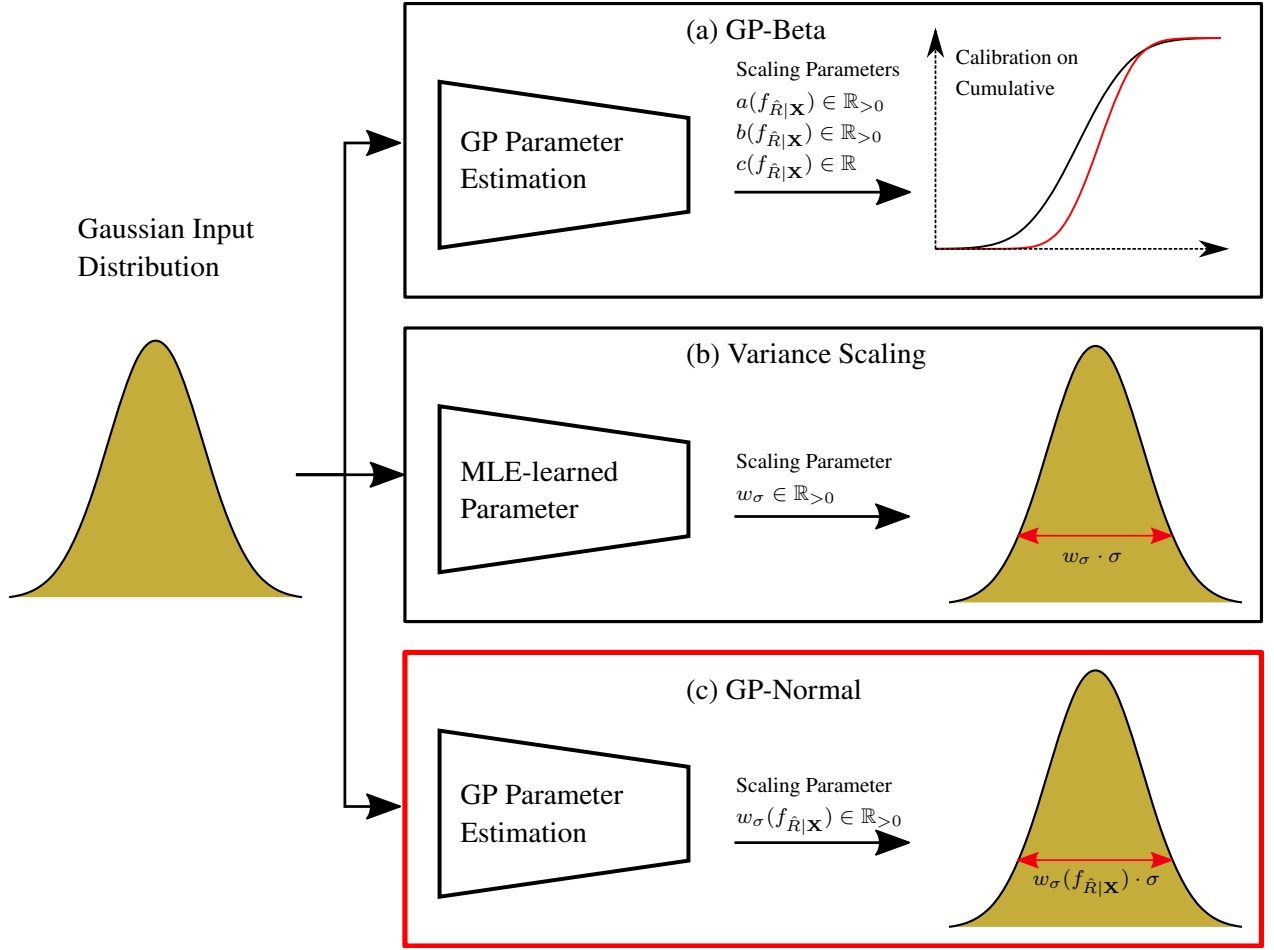
\begin{figure}[t!]
    \centering
    \begin{overpic}[width=1.0\linewidth]{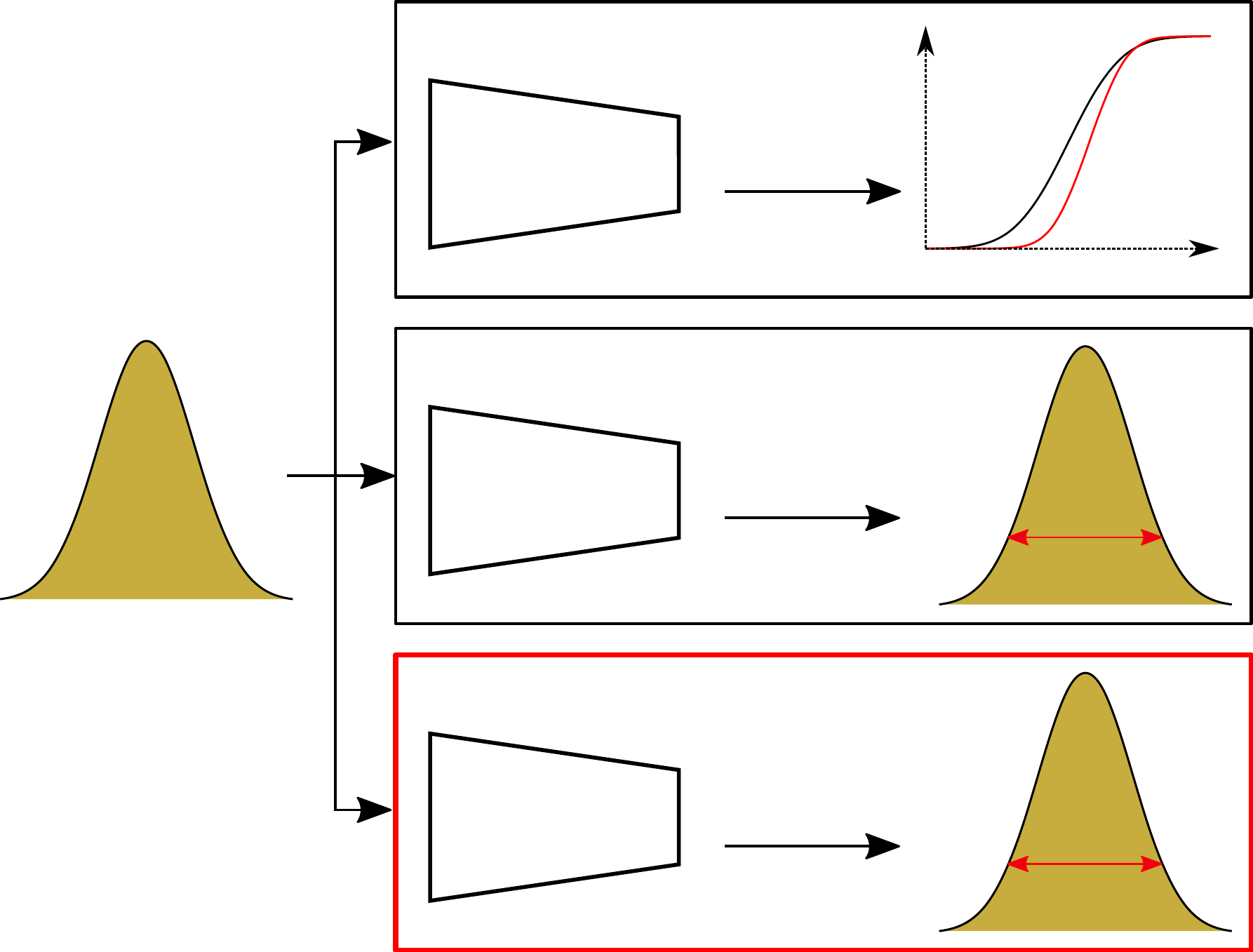}
        \put(5.5, 57){Gaussian Input}
        \put(5.5, 54){Distribution}
        
        \put(54, 73){(a) GP-Beta}
        \put(54, 47){(b) Variance Scaling}
        \put(54, 20){(c) GP-Normal}
        
        \put(36, 64){GP Parameter}
        \put(36, 61){Estimation}
        
        \put(36, 37.5){MLE-learned}
        \put(36, 34.5){Parameter}
        
        \put(36, 11.5){GP Parameter}
        \put(36, 8.5){Estimation}
        
        \put(75.5, 71){\scriptsize Calibration on}
        \put(75.5, 68.5){\scriptsize Cumulative}
        
        \put(58, 70){\scriptsize Scaling Parameters}
        \put(58, 67.5){\scriptsize $\betaparama(\pdf_{\predbboxvariate|\allinputvariates}) \in \realdigitspositive$}
        \put(58, 65){\scriptsize $\betaparamb(\pdf_{\predbboxvariate|\allinputvariates}) \in \realdigitspositive$}
        \put(58, 62.5){\scriptsize $\betaparamc(\pdf_{\predbboxvariate|\allinputvariates}) \in \realdigits$}
        
        \put(58, 38.5){\scriptsize Scaling Parameter}
        \put(58, 36){\scriptsize $\scaleweight_{\stddev} \in \realdigitspositive$}
        
        \put(58, 12.5){\scriptsize Scaling Parameter}
        \put(58, 10){\scriptsize $\scaleweight_{\stddev}(\pdf_{\predbboxvariate|\allinputvariates}) \in \realdigitspositive$}
        
        \put(84, 31){\scriptsize $\scaleweight_{\stddev} \cdot \stddev$}
        \put(81.65, 5){\scriptsize $\scaleweight_{\stddev}(\pdf_{\predbboxvariate|\allinputvariates}) \cdot \stddev$}
    \end{overpic}
    \caption[Schematic representation of the work principle of various calibration methods and how we derive the GP-Normal method.]{
        Schematic representation of the work principle of various calibration methods and how we derive the GP-Normal method.
        (a) The GP-Beta \cite{Song2019} is a non-parametric calibration method that uses a Gaussian Process (GP) to flexibly obtain the calibration parameters based on the input distribution.
        (b) In contrast, the Variance Scaling \cite{Levi2019,Laves2020} yields a parametric normal distribution as calibration output but only uses a single scaling parameter for the recalibration of the input variance.
        (c) Our GP-Normal method also yields a Gaussian as calibration output but uses the flexible GP parameter estimation scheme of the GP-Beta.
        In this way, it is possible to keep a parametric Gaussian representation of the input distribution as well as to perform input-sensitive uncertainty calibration.
    }
    \label{fig:regression:blockimage}
\end{figure}

\begin{figure}[t!]
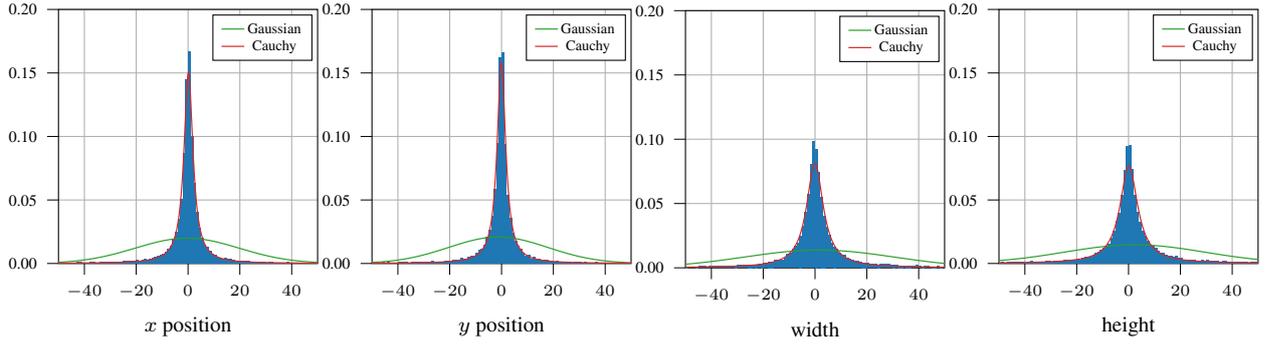

    \centering
    \def\stackalignment{r}
        \begin{subfigure}[t]{0.25\textwidth}
            \input{images/regression/error_distribution/error_histogram_dim_0.tikz}
        \end{subfigure}%
        \begin{subfigure}[t]{0.25\textwidth}
            \input{images/regression/error_distribution/error_histogram_dim_1.tikz}
        \end{subfigure}%
        \begin{subfigure}[t]{0.25\textwidth}
            \input{images/regression/error_distribution/error_histogram_dim_2.tikz}
        \end{subfigure}%
        \begin{subfigure}[t]{0.25\textwidth}
            \input{images/regression/error_distribution/error_histogram_dim_3.tikz}
        \end{subfigure}
    \caption[Error distribution of the regression output obtained by a \fasterrcnn{} object detector on the MS COCO data set] {
        Error distribution of the regression output $\allpredbboxvariates$ obtained by a \fasterrcnn{} object detector on the MS COCO data set \cite[p. 8, Fig.~3]{Kueppers2022b}.
        The Gaussian fit does not yield a good representation of the error distribution.
        In contrast, the Cauchy distribution is more suitable as it allows for heavier tails compared to a Gaussian.
    }
    \label{fig:regression:methods:error_dist}
\end{figure}

On the one hand, we seek to keep parameterized probability distributions after calibration.
Although non-parametric distributions might have a better representational power of the underlying data distribution, it is often advantageous to use parametric distributions especially for subsequent processes such as object tracking (cf. \chapref{chapter:tracking}).
On the other hand, we seek for more flexibility during calibration compared to the previously introduced Variance Scaling \cite{Levi2019,Laves2020} to finally achieve \distributioncalibration{}.
For this reason, we adapt the recalibration framework provided by \cite{Song2019} which is based on the \ac{GP} recalibration.
Instead of a non-parametric Beta Calibration on the \ac{CDF}, we also adapt the Variance Scaling scheme by \cite{Levi2019,Laves2020} but seek to obtain the scaling parameter $\scaleweight_{\stddev}(\pdf_{\predbboxvariate|\allinputvariates})$ by a \ac{GP} as a function of the input distribution $\pdf_{\predbboxvariate|\allinputvariates}$ to the calibration method, so that
\begin{align}
    \log\big(\scaleweight_{\stddev}(\cdot)\big) \sampledfrom \gp\big(0, \kernel(\cdot, \cdot)\big) ,
\end{align}
which finally yields a similar recalibrated \ac{PDF} as within (\ref{eq:regression:methods:varscaling:pdf}) but with an input-dependent scaling weight
\begin{align}
    \label{eq:regression:methods:gpnormal:pdf}
    \pdfcalibrated_{\predbboxvariate|\allinputvariates}(\predbbox) = \normaldistribution\Big(\predbbox; \mean_{\predbboxvariate|\allinputvariates}, (\scaleweight_{\stddev}(\pdf_{\predbboxvariate|\allinputvariates}) \cdot \stddev_{\predbboxvariate|\allinputvariates})^2\Big) .
\end{align}
This method is further denoted by GP-Normal.
We use the exponential function to guarantee $\scaleweight_{\stddev}(\pdf_{\predbboxvariate|\allinputvariates}) \in \realdigitspositive$.
Since we only infer a single parameter, a coregionalization matrix $\coregion$ is not necessary any more.
This method offers the flexibility of a \ac{GP} towards \distributioncalibration{} and preserves a parametric Gaussian distribution for subsequent processes.
Therefore, the GP-Normal can be seen as an extension of the definition for \variancecalibration{} towards \distributioncalibration{} but for normally distributed data.
Thus, under the assumption of normally distributed data, the GP-Normal seeks to achieve
\begin{align}
    \expectation_{\allinputvariates,\groundtruthbboxvariate} \big[ (\groundtruthbboxvariate - \mean_{\predbboxvariate|\allinputvariates})^2 \big| \distvariate_{\groundtruthbboxvariate} = \normaldistribution(\mean, \variance) \big] = \variance ,
\end{align}
for all $\mean \in \realdigits$ and for all $\variance \in \realdigitspositive$.
The concept for the GP-Normal method is schematically shown in \figref{fig:regression:blockimage}.
Basically, we can plug in any desired parametric probability distribution.
For our probabilistic object detection model from the experiments in \secref{section:regression:experiments}, we observe error distributions between the predicted mean and the ground-truth scores that are rather Cauchy-distributed than normally distributed.
This is shown in \figref{fig:regression:methods:error_dist}.
Therefore, we can also use the \ac{GP} to construct a calibrated Cauchy distribution with location $\cauchymode \in \realdigits$ and scale $\cauchyscale \in \realdigitspositive$, where the location parameter is approximated using the uncalibrated mean $\mean_{\predbboxvariate|\allinputvariates}$ and the scale parameter is obtained by
\begin{align}
    \log\big(\scaleweight_{\cauchyscale}(\cdot)\big) \sampledfrom \gp\big(0, \kernel(\cdot, \cdot)\big) ,
\end{align}
so that a calibrated \ac{PDF} is given by
\begin{align}
    \pdfcalibrated_{\predbboxvariate|\allinputvariates}(\predbbox) = \cauchydistribution\big(\predbbox; \cauchymode = \mean_{\predbboxvariate|\allinputvariates}, \cauchyscale = (\scaleweight_\cauchyscale(\pdf_{\predbboxvariate|\allinputvariates}) \cdot \stddev_{\predbboxvariate|\allinputvariates})\big) .
\end{align}
This method is denoted by GP-Cauchy.

\subsection{Joint Multivariate Calibration}
\label{section:regression:methods:multivariate}

For the task of object detection, it is required to determine the object position and shape which results in a multivariate regression problem.
Most approaches for probabilistic object detection use independent normal distributions for each bounding box dimension \cite{Hall2020,He2019,Harakeh2020}.
Therefore, we further extend the \ac{GP} recalibration methods GP-Beta, GP-Normal, and GP-Cauchy to jointly infer the calibrated distributions using all available input information.
Similar to the position-dependent confidence calibration in \secref{section:confidence:methods}, the joint recalibration allows to capture possible correlations between the input quantities.
For this reason, we adapt the \ac{GP} recalibration scheme by \cite{Song2019} and use a multi-output \ac{GP} \cite{Alvarez2008,Moreno2018,Skolidis2011} which estimates the latent functions for each dimension given the uncalibrated input distributions, where $\numbboxdims$ is the number of bounding box dimensions.
For the GP-Beta method \cite{Song2019}, the parameter estimation for the multivariate case extends to 
\begin{align}
    \scalevec_\betaparama(\cdot), \scalevec_\betaparamb(\cdot), \scalevec_\betaparamc(\cdot) \sampledfrom \gp\big(0, \mvkernel(\cdot, \cdot), \coregion_\beta\big) ,
\end{align}
with scaling functions $\scalevec_\betaparama(\pdf_{\allpredbboxvariates|\allinputvariates}), \scalevec_\betaparamb(\pdf_{\allpredbboxvariates|\allinputvariates}), \scalevec_\betaparamc(\pdf_{\allpredbboxvariates|\allinputvariates}) \in \realdigits^{\numbboxdims}$ that use the uncalibrated multivariate distributions as input and return a scaling weight for each bounding box dimension $\numbboxdims$.
In this case, the coregionalization matrix $\coregion_\beta$ captures the dependencies between all parameters and all dimensions, so that it is given by $\coregion_\beta \in \realdigits^{3\numbboxdims \times 3\numbboxdims}$.
Furthermore, since we now work with a multivariate input with mean vector $\meanvec_{\allpredbboxvariates|\allinputvariates}$ and (diagonal) covariance matrix $\cov_{\allpredbboxvariates|\allinputvariates}$, we use the original (multivariate) formulation for the kernel function $\mvkernel$ \cite{Song2008} which is given by
\begin{align}
    \mvkernel\big((\meanvec_i, \cov_i), (\meanvec_j, \cov_j)\big) = \parameter^{\numbboxdims} |\cov_{ij}|^{-\frac{1}{2}} \exp\Big(-\frac{1}{2}(\meanvec_i - \meanvec_j)^\T\cov_{ij}^{-1}(\meanvec_i-\meanvec_j)\Big) ,
\end{align}
where $\cov_{ij} = \cov_i + \cov_j + \parameter^2\identity$ with identity matrix $\identity \in \realdigits^{\numbboxdims \times \numbboxdims}$ \cite{Song2008}.
Similarly, the parameter estimation for the GP-Normal in the multivariate case extends to
\begin{align}
    \label{eq:regression:methods:gpnormal}
    \log\big(\scalevec_{\cov}(\cdot)\big) \sampledfrom \gp\big(0, \mvkernel(\cdot, \cdot), \coregion_{\normaldistribution}\big) ,
\end{align}
as well as for the GP-Cauchy by
\begin{align}
    \log\big(\scalevec_{\cauchyscalevec}(\cdot)\big) \sampledfrom \gp\big(0, \mvkernel(\cdot, \cdot), \coregion_{\text{C}}\big) ,
\end{align}
with $\scalevec_{\cov}(\pdf_{\allpredbboxvariates|\allinputvariates}), \scalevec_{\cauchyscalevec}(\pdf_{\allpredbboxvariates|\allinputvariates}) \in \realdigits^{\numbboxdims}$ as the functions that return a scale weight $\scaleweight_{\stddev}^{(\indexbboxdims)}$ and $\scaleweight_{\cauchyscale}^{(\indexbboxdims)}$ for each bounding box dimension $\indexdims \in \{1, \ldots, \numbboxdims\}$.
Here, the coregionalization matrices are given by $\coregion_{\normaldistribution}, \coregion_{\text{C}} \in \realdigits^{\numbboxdims \times \numbboxdims}$.
Although we aim to capture possible correlations of the input for calibration, the resulting calibrated probability distributions are still modeled using independent distributions, so that e.g. the calibrated normal distribution is given by
\begin{align}
    \pdfcalibrated_{\allpredbboxvariates|\allinputvariates}(\allpredbboxes) = \prod^\numbboxdims_{\indexbboxdims=1} \normaldistribution\Big(\predbbox_\indexbboxdims; \mean_{\predbboxvariate_\indexbboxdims|\allinputvariates}, \big(\scalevec_{\cov}(\pdf_{\allpredbboxvariates|\allinputvariates})^{(\indexbboxdims)} \cdot \stddev_{\predbboxvariate_\indexbboxdims|\allinputvariates}\big)^2\Big) .
\end{align}
This holds for all cases, the non-parametric GP-Beta as well as for the parametric GP-Normal or GP-Cauchy.
Thus, the likelihood of the \ac{GP} for parameter training is simply the product of the likelihood functions for each bounding box dimension. \big(cf. (\ref{eq:regression:gpbeta:likelihood})\big).
Therefore, it is now possible to jointly infer multiple dimensions using the \ac{GP} recalibration scheme.

\subsection{Correlation Estimation and Recalibration}
\label{section:regression:methods:correlations}

Besides the inference of multiple independent probability distributions, it is also possible to infer parametric probability distributions with correlations between the dimensions using the \ac{GP} recalibration scheme.
This allows for a post-hoc introduction of correlations between independently inferred random variables.
In this section, we use the multivariate normal distribution to introduce a correlation estimation scheme.
In a first step, we compute the marginal correlation coefficients $\correlation_{ij} \in [-1, 1]$ between all di\-men\-sions $i, j \in \{1, \ldots, \numbboxdims\}$.
The correlation coefficients are used to compute the full covariance matrices $\cov_{\allpredbboxvariates|\allinputvariates}$ with covariances ${\stddev_{ij} = \correlation_{ij} \stddev_i \stddev_j}$ for all samples which can be interpreted as a prior for the calibrated covariances.
Afterwards, we use the $\decomposed\diagonal\decomposed^\T$ decomposition of $\cov$ where $\decomposed$ is a lower triangular matrix and $\diagonal$ the respective diagonal.
Using the decomposed representation, we can obtain the input-dependent scale weights $\scalevec_\decomposed(\pdf_{\allpredbboxvariates|\allinputvariates}) \in \realdigits^{\numbboxdims \times \numbboxdims}$ and $\scalevec_\diagonal(\pdf_{\allpredbboxvariates|\allinputvariates}) \in \realdigitspositive^{\numbboxdims \times \numbboxdims}$ for the lower triangular and the diagonal matrix, respectively, by
\begin{align}
    \log\big(\scalevec_\diagonal(\cdot)\big), \scalevec_\decomposed(\cdot) \sampledfrom \gp\big(0, \mvkernel(\cdot, \cdot), \coregion_{\normaldistribution}\big) .
\end{align}
This allows for a reconstruction of a calibrated covariance matrix by
\begin{align}
    \cov = \big(\scalevec_\decomposed(\pdf_{\allpredbboxvariates|\allinputvariates}) \elementwiseprod \decomposed\big) \big(\scalevec_\diagonal(\pdf_{\allpredbboxvariates|\allinputvariates}) \elementwiseprod \diagonal\big)  \big(\scalevec_\decomposed(\pdf_{\allpredbboxvariates|\allinputvariates}) \elementwiseprod \decomposed\big)^\T ,
\end{align}
where $\elementwiseprod$ denotes the element-wise product.
The rescaling of a $\decomposed\diagonal\decomposed^\T$ decomposed matrix is numerically more stable as it preserves a symmetric and positive semidefinit covariance matrix after calibration.
The \ac{NLL} of the multivariate \ac{GP} is simply obtained by the likelihood of the multivariate normal distribution.
Therefore, it is now possible not only to jointly recalibrate multiple dimensions but also to introduce correlations between independently inferred random variables.
This method can also be used for a covariance recalibration as well, if the input is modeled with a non-diagonal covariance matrix.
In this case, we can omit the first step of computing the marginal correlation coefficients and directly use the provided covariances as the priors for the \ac{GP} recalibration.
Similar to the standard GP-Normal method, covariance estimation and covariance recalibration can be seen as the multivariate extension of the definition for \variancecalibration{} towards multivariate \distributioncalibration{} using normal distributions, so that
\begin{align}
	\expectation_{\allinputvariates,\groundtruthbboxvariate} \big[ ( \allgroundtruthbboxvariates - \meanvec_{\allpredbboxvariates | \allinputvariates} )(\allgroundtruthbboxvariates - \meanvec_{\allpredbboxvariates | \allinputvariates} )^\T \big| \distvariate_{\allgroundtruthbboxvariates} = \normaldistribution(\meanvec, \cov) \big] = \cov ,
\end{align}
is fulfilled for all $\meanvec \in \bboxset$ and all $\cov \in \realdigits^{\numdims \times \numdims}$, where $\cov = \cov^\T$ and $\cov \succeq 0$.
\newpage
\section{Experiments for Spatial Uncertainty Calibration} 
\label{section:regression:experiments}

The evaluations presented in this section are part of our publication in \cite[pp.~11]{Kueppers2022b}.
We use a probabilistic \fasterrcnn{} as well as a \retinanet{} that output Gaussian distributions with mean and variance for each bounding box quantity $\centerx$, $\centery$, width, and height.
The basic architectures for both networks have been adapted from \cite{Wu2019} and extended by a probabilistic bounding box regression output (cf. \secref{section:introduction:spatial}).
We trained both networks on the MS COCO \cite{Lin2014} and Berkeley DeepDrive \cite{Yu2018} training data sets following the standard training configuration by \cite{Wu2019}.
Similar to our previous experiments for semantic confidence calibration in \secref{section:confidence:experiments}, we use the predictions on the respective validation data sets for calibration training and evaluation.
For this reason, the images of the validation sets are divided randomly which yields equally sized sets with probabilistic predictions for calibration training and evaluation, respectively.
We start by examining if the object detection model makes unbiased predictions for the object bounding boxes.
The results are shown in \tabref{tab:regression:bias}.
\begin{table}[b!]
    \centering
    \caption[Bias and the respective error standard deviation of the object detection models on the data sets for all bounding box quantities (in pixels).]{
        Bias and the respective error standard deviation of the object detection models on the data sets for all bounding box quantities (in pixels).
        We observe a small bias in each model which, however, is in a range where it does not have a large impact on our uncertainty evaluation results.
    }
     \begin{tabular}{l|c||c|c|c|c}
    & DB & Num. Detections & Quantity & Bias & RMSE \\ \hline \hline
    \multirow{8}{*}{\rotatebox[origin=c]{90}{\fasterrcnn{}}} & \multirow{4}{*}{BDD} & \multirow{4}{*}{69.775} & $\centerx$ & 0.165 & 7.637 \\
    & & & $\centery$ & 0.190 & 7.857 \\
    & & & $\width$ & -0.300 & 10.602 \\
    & & & $\height$ & -0.439 & 10.441 \\ \cline{2-6}
    & \multirow{4}{*}{COCO} & \multirow{4}{*}{24.807} & $\centerx$ & 0.604 & 20.406 \\
    & & & $\centery$ & 1.455 & 20.085 \\
    & & & $\width$ & -1.716 & 30.598 \\
    & & & $\height$ & -2.509 & 27.672 \\ \hline
    \multirow{8}{*}{\rotatebox[origin=c]{90}{\retinanet{}}} & \multirow{4}{*}{BDD} & \multirow{4}{*}{108.133} & $\centerx$ & 0.351 & 9.531 \\
    & & & $\centery$ & 0.196 & 8.199 \\
    & & & $\width$ & -0.640 & 12.467 \\
    & & & $\height$ & -0.801 & 11.763 \\ \cline{2-6}
    & \multirow{4}{*}{COCO} & \multirow{4}{*}{50.274} & $\centerx$ & 1.040 & 26.200 \\
    & & & $\centery$ & 3.180 & 26.622 \\
    & & & $\width$ & -1.849 & 37.102 \\
    & & & $\height$ & -5.187 & 34.999 \\ \hline
\end{tabular}

    \label{tab:regression:bias}
\end{table}
For each object detection model, we observe a small bias in the estimated object position.
However, compared to the error variance, the bias is in a range where it does not have a large impact on our further evaluations.

For our experiments, we adopt Isotonic Regression \cite{Kuleshov2018}, Variance Scaling \cite{Levi2019,Laves2020}, GP-Beta \cite{Song2019}, and our new parametric GP-Normal and GP-Cauchy methods.
As calibration metrics, we use the \ac{NLL}, Pinball loss, \ac{C-QCE}, \ac{UCE} \cite{Laves2020}, and \ac{ENCE} \cite{Levi2019}.
Each metric is reported as the average over all bounding box quantities.
Since the Pinball loss and \ac{C-QCE} are computed for certain quantiles, we further report a mean Pinball loss and a mean \ac{C-QCE} for quantile levels from $0.05$ to $0.95$ with steps of $0.05$.
Furthermore, the \ac{C-QCE}, \ac{UCE}, and \ac{ENCE} require a binning scheme over the predicted variance/standard deviation, so that we use $\numbins=20$ bins for binning.
Note that the GP-Cauchy calibration method yields a Cauchy distribution which, however, has no statistical moments such as expectation and variance.
Thus, we can not compute the \ac{UCE} and \ac{ENCE} after calibration with GP-Cauchy.

Subsequently, we use the GP-Normal method to perform covariance estimation (cf. \secref{section:regression:methods:correlations}).
This variant is denoted by GP-Normal (mv.).
In this way, it is possible to capture possible correlations between the bounding box quantities.
For the multivariate evaluation, we also use the \ac{NLL} and the \ac{C-QCE}, where the latter one is currently only defined for Gaussian distributions.
Except for the multivariate GP-Normal, the \ac{NLL} is obtained by assuming independent probability distributions for each bounding box quantity.
The calibration results for the univariate as well as for the multivariate case are presented in \tabref{tab:regression:evaluation}.
Furthermore, we show the reliability diagram in \figref{fig:regression:evaluation:reliability} which allows for an evaluation of the probabilistic forecaster by means of \quantilecalibration{}.
\begin{table}[t!]
    \centering
    \caption[Calibration results for the probabilistic bounding boxes of \fasterrcnn{} and \retinanet{} object detectors before and after regression uncertainty calibration.]{
        Calibration results for the probabilistic bounding boxes of \fasterrcnn{} and \retinanet{} object detectors before and after regression uncertainty calibration \cite[p.~12, Tab.~1]{Kueppers2022b}.
        The best calibration scores are highlighted in bold.
        In all cases, we observe a miscalibration for the uncalibrated regression uncertainty.
        Furthermore, Isotonic Regression is able to achieve the best calibration performance in terms of \quantilecalibration{} (cf. Pinball loss and \ac{C-QCE}) compared to the remaining calibration methods.
        In contrast, Variance Scaling, GP-Normal, and GP-Beta achieve the best results for \variancecalibration{} (cf. UCE and ENCE).
    }
    \begin{tabular}{l|c|l||c|c|c|c|c||c|c}
    \multicolumn{3}{c||}{Setup} & \multicolumn{5}{c||}{Univariate} & \multicolumn{2}{c}{Multivariate}\\ \hline
    & DB & Method & NLL & $\overline{\text{C-QCE}}$ & $\overline{\loss_\text{Pin}}$ & UCE & ENCE & NLL & $\overline{\text{C-QCE}}$ \\ \hline \hline
    \multirow{14}{*}{\rotatebox[origin=c]{90}{\fasterrcnn{}}} & \multirow{7}{*}{BDD} & Uncalibrated & 3.053 & 0.040 & 1.079 & 19.683 & 0.454 & 12.210 & \textbf{0.071} \\
    &  & Isotonic Reg.& \textbf{2.895} & \textbf{0.017} & \textbf{1.059} & 39.157 & 0.303 & \textbf{11.579} & - \\
    &  & GP-Beta & 2.941 & 0.057 & 1.077 & 3.635 & 0.199 & 11.764 & - \\
    &  & Var. Scaling & 2.962 & 0.061 & 1.086 & 3.361 & \textbf{0.175} & 11.848 & 0.131\\
    &  & GP-Normal & 2.962 & 0.059 & 1.084 & 3.289 & 0.188  & 11.848 & 0.128\\
    &  & GP-Normal (mv.) & 2.968 & 0.054 & 1.150 & \textbf{3.234} & 0.191 & 11.584 & 0.133 \\
    &  & GP-Cauchy & 3.011 & 0.050 & 1.189 & - & - & 12.045 & - \\ \cline{2-10}
    & \multirow{7}{*}{COCO} & Uncalibrated & 3.561 & 0.154 & 3.055 & 32.899 & 0.096 & 14.245 & 0.256 \\
    &  & Isotonic Reg.& \textbf{3.340} & \textbf{0.020} & \textbf{2.715} & 42.440 & 0.121 & \textbf{13.360} & - \\
    &  & GP-Beta & 3.412 & 0.074 & 2.750 & 51.455 & 0.140 & 13.649 & - \\
    &  & Var. Scaling & 3.554 & 0.131 & 2.952 & 33.155 & 0.093 & 14.216 & 0.222\\
    &  & GP-Normal & 3.554 & 0.130 & 2.949 & 48.167 & 0.132 & 14.235 & \textbf{0.200}\\
    &  & GP-Normal (mv.) & 3.562 & 0.121 & 3.298 & \textbf{28.382} & \textbf{0.087} & 13.955 & 0.216\\ 
    &  & GP-Cauchy & 3.406 & 0.039 & 2.897 & - & - & 13.624 & - \\ \hline \hline
    \multirow{14}{*}{\rotatebox[origin=c]{90}{\retinanet{}}} & \multirow{7}{*}{BDD} & Uncalibrated & 4.052 & 0.130 & 1.847 & 39.933 & 0.491 & 16.208 & 0.234 \\
    & & Isotonic Reg.& \textbf{3.224} & \textbf{0.068} & \textbf{1.814} & 39.498 & 0.252 & \textbf{12.898} &  - \\
    & & GP-Beta & 3.419 & 0.091 & 2.169 & \textbf{14.892} & \textbf{0.173} & 13.677 & - \\
    & & Var. Scaling & 3.392 & 0.095 & 2.203 & 15.833 & 0.180 & 13.568 & \textbf{0.131} \\
    & & GP-Normal & 3.392 & 0.095 & 2.205 & 15.820 & 0.180 & 13.568 & \textbf{0.131} \\
    & & GP-Normal (mv.) & 3.434 & 0.097 & 3.356 & 15.342 & 0.200 & 13.353 & 0.133 \\
    & & GP-Cauchy & 3.316 & 0.097 & 1.944 & - & - & 13.262 & - \\ \cline{2-10}
    & \multirow{7}{*}{COCO} & Uncalibrated & 4.694 & 0.115 & 4.999 & 553.244 & 0.530 & 18.778 & \textbf{0.153} \\
    & & Isotonic Reg.& \textbf{3.886} & \textbf{0.029} & \textbf{4.534} & 105.343 & 0.113 & \textbf{15.544} & - \\
    & & GP-Beta & 4.169 & 0.142 & 5.093 & \textbf{77.017} & \textbf{0.071} & 16.677 & - \\
    & & Var. Scaling & 4.204 & 0.167 & 5.606 & 83.653 & 0.072 & 16.815 & 0.207 \\
    & & GP-Normal & 4.204 & 0.167 & 5.606 & 84.367 & 0.072 & 16.815 & 0.207 \\
    & & GP-Normal (mv.) & 4.236 & 0.158 & 6.233 & 82.074 & 0.087 & 16.455 & 0.194 \\ 
    & & GP-Cauchy & 3.936 & 0.043 & 4.716 & - & - & 15.745 & - \\
    \hline
\end{tabular}
    
    \label{tab:regression:evaluation}
\end{table}
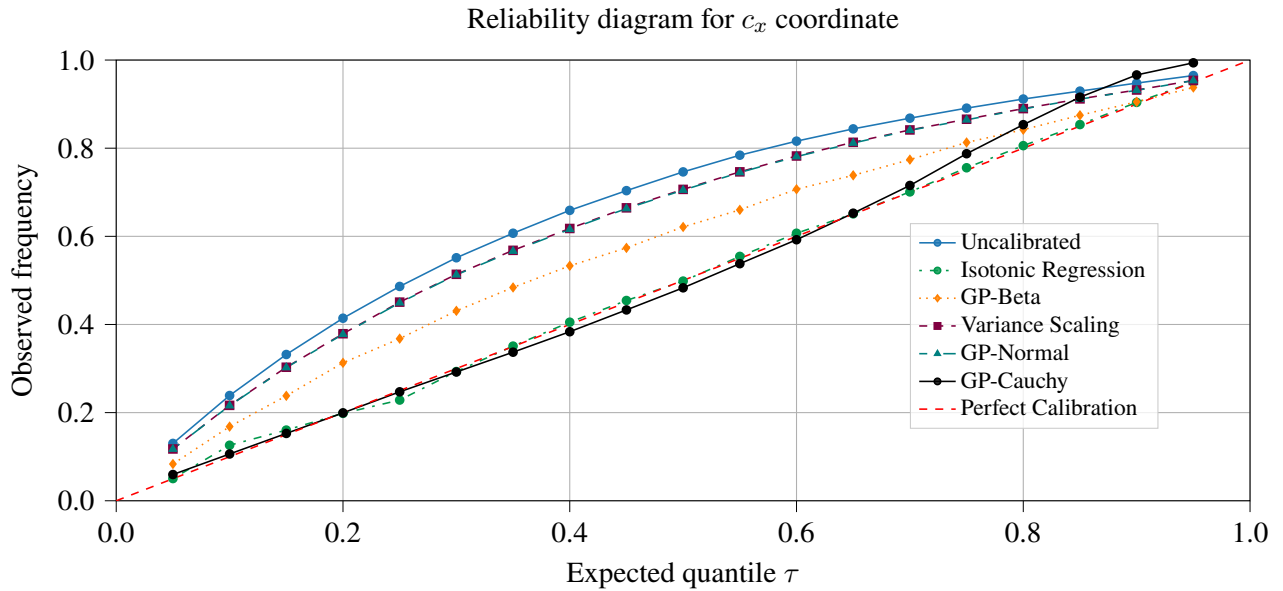
\begin{figure}[t]
    \centering
\begin{tikzpicture}

\definecolor{color0}{rgb}{0.12156862745098,0.466666666666667,0.705882352941177}
\definecolor{color1}{rgb}{0.0,0.597656,0.296875}
\definecolor{color2}{rgb}{1.0,0.5,0.0}
\definecolor{color3}{rgb}{0.5,0.0,0.25}
\definecolor{color4}{rgb}{0.0,0.5,0.5}
\definecolor{color5}{rgb}{0.0,0.0,0.0}

\begin{axis}[
width=\linewidth,
height=0.45\linewidth,
legend cell align={left},
legend style={
    fill opacity=0.8,
    draw opacity=1,
    text opacity=1,
    at={(0.7,0.63)},
    anchor=north west,
    draw=white!80!black,
    nodes={scale=0.8, transform shape},
},
legend image post style={scale=0.8},
tick align=outside,
tick pos=left,
title={Reliability diagram for $\centerx$ coordinate},
x grid style={white!69.0196078431373!black},
xlabel={Expected quantile $\quantile$},
xmajorgrids,
xmin=0, xmax=1,
xtick style={color=black},
xtick={0,0.2,0.4,0.6,0.8,1},
xticklabels={0.0,0.2,0.4,0.6,0.8,1.0},
y grid style={white!69.0196078431373!black},
ylabel={Observed frequency},
ymajorgrids,
ymin=0, ymax=1,
ytick style={color=black},
ytick={0,0.2,0.4,0.6,0.8,1},
yticklabels={0.0,0.2,0.4,0.6,0.8,1.0}
]
\addplot [semithick, color0, solid, mark=*, mark size=1.5, mark options={solid}]
table {%
    0.05 0.130131616949292
    0.1 0.23880843338697
    0.15 0.331889260221975
    0.2 0.414201427569419
    0.25 0.486322564673846
    0.3 0.551388373148492
    0.35 0.60704707678343
    0.4 0.658951190328836
    0.45 0.703676197549202
    0.5 0.746214465486653
    0.55 0.784090440236003
    0.6 0.815983826381153
    0.65 0.843833807814498
    0.7 0.868094236085324
    0.75 0.890869332013038
    0.8 0.911540207121343
    0.85 0.929529232165697
    0.9 0.947683294137063
    0.95 0.964764616082849
};
\addlegendentry{Uncalibrated}

\addplot [semithick, dash pattern=on 1pt off 3pt on 3pt off 3pt, color1, mark=*, mark size=1.5, mark options={solid}]
table {%
    0.05 0.0503775219705409
    0.1 0.125964434542229
    0.15 0.160374633824318
    0.2 0.198333127037175
    0.25 0.228493625448694
    0.3 0.293188100837562
    0.35 0.350909766060156
    0.4 0.405165655815489
    0.45 0.454387919296943
    0.5 0.498452778809259
    0.55 0.554482815529975
    0.6 0.607005817551677
    0.65 0.650988158600487
    0.7 0.700829310558237
    0.75 0.755580311094607
    0.8 0.805751536906383
    0.85 0.853612245739984
    0.9 0.903907249247019
    0.95 0.950158848042249
};
\addlegendentry{Isotonic Regression}

\addplot [semithick, dotted, color2, mark=diamond*, mark size=1.5, mark options={solid}]
table {%
    0.05 0.0833849073730247
    0.1 0.168255147089161
    0.15 0.238107026447168
    0.2 0.312992532079053
    0.25 0.367867310310682
    0.3 0.430952675661179
    0.35 0.484012047695672
    0.4 0.533110533481867
    0.45 0.573750876758675
    0.5 0.621281511738251
    0.55 0.66006518958617
    0.6 0.706853158394191
    0.65 0.738416470685316
    0.7 0.774064446919998
    0.75 0.812971902463176
    0.8 0.841935883153856
    0.85 0.874695713165821
    0.9 0.905268803894872
    0.95 0.937739819284565
};
\addlegendentry{GP-Beta}

\addplot [semithick, color3, opacity=1.0, dashed, mark=square*, mark size=1.5, mark options={solid}]
table {%
    0.05 0.117506292032842
    0.1 0.216487189008541
    0.15 0.302966538763048
    0.2 0.37900730288402
    0.25 0.450715847670916
    0.3 0.513842472253167
    0.35 0.568222139703759
    0.4 0.617815736270991
    0.45 0.664644964310765
    0.5 0.706523084540166
    0.55 0.746296983950159
    0.6 0.782275034038866
    0.65 0.813343235548954
    0.7 0.841399513141065
    0.75 0.865701200643644
    0.8 0.889796591987457
    0.85 0.911870280975368
    0.9 0.932004786070883
    0.95 0.953707141973016
};
\addlegendentry{Variance Scaling}

\addplot [semithick, color4, opacity=1.0, dash pattern=on 3pt off 6pt on 6pt off 6pt, mark=triangle*, mark size=1.5, mark options={solid}]
table {%
    0.05 0.117176218178818
    0.1 0.215785782068738
    0.15 0.302223872591492
    0.2 0.377728266699674
    0.25 0.449230515327805
    0.3 0.51268721376408
    0.35 0.566447992738375
    0.4 0.615876552378595
    0.45 0.66299459504064
    0.5 0.704913974501795
    0.55 0.744811651607047
    0.6 0.780954738622767
    0.65 0.811651607047077
    0.7 0.840574328506003
    0.75 0.864339645995792
    0.8 0.889053925815901
    0.85 0.911375170194331
    0.9 0.931262119899328
    0.95 0.953335808887239
};
\addlegendentry{GP-Normal}

\addplot [semithick, color5, mark=*, mark size=1.5, mark options={solid}]
table {%
    0.05 0.0595370714197302
    0.1 0.106118744068985
    0.15 0.152782935181747
    0.2 0.199447126294508
    0.25 0.247184057432851
    0.3 0.292156620043735
    0.35 0.337129182654619
    0.4 0.383545818376862
    0.45 0.433015637248834
    0.5 0.48318686306061
    0.55 0.53793786359698
    0.6 0.592276271815819
    0.65 0.652514750175352
    0.7 0.715435078598836
    0.75 0.787473697239757
    0.8 0.853364690349466
    0.85 0.915707389528407
    0.9 0.966208689194207
    0.95 0.993728596773528
};
\addlegendentry{GP-Cauchy}

\addplot [semithick, red, dashed]
table {%
    0 0
    1 1
};
\addlegendentry{Perfect Calibration}
\end{axis}

\end{tikzpicture}
 
    \caption[Reliability diagram of the regression uncertainty for different quantile levels and for different calibration methods.]{
        Reliability diagram of the regression uncertainty for different quantile levels and for different calibration methods \cite[p.~13, Fig.~4]{Kueppers2022b}.
        The expected quantile level (x~axis) is compared to the observed prediction interval coverage frequency of the ground-truth scores over all samples (y~axis).
        The diagram shows the miscalibration in terms of \quantilecalibration{} for the predicted uncertainty of the $\centerx$ coordinates before and after calibration.
        We can observe an underconfidence of the uncalibrated uncertainty estimates.
        In this case, Isotonic Regression and GP-Cauchy achieve the best calibration performance.
    }
    \label{fig:regression:evaluation:reliability}
\end{figure}

The results show that Isotonic Regression calibration \cite{Kuleshov2018} achieves the best results for \ac{NLL}, Pinball loss, and \ac{C-QCE}.
This shows that Isotonic Regression achieves the best results for \quantilecalibration{} in our experiments.
As already mentioned in \secref{section:regression:methods}, the error distributions of the examined object detectors rather follow a Cauchy distribution than a Gaussian.
This can be seen in \figref{fig:regression:methods:error_dist}.
Thus, the GP-Cauchy method is also able to achieve qualitatively good calibration results.
This is also underlined by the observations within the reliability diagram in \figref{fig:regression:evaluation:reliability}.
In contrast to the semantic confidence calibration, we observe an underconfidence of the predicted uncertainty estimates.
Interestingly, Isotonic Regression also outperforms the non-parametric GP-Beta \cite{Song2019} which is originally designed to achieve \distributioncalibration{}.
On the one hand, GP-Beta is restricted to the beta calibration family of functions which, however, limits the calibration power of this method.
On the other hand, we can not find a strong connection between miscalibration and position/shape information.
This can also be seen by the minor differences in the calibration between Isotonic Regression and GP-Beta, as well as between Variance Scaling \cite{Levi2019,Laves2020} and GP-Normal.
Both methods, Variance Scaling and GP-Normal, yield parametric Gaussians as calibration output, whereas GP-Normal also has the flexibility to capture possible correlations between miscalibration and position/shape information.
However, we only observe minor differences in calibration which leads to the assumption of only minor correlations between position/shape information and miscalibration.

The best calibration results for \ac{UCE} and \ac{ENCE} are obtained by Variance Scaling \cite{Levi2019,Laves2020} and GP-Normal which perform equally in our experiments.
Thus, both methods achieve the best results in terms of \variancecalibration{}.
In addition, the multivariate GP-Normal is able to further improve the multivariate \ac{NLL} compared to the (independent) Variance Scaling and GP-Normal recalibration. 
As described in \secref{section:regression:definition:variance}, a \variancecalibrated{} forecaster is also \quantilecalibrated{} if the predicted estimates are unbiased and the ground-truth data are normally distributed.
However, we already stated that the observed error distributions are not optimally fit by Gaussians so that \variancecalibration{} does not lead to optimal \quantilecalibration{} in our experiments.
Nevertheless, this might not be an issue if subsequent processes such as Kalman filtering assert Gaussian distributions as input.
In this case, calibration with Variance Scaling and GP-Normal are advantageous since they are designed to provide an optimal Gaussian fit.
The influence of regression uncertainty calibration for a subsequent Kalman filtering is part of the experiments within the next \chapref{chapter:tracking}.

Therefore, we conclude that the simple Isotonic Regression \cite{Kuleshov2018} achieves the best calibration results and thus is sufficient to achieve \quantilecalibrated{} uncertainty estimates.
If the estimated quantiles of a probabilistic object detector are of interest in application, then Isotonic Regression should be the preferred method.
In contrast, if a parametric distribution is required, the GP-Cauchy leads to an optimal fit for the observed error distributions in our examinations.
Further investigations are necessary to gain more evidence if this also holds for other detection architectures.
Finally, the Variance Scaling \cite{Levi2019,Laves2020} and (multivariate) GP-Normal lead to the best results regarding \variancecalibration{} and thus are able to achieve the best Gaussian fit in our experiments.
These methods should be preferred if a Gaussian representation of the spatial uncertainty is required in a subsequent application.

\section{Conclusion for Spatial Uncertainty Calibration} 
\label{section:regression:conclusion}

Since not only semantic class but also spatial position uncertainty is an important part of environment perception, we examine and evaluate the spatial position uncertainty of common probabilistic object detectors in this chapter.
For this reason, we give an overview of the recent definitions for uncertainty calibration of regression tasks and set these definitions into a common mathematical context.
Thus, we are able to show that a \variancecalibrated{} forecaster \cite{Levi2019,Laves2020} is also \quantilecalibrated{} \cite{Kuleshov2018} for unbiased predictions and normally distributed data.
Furthermore, we show the connection between \distributioncalibration{} \cite{Song2019} and \variancecalibration{} \cite{Levi2019,Laves2020}.
Since an object detection model needs to jointly estimate the position and shape information of detected objects, we extend these definitions to the multivariate case.

In the next step, we present the most recent methods for regression uncertainty calibration.
In this scope, we provide detailed descriptions for the non-parametric Isotonic Regression \cite{Kuleshov2018} and GP-Beta \cite{Song2019} for \quantilecalibration{} and \distributioncalibration{}, respectively, as well as for the parametric Variance Scaling \cite{Levi2019,Laves2020} for \variancecalibration{}.
We extend the existing calibration methods and propose the GP-Normal and GP-Cauchy methods \cite{Kueppers2022b} which both adapt the approach of \distributioncalibration{} using a Gaussian process (GP) for parameter estimation.
In contrast to the non-parametric GP-Beta \cite{Song2019}, the GP-Normal and GP-Cauchy yield parametric normal and Cauchy distributions as calibration output.
This might be advantageous for subsequent applications such as Kalman filtering that require parametric distributions as input.
We provide more detailed experiments on the effect of calibration on object tracking in \chapref{chapter:tracking}.
Furthermore, we use the \ac{GP} recalibration framework to jointly calibrate all dimensions that are necessary for bounding box inference.
This allows to capture possible correlations between position/shape information and miscalibration.
We also use the GP-Normal to derive a covariance estimation scheme which allows for a post-hoc introduction of correlations between independently inferred random variables.

Our experiments show that the simple Isotonic Regression \cite{Kuleshov2018} is able to achieve the best calibration results in terms of \quantilecalibration{}.
We can not find a strong connection between position/shape information and miscalibration so that the (marginal) recalibration using Isotonic Regression is sufficient within our experiments.
In contrast, the Variance Scaling \cite{Levi2019,Laves2020} and GP-Normal achieve the best results for a Gaussian fit of the predicted uncertainty to the observed error distribution which, in turn, leads to the best results for \variancecalibration{}.
Furthermore, the GP-Normal method also allows for a post-hoc introduction of correlations between independently inferred random variables.
This allows for a further improvement in the multivariate calibration case.
Therefore, we conclude that Isotonic Regression is the method of choice if the predicted quantiles are of special interest.
However, if a parametric normal distribution is required, we recommend to use Variance Scaling \cite{Levi2019,Laves2020} or the (multivariate) GP-Normal.
This assumption is underlined by our examinations for uncertainty calibration on object tracking which can be found in \chapref{chapter:tracking}.

\acresetall
\chapter{Application of Calibration to Object Tracking}
\label{chapter:tracking}

In the previous chapters, we focused on the task of object detection whose target is to identify objects within a single frame.
In contrast, the task of \ac{MOT} is to identify the same objects in subsequent frames within a sequence of images.
In object tracking, we are interested in the position information over time as well as the belief that the actual track matches a ground-truth object, i.e., if the tracked object exists.
In this chapter, we demonstrate how the uncertainty calibration methods for object detection from the previous \chapref{chapter:confidence} and \chapref{chapter:regression} can be integrated into an object tracking framework.
For this reason, we elaborate the mathematical context of object tracking and further show how to include semantic confidence calibration as well as spatial uncertainty calibration in the estimation of the object's existence as well as for its position/shape information, respectively.
We do not provide a state-of-the-art tracking framework but rather aim to demonstrate the usefulness of calibration for subsequent applications such as object tracking.

In \secref{section:tracking:introduction}, we start by introducing the basic concept of tracking-by-detection algorithms which utilizes object detectors to generate the observations for the tracking process \cite{Andriluka2008,Breitenstein2009}.
Furthermore, we introduce recursive Bayesian filtering in \secref{section:tracking:concept} that is the underlying framework for object tracking.
Subsequently, we use this concept in \secref{section:tracking:existence} to derive a mathematical expression for the estimation of the object existence within a sequence of frames.
On the one hand, this expression allows for a direct integration of the frame-wise detector confidence into the estimation of the object's confidence over time.
On the other hand, we can show how to directly include the semantic confidence calibration methods into object tracking.
Similarly, we introduce the Kalman filter in \secref{section:tracking:position} that is an implementation of recursive filtering for the object position using Gaussian distributions.
We show how to include the uncertainty information for the object position provided by the underlying probabilistic object detector.
This uncertainty can also be recalibrated using the calibration methods for spatial regression uncertainty.
In \secref{section:tracking:association}, the basic functionality for the initialization of new tracks as well as the association between existing tracks and new detections is shown so that we have everything together to run our evaluations for object tracking in \secref{section:tracking:experiments}.
Finally, we summarize our contributions and findings in \secref{section:tracking:conclusion}.

\textbf{Contributions:} In summary, we provide the following contributions to the field of object tracking:
\begin{itemize}
    \item Derivation of a confidence likelihood to estimate the probability of object existence over time using the detector's confidence.
    \item Integration of semantic confidence calibration methods into the estimation of object existence.
    \item Integration of time-varying observation uncertainty for the object position which is provided by probabilistic detection models.
    \item Integration of spatial uncertainty calibration methods into the estimation of object position.
    \item Evaluation of the performance and calibration properties of the object tracking when using calibrated uncertainty.
\end{itemize}

\section{Object Tracking by Detection}
\label{section:tracking:introduction}

In the scope of object tracking, a relationship between objects detected on different frames over time is established.
Common literature on object tracking offers several ways of identifying and tracking single objects in subsequent frames \cite{Van2005,Bar2004,Bar2011}.
In this work, we will focus on the tracking-by-detection paradigm \cite{Andriluka2008,Breitenstein2009}.
This approach utilizes an object detector to identify multiple objects within a single frame.
Subsequently, each detected object is assigned to an object already known by the tracker.
If no appropriate object has been found, a new track is generated (cf. \secref{section:tracking:association}).
In this scope, each track is within a dedicated state that consists of the position/shape information, the object's velocity and/or acceleration, as well as a confidence score that represents the tracker's belief of matching a real ground-truth object.

\begin{figure}[th]
    \centering
    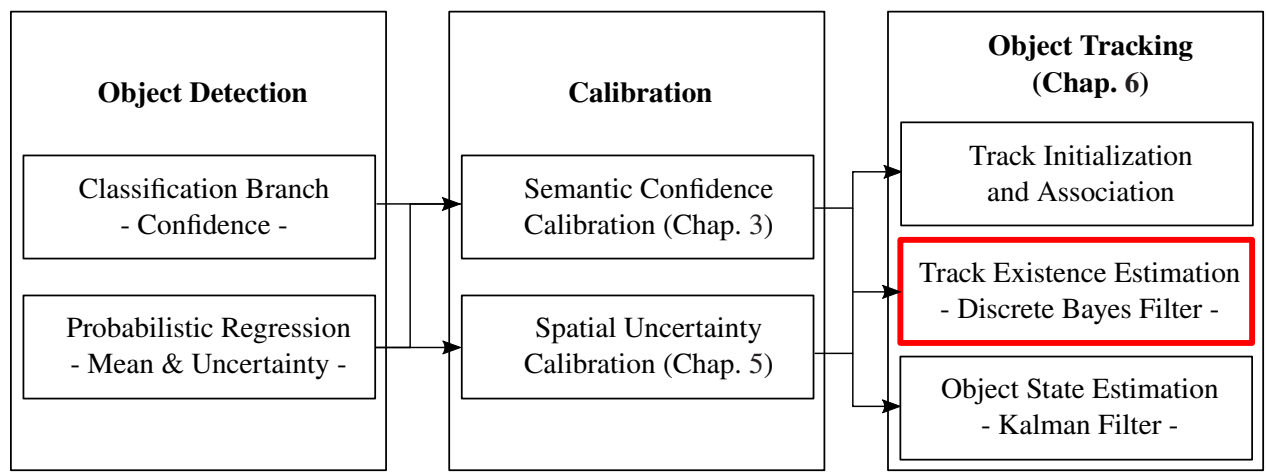%
    \caption[Concept of object tracking by detection with additional uncertainty calibration.]{
        Concept of object tracking by detection with additional uncertainty calibration.
        First, a probabilistic object detector predicts multiple objects in a single frame with uncertainty information for the object existence and position in\-for\-ma\-tion.
        This information is used during object tracking to establish a relationship between objects in consecutive frames within a sequence of images.
        To study the effect of our proposed uncertainty calibration methods from \chapref{chapter:confidence} and \chapref{chapter:regression}, we add an additional uncertainty calibration step between object detection and object tracking.
    }
    \label{fig:tracking:blockimage}
\end{figure}

As already mentioned in the previous chapters, common object detectors provide the position/shape in\-for\-ma\-tion as well as a confidence for each detected object.
The confidence represents the detector's belief of correctness about the individual detection.
If we utilize a probabilistic object detector as shown in \chapref{chapter:regression}, it is also possible to obtain uncertainty information for the position/shape in\-for\-ma\-tion as well.
Thus, we can provide the object position/shape as well as the semantic and spatial detection uncertainty to the tracking algorithm.
In this chapter, we study the effect of uncertainty calibration on the task of object tracking.
For this reason, we adapt our calibration methods for semantic confidence (\chapref{chapter:confidence}) and spatial uncertainty calibration (\chapref{chapter:regression}) and use them as an intermediate calibration step before passing the uncertainty information to the tracking algorithm.
This concept is schematically shown in \figref{fig:tracking:blockimage}.

\section{Recursive Bayesian Filtering}
\label{section:tracking:concept}

In image-based object tracking-by-detection, we seek to infer the state of an object (e.g., its position and shape, velocity, acceleration) at time step $\timestep$ given the predictions by an object detection model.
In this scope, an object detection model is used to identify objects within a single frame.
The detection model outputs a label $\predoutputvariate_{\timestep} \in \outputset$ with an according confidence $\predconfidencevariate_{\timestep} \in \probset$ indicating its belief of matching real-world object which we denote as $\matchedvariate_{\timestep} \sampledfrom \bernoullidistribution(\predconfidencevariate_{\timestep})$. 
Furthermore, the detector outputs a probabilistic (Gaussian) estimate for the object position $\allpredbboxvariates_{\timestep} \sampledfrom \normaldistribution(\meanvec_{\allpredbboxvariates, \timestep}, \cov_{\allpredbboxvariates, \timestep})$ with mean $\meanvec_{\allpredbboxvariates, \timestep} \in \bboxset$ and measurement noise covariance matrix $\cov_{\allpredbboxvariates, \timestep} \in \realdigits^{\numbboxdims \times \numbboxdims}$, where $\numbboxdims$ is the size of the box encoding.
Thus, the aggregated observations at time step $\timestep$ are denoted by $\allcollectvariates_{\timestep} \in \collectset$, where $\allcollectvariates_{\timestep} = (\predconfidencevariate_{\timestep}, \predoutputvariate_{\timestep}, \allpredbboxvariates_{\timestep})^\T$.

A standard detection model does not establish a relationship between objects in consecutive frames.
Furthermore, the complete state $\hiddenstatevariate_{\timestep} \in \hiddenstateset$ of an object at time step $\timestep$ is not directly observable given a prediction within a single frame \cite{Bar2004,Van2005,Bar2011}.
For example, it is possible to denote the position of an object given the output of the underlying object detector.
However, we can not quantify a velocity and an acceleration of an object given a single frame \cite{Bar2004,Van2005,Bar2011} as these quantities require more observations of the same object over time.
In our setup, the internal object state for the position information is represented by a Gaussian kinematic state model of second order \cite[p. 268 ff.]{Bar2004}, i.e., the state information does not only contain the actual position/shape but also its velocity (first order derivative) and acceleration (second order derivative).
These information are obtained during the tracking process.

In object tracking, we are interested in the (joint) probability distribution for the actual state $\hiddenstate_{\timestep}$ given all previous observations $\allcollect_1, \ldots, \allcollect_{\timestep-1}$ as well as the actual observation $\allcollect_{\timestep}$, so that the \ac{PDF} for the actual state is denoted by $\pdf_{\hiddenstatevariate}(\hiddenstate_{\timestep} | \allcollect_1, \ldots, \allcollect_{\timestep})$. 
We further use $\pdf_{\hiddenstatevariate}(\hiddenstate_{\timestep} | \allcollect_{1:\timestep})$ for notation simplicity.
Under the Markov assumption, that the actual state $\hiddenstate_{\timestep}$ solely depends on the last state $\hiddenstate_{\timestep-1}$ as well as on the actual observation $\allcollect_{\timestep}$ \cite{Bar2004}, the \ac{PDF} of the updated state variables is given by
\begin{align}
    \label{eq:tracking:general:update}
    \pdf_{\hiddenstatevariate}(\hiddenstate_{\timestep} | \allcollect_{1:\timestep}) = 
    \frac{ \pdf_{\allcollectvariates}(\allcollect_{\timestep} | \hiddenstate_{\timestep}) \pdf_{\hiddenstatevariate}(\hiddenstate_{\timestep} | \allcollect_{1:\timestep-1})}
    {\int_{\hiddenstateset} \pdf_{\allcollectvariates}(\allcollect_{\timestep} | \hiddenstate_{\timestep}^{\ast}) \pdf_{\hiddenstatevariate}(\hiddenstate_{\timestep}^{\ast} | \allcollect_{1:\timestep-1}) \diff \hiddenstate_{\timestep}^{\ast}} .
\end{align}
The term $\pdf_{\allcollectvariates}(\allcollect_{\timestep} | \hiddenstate_{\timestep})$ denotes the likelihood of the observation $\allcollect_{\timestep}$ given the state $\hiddenstate_{\timestep}$, whereas $\pdf_{\hiddenstatevariate}(\hiddenstate_{\timestep} | \allcollect_{1:\timestep-1})$ denotes the probability distribution of the (new) state $\hiddenstate_{\timestep}$ given all previous observations.
The latter term can be rewritten using the Chapman–Kolmogorov equation \cite{Bar2004}, so that it is given by
\begin{align}
    \label{eq:tracking:general:prediction}
    \pdf_{\hiddenstatevariate}(\hiddenstate_{\timestep} | \allcollect_{1:\timestep-1}) = \int_{\hiddenstateset} \pdf_{\hiddenstatevariate}(\hiddenstate_{\timestep} | \hiddenstate_{\timestep-1})  \pdf_{\hiddenstatevariate}(\hiddenstate_{\timestep-1} | \allcollect_{1:\timestep-1}) \diff \hiddenstate_{\timestep-1} ,
\end{align}
where $\pdf_{\hiddenstatevariate}(\hiddenstate_{\timestep} | \hiddenstate_{\timestep-1})$ denotes the probability of a transition from the previous state~$\hiddenstate_{\timestep-1}$ to the actual one~$\hiddenstate_{\timestep}$. 
Furthermore, $\pdf_{\hiddenstatevariate}(\hiddenstate_{\timestep-1} | \allcollect_{1:\timestep-1})$ is nothing else but the state probability distribution of the last time step.
Therefore, it is required to provide an appropriate modeling for the observation likelihood function as well as for the state transition distribution.
This concept is known as recursive Bayesian filtering.
The term in (\ref{eq:tracking:general:prediction}) is also known as the prediction step as it can be used to generate predictions for the consecutive step without observing any new data, whereas the equation in (\ref{eq:tracking:general:update}) is known as the update step that updates the predictions for the internal state representations by new incoming observations.

We denote the object state of the estimated position by $\hiddenbboxvariate_{\timestep} \sampledfrom \normaldistribution(\meanvec_{\hiddenbboxvariate, \timestep}, \cov_{\hiddenbboxvariate, \timestep})$ with mean vector $\meanvec_{\hiddenbboxvariate, \timestep} \in \realdigits^{\numstatedims}$ and error covariance matrix $\cov_{\hiddenbboxvariate, \timestep} \in \realdigits^{\numstatedims \times \numstatedims}$, where $\numstatedims$ is the size of state's position information, in this case $\numstatedims = 3 \cdot \numbboxdims$ (since the state consists of the actual position/shape, the velocity, and the acceleration).
For the state estimation of the object position, a Kalman filter is commonly used which will be explained in \secref{section:tracking:position} in more detail.

Furthermore, we are interested in the belief that the tracked object matches a real ground-truth object given all observations $\allcollectvariates_1, \ldots, \allcollectvariates_{\timestep}$ until time step $\timestep$ which we denote as $\trackmatchedvariate_{\timestep} | \allcollectvariates_{1:\timestep} \sampledfrom \bernoullidistribution(\trackconfidencevariate_{\timestep})$ with the track confidence $\trackconfidencevariate_{\timestep} \in \probset$.
The estimation of the object existence $\trackmatchedvariate_{\timestep}$ can be realized using a discrete Bayes filter.
Both concepts are a realization of the Bayesian filter framework in (\ref{eq:tracking:general:update}) and (\ref{eq:tracking:general:prediction}) for continuous and discrete random variables, respectively.
We further show how to include the calibration methods for semantic confidence calibration (\chapref{chapter:confidence}) and spatial regression uncertainty calibration (\chapref{chapter:regression}) into the tracking environment.

\section{Discrete Bayes Filter for Object Existence Estimation}
\label{section:tracking:existence}

For a proper track management, it is necessary to decide when a track no longer exists, e.g., when it has not been recognized correctly or the object has left the image area.
Commonly, simple techniques such as exponential moving average (EMA) filters are used to implement a basic track management \cite{Huang2019}.
However, such techniques are not aware of the uncertainty that is indicated by the underlying detection model as they commonly use the information if an appropriate detection has been found for an existing track or not.
In this section, we therefore propose a new track management using the detector confidence for the estimation of the object existence.

As already mentioned in \secref{section:confidence:definition:detection}, the probability of a match $\prob(\matchedvariate_{\timestep}=1)$ for an object within a single frame is a shorthand notation for $\prob(\predoutputvariate_{\timestep} = \groundtruthvariate_{\timestep}, \allpredbboxvariates_{\timestep}=\allgroundtruthbboxvariates_{\timestep})$, where $\predoutputvariate_{\timestep} \in \outputset$ and $\allpredbboxvariates_{\timestep} \in \bboxset$ are the predicted label and position of an object at time step $\timestep$, respectively.
An predicted object is considered to match a real object if the IoU between predicted bounding box and a ground truth box is above a certain \ac{IoU} threshold (in our case above $0.5$).
Furthermore, let $\groundtruthvariate_{\timestep} \in \outputset$ and $\allgroundtruthbboxvariates_{\timestep} \in \bboxset$ denote the respective ground-truth information for the object label and position, respectively.
For each detection, the underlying object detector outputs a confidence score $\predconfidencevariate_{\timestep}$ indicating its belief that the prediction matches a ground-truth object, i.e., for $\matchedvariate_{\timestep} = 1$, so that $\matchedvariate_{\timestep} \sampledfrom \bernoullidistribution(\predconfidencevariate_{\timestep})$.
When using a position-dependent confidence calibration model $\calmodel$, it is also possible to construct a Bernoulli distribution for $\matchedvariate_{\timestep}$ whose probability parameter depends on the complete model output $\allcollectvariates_{\timestep} = (\predconfidencevariate_{\timestep}, \predoutputvariate_{\timestep}, \allpredbboxvariates_{\timestep})^\T$, so that $\matchedvariate_{\timestep} \sampledfrom \bernoullidistribution\big(\calmodel(\allcollectvariates_{\timestep})\big)$ (cf. (\ref{eq:confidence:calibrated}) in \secref{section:confidence:methods}).

For the estimation of the object existence, we are interested in the probability $\trackconfidencevariate_{\timestep} \in \probset$ that a track matches a ground-truth object given all previous confidence, label, and bounding box information, where $\allcollectvariates_{1:\timestep} = (\predconfidencevariate_{1:\timestep}, \predoutputvariate_{1:\timestep}, \allpredbboxvariates_{1:\timestep})^\T$ denotes the aggregated observations provided by the detection model.
Thus, we can interpret the random variable $\trackmatchedvariate_{\timestep} | \allcollectvariates_{1:\timestep} \sampledfrom \bernoullidistribution(\trackconfidencevariate_{\timestep})$ for the track confidence similarly.
Note that we can interpret $\allcollectvariates_{1:\timestep}$ either as the aggregated model output, if we use position-dependent confidence calibration during object existence estimation.
Alternatively, it is also possible to assume that $\allcollectvariates_{1:\timestep}$ only represents the detector's confidence information, if no calibration or standard calibration methods are used.

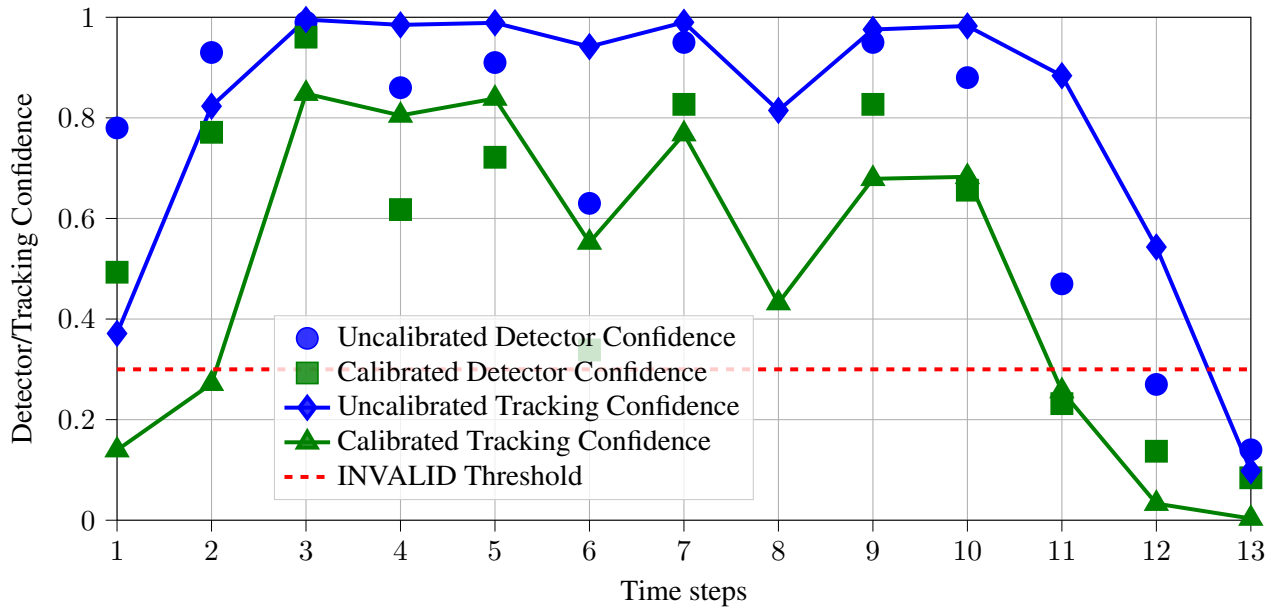
\begin{figure}[t!]
    \centering
\begin{tikzpicture}

\definecolor{darkgray176}{RGB}{176,176,176}
\definecolor{green}{RGB}{0,128,0}
\definecolor{lightgray204}{RGB}{204,204,204}

\begin{axis}[
legend cell align={left},
legend style={
  fill opacity=0.8,
  draw opacity=1,
  text opacity=1,
  at={(0.35,0.04)},
  anchor=south,
  draw=lightgray204
},
width=\linewidth,
height=0.5\linewidth,
tick align=outside,
tick pos=left,
unbounded coords=jump,
x grid style={darkgray176},
xlabel={Time steps},
xmajorgrids,
xmin=1, xmax=13,
xtick style={color=black},
y grid style={darkgray176},
ylabel={Detector/Tracking Confidence},
ymajorgrids,
ymin=0, ymax=1,
ytick style={color=black}
]
\addplot [semithick, blue, mark=*, mark size=4, mark options={solid}, only marks]
table {%
0 nan
1 0.78
2 0.93
3 0.99
4 0.86
5 0.91
6 0.63
7 0.95
8 nan
9 0.95
10 0.88
11 0.47
12 0.27
13 0.14
};
\addlegendentry{Uncalibrated Detector Confidence}
\addplot [semithick, green, mark=square*, mark size=4, mark options={solid}, only marks]
table {%
0 nan
1 0.493092117642894
2 0.771318687482444
3 0.960508544648195
4 0.617636663777138
5 0.721654782263598
6 0.338542064169189
7 0.826849158270287
8 nan
9 0.826849158270287
10 0.656223379378054
11 0.2318070416234
12 0.137146639232211
13 0.0843990076326762
};
\addlegendentry{Calibrated Detector Confidence}
\addplot [semithick, blue, mark=diamond*, mark size=4, mark options={solid}, line width=1.5]
table {%
0 0.206672140131436
1 0.371263872287046
2 0.823185980776072
3 0.995450021604392
4 0.984890045653144
5 0.988976154974625
6 0.941646387756189
7 0.98997000325494
8 0.814813130283839
9 0.975677735444178
10 0.982696114493045
11 0.883802983849694
12 0.543194718146927
13 0.0982508036088294
};
\addlegendentry{Uncalibrated Tracking Confidence}
\addplot [semithick, green, mark=triangle*, mark size=4, mark options={solid}, line width=1.5]
table {%
0 0.206672140131436
1 0.139422105945946
2 0.271486549887556
3 0.848187152418289
4 0.80479669221432
5 0.838466674157037
6 0.552237930812452
7 0.767720300500548
8 0.431825660844245
9 0.678827793830604
10 0.682793909006842
11 0.25697175916352
12 0.0329955876707568
13 0.00338719824222269
};
\addlegendentry{Calibrated Tracking Confidence}
\addplot [semithick, red, dashed, line width=1.5]
table {%
1 0.3
13 0.3
};
\addlegendentry{INVALID Threshold}
\end{axis}

\end{tikzpicture}
    \caption[Confidence diagram to qualitatively demonstrate the influence of the detector confidence before and after calibration to the estimation of the existence score over the time.]{
        Confidence diagram to qualitatively demonstrate the influence of the detector confidence before and after calibration to the estimation of the existence score over the time.
        During tracking, objects with an existence score below the invalid threshold are discarded.
    }
    \label{fig:tracking:existence:confidence}
\end{figure}

From equation (\ref{eq:confidence:definition:bernoulli}) in \secref{section:confidence:definition:classification}, we know that the confidence $\predconfidencevariate_{\timestep}$ (or its calibrated variants) can be interpreted as the direct estimation of the probability parameter of the Bernoulli distribution of $\matchedvariate_{\timestep}$.
However, we can not directly adapt the detector confidence as a raw estimate for the track confidence $\trackconfidencevariate_{\timestep}$ as the underlying random variable $\trackmatchedvariate_{\timestep} | \allcollectvariates_{1:\timestep}$ is conditioned on all previous observations.
Thus, the target now is to update the model's belief of matching a real ground-truth object over time given all previous observations $\allcollectvariates_{1:\timestep}$.
Since $\trackmatchedvariate_{\timestep}$ is a discrete random variable, the probability distribution is given as a \ac{PMF}, and we can adapt the Bayesian filtering equation (\ref{eq:tracking:general:update}) for the existence estimation to
\begin{align}
    \label{eq:tracking:existence:update}
    \prob(\trackmatched_{\timestep} | \allcollect_{1:\timestep}) = \frac{ \pdf_{\allcollectvariates}(\allcollect_{\timestep} | \trackmatched_{\timestep}) \prob(\trackmatched_{\timestep} | \allcollect_{1:\timestep-1})}
    {\sum_{\trackmatched^\ast_{\timestep} \in \{0, 1\}} \pdf_{\allcollectvariates}(\allcollect_{\timestep} | \trackmatched_{\timestep}^\ast) \prob(\trackmatched_{\timestep}^\ast | \allcollect_{1:\timestep-1})} ,
\end{align}
with
\begin{align}
    \prob(\trackmatched_{\timestep} | \allcollect_{1:\timestep-1}) = \sum_{\trackmatched_{\timestep-1} \in \{0, 1\}} \prob(\trackmatched_{\timestep} | \trackmatched_{\timestep-1}) \prob(\trackmatched_{\timestep-1} | \allcollect_{1:\timestep-1}) ,
\end{align}
as the probability for the new state $\trackmatched_{\timestep}$ given the last observations $\allcollect_{1}, \ldots \allcollect_{\timestep-1}$, and $\pdf_{\allcollectvariates}(\allcollect_{\timestep} | \trackmatched_{\timestep})$ as the likelihood for the observation given the actual state.
Here we can see that applying Bayes' theorem under the Markov assumption in (\ref{eq:tracking:existence:update}) leads to a likelihood function only for the actual observation.
We can rewrite this likelihood to
\begin{align}
    \pdf_{\allcollectvariates}(\allcollect_{\timestep} | \trackmatched_{\timestep}) = \frac{\prob(\trackmatched_{\timestep} | \allcollect_{\timestep}) \pdf_{\allcollectvariates}(\allcollect_{\timestep})}{\prob(\trackmatched_{\timestep})} .
\end{align}
Since the probability solely depends on the actual observation, the random variable $\trackmatchedvariate_{\timestep} | \allcollectvariates_{\timestep}$ stands for the same intuitive interpretation of matching a ground-truth object given the actual prediction, which is also reflected by the detector output $\matchedvariate_{\timestep} | \allcollectvariates_{\timestep}$.
Therefore, we set $\matchedvariate_{\timestep} | \allcollectvariates_{\timestep} = \trackmatchedvariate_{\timestep} | \allcollectvariates_{\timestep}$ so that the probability $\prob(\matched_{\timestep} | \allcollect_{\timestep}) = \prob(\matched_{\timestep} | \predconfidence_{\timestep}, \predoutput_{\timestep}, \allpredbboxes_{\timestep})$ is our calibrated confidence known from equation (\ref{eq:confidence:definition:detection}) in \secref{section:confidence:definition:detection}.
The marginal probability $\prob(\matched_{\timestep})$ is nothing else but the average precision of the object detector given a certain \ac{IoU} threshold.
If we now plug-in the likelihood equation in (\ref{eq:tracking:existence:update}), the final filter equation for the track existence is given by
\begin{align}
    \prob(\trackmatched_{\timestep} | \allcollect_{1:\timestep}) &= \frac{ \prob(\matched_{\timestep} | \allcollect_{\timestep}) \pdf_{\allcollectvariates}(\allcollect_{\timestep})\prob(\matched_{\timestep})^{-1} \prob(\trackmatched_{\timestep} | \allcollect_{1:\timestep-1}) }
    {\sum_{\matched^\ast_{\timestep} \in \{0, 1\}} \prob(\matched_{\timestep}^\ast | \allcollect_{\timestep}) \pdf_{\allcollectvariates}(\allcollect_{\timestep})\prob(\matched_{\timestep}^\ast)^{-1} \prob(\trackmatched_{\timestep}^\ast | \allcollect_{1:\timestep-1}) } \\
    &= \frac{ \prob(\matched_{\timestep} | \allcollect_{\timestep}) \prob(\matched_{\timestep})^{-1} \prob(\trackmatched_{\timestep} | \allcollect_{1:\timestep-1}) }
    {\sum_{\matched^\ast_{\timestep} \in \{0, 1\}} \prob(\matched_{\timestep}^\ast | \allcollect_{\timestep}) \prob(\matched_{\timestep}^\ast)^{-1} \prob(\trackmatched_{\timestep}^\ast | \allcollect_{1:\timestep-1}) } ,
\end{align}
because the prior $\pdf_{\allcollectvariates}(\allcollect_{\timestep})$ for the detector output is canceled out.
In the standard case, it would be sufficient to use the detector confidence $\predconfidencevariate_{\timestep}$ as the Bernoulli parameter for $\matchedvariate_{\timestep} | \allcollectvariates_{\timestep}$.
According to equation (\ref{eq:confidence:calibrated}) in \secref{section:confidence:methods}, the distribution for $\matchedvariate_{\timestep} | \allcollectvariates_{\timestep}$ is a Bernoulli using the position-dependent calibration function $\calmodel: \collectset \rightarrow \probset$ that maps the detector output $\predconfidencevariate_{\timestep}, \predoutputvariate_{\timestep}, \allpredbboxvariates_{\timestep}$ to a calibrated confidence estimate $\calibratedvariate_{\timestep} \in [0, 1]$, so that $\matchedvariate_{\timestep} | \allcollectvariates_{\timestep} \sampledfrom \bernoullidistribution\big(\calmodel(\predconfidencevariate_{\timestep}, \predoutputvariate_{\timestep}, \allpredbboxvariates_{\timestep})\big)$.
In summary, we need to determine the state transition probabilities $\prob(\trackmatched_{\timestep} | \trackmatched_{\timestep-1})$, the detector precision $\prob(\matched_\timestep)$ on a dedicated data set, and a calibration function $\calmodel(\cdot)$ to model the track existence over time using our position-dependent confidence calibration methods from \secref{section:confidence:methods}.
Additionally, a confidence threshold is required to decide when to discard an existing track.
The influence of the detector confidence on the object existence estimation before and after calibration is qualitatively demonstrated in \figref{fig:tracking:existence:confidence} using the precision and transition probabilities from our experiments in \secref{section:tracking:experiments}.

\section{Kalman Filter for Object Position Tracking}
\label{section:tracking:position}

The Kalman filter is an implementation of the recursive Bayes filter in (\ref{eq:tracking:general:update}) for normally distributed state models and observations.
In common applications, the Kalman filter is used to track the position and size of an object within a sequence of images.
The random variables for the internal state as well as the ones for the observations are represented by multivariate normal distributions.
Since a Gaussian is a conjugate prior to itself, the prediction and update steps in (\ref{eq:tracking:general:prediction}) and (\ref{eq:tracking:general:update}) can be solved analytically which is advantageous especially for real-time applications.
In this section, we give a short overview of the basic Kalman filter concept and propose to use the spatial uncertainty of a probabilistic object detector as a time-varying noise for the observation likelihood.
We further show how to use the spatial calibration methods from \chapref{chapter:regression} to perform calibration of the time-varying observation noise.

As already mentioned, we use a Gaussian kinematic state model of second order \cite[p. 268 ff.]{Bar2004} which represents the position/shape information as well as the velocity and acceleration of the object's position/shape as Gaussian random variables for the state representation.
The state for the position/shape follows a multivariate normal distribution so that $\hiddenbboxvariate_{\timestep} \sampledfrom \normaldistribution(\meanvec_{\hiddenbboxvariate, \timestep}, \cov_{\hiddenbboxvariate, \timestep})$ with mean $\meanvec_{\hiddenbboxvariate, \timestep} \in \realdigits^{\numstatedims}$ and error covariance $\cov_{\hiddenbboxvariate, \timestep} \in \realdigits^{\numstatedims \times \numstatedims}$, where $\numstatedims$ is the state size.
The observations for each time step are the predicted bounding boxes $\allpredbboxvariates_{\timestep} \sampledfrom \normaldistribution(\meanvec_{\allpredbboxvariates, \timestep}, \cov_{\allpredbboxvariates, \timestep})$ obtained by a probabilistic object detector with mean $\meanvec_{\allpredbboxvariates, \timestep} \in \bboxset$ and estimated covariance $\cov_{\allpredbboxvariates, \timestep} \in \realdigits^{\numbboxdims \times \numbboxdims}$, where $\numbboxdims$ is the size of the box encoding.

As we can see in (\ref{eq:tracking:general:update}), we need to define appropriate functions for the state transition and the observation likelihood.
In the setting of a Kalman filter, the state transition is a linear function with transition matrix $\statetransition_{\timestep} \in \realdigits^{\numstatedims \times \numstatedims}$ that defines the transition from $\hiddenbbox_{\timestep-1}$ to $\hiddenbbox_{\timestep}$. The state transition matrix is used to specify the relationship between position/shape information and the respective velocities/accelerations and is further used to generate predictions for consecutive time steps.
To construct a multivariate normal distribution for the state, we seek to introduce Gaussian noise with zero mean and covariance $\statenoisecov_{\timestep} \in \realdigits^{\numstatedims \times \numstatedims}$ that is also known as the system noise of the Kalman filter.
Thus, the state transition model is given by
\begin{align}
    \hiddenbbox_{\timestep} = \statetransition_{\timestep} \hiddenbbox_{\timestep-1} + \statenoise_{\timestep} ,\\
    \text{where } \statenoise_{\timestep} \sampledfrom \normaldistribution(0, \statenoisecov_{\timestep} ) .
\end{align}
This yields a multivariate normal distribution for the state transition distribution given by
\begin{align}
    \pdf_{\hiddenbboxvariate}(\hiddenbbox_{\timestep} | \hiddenbbox_{\timestep-1}) = \normaldistribution(\hiddenbbox_{\timestep}; \statetransition_{\timestep} \meanvec_{\hiddenbboxvariate, \timestep-1}, \statenoisecov_{\timestep})
\end{align}
If the initial prior belief for $\pdf_{\hiddenbboxvariate}(\hiddenbbox_{0})$ is also normally distributed, the posterior $\pdf_{\hiddenbboxvariate}(\hiddenbbox_{\timestep} | \allpredbboxes_{0}, \ldots, \allpredbboxes_{\timestep})$ is always normally distributed as well.

Furthermore, the likelihood is also defined by a linear function with transition matrix $\observationtransition_{\timestep} \in \realdigits^{\numbboxdims \times \numstatedims}$ that translates from the state space to the observation space. Similar to the state transition, a Gaussian noise with zero mean and covariance $\observationnoisecov_{\timestep} \in \realdigits^{\numbboxdims \times \numbboxdims}$ is added that represents the observation noise within the Kalman filter, so that
\begin{align}
    \label{eq:tracking:kalman:observation}
    \allpredbboxes_{\timestep} &= \observationtransition_{\timestep} \hiddenbbox_{\timestep} + \observationnoise_{\timestep} ,\\
    \text{where } \observationnoise_{\timestep} &\sampledfrom \normaldistribution(0, \observationnoisecov_{\timestep}) .
\end{align}
In this case, the density function for the observation likelihood is a Gaussian of the form
\begin{align}
    \pdf_{\allpredbboxvariates}(\allpredbboxes_{\timestep} | \hiddenbbox_{\timestep}) = \normaldistribution(\allpredbboxes_{\timestep}; \observationtransition_{\timestep} \meanvec_{\hiddenbboxvariate, \timestep}, \observationnoisecov_{\timestep} ) .
\end{align}
For common object detectors, which only provide deterministic predictions for the object's position and shape, the covariance matrix $\observationnoisecov_{\timestep}$ of the observation noise is fixed.
The variance for each random variable in the observation space is commonly set to the mean squared error of the object detector that is obtained on a dedicated training set \cite[p. 16]{Bar2011}.

Instead, if we use a probabilistic object detector, we can directly provide the estimated uncertainty $\cov_{\allpredbboxvariates_{\timestep}}$ to the filter, so that the observation noise is time-varying.
Therefore, we set $\observationnoisecov_{\timestep} = \cov_{\allpredbboxvariates_{\timestep}}$ if our predictions are obtained by a probabilistic object detector.
At this point, we can now integrate our regression calibration methods from \chapref{chapter:regression} to recalibrate the spatial uncertainty in each frame before the application of the filter update.
Let $\calmodel(\cdot)$ denote a regression calibration method that takes an uncalibrated (multivariate) Gaussian as input and outputs a recalibrated (multivariate) Gaussian distribution as well.
The observation noise then changes to $\observationnoisecov_{\timestep} = \calmodel(\cov_{\allpredbboxvariates_{\timestep}})$.
Note that for the non-parametric uncertainty calibration methods such as Isotonic Regression (\secref{section:regression:methods:nonparametric:isotonic}) and GP-Beta (\secref{section:regression:methods:nonparametric:gp}), a moment-matching from the non-parametric distributions to a normal distribution is necessary to integrate the recalibrated uncertainty into object tracking using Kalman filtering.
In contrast, the output of the parametric Variance Scaling (\secref{section:regression:methods:parametric:variance}) and GP-Normal (\secref{section:regression:methods:parametric:gp}) can be directly used as these methods already return a Gaussian representation of the uncertainty.
Furthermore, we directly use the multivariate extension of the GP-Normal (\secref{section:regression:methods:multivariate}) for recalibration.
Since the probabilistic object detector only estimates the position/shape information independently, we further integrate our covariance estimation scheme presented in \secref{section:regression:methods:correlations} to include possible correlations between the observation quantities into the observation noise.

In summary, for Kalman filter implementation, the definition of the transition model $\statetransition_{\timestep}$ is given by the used kinematic state model, whereas the observation matrix $\observationtransition_{\timestep}$ defines the translation from the state space to the observation space.
For the observation noise $\observationnoisecov_{\timestep}$, we adapt the Gaussian uncertainty of the underlying probabilistic object detector as our observation uncertainty.
In this way, we are able to integrate our spatial uncertainty calibration methods from \chapref{chapter:regression} into object tracking.

\section{Track Initialization and Association}
\label{section:tracking:association}

At each time step $\timestep$, it is necessary to either assign an observation by the object detector to an existing track or to generate a new track if no appropriate track has been found.
The assignment is commonly performed by calculating a distance metric between each observation and each track \cite{Bar2004,Bar2011}.
As a distance metric, the \ac{NIS} is commonly used which is related to the \ac{NEES} (cf. \secref{section:regression:definition:quantile}) \cite{Bar2004,Bar2011}.
The \ac{NIS} is the squared Mahalanobis distance between a prediction by the object detector and the predicted position of an existing track for the actual time step $\timestep$
\begin{align}
    \text{NIS} := \big(\mean_{\allpredbboxvariates_\timestep} - \observationtransition_{\timestep} \mean_{\hiddenbboxvariate_\timestep}\big)^\T \big(\observationtransition_{\timestep} \cov_{\hiddenbboxvariate_\timestep} \observationtransition_{\timestep}^\T + \cov_{\allpredbboxvariates_\timestep} \big)^{-1} \big(\mean_{\allpredbboxvariates_\timestep} - \observationtransition_{\timestep} \mean_{\hiddenbboxvariate_\timestep}\big) ,
\end{align}
where $\mean_{\allpredbboxvariates_\timestep}$, $\mean_{\hiddenbboxvariate_\timestep}$ are the mean and $\cov_{\hiddenbboxvariate_\timestep}$, $\cov_{\allpredbboxvariates_\timestep}$ are the predicted track mean and covariance by the object tracker, and the observation mean and covariance by the object detector, respectively.
Similar to the \ac{NEES} (cf. \secref{section:regression:definition:quantile}), the \ac{NIS} can be interpreted as an equation for an ellipsoid that - given a certain quantile level $\quantile \in [0, 1]$ - spans a \ac{HPDR} for a certain $\quantile$ and for the aggregated state and observation uncertainty.

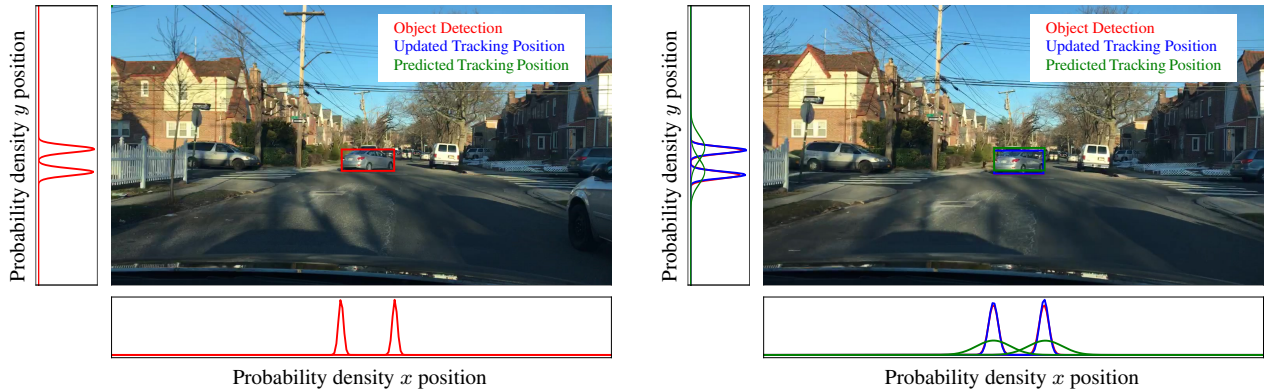
\begin{figure}[t!]
    \centering
    \begin{overpic}[width=0.48\textwidth]{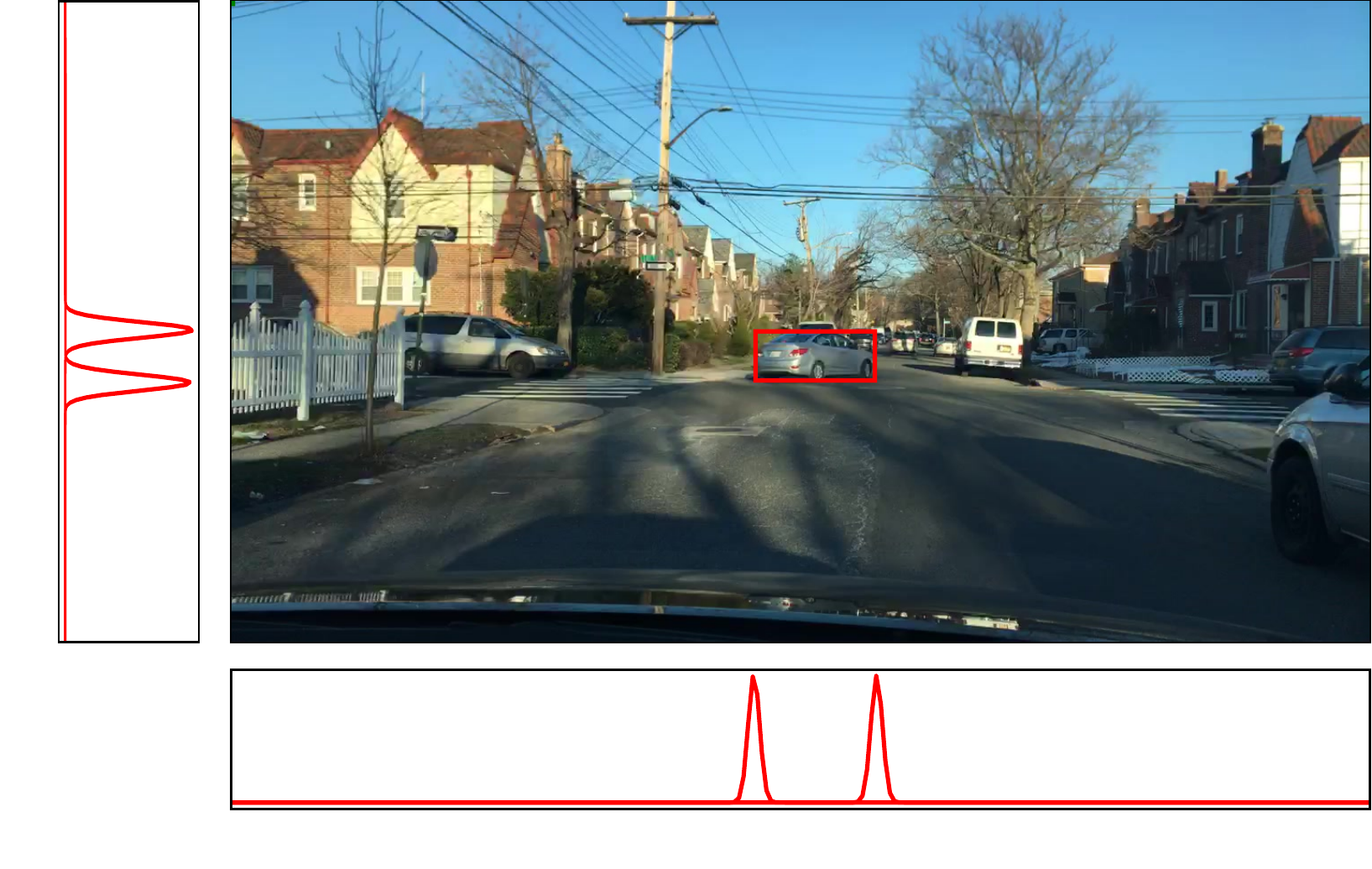}
        \put(37, 0){\scriptsize Probability density $x$ position}
        \put(0, 17){\rotatebox{90}{\scriptsize Probability density $y$ position }}
        \put(60, 55){
            \colorbox{white}{
                \begin{minipage}[l]{6em}
                    \tiny
                    \definecolor{trackinggreen}{RGB}{0,128,0}
                    {\color{red} Object Detection}\\
                    {\color{blue} Updated Tracking Position}\\
                    {\color{trackinggreen} Predicted Tracking Position}
                \end{minipage}
            }
        }
    \end{overpic}%
    \hfill%
    \begin{overpic}[width=0.48\textwidth]{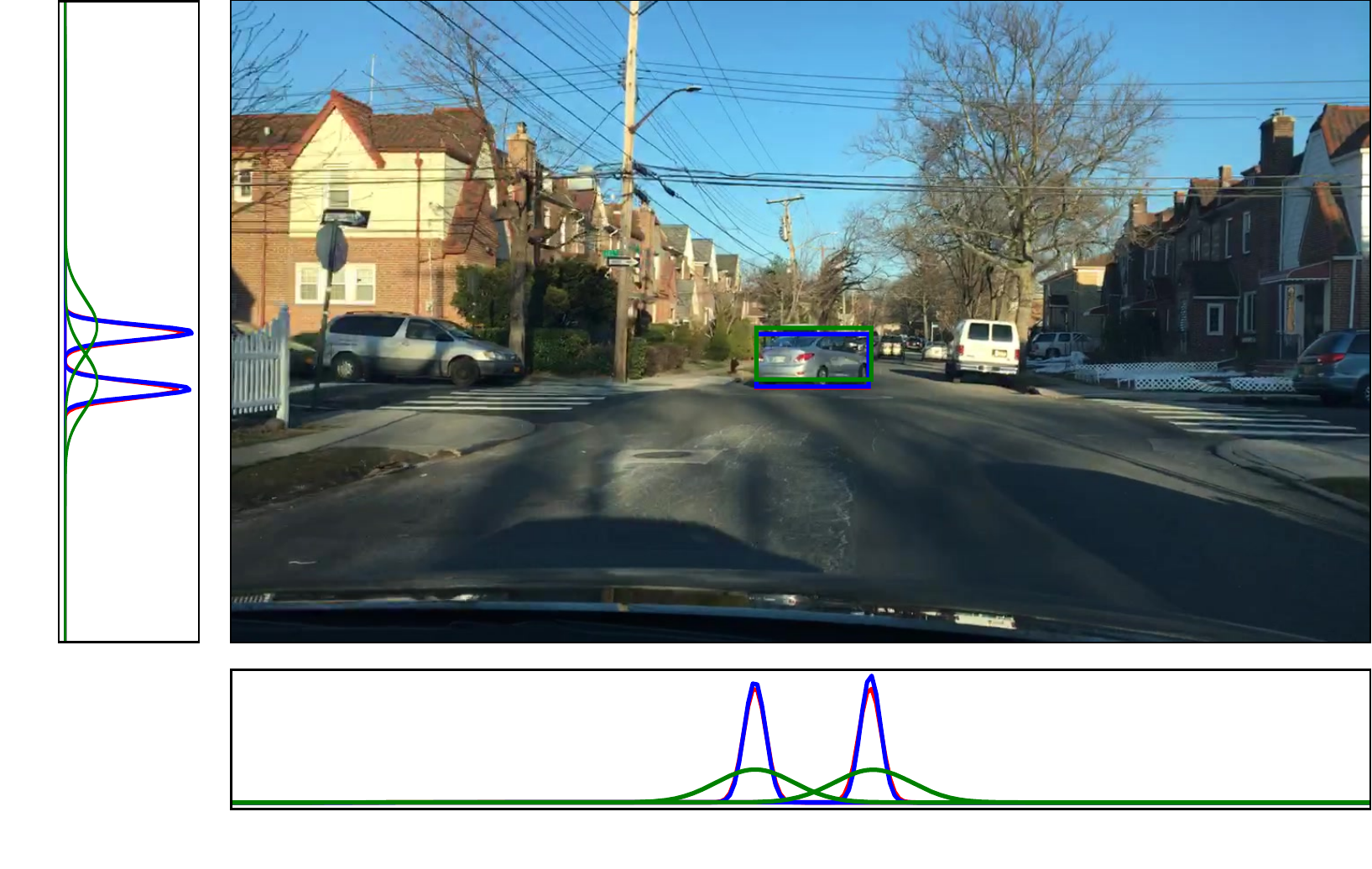}
        \put(37, 0){\scriptsize Probability density $x$ position}
        \put(0, 17){\rotatebox{90}{\scriptsize Probability density $y$ position }}
        \put(60, 55){
            \colorbox{white}{
                \begin{minipage}[l]{6em}
                    \tiny
                    \definecolor{trackinggreen}{RGB}{0,128,0}
                    {\color{red} Object Detection}\\
                    {\color{blue} Updated Tracking Position}\\
                    {\color{trackinggreen} Predicted Tracking Position}
                \end{minipage}
            }
        }
    \end{overpic}%
    
    \caption[Concept of Kalman filtering illustrated by the example of a single vehicle.]{
        Concept of Kalman filtering illustrated by the example of a single vehicle.
        On the first frame (left), an object detector (red) detects a vehicle with a certain spatial uncertainty for the bounding box edges.
        However, a track has not been initialized so far.
        In the consecutive frame (right), the object detector still recognizes the object and produces a prediction.
        In the meantime, the object tracker initialized a track and predicts its next position (green).
        Finally, it associates the detector observation to the actual track and updates the internal position (blue).
    }
    \label{fig:tracking:kalman:qualitative}
\end{figure}

Thus, we can construct a distance matrix $\distancemat_{\text{NIS}} \in \realdigitspositive^{\numstatedims \times \numbboxdims}$ containing the \ac{NIS} scores between all existing tracks and all observations.
Commonly, a quantile threshold is set for the \ac{NIS} using a $\chi^2_{\numbboxdims}$ distribution with $\numbboxdims$ degrees of freedom that determines the region in which an appropriate detection is searched for the assignment.
This threshold is used to mark single entries as invalid for the final assignment.
For the final association step, we use the Hungarian method \cite{Kuhn1955} which is a combinatorial optimization algorithm that is used to obtain the optimal observation-to-track assignment by minimizing the overall assignment costs given a cost matrix which is the matrix $\distancemat_{\text{NIS}}$ in our case.
If no appropriate existing track can be found for an observation, a new track is initialized at the position of the observation.
The concept of Kalman filtering is qualitatively shown in \figref{fig:tracking:kalman:qualitative}

\section{Experiments for Calibration in Object Tracking} 
\label{section:tracking:experiments}

To evaluate the influence of calibration on the task of object tracking, we use the Berkeley DeepDrive data set as well as the same \fasterrcnn{} network architecture with a probabilistic regression branch for the object position as within our evaluations for spatial uncertainty calibration (cf.~\secref{section:regression:experiments}).
For object tracking evaluation, the Berkeley DeepDrive tracking data set provides 200 sequences with approx. 200 frames per sequence on average with a frame rate of 5 Hz.
We split these sequences and use the first half as a training set to train the calibration methods as well as to get the average precision $\prob(\matched_{\timestep})$ and the state transition probabilities $\prob(\matched_{\timestep} | \matched_{\timestep-1})$ which are both required to set up the object tracking model.
For precision calculation and calibration training, we use an \ac{IoU} threshold of 0.5.

As described in \secref{section:tracking:association}, the \ac{NIS} score between existing tracks and incoming detections in conjunction with the Hungarian algorithm \cite{Kuhn1955} is used for a track-to-detection association.
We further use a \ac{HPDR} with a quantile level of $0.95$ to search for appropriate observations for each track.
During object tracking, each track with an existence probability below $0.3$ is dropped.
For object tracking evaluation, we use the \ac{MOT} metrics accuracy $\text{MOTA}$, precision $\text{MOTP}$, and the ratio of correctly identified detections over the average number of ground-truths and detections which is denoted by $\text{IDF1}$ \cite{Bernardin2008,Ristani2016}.
The $\text{MOTA}$ metric reflects the error ratios of false positives, false negatives, and identity switches and is given by
\begin{align}
    \text{MOTA} := 1 - \Bigg[\frac{\sum^{\numsteps}_{\timestep=1} \big[ \text{FP}_{\timestep} + \text{FN}_{\timestep} + \text{IDSw}_{\timestep} \big]}{\sum^{\numsteps}_{\timestep=1} \numsamples_{\timestep}}\Bigg] ,
\end{align}
with $\numsteps$ as the number of frames, where $\numsamples_{\timestep}$ is the number of ground-truth samples present at time step $\timestep$.
Furthermore, $\text{FP}_{\timestep}$, $\text{FN}_{\timestep}$, and $\text{IDSw}_{\timestep}$ are the false positives, false negatives, and identity switches at time step $\timestep$, respectively.
The $\text{MOTP}$ metric denotes the misalignment between the tracks and the respective ground-truth objects and is given by
\begin{align}
    \text{MOTP} := \frac{\sum^{\numsteps}_{\timestep=1} \sum^{\numsamples}_{\indexsamples=1} \big[\ind(\matched_{\timestep, \indexsamples} = 1) \cdot ||\hiddenbbox_{\timestep, \indexsamples} - \allgroundtruthbboxes_{\timestep, \indexsamples}||_2 \big]}{\sum^{\numsteps}_{\timestep=1} \text{TP}_{\timestep}},
\end{align}
where $\text{TP}_{\timestep}$ are the number of true positives at time step $\timestep$ and $||\hiddenbbox_{\timestep, \indexsamples} - \allgroundtruthbboxes_{\timestep, \indexsamples}||_2$ is the Euclidean distance between the estimated track position $\hiddenbbox_{\timestep, \indexsamples}$ and the ground-truth position $\allgroundtruthbboxes_{\timestep, \indexsamples}$ for object $\indexsamples$ at time step $\timestep$, respectively.
Besides these metrics, we further calculate the average false positives ($\text{FP}$) and average false negatives ($\text{FN}$) per frame as well as the average ID switches ($\text{IDSw}$) per object.
Finally, we denote the fraction of objects whose trajectories have been covered more than $80\%$ (mostly tracked $\text{MT}$), whose trajectories have been covered between $20\%$ and $80\%$ (partially tracked $\text{PT}$), and which are covered only below $20\%$ (mostly lost $\text{ML}$).
For the final evaluation of the \ac{MOT} metrics, we filter all tracks that have an existence probability above $0.5$.
The tracks with a confidence above the invalid threshold of $0.3$ and below the evaluation threshold of $0.5$ are used during object tracking but are discarded during \ac{MOT} evaluation.

In contrast, we evaluate the calibration metrics using all available information to measure the calibration properties on the complete confidence range.
Thus, we only apply the invalid threshold of $0.3$ only during object tracking.
Similar to the evaluations for semantic confidence calibration in \secref{section:confidence:experiments}, we use the \ac{ECE}, the position-dependent \ac{D-ECE}, the Brier score, and the \ac{NLL} with the same setup as within \secref{section:confidence:experiments} to evaluate the calibration properties of the object existence score.
For the evaluation of the spatial uncertainty in object tracking, common literature uses the \ac{NEES} as an evaluation metric \cite{Bar2004,Van2005}.
In \secref{section:regression:definition:quantile}, we use the \ac{NEES} to construct the \ac{M-QCE} metric which is more interpretable as it denotes the average error (in percent) between predicted and observed quantile.
Therefore, we use the \ac{M-QCE} as well as the \ac{NLL} with the same setup as within \secref{section:regression:experiments} to evaluate the spatial uncertainty calibration.
Note that the evaluation of the \ac{NEES} is commonly performed on the estimated states directly which requires ground-truth information for the whole state vector.
However, no ground-truth information for the velocity and acceleration are available by the used data set.
Furthermore, we perform the state estimation in image coordinates in our setup.
To mitigate these limitations, we transform the estimated states to the observation space using (\ref{eq:tracking:kalman:observation}) and evaluate the state estimation using the available ground-truth information for the object location and size in image coordinates.
In the next section, we present our results using uncertainty calibration for object tracking.

\subsection{Evaluations for intermediate Semantic Confidence Calibration}
\label{section:tracking:experiments:semantic}

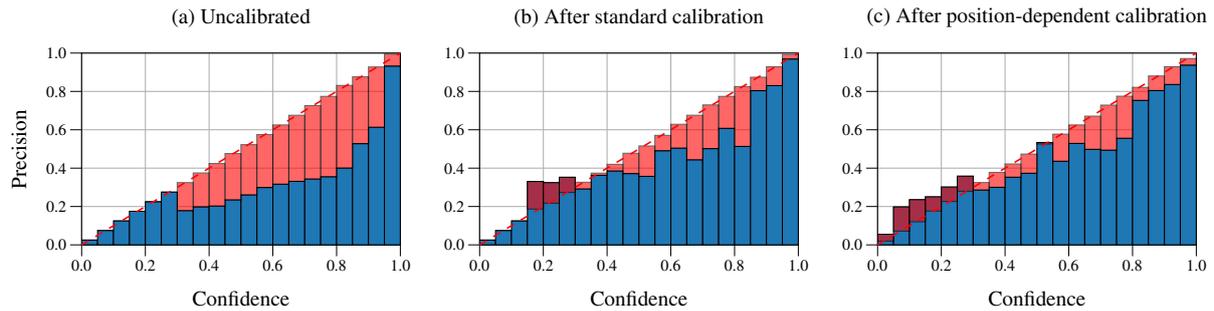
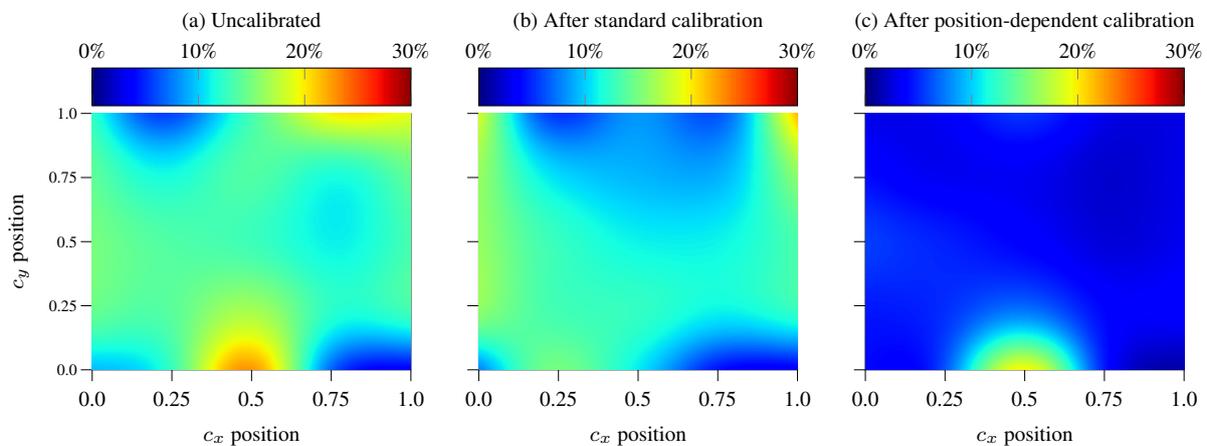
\begin{figure}[ht!]
    \centering
    \begin{subfigure}{\textwidth}
\begin{tikzpicture}

\tikzstyle{every node}=[font=\scriptsize]
\pgfplotsset{every x tick label/.append style={font=\tiny, yshift=0.5ex}}
\pgfplotsset{every y tick label/.append style={font=\tiny, xshift=0.5ex}}

\definecolor{darkgray176}{RGB}{176,176,176}
\definecolor{gray}{RGB}{128,128,128}
\definecolor{lightgray204}{RGB}{204,204,204}
\definecolor{steelblue31119180}{RGB}{31,119,180}

\begin{axis}[
legend cell align={left},
legend style={
  fill opacity=0.8,
  draw opacity=1,
  text opacity=1,
  at={(0.03,0.97)},
  anchor=north west,
  draw=lightgray204
},
width=0.35\textwidth,
height=0.25\textwidth,
tick align=outside,
tick pos=left,
title={(a) Uncalibrated},
x grid style={white!69.0196078431373!black},
xmajorgrids,
xlabel={Confidence},
xmin=0, xmax=1,
xtick style={color=black},
xtick={0,0.2,0.4,0.6,0.8,1},
xticklabels={0.0,0.2,0.4,0.6,0.8,1.0},
y grid style={white!69.0196078431373!black},
ymajorgrids,
ylabel={Precision},
ymin=0, ymax=1,
ytick style={color=black},
ytick={0,0.2,0.4,0.6,0.8,1},
yticklabels={0.0,0.2,0.4,0.6,0.8,1.0}
]
\draw[draw=black,fill=steelblue31119180] (axis cs:-3.46944695195361e-18,0) rectangle (axis cs:0.05,0.025);
\draw[draw=black,fill=steelblue31119180] (axis cs:0.05,0) rectangle (axis cs:0.1,0.075);
\draw[draw=black,fill=steelblue31119180] (axis cs:0.1,0) rectangle (axis cs:0.15,0.125);
\draw[draw=black,fill=steelblue31119180] (axis cs:0.15,0) rectangle (axis cs:0.2,0.175);
\draw[draw=black,fill=steelblue31119180] (axis cs:0.2,0) rectangle (axis cs:0.25,0.225);
\draw[draw=black,fill=steelblue31119180] (axis cs:0.25,0) rectangle (axis cs:0.3,0.275);
\draw[draw=black,fill=steelblue31119180] (axis cs:0.3,0) rectangle (axis cs:0.35,0.178538872248991);
\draw[draw=black,fill=steelblue31119180] (axis cs:0.35,0) rectangle (axis cs:0.4,0.198190900499527);
\draw[draw=black,fill=steelblue31119180] (axis cs:0.4,0) rectangle (axis cs:0.45,0.202577127989097);
\draw[draw=black,fill=steelblue31119180] (axis cs:0.45,0) rectangle (axis cs:0.5,0.234699379886762);
\draw[draw=black,fill=steelblue31119180] (axis cs:0.5,0) rectangle (axis cs:0.55,0.260625701109311);
\draw[draw=black,fill=steelblue31119180] (axis cs:0.55,0) rectangle (axis cs:0.6,0.299094707520891);
\draw[draw=black,fill=steelblue31119180] (axis cs:0.6,0) rectangle (axis cs:0.65,0.31646030589949);
\draw[draw=black,fill=steelblue31119180] (axis cs:0.65,0) rectangle (axis cs:0.7,0.331077651835561);
\draw[draw=black,fill=steelblue31119180] (axis cs:0.7,0) rectangle (axis cs:0.75,0.34341252699784);
\draw[draw=black,fill=steelblue31119180] (axis cs:0.75,0) rectangle (axis cs:0.8,0.355199115044248);
\draw[draw=black,fill=steelblue31119180] (axis cs:0.8,0) rectangle (axis cs:0.85,0.400426807119506);
\draw[draw=black,fill=steelblue31119180] (axis cs:0.85,0) rectangle (axis cs:0.9,0.527704485488127);
\draw[draw=black,fill=steelblue31119180] (axis cs:0.9,0) rectangle (axis cs:0.95,0.613120131201312);
\draw[draw=black,fill=steelblue31119180] (axis cs:0.95,0) rectangle (axis cs:1,0.932117292394321);
\draw[draw=black,fill=red,opacity=0.6] (axis cs:-3.46944695195361e-18,0.025) rectangle (axis cs:0.05,0.025);
\draw[draw=black,fill=red,opacity=0.6] (axis cs:0.05,0.075) rectangle (axis cs:0.1,0.075);
\draw[draw=black,fill=red,opacity=0.6] (axis cs:0.1,0.125) rectangle (axis cs:0.15,0.125);
\draw[draw=black,fill=red,opacity=0.6] (axis cs:0.15,0.175) rectangle (axis cs:0.2,0.175);
\draw[draw=black,fill=red,opacity=0.6] (axis cs:0.2,0.225) rectangle (axis cs:0.25,0.225);
\draw[draw=black,fill=red,opacity=0.6] (axis cs:0.25,0.275) rectangle (axis cs:0.3,0.275);
\draw[draw=black,fill=red,opacity=0.6] (axis cs:0.3,0.178538872248991) rectangle (axis cs:0.35,0.324505532184761);
\draw[draw=black,fill=red,opacity=0.6] (axis cs:0.35,0.198190900499527) rectangle (axis cs:0.4,0.374773837584765);
\draw[draw=black,fill=red,opacity=0.6] (axis cs:0.4,0.202577127989097) rectangle (axis cs:0.45,0.42401200058236);
\draw[draw=black,fill=red,opacity=0.6] (axis cs:0.45,0.234699379886762) rectangle (axis cs:0.5,0.47602226526737);
\draw[draw=black,fill=red,opacity=0.6] (axis cs:0.5,0.260625701109311) rectangle (axis cs:0.55,0.522444604765319);
\draw[draw=black,fill=red,opacity=0.6] (axis cs:0.55,0.299094707520891) rectangle (axis cs:0.6,0.574781629647725);
\draw[draw=black,fill=red,opacity=0.6] (axis cs:0.6,0.31646030589949) rectangle (axis cs:0.65,0.625036387744462);
\draw[draw=black,fill=red,opacity=0.6] (axis cs:0.65,0.331077651835561) rectangle (axis cs:0.7,0.675545726120191);
\draw[draw=black,fill=red,opacity=0.6] (axis cs:0.7,0.34341252699784) rectangle (axis cs:0.75,0.725992895822082);
\draw[draw=black,fill=red,opacity=0.6] (axis cs:0.75,0.355199115044248) rectangle (axis cs:0.8,0.773996728688427);
\draw[draw=black,fill=red,opacity=0.6] (axis cs:0.8,0.400426807119506) rectangle (axis cs:0.85,0.831070820383565);
\draw[draw=black,fill=red,opacity=0.6] (axis cs:0.85,0.527704485488127) rectangle (axis cs:0.9,0.876016153030328);
\draw[draw=black,fill=red,opacity=0.6] (axis cs:0.9,0.613120131201312) rectangle (axis cs:0.95,0.927483777465701);
\draw[draw=black,fill=red,opacity=0.6] (axis cs:0.95,0.932117292394321) rectangle (axis cs:1,0.994435006648777);
\addplot [semithick, red, dashed]
table {%
0 0
1 1
};
\end{axis}

\end{tikzpicture}
\begin{tikzpicture}

\tikzstyle{every node}=[font=\scriptsize]
\pgfplotsset{every x tick label/.append style={font=\tiny, yshift=0.5ex}}
\pgfplotsset{every y tick label/.append style={font=\tiny, xshift=0.5ex}}

\definecolor{darkgray176}{RGB}{176,176,176}
\definecolor{gray}{RGB}{128,128,128}
\definecolor{lightgray204}{RGB}{204,204,204}
\definecolor{steelblue31119180}{RGB}{31,119,180}

\begin{axis}[
legend cell align={left},
legend style={
    fill opacity=0.8,
    draw opacity=1,
    text opacity=1,
    at={(0.03,0.97)},
    anchor=north west,
    draw=lightgray204
},
width=0.35\textwidth,
height=0.25\textwidth,
tick align=outside,
tick pos=left,
title={(b) After standard calibration},
x grid style={white!69.0196078431373!black},
xmajorgrids,
xlabel={Confidence},
xmin=0, xmax=1,
xtick style={color=black},
xtick={0,0.2,0.4,0.6,0.8,1},
xticklabels={0.0,0.2,0.4,0.6,0.8,1.0},
y grid style={white!69.0196078431373!black},
ymajorgrids,
ymin=0, ymax=1,
ytick style={color=black},
ytick={0,0.2,0.4,0.6,0.8,1},
yticklabels={0.0,0.2,0.4,0.6,0.8,1.0}
]
\draw[draw=black,fill=steelblue31119180] (axis cs:-3.46944695195361e-18,0) rectangle (axis cs:0.05,0.025);
\draw[draw=black,fill=steelblue31119180] (axis cs:0.05,0) rectangle (axis cs:0.1,0.075);
\draw[draw=black,fill=steelblue31119180] (axis cs:0.1,0) rectangle (axis cs:0.15,0.125);
\draw[draw=black,fill=steelblue31119180] (axis cs:0.15,0) rectangle (axis cs:0.2,0.330357142857143);
\draw[draw=black,fill=steelblue31119180] (axis cs:0.2,0) rectangle (axis cs:0.25,0.324299494717501);
\draw[draw=black,fill=steelblue31119180] (axis cs:0.25,0) rectangle (axis cs:0.3,0.352331606217617);
\draw[draw=black,fill=steelblue31119180] (axis cs:0.3,0) rectangle (axis cs:0.35,0.291544374563243);
\draw[draw=black,fill=steelblue31119180] (axis cs:0.35,0) rectangle (axis cs:0.4,0.362745098039216);
\draw[draw=black,fill=steelblue31119180] (axis cs:0.4,0) rectangle (axis cs:0.45,0.385324232081911);
\draw[draw=black,fill=steelblue31119180] (axis cs:0.45,0) rectangle (axis cs:0.5,0.371690031152648);
\draw[draw=black,fill=steelblue31119180] (axis cs:0.5,0) rectangle (axis cs:0.55,0.357163573085847);
\draw[draw=black,fill=steelblue31119180] (axis cs:0.55,0) rectangle (axis cs:0.6,0.490774023484304);
\draw[draw=black,fill=steelblue31119180] (axis cs:0.6,0) rectangle (axis cs:0.65,0.504373177842566);
\draw[draw=black,fill=steelblue31119180] (axis cs:0.65,0) rectangle (axis cs:0.7,0.443502824858757);
\draw[draw=black,fill=steelblue31119180] (axis cs:0.7,0) rectangle (axis cs:0.75,0.501749331138094);
\draw[draw=black,fill=steelblue31119180] (axis cs:0.75,0) rectangle (axis cs:0.8,0.607450305212083);
\draw[draw=black,fill=steelblue31119180] (axis cs:0.8,0) rectangle (axis cs:0.85,0.513204146011717);
\draw[draw=black,fill=steelblue31119180] (axis cs:0.85,0) rectangle (axis cs:0.9,0.80391061452514);
\draw[draw=black,fill=steelblue31119180] (axis cs:0.9,0) rectangle (axis cs:0.95,0.830238147258029);
\draw[draw=black,fill=steelblue31119180] (axis cs:0.95,0) rectangle (axis cs:1,0.969016461272605);
\draw[draw=black,fill=red,opacity=0.6] (axis cs:-3.46944695195361e-18,0.025) rectangle (axis cs:0.05,0.025);
\draw[draw=black,fill=red,opacity=0.6] (axis cs:0.05,0.075) rectangle (axis cs:0.1,0.075);
\draw[draw=black,fill=red,opacity=0.6] (axis cs:0.1,0.125) rectangle (axis cs:0.15,0.125);
\draw[draw=black,fill=red,opacity=0.6] (axis cs:0.15,0.330357142857143) rectangle (axis cs:0.2,0.186672825418441);
\draw[draw=black,fill=red,opacity=0.6] (axis cs:0.2,0.324299494717501) rectangle (axis cs:0.25,0.21718514183212);
\draw[draw=black,fill=red,opacity=0.6] (axis cs:0.25,0.352331606217617) rectangle (axis cs:0.3,0.273158980883506);
\draw[draw=black,fill=red,opacity=0.6] (axis cs:0.3,0.291544374563243) rectangle (axis cs:0.35,0.326432117487968);
\draw[draw=black,fill=red,opacity=0.6] (axis cs:0.35,0.362745098039216) rectangle (axis cs:0.4,0.373973752157069);
\draw[draw=black,fill=red,opacity=0.6] (axis cs:0.4,0.385324232081911) rectangle (axis cs:0.45,0.419919526967614);
\draw[draw=black,fill=red,opacity=0.6] (axis cs:0.45,0.371690031152648) rectangle (axis cs:0.5,0.477629881062415);
\draw[draw=black,fill=red,opacity=0.6] (axis cs:0.5,0.357163573085847) rectangle (axis cs:0.55,0.515483081290209);
\draw[draw=black,fill=red,opacity=0.6] (axis cs:0.55,0.490774023484304) rectangle (axis cs:0.6,0.570745261482695);
\draw[draw=black,fill=red,opacity=0.6] (axis cs:0.6,0.504373177842566) rectangle (axis cs:0.65,0.628551886439356);
\draw[draw=black,fill=red,opacity=0.6] (axis cs:0.65,0.443502824858757) rectangle (axis cs:0.7,0.675629061385108);
\draw[draw=black,fill=red,opacity=0.6] (axis cs:0.7,0.501749331138094) rectangle (axis cs:0.75,0.72944836877044);
\draw[draw=black,fill=red,opacity=0.6] (axis cs:0.75,0.607450305212083) rectangle (axis cs:0.8,0.774019207130162);
\draw[draw=black,fill=red,opacity=0.6] (axis cs:0.8,0.513204146011717) rectangle (axis cs:0.85,0.825265335576543);
\draw[draw=black,fill=red,opacity=0.6] (axis cs:0.85,0.80391061452514) rectangle (axis cs:0.9,0.873921370355021);
\draw[draw=black,fill=red,opacity=0.6] (axis cs:0.9,0.830238147258029) rectangle (axis cs:0.95,0.92839090583682);
\draw[draw=black,fill=red,opacity=0.6] (axis cs:0.95,0.969016461272605) rectangle (axis cs:1,0.992392966761829);
\addplot [semithick, red, dashed]
table {%
0 0
1 1
};
\end{axis}

\end{tikzpicture}
\begin{tikzpicture}

\tikzstyle{every node}=[font=\scriptsize]
\pgfplotsset{every x tick label/.append style={font=\tiny, yshift=0.5ex}}
\pgfplotsset{every y tick label/.append style={font=\tiny, xshift=0.5ex}}

\definecolor{darkgray176}{RGB}{176,176,176}
\definecolor{gray}{RGB}{128,128,128}
\definecolor{lightgray204}{RGB}{204,204,204}
\definecolor{steelblue31119180}{RGB}{31,119,180}

\begin{axis}[
legend cell align={left},
legend style={
  fill opacity=0.8,
  draw opacity=1,
  text opacity=1,
  at={(0.03,0.97)},
  anchor=north west,
  draw=lightgray204
},
width=0.35\textwidth,
height=0.25\textwidth,
tick align=outside,
tick pos=left,
title={(c) After position-dependent calibration},
x grid style={white!69.0196078431373!black},
xmajorgrids,
xlabel={Confidence},
xmin=0, xmax=1,
xtick style={color=black},
xtick={0,0.2,0.4,0.6,0.8,1},
xticklabels={0.0,0.2,0.4,0.6,0.8,1.0},
y grid style={white!69.0196078431373!black},
ymajorgrids,
ymin=0, ymax=1,
ytick style={color=black},
ytick={0,0.2,0.4,0.6,0.8,1},
yticklabels={0.0,0.2,0.4,0.6,0.8,1.0}
]
\draw[draw=black,fill=steelblue31119180] (axis cs:-3.46944695195361e-18,0) rectangle (axis cs:0.05,0.0551751592356688);
\draw[draw=black,fill=steelblue31119180] (axis cs:0.05,0) rectangle (axis cs:0.1,0.197667638483965);
\draw[draw=black,fill=steelblue31119180] (axis cs:0.1,0) rectangle (axis cs:0.15,0.235625109899771);
\draw[draw=black,fill=steelblue31119180] (axis cs:0.15,0) rectangle (axis cs:0.2,0.250508411906082);
\draw[draw=black,fill=steelblue31119180] (axis cs:0.2,0) rectangle (axis cs:0.25,0.301499605367009);
\draw[draw=black,fill=steelblue31119180] (axis cs:0.25,0) rectangle (axis cs:0.3,0.358063031976075);
\draw[draw=black,fill=steelblue31119180] (axis cs:0.3,0) rectangle (axis cs:0.35,0.28600061671292);
\draw[draw=black,fill=steelblue31119180] (axis cs:0.35,0) rectangle (axis cs:0.4,0.300028918449971);
\draw[draw=black,fill=steelblue31119180] (axis cs:0.4,0) rectangle (axis cs:0.45,0.352776424744277);
\draw[draw=black,fill=steelblue31119180] (axis cs:0.45,0) rectangle (axis cs:0.5,0.373246393993282);
\draw[draw=black,fill=steelblue31119180] (axis cs:0.5,0) rectangle (axis cs:0.55,0.533611383709519);
\draw[draw=black,fill=steelblue31119180] (axis cs:0.55,0) rectangle (axis cs:0.6,0.436018957345972);
\draw[draw=black,fill=steelblue31119180] (axis cs:0.6,0) rectangle (axis cs:0.65,0.528828828828829);
\draw[draw=black,fill=steelblue31119180] (axis cs:0.65,0) rectangle (axis cs:0.7,0.498145675715769);
\draw[draw=black,fill=steelblue31119180] (axis cs:0.7,0) rectangle (axis cs:0.75,0.493767313019391);
\draw[draw=black,fill=steelblue31119180] (axis cs:0.75,0) rectangle (axis cs:0.8,0.555797022709986);
\draw[draw=black,fill=steelblue31119180] (axis cs:0.8,0) rectangle (axis cs:0.85,0.753153484946138);
\draw[draw=black,fill=steelblue31119180] (axis cs:0.85,0) rectangle (axis cs:0.9,0.804488028977825);
\draw[draw=black,fill=steelblue31119180] (axis cs:0.9,0) rectangle (axis cs:0.95,0.835997984332768);
\draw[draw=black,fill=steelblue31119180] (axis cs:0.95,0) rectangle (axis cs:1,0.937059661048087);
\draw[draw=black,fill=red,opacity=0.6] (axis cs:-3.46944695195361e-18,0.0551751592356688) rectangle (axis cs:0.05,0.0196635941192884);
\draw[draw=black,fill=red,opacity=0.6] (axis cs:0.05,0.197667638483965) rectangle (axis cs:0.1,0.0717641514079814);
\draw[draw=black,fill=red,opacity=0.6] (axis cs:0.1,0.235625109899771) rectangle (axis cs:0.15,0.120444151742688);
\draw[draw=black,fill=red,opacity=0.6] (axis cs:0.15,0.250508411906082) rectangle (axis cs:0.2,0.176279877315525);
\draw[draw=black,fill=red,opacity=0.6] (axis cs:0.2,0.301499605367009) rectangle (axis cs:0.25,0.226323880140955);
\draw[draw=black,fill=red,opacity=0.6] (axis cs:0.25,0.358063031976075) rectangle (axis cs:0.3,0.279604789845694);
\draw[draw=black,fill=red,opacity=0.6] (axis cs:0.3,0.28600061671292) rectangle (axis cs:0.35,0.324907614746669);
\draw[draw=black,fill=red,opacity=0.6] (axis cs:0.35,0.300028918449971) rectangle (axis cs:0.4,0.377753524054863);
\draw[draw=black,fill=red,opacity=0.6] (axis cs:0.4,0.352776424744277) rectangle (axis cs:0.45,0.421791933515126);
\draw[draw=black,fill=red,opacity=0.6] (axis cs:0.45,0.373246393993282) rectangle (axis cs:0.5,0.473384417998322);
\draw[draw=black,fill=red,opacity=0.6] (axis cs:0.5,0.533611383709519) rectangle (axis cs:0.55,0.526055800556775);
\draw[draw=black,fill=red,opacity=0.6] (axis cs:0.55,0.436018957345972) rectangle (axis cs:0.6,0.577214155213082);
\draw[draw=black,fill=red,opacity=0.6] (axis cs:0.6,0.528828828828829) rectangle (axis cs:0.65,0.625083489639577);
\draw[draw=black,fill=red,opacity=0.6] (axis cs:0.65,0.498145675715769) rectangle (axis cs:0.7,0.670994906078049);
\draw[draw=black,fill=red,opacity=0.6] (axis cs:0.7,0.493767313019391) rectangle (axis cs:0.75,0.728475510988035);
\draw[draw=black,fill=red,opacity=0.6] (axis cs:0.75,0.555797022709986) rectangle (axis cs:0.8,0.774871740206531);
\draw[draw=black,fill=red,opacity=0.6] (axis cs:0.8,0.753153484946138) rectangle (axis cs:0.85,0.821040975881492);
\draw[draw=black,fill=red,opacity=0.6] (axis cs:0.85,0.804488028977825) rectangle (axis cs:0.9,0.880420327942746);
\draw[draw=black,fill=red,opacity=0.6] (axis cs:0.9,0.835997984332768) rectangle (axis cs:0.95,0.928708507071874);
\draw[draw=black,fill=red,opacity=0.6] (axis cs:0.95,0.937059661048087) rectangle (axis cs:1,0.971315817928044);
\addplot [semithick, red, dashed]
table {%
0 0
1 1
};
\end{axis}

\end{tikzpicture}
        \caption{
            Reliability diagrams w.r.t. the confidence only (0d).
            The uncalibrated object tracking is consistently overconfident for all confidence levels.
            This is mitigated by standard Histogram Binning \cite{Zadrozny2001} and even further improved using our position-dependent Histogram Binning (cf. \secref{section:confidence:methods:binning}).
        }
        \label{fig:tracking:evaluation:semantic:reliability:confidence:0d}
    \end{subfigure}
    \vspace{1cm}
    
    \begin{subfigure}{\textwidth}
\begin{tikzpicture}

\tikzstyle{every node}=[font=\scriptsize]
\pgfplotsset{every y tick label/.append style={font=\tiny, xshift=0.5ex}}
\pgfplotsset{ every non boxed x axis/.append style={x axis line style=-}, every non boxed y axis/.append style={y axis line style=-}}

\definecolor{darkgray176}{RGB}{176,176,176}

\begin{axis}[
width=0.35\textwidth,
colorbar horizontal,
colormap={mymap}{[1pt]
    rgb(0pt)=(0,0,0.5);
    rgb(22pt)=(0,0,1);
    rgb(25pt)=(0,0,1);
    rgb(68pt)=(0,0.86,1);
    rgb(70pt)=(0,0.9,0.967741935483871);
    rgb(75pt)=(0.0806451612903226,1,0.887096774193548);
    rgb(128pt)=(0.935483870967742,1,0.0322580645161291);
    rgb(130pt)=(0.967741935483871,0.962962962962963,0);
    rgb(132pt)=(1,0.925925925925926,0);
    rgb(178pt)=(1,0.0740740740740741,0);
    rgb(182pt)=(0.909090909090909,0,0);
    rgb(200pt)=(0.5,0,0)
},
colorbar style={
    title style={at={(0.5,0)},anchor=north,yshift=3em},
    title={(a) Uncalibrated},
    xticklabels={0, 0\%, 10\%, 20\%, 30\%},
    x tick scale label style={yshift=0.5cm},
    at={(1,1.03)},anchor=south east,
    width=1.0*\pgfkeysvalueof{/pgfplots/parent axis width},
    xticklabel pos=upper,
},
point meta max=0.3,
point meta min=0,
tick align=outside,
tick pos=left,
x grid style={white!69.0196078431373!black},
xlabel={$c_x$ position},
xmin=0, xmax=8,
xtick style={color=black},
xtick={0,2,4,6,8},
xticklabels={0.0,0.25,0.5,0.75,1.0},
y grid style={white!69.0196078431373!black},
ylabel={$c_y$ position},
ymin=0, ymax=8,
ytick style={color=black},
ytick={0,2,4,6,8},
yticklabels={0.0,0.25,0.5,0.75,1.0}
]
\addplot graphics [includegraphics cmd=\pgfimage,xmin=-0.5, xmax=8.5, ymin=-0.5, ymax=8.5] {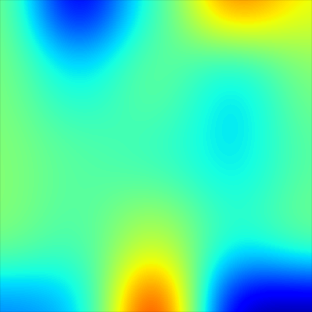};
\end{axis}

\end{tikzpicture}
\begin{tikzpicture}

\tikzstyle{every node}=[font=\scriptsize]
\pgfplotsset{every y tick label/.append style={font=\tiny, xshift=0.5ex}}
\pgfplotsset{ every non boxed x axis/.append style={x axis line style=-}, every non boxed y axis/.append style={y axis line style=-}}

\definecolor{darkgray176}{RGB}{176,176,176}

\begin{axis}[
width=0.35\textwidth,
colorbar horizontal,
colormap={mymap}{[1pt]
    rgb(0pt)=(0,0,0.5);
    rgb(22pt)=(0,0,1);
    rgb(25pt)=(0,0,1);
    rgb(68pt)=(0,0.86,1);
    rgb(70pt)=(0,0.9,0.967741935483871);
    rgb(75pt)=(0.0806451612903226,1,0.887096774193548);
    rgb(128pt)=(0.935483870967742,1,0.0322580645161291);
    rgb(130pt)=(0.967741935483871,0.962962962962963,0);
    rgb(132pt)=(1,0.925925925925926,0);
    rgb(178pt)=(1,0.0740740740740741,0);
    rgb(182pt)=(0.909090909090909,0,0);
    rgb(200pt)=(0.5,0,0)
},
colorbar style={
    title style={at={(0.5,0)},anchor=north,yshift=3em},
    title=(b) After standard calibration,
    xticklabels={0, 0\%, 10\%, 20\%, 30\%},
    x tick scale label style={yshift=0.5cm},
    at={(1,1.03)},anchor=south east,
    width=1.0*\pgfkeysvalueof{/pgfplots/parent axis width},
    xticklabel pos=upper,
},
point meta max=0.3,
point meta min=0,
tick align=outside,
tick pos=left,
x grid style={white!69.0196078431373!black},
xlabel={$c_x$ position},
xmin=0, xmax=8,
xtick style={color=black},
xtick={0,2,4,6,8},
xticklabels={0.0,0.25,0.5,0.75,1.0},
y grid style={white!69.0196078431373!black},
ymin=0, ymax=8,
ytick style={color=black},
ytick={},
yticklabels={},
]
\addplot graphics [includegraphics cmd=\pgfimage,xmin=-0.5, xmax=8.5, ymin=-0.5, ymax=8.5] {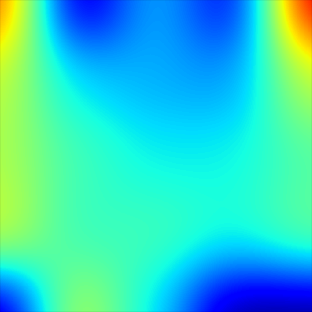};
\end{axis}

\end{tikzpicture}
\begin{tikzpicture}

\tikzstyle{every node}=[font=\scriptsize]
\pgfplotsset{every y tick label/.append style={font=\tiny, xshift=0.5ex}}
\pgfplotsset{ every non boxed x axis/.append style={x axis line style=-}, every non boxed y axis/.append style={y axis line style=-}}

\definecolor{darkgray176}{RGB}{176,176,176}

\begin{axis}[
width=0.35\textwidth,
colorbar horizontal,
colormap={mymap}{[1pt]
    rgb(0pt)=(0,0,0.5);
    rgb(22pt)=(0,0,1);
    rgb(25pt)=(0,0,1);
    rgb(68pt)=(0,0.86,1);
    rgb(70pt)=(0,0.9,0.967741935483871);
    rgb(75pt)=(0.0806451612903226,1,0.887096774193548);
    rgb(128pt)=(0.935483870967742,1,0.0322580645161291);
    rgb(130pt)=(0.967741935483871,0.962962962962963,0);
    rgb(132pt)=(1,0.925925925925926,0);
    rgb(178pt)=(1,0.0740740740740741,0);
    rgb(182pt)=(0.909090909090909,0,0);
    rgb(200pt)=(0.5,0,0)
},
colorbar style={
    title style={at={(0.5,0)},anchor=north,yshift=3em},
    title=(c) After position-dependent calibration,
    xticklabels={0, 0\%, 10\%, 20\%, 30\%},
    x tick scale label style={yshift=0.5cm},
    at={(1,1.03)},anchor=south east,
    width=1.0*\pgfkeysvalueof{/pgfplots/parent axis width},
    xticklabel pos=upper,
},
point meta max=0.3,
point meta min=0,
tick align=outside,
tick pos=left,
x grid style={white!69.0196078431373!black},
xlabel={$c_x$ position},
xmin=0, xmax=8,
xtick style={color=black},
xtick={0,2,4,6,8},
xticklabels={0.0,0.25,0.5,0.75,1.0},
y grid style={white!69.0196078431373!black},
ymin=0, ymax=8,
ytick style={color=black},
ytick={},
yticklabels={},
]
\addplot graphics [includegraphics cmd=\pgfimage,xmin=-0.5, xmax=8.5, ymin=-0.5, ymax=8.5] {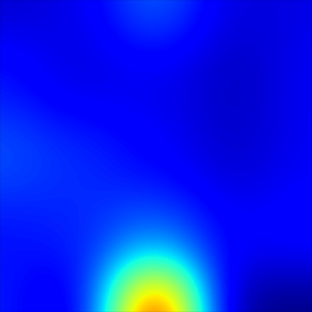};
\end{axis}

\end{tikzpicture}
        \caption{
            Reliability diagrams w.r.t. the $\centerx$ and $\centerx$ position of the tracked objects (2d).
            The standard Histogram Binning \cite{Zadrozny2001} improves the calibration properties of the uncalibrated object tracking.
            However, our position-dependent Histogram Binning is able to reduce miscalibration consistently for nearly all image regions.
        }
        \label{fig:tracking:evaluation:semantic:reliability:confidence:2d}
    \end{subfigure}
    \vspace{1cm}
    
    \caption[Reliability diagrams (object tracking) of the semantic confidence for a \fasterrcnn{} on the Berkeley DeepDrive tracking calibration validation set before and after calibration by Histogram Binning.]{
        Reliability diagrams (object tracking) of the semantic confidence for a \fasterrcnn{} on the Berkeley DeepDrive tracking calibration validation set before and after calibration by Histogram Binning.
        For each frame, we determine for each track if it matches a real ground-truth annotation.
        In combination with the track confidence for each object at a certain time step, we are able to compute a reliability diagram over all objects and time steps.
        As already known from the evaluations for object detection calibration in \secref{section:confidence:experiments:detection}, the \fasterrcnn{} is consistently too overconfident in its predictions.
        Accordingly, we can observe the same phenomenon for the track confidence.
        If we now apply intermediate confidence calibration by standard Histogram Binning, we can also observe a better calibration for the track confidence.
        The calibration properties are further improved by our position-dependent Histogram Binning (cf. \secref{section:confidence:methods:binning}).
    }
    \label{fig:tracking:evaluation:semantic:reliability:confidence}
\end{figure}

\begin{figure}[t!]
    \centering
    \begin{subfigure}{0.5\textwidth}
\begin{tikzpicture}

\tikzstyle{every node}=[font=\scriptsize]
\pgfplotsset{every x tick label/.append style={font=\scriptsize, yshift=0.5ex}}
\pgfplotsset{every y tick label/.append style={font=\scriptsize, xshift=0.5ex}}

\definecolor{darkgray176}{RGB}{176,176,176}
\definecolor{steelblue31119180}{RGB}{31,119,180}

\begin{axis}[
width=0.9\textwidth,
height=0.75\textwidth,
tick align=outside,
tick pos=left,
title=Object Tracking Accuracy,
x grid style={darkgray176},
xmajorgrids,
xmin=-0.84, xmax=8.84,
xtick style={color=black},
xtick={0,1,2,3,4,5,6,7,8},
xticklabel style={rotate=40.0,anchor=east},
xticklabels={
  Uncalibrated,
  Histogram Binning (conf.),
  Histogram Binning (mv.),
  Logistic Calibration (conf.),
  Logistic Calibration (indep.),
  Logistic Calibration (dep.),
  Beta Calibration (conf.),
  Beta Calibration (indep.),
  Beta Calibration (dep.)
},
y grid style={darkgray176},
ylabel={Multiple Object Tracking Accuracy (MOTA)},
ymajorgrids,
ymin=30, ymax=50,
ytick style={color=black}
]
\draw[draw=none,fill=steelblue31119180] (axis cs:-0.4,0) rectangle (axis cs:0.4,36.845);
\draw[draw=none,fill=steelblue31119180] (axis cs:0.6,0) rectangle (axis cs:1.4,44.201);
\draw[draw=none,fill=steelblue31119180] (axis cs:1.6,0) rectangle (axis cs:2.4,47.404);
\draw[draw=none,fill=steelblue31119180] (axis cs:2.6,0) rectangle (axis cs:3.4,43.581);
\draw[draw=none,fill=steelblue31119180] (axis cs:3.6,0) rectangle (axis cs:4.4,42.086);
\draw[draw=none,fill=steelblue31119180] (axis cs:4.6,0) rectangle (axis cs:5.4,43.226);
\draw[draw=none,fill=steelblue31119180] (axis cs:5.6,0) rectangle (axis cs:6.4,43.505);
\draw[draw=none,fill=steelblue31119180] (axis cs:6.6,0) rectangle (axis cs:7.4,42.941);
\draw[draw=none,fill=steelblue31119180] (axis cs:7.6,0) rectangle (axis cs:8.4,43.661);
\end{axis}

\end{tikzpicture}
        \caption{Tracking accuracy before and after calibration.}
        \label{fig:tracking:evaluation:semantic:mota:mota}
    \end{subfigure}%
    \begin{subfigure}{0.5\textwidth}
\begin{tikzpicture}

\tikzstyle{every node}=[font=\scriptsize]
\pgfplotsset{every x tick label/.append style={font=\scriptsize, yshift=0.5ex}}
\pgfplotsset{every y tick label/.append style={font=\scriptsize, xshift=0.5ex}}

\definecolor{darkgray176}{RGB}{176,176,176}
\definecolor{green}{RGB}{0,128,0}
\definecolor{lightgray204}{RGB}{204,204,204}
\definecolor{orange}{RGB}{255,165,0}

\begin{axis}[
width=0.9\textwidth,
height=0.75\textwidth,
legend cell align={left},
legend style={fill opacity=1.0, draw opacity=1, text opacity=1, draw=lightgray204, at={(0.375, 0.65)}, anchor=north west},
tick align=outside,
tick pos=left,
title=Ground-Truth Coverage,
x grid style={darkgray176},
xmajorgrids,
xmin=-0.6, xmax=8.6,
xtick style={color=black},
xtick={0,1,2,3,4,5,6,7,8},
xticklabel style={rotate=40.0,anchor=east},
xticklabels={
  Uncalibrated,
  Histogram Binning (conf.),
  Histogram Binning (mv.),
  Logistic Calibration (conf.),
  Logistic Calibration (indep.),
  Logistic Calibration (dep.),
  Beta Calibration (conf.),
  Beta Calibration (indep.),
  Beta Calibration (dep.)
},
y grid style={darkgray176},
ylabel={Relative Amount of Ground-Truth Objects},
ymajorgrids,
ymin=0, ymax=1,
ytick style={color=black},
yticklabels={0, 0.0,0.2,0.4,0.6,0.8,1.0},
]
\draw[draw=none,fill=green] (axis cs:-0.4,0) rectangle (axis cs:0.4,0.382);
\draw[draw=none,fill=green] (axis cs:0.6,0) rectangle (axis cs:1.4,0.273);
\draw[draw=none,fill=green] (axis cs:1.6,0) rectangle (axis cs:2.4,0.266);
\draw[draw=none,fill=green] (axis cs:2.6,0) rectangle (axis cs:3.4,0.263);
\draw[draw=none,fill=green] (axis cs:3.6,0) rectangle (axis cs:4.4,0.261);
\draw[draw=none,fill=green] (axis cs:4.6,0) rectangle (axis cs:5.4,0.260);
\draw[draw=none,fill=green] (axis cs:5.6,0) rectangle (axis cs:6.4,0.259);
\draw[draw=none,fill=green] (axis cs:6.6,0) rectangle (axis cs:7.4,0.270);
\draw[draw=none,fill=green] (axis cs:7.6,0) rectangle (axis cs:8.4,0.267);
\draw[draw=none,fill=orange] (axis cs:-0.4,0.382) rectangle (axis cs:0.4,0.820);
\draw[draw=none,fill=orange] (axis cs:0.6,0.273) rectangle (axis cs:1.4,0.729);
\draw[draw=none,fill=orange] (axis cs:1.6,0.266) rectangle (axis cs:2.4,0.763);
\draw[draw=none,fill=orange] (axis cs:2.6,0.263) rectangle (axis cs:3.4,0.721);
\draw[draw=none,fill=orange] (axis cs:3.6,0.261) rectangle (axis cs:4.4,0.719);
\draw[draw=none,fill=orange] (axis cs:4.6,0.260) rectangle (axis cs:5.4,0.721);
\draw[draw=none,fill=orange] (axis cs:5.6,0.259) rectangle (axis cs:6.4,0.718);
\draw[draw=none,fill=orange] (axis cs:6.6,0.270) rectangle (axis cs:7.4,0.720);
\draw[draw=none,fill=orange] (axis cs:7.6,0.267) rectangle (axis cs:8.4,0.723);
\draw[draw=none,fill=red] (axis cs:-0.4,0.820) rectangle (axis cs:0.4,1);
\draw[draw=none,fill=red] (axis cs:0.6,0.729) rectangle (axis cs:1.4,1);
\draw[draw=none,fill=red] (axis cs:1.6,0.763) rectangle (axis cs:2.4,1);
\draw[draw=none,fill=red] (axis cs:2.6,0.721) rectangle (axis cs:3.4,1);
\draw[draw=none,fill=red] (axis cs:3.6,0.719) rectangle (axis cs:4.4,1);
\draw[draw=none,fill=red] (axis cs:4.6,0.721) rectangle (axis cs:5.4,1);
\draw[draw=none,fill=red] (axis cs:5.6,0.718) rectangle (axis cs:6.4,1);
\draw[draw=none,fill=red] (axis cs:6.6,0.720) rectangle (axis cs:7.4,1);
\draw[draw=none,fill=red] (axis cs:7.6,0.723) rectangle (axis cs:8.4,1);

\addlegendimage{area legend, black, fill=green}
\addlegendentry{Mostly Tracked (MT)}
\addlegendimage{area legend, black, fill=orange}
\addlegendentry{Partially Tracked (PT)}
\addlegendimage{area legend, black, fill=red}
\addlegendentry{Mostly Lost (ML)}

\end{axis}

\end{tikzpicture}
        \caption{Ground-truth coverage before and after calibration.}
        \label{fig:tracking:evaluation:semantic:mota:mt}
    \end{subfigure}
    \caption[Visualization of the Multiple Object Tracking (MOT) metrics before and after semantic confidence calibration.]{
        Visualization of the \ac{MOT} metrics before and after semantic confidence calibration.
        In this case, the position-dependent (mv.) Histogram Binning achieves the best results in the $\text{MOTA}$ metric (left diagram).
        Furthermore, it is almost able to preserve the tracking coverage of the uncalibrated baseline model (right diagram).
    }
    \label{fig:tracking:evaluation:semantic:mota}
\end{figure}
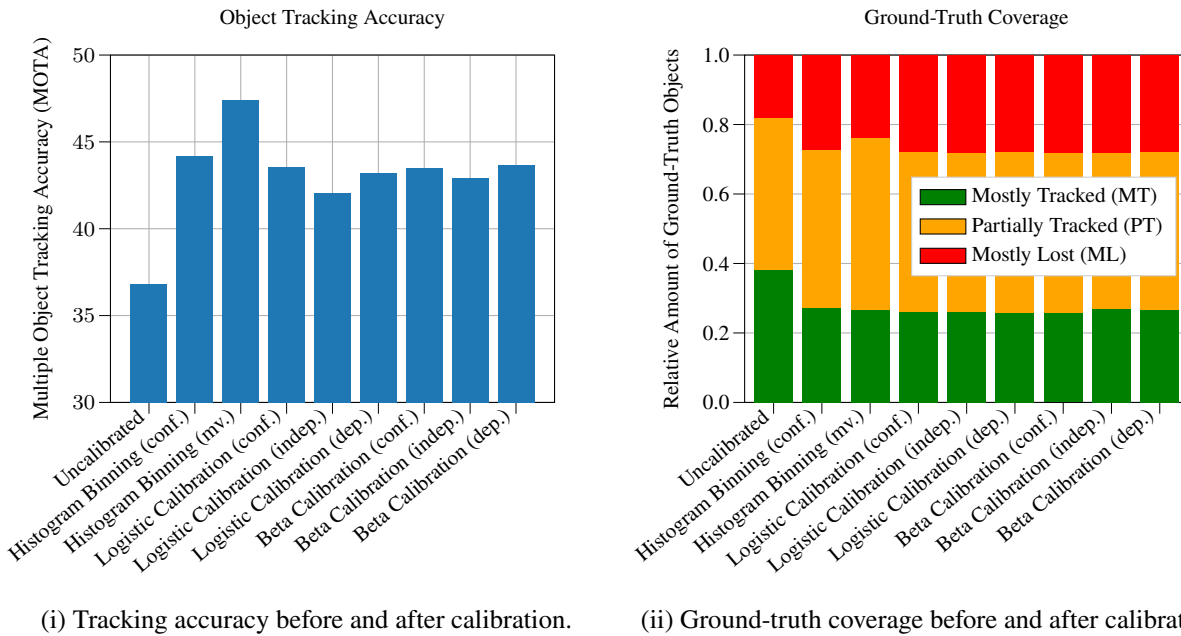

We start with including the semantic confidence calibration methods into the object tracking that have been presented in \secref{section:confidence:methods}.
For confidence calibration, we use the standard Histogram Binning \cite{Zadrozny2001}, Logistic Calibration \cite{Platt1999}, and Beta Calibration \cite{Kull2017} as well as their multivariate and position-dependent counter parts presented in \secref{section:confidence:methods:binning} and \ref{section:confidence:methods:scaling}, respectively.
In the following, the position-dependent methods are denoted by ``mv.'' to distinguish between the standard confidence-only and the multivariate calibration methods.
For the position-dependent Logistic Calibration and Beta Calibration, we further distinguish between the conditional independent (indep.) and the conditional dependent (dep.) variants which also model correlations between the position information (cf. \secref{section:confidence:methods:scaling}).
The results of the object tracking with semantic confidence calibration are presented in \tabref{tab:tracking:evaluation:existence:mot} and \tabref{tab:tracking:evaluation:existence:stats} with the \ac{MOT} and calibration metrics, respectively.
Furthermore, the calibration results are presented in \figref{fig:tracking:evaluation:semantic:reliability:confidence} and \ref{fig:tracking:evaluation:semantic:mota} as reliability diagrams and as a visualization of the track coverage, respectively.
\begin{table}[b!]
    \centering
    \caption[Results in the Multiple Object Tracking (MOT) metrics for object existence estimation using semantic confidence calibration.]{
        Results in the \ac{MOT} metrics for object existence estimation using semantic confidence calibration
        The best scores are highlighted in bold.
        The multivariate Histogram Binning as well as the multivariate and conditionally dependent scaling methods achieve the best results.
    }
\begin{tabular}{l|l|ccc|ccc|ccc}
    \hline
    \makecell[l]{Calibration\\method} & \makecell[l]{Calibration\\type} & MOTA & MOTP & IDF1 & IDSw & FP & FN & MT & PT & ML \\ \hline \hline
    Uncalibrated & \makecell{-} & 36.845 & 82.679 & 47.88 & 2.23 & 3.06 & \textbf{3.16} & \textbf{0.382} & 0.438 & \textbf{0.180} \\ \hline
    \multirow{2}{*}{\makecell[l]{Histogram\\Binning}} & conf. only & 44.201 & 84.212 & 50.11 & 1.68 & 1.52 & 4.12 & 0.273 & 0.456 & 0.270 \\
    & mv. & \textbf{47.404} & 83.805 & \textbf{50.98} & 1.64 & \textbf{1.41} & 3.88 & 0.266 & 0.497 & 0.237 \\ \hline
    \multirow{3}{*}{\makecell[l]{Logistic\\Calibration}} & conf. only &  43.581 & 84.342 & 50.06 & 1.61 & 1.52 & 4.22 & 0.263 & 0.458 & 0.279 \\
    & mv. (indep.) & 42.086 & 84.396 & 49.53 & 1.71 & 1.64 & 4.23 & 0.261 & 0.458 & 0.281 \\
    & mv. (dep.) & 43.226 & \textbf{84.422} & 49.97 & \textbf{1.58} & 1.55 & 4.25 & 0.260 & 0.461 & 0.279 \\ \hline
    \multirow{3}{*}{\makecell[l]{Beta\\Calibration}} & conf. only & 43.505 & 84.371 & 50.10 & 1.59 & 1.53 & 4.24 & 0.259 & 0.459 & 0.282 \\
    & mv. (indep.) & 42.941 & 84.329 & 49.71 & 1.67 & 1.59 & 4.20 & 0.270 & 0.450 & 0.280 \\
    & mv. (dep.) & 43.661 & 84.323 & 49.87 & \textbf{1.58} & 1.57 & 4.17 & 0.267 & 0.456 & 0.277 \\ \hline
\end{tabular}%
    \label{tab:tracking:evaluation:existence:mot}
\end{table}
\begin{table}[b!]
    \centering
    \caption[Results in the calibration metrics for object existence estimation using semantic confidence calibration.]{
        Results in the calibration metrics for object existence estimation using semantic confidence calibration.
        The multivariate dependent Beta Calibration offers the best calibration performance.
    }
\begin{tabular}{l|l|cccc|cc}
    \hline
    \multirow{2}{*}{Calibration method} & \multirow{2}{*}{\makecell[l]{Calibration\\type}} & \multicolumn{4}{c|}{Semantic metrics} &  \multicolumn{2}{c}{Spatial metrics} \\
    &  & ECE & D-ECE & Brier & NLL & M-QCE & NLL \\ \hline \hline
    Uncalibrated & \makecell{-} & 0.174 & 0.174 & 0.177 & 0.560 & \textbf{0.107} & 19.095 \\ \hline
    \multirow{2}{*}{Histogram Binning} & conf. only & 0.075 & 0.080 & 0.130 & 0.403 & 0.117 & 18.027 \\
    & mv. & \textbf{0.070} & \textbf{0.077} & 0.144 & 0.456 & 0.119 & \textbf{15.227} \\ \hline
    \multirow{3}{*}{Logistic Calibration} & conf. only &  0.073 & 0.078 & 0.130 & 0.390 & 0.117 & 19.153 \\
    & mv. (indep.) & 0.083 & 0.088 & 0.134 & 0.444 & 0.116 & 19.870 \\
    & mv. (dep.) & 0.071 & \textbf{0.077} & 0.128 & 0.384 & 0.117 & 19.146 \\ \hline
    \multirow{3}{*}{Beta Calibration} & conf. only & 0.073 & 0.079 & 0.127 & 0.384 & 0.117 & 19.286 \\
    & mv. (indep.)& 0.080 & 0.085 & 0.130 & 0.424 & 0.116 & 19.010 \\
    & mv. (dep.) & 0.079 & \textbf{0.077} & \textbf{0.114} & \textbf{0.357} & 0.117 & 19.239 \\ \hline
\end{tabular}%

    \label{tab:tracking:evaluation:existence:stats}
\end{table}

Although we observe a coverage drop in the track trajectories after calibration (cf. $\text{MT}$, $\text{PT}$, and $\text{ML}$ in \figref{fig:tracking:evaluation:semantic:mota}), we are able to achieve a significant improvement in the overall $\text{MOTA}$ tracking accuracy, the $\text{MOTP}$ tracking precision and the $\text{IDF1}$ score.
All calibration methods lead to significant improvements in the overall tracking performance, whereas the position-dependent Histogram Binning is able to achieve the best results.
We observe a major decrease in the average scores of false positives per frame ($\text{FP}$) but also an increasing score of average false negatives ($\text{FN}$) which, however, is not as large as the drop of false positives.
In safety-critical applications, the false negatives are particularly important since objects that are not detected pose a potential safety risk.
Changing the confidence threshold value during object tracking for filtering the objects could counteract this phenomenon in order to increase the sensitivity of the model.
Furthermore, we observe that the average number of ID switches per object is decreasing ($\text{IDSw}$).
If we further inspect the calibration results given in \tabref{tab:tracking:evaluation:existence:stats}, we can see an overall improvement of the confidence calibration properties with intermediate calibration. 
In this case, the standard confidence methods but also our multivariate and conditionally dependent Logistic Calibration and Beta Calibration (cf. \secref{section:confidence:methods:scaling}) are able to achieve the best calibration performance.
In \figref{fig:tracking:evaluation:semantic:reliability:confidence}, we show the reliability diagrams for the track confidence before and after calibration by a standard Histogram Binning \cite{Zadrozny2001} as well as by our position-dependent Histogram Binning (cf. \ref{section:confidence:methods:binning}).
By examining these diagrams, we can see the benefit of position-dependent calibration as it further improves the calibration properties of the object tracking scores.
Especially in \figref{fig:tracking:evaluation:semantic:reliability:confidence:2d}, we observe that the position-dependent Histogram Binning leads to an improvement in the calibrated tracking scores across all image regions.

\subsection{Evaluations for intermediate Spatial Uncertainty Calibration}
\label{section:tracking:experiments:spatial}

Subsequently, we investigate the effect of spatial uncertainty calibration on object tracking.
For this reason, we use the Isotonic Regression \cite{Kuleshov2018}, Variance Scaling \cite{Levi2019,Laves2020}, and GP-Beta \cite{Song2019} for uncertainty recalibration.
Since Isotonic Regression and GP-Beta yield non-parametric distributions as calibration output, it is necessary to extract the expectation and the variance of the recalibrated distributions which allows for a Gaussian approximation (moment-matching).
This is mandatory as the Kalman filter only works with normal distributions.
In addition to these methods, we also adapt our GP-Normal uncertainty calibration method presented in \secref{section:regression:methods:parametric:gp} which directly yields Gaussian distributions after calibration.
In our experiments, we use the GP-Normal under the assumption of independent output variables as well as the multivariate (mv.) GP-Normal that performs covariance estimation of independently learned quantities (cf. \secref{section:regression:methods:correlations}).
In this way, we can also evaluate the effect of covariance estimation calibration.
The results of the object tracking with spatial uncertainty calibration are presented in \tabref{tab:tracking:evaluation:position:mot} and \tabref{tab:tracking:evaluation:position:stats} with the \ac{MOT} and calibration metrics, respectively.
Furthermore, we present the reliability diagrams as well as a visualization of the trajectory coverage in \figref{fig:tracking:evaluation:spatial:reliability:regression} and \ref{fig:tracking:evaluation:spatial:mota}, respectively.
\begin{table}[b!]
    \centering
    \caption[Results in the Multiple Object Tracking (MOT) metrics of object tracking using spatial uncertainty calibration of the position uncertainty.]{
        Results in the \ac{MOT} metrics of object tracking using spatial uncertainty calibration of the position uncertainty.
        The best scores are highlighted in bold.
        Our (independent) GP-Normal method is able to achieve the overall best tracking accuracy $\text{MOTA}$.
        The non-parametric Isotonic Regression \cite{Kuleshov2018} is able to achieve the best results regarding the ID switches.
    }
\begin{tabular}{l|ccc|ccc|ccc}
    \hline
    \makecell[l]{Calibration\\method} & MOTA & MOTP & IDF1 & IDSw & FP & FN & MT & PT & ML \\ \hline \hline
    Uncalibrated & 36.845 & 82.679 & 47.88 & 2.23 & 3.06 & 3.16 & 0.382 & 0.438 & 0.180 \\ \hline
    Isotonic Regression & 37.315 & 81.777 & \textbf{49.76} & \textbf{2.00} & 3.05 & 3.18 & 0.383 & 0.434 & 0.183 \\ 
    Variance Scaling & 37.557 & 82.611 & 48.30 & 2.18 & \textbf{3.03} & \textbf{3.13} & 0.391 & 0.432 & \textbf{0.178} \\ 
    GP-Beta & 37.316 & 82.629 & 48.30 & 2.20 & 3.04 & \textbf{3.13} & 0.388 & 0.434 & \textbf{0.178} \\ 
    GP-Normal & \textbf{37.702} & 82.608 & 48.43 & 2.21 & 3.04 & \textbf{3.13} & \textbf{0.392} & 0.429 & 0.179 \\ 
    GP-Normal (mv.) & 34.917 & \textbf{82.759} & 46.20 & 2.38 & 3.18 & 3.19 & 0.373 & 0.446 & 0.181 \\ \hline
\end{tabular}%
    \label{tab:tracking:evaluation:position:mot}
\end{table}
\begin{table}[b!]
    \centering
    \caption[Results in the calibration metrics of object tracking using spatial uncertainty calibration of the position uncertainty.]{
        Results in the calibration metrics of object tracking using spatial uncertainty calibration of the position uncertainty.
        The best scores are highlighted in bold.
        The GB-Beta is the only calibration method that leads to slight improvements in the metrics for semantic confidence calibration.
        In contrast, the non-parametric Isotonic Regression \cite{Kuleshov2018} leads to a significant improvement in the spatial \ac{NLL} score.
        The multivariate GP-Normal achieves the best multivariate quantile coverage.
    }
\begin{tabular}{l|cccc|cc}
    \hline
    \multirow{2}{*}{Calibration method} & \multicolumn{4}{c|}{Semantic metrics} &  \multicolumn{2}{c}{Spatial metrics} \\
    &  ECE & D-ECE & Brier & NLL & M-QCE & NLL \\ \hline \hline
    Uncalibrated & 0.174 & 0.174 & 0.177 & \textbf{0.560} & 0.107 & 19.095 \\ \hline
    Isotonic Regression & 0.188 & 0.188 & 0.186 & 0.605 & 0.204 & \textbf{15.561} \\
    Variance Scaling & 0.175 & 0.175 & 0.177 & 0.562 & 0.152 & 17.271 \\
    GP-Beta & \textbf{0.173} & \textbf{0.173} & \textbf{0.176} & \textbf{0.560} & 0.135 & 17.916 \\
    GP-Normal & 0.175 & 0.175 & 0.177 & 0.563 & 0.151 & 17.279 \\
    GP-Normal (mv.) & 0.176 & 0.176 & 0.179 & 0.564 & \textbf{0.073} & 21.894 \\ \hline
\end{tabular}%
    \label{tab:tracking:evaluation:position:stats}
\end{table}

\begin{figure}[ht!]
    \centering
    \begin{subfigure}{0.5\textwidth}
\begin{tikzpicture}

\pgfplotsset{every axis/.append style={label style={font=\scriptsize}, tick label style={font=\tiny}}}

\definecolor{darkgray176}{RGB}{176,176,176}
\definecolor{lightgray204}{RGB}{204,204,204}
\definecolor{steelblue31119180}{RGB}{31,119,180}

\definecolor{color0}{rgb}{0.12156862745098,0.466666666666667,0.705882352941177}
\definecolor{color1}{rgb}{0.0,0.597656,0.296875}
\definecolor{color2}{rgb}{1.0,0.5,0.0}
\definecolor{color3}{rgb}{0.5,0.0,0.25}
\definecolor{color4}{rgb}{0.0,0.5,0.5}
\definecolor{color5}{rgb}{0.0,0.0,0.0}

\begin{axis}[
legend cell align={left},
legend style={
  fill opacity=0.8,
  draw opacity=1,
  text opacity=1,
  at={(0.6,0.67)},
  anchor=north west,
  draw=lightgray204,
  nodes={scale=0.6, transform shape},
},
legend image post style={scale=0.6},
width=\linewidth,
height=0.6\linewidth,
tick align=outside,
tick pos=left,
x grid style={darkgray176},
xlabel={Expected quantile $\quantile$},
xmajorgrids,
xmin=0, xmax=1,
xtick style={color=black},
y grid style={darkgray176},
ylabel={Observed frequency},
ymajorgrids,
ymin=0, ymax=1,
ytick style={color=black},
xticklabels={0, 0.0, 0.2, 0.4, 0.6, 0.8, 1.0},
yticklabels={0, 0.0, 0.2, 0.4, 0.6, 0.8, 1.0},
]
\addplot [semithick, color0, solid, mark=*, mark size=1.5, mark options={solid}]
table {%
    0.05 0.0824993468210062
    0.095 0.156717519964558
    0.14 0.22726653110836
    0.185 0.297247560518454
    0.23 0.363939975690382
    0.275 0.427775442741761
    0.32 0.489077711260806
    0.365 0.546756256318797
    0.41 0.599862548420443
    0.455 0.647817246197362
    0.5 0.692557167361498
    0.545 0.732622598857221
    0.59 0.768024900319206
    0.635 0.800644091286024
    0.68 0.829395326646295
    0.725 0.855624723108905
    0.77 0.878809737478843
    0.815 0.899046926650839
    0.86 0.916387408980927
    0.905 0.931592279992275
    0.95 0.945314718678647
};
\addlegendentry{Uncalibrated}

\addplot [semithick, dash pattern=on 1pt off 3pt on 3pt off 3pt, color1, mark=*, mark size=1.5, mark options={solid}]
table {%
    0.05 0.173025752429556
    0.095 0.313776199260932
    0.14 0.432739665408068
    0.185 0.531249461216186
    0.23 0.610845790015115
    0.275 0.677091774279754
    0.32 0.730516140526313
    0.365 0.775268242501566
    0.41 0.812876789480641
    0.455 0.843111900369534
    0.5 0.867841359057028
    0.545 0.888875479158405
    0.59 0.906288972029218
    0.635 0.920593323103625
    0.68 0.932075883749131
    0.725 0.941771118888755
    0.77 0.949644546358396
    0.815 0.956167422400763
    0.86 0.961425952426108
    0.905 0.96599483916944
    0.95 0.970155686970914
};
\addlegendentry{Isotonic Regression}

\addplot [semithick, dotted, color2, mark=diamond*, mark size=1.5, mark options={solid}]
table {%
    0.05 0.0925333514838007
    0.095 0.174136718395499
    0.14 0.253233052001089
    0.185 0.328903484889736
    0.23 0.401675515019512
    0.275 0.469983664579363
    0.32 0.533425220074417
    0.365 0.591784644704601
    0.41 0.645390915691079
    0.455 0.693580406570469
    0.5 0.73556470641619
    0.545 0.773272302386786
    0.59 0.806550049913785
    0.635 0.835017923586532
    0.68 0.860922043742626
    0.725 0.883445639350213
    0.77 0.902877983483075
    0.815 0.918986523277974
    0.86 0.932905662945821
    0.905 0.944261049097014
    0.95 0.954482031037299
};
\addlegendentry{Variance Scaling}

\addplot [semithick, color3, opacity=1.0, dashed, mark=square*, mark size=1.5, mark options={solid}]
table {%
    0.05 0.0889443765004303
    0.095 0.166627032658423
    0.14 0.243064048557322
    0.185 0.315781129682475
    0.23 0.38559360420347
    0.275 0.452473162114418
    0.32 0.515689405263396
    0.365 0.57358336730534
    0.41 0.627151560447525
    0.455 0.674904878380215
    0.5 0.717896453322462
    0.545 0.756335779317842
    0.59 0.791089142546542
    0.635 0.821409158853105
    0.68 0.848230058431852
    0.725 0.871936857362866
    0.77 0.892880599719165
    0.815 0.910795171445396
    0.86 0.925878742582778
    0.905 0.939139149340943
    0.95 0.950661321737555
};
\addlegendentry{GP-Beta}

\addplot [semithick, color4, opacity=1.0, dash pattern=on 3pt off 6pt on 6pt off 6pt, mark=triangle*, mark size=1.5, mark options={solid}]
table {%
    0.05 0.0920842905357497
    0.095 0.17343656938654
    0.14 0.252287359255793
    0.185 0.328602626279815
    0.23 0.401015343600216
    0.275 0.469049037125273
    0.32 0.532788791514223
    0.365 0.591553929493179
    0.41 0.644890666212882
    0.455 0.693179046484586
    0.5 0.735165489662214
    0.545 0.772472276581866
    0.59 0.805626932130803
    0.635 0.835088913468902
    0.68 0.860648345103378
    0.725 0.883212796732749
    0.77 0.903179330100116
    0.815 0.91844919027766
    0.86 0.932573243710826
    0.905 0.944088034260756
    0.95 0.954547775036161
};
\addlegendentry{GP-Normal}

\addplot [semithick, color5, mark=*, mark size=1.5, mark options={solid}]
table {%
    0.05 0.0758717842394361
    0.095 0.143671860355473
    0.14 0.209285020477486
    0.185 0.27375644004158
    0.23 0.336228393553994
    0.275 0.396729282521145
    0.32 0.45409464517998
    0.365 0.5089663555755
    0.41 0.559850494470227
    0.455 0.60729805108863
    0.5 0.651786170738496
    0.545 0.692991076246684
    0.59 0.730577629837487
    0.635 0.764619675427585
    0.68 0.79469119042074
    0.725 0.822354257669827
    0.77 0.84793833464927
    0.815 0.871863763654024
    0.86 0.893244417683914
    0.905 0.912352951200532
    0.95 0.929956204877107
};
\addlegendentry{GP-Normal (mv.)}

\addplot [semithick, red, dashed]
table {%
0 0
1 1
};
\addlegendentry{Perfect Calibration}
\end{axis}

\end{tikzpicture}
        \caption{Reliability diagram for the track $\centerx$ coordinate.}
        \label{fig:tracking:evaluation:spatial:reliability:regression:cx}
    \end{subfigure}%
    \begin{subfigure}{0.5\textwidth}
\begin{tikzpicture}

\pgfplotsset{every axis/.append style={label style={font=\scriptsize}, tick label style={font=\tiny}}}

\definecolor{darkgray176}{RGB}{176,176,176}
\definecolor{lightgray204}{RGB}{204,204,204}
\definecolor{steelblue31119180}{RGB}{31,119,180}

\definecolor{color0}{rgb}{0.12156862745098,0.466666666666667,0.705882352941177}
\definecolor{color1}{rgb}{0.0,0.597656,0.296875}
\definecolor{color2}{rgb}{1.0,0.5,0.0}
\definecolor{color3}{rgb}{0.5,0.0,0.25}
\definecolor{color4}{rgb}{0.0,0.5,0.5}
\definecolor{color5}{rgb}{0.0,0.0,0.0}

\begin{axis}[
legend cell align={left},
legend style={
    fill opacity=0.8,
    draw opacity=1,
    text opacity=1,
    at={(0.6,0.67)},
    anchor=north west,
    draw=lightgray204,
    nodes={scale=0.6, transform shape},
},
legend image post style={scale=0.6},
width=\linewidth,
height=0.6\linewidth,
tick align=outside,
tick pos=left,
x grid style={darkgray176},
xlabel={Expected quantile $\quantile$},
xmajorgrids,
xmin=0, xmax=1,
xtick style={color=black},
y grid style={darkgray176},
ylabel={Observed frequency},
ymajorgrids,
ymin=0, ymax=1,
ytick style={color=black},
xticklabels={0, 0.0, 0.2, 0.4, 0.6, 0.8, 1.0},
yticklabels={0, 0.0, 0.2, 0.4, 0.6, 0.8, 1.0},
]
\addplot [semithick, color0, solid, mark=*, mark size=1.5, mark options={solid}]
table {%
    0.05 0.0782451636355375
    0.095 0.148561302268519
    0.14 0.218741125285411
    0.185 0.286915972782315
    0.23 0.353920777907782
    0.275 0.418880848791903
    0.32 0.481199804614284
    0.365 0.540025672774364
    0.41 0.595006304597244
    0.455 0.646573366200543
    0.5 0.692977473844441
    0.545 0.736706387522577
    0.59 0.775772171166975
    0.635 0.811373266235758
    0.68 0.843089366245982
    0.725 0.870897751928298
    0.77 0.89599686474083
    0.815 0.917148504504095
    0.86 0.935312560348059
    0.905 0.951891947154979
    0.95 0.966080130862991
};
\addlegendentry{Uncalibrated}

\addplot [semithick, dash pattern=on 1pt off 3pt on 3pt off 3pt, color1, mark=*, mark size=1.5, mark options={solid}]
table {%
    0.05 0.151727269070074
    0.095 0.279242311914162
    0.14 0.396855226634024
    0.185 0.501502847652052
    0.23 0.588817434182169
    0.275 0.662729952931846
    0.32 0.724510496945455
    0.365 0.774245271633248
    0.41 0.815405481514687
    0.455 0.849801440205054
    0.5 0.876588334683885
    0.545 0.89861669051683
    0.59 0.915794555266289
    0.635 0.930460969063752
    0.68 0.942845813003224
    0.725 0.953046786549657
    0.77 0.961730544875663
    0.815 0.969109727993196
    0.86 0.975609615926162
    0.905 0.981144003264312
    0.95 0.98605771164865
};
\addlegendentry{Isotonic Regression}

\addplot [semithick, dotted, color2, mark=diamond*, mark size=1.5, mark options={solid}]
table {%
    0.05 0.0834411017333696
    0.095 0.158799573464017
    0.14 0.232723023867865
    0.185 0.305126372629095
    0.23 0.3756863145476
    0.275 0.443529358380978
    0.32 0.507793356928941
    0.365 0.567928124149197
    0.41 0.623525274525819
    0.455 0.67529948271168
    0.5 0.721350621653508
    0.545 0.763181776930756
    0.59 0.800980125238225
    0.635 0.834172792449406
    0.68 0.863151148017061
    0.725 0.888902123604683
    0.77 0.911011661675288
    0.815 0.929371993828841
    0.86 0.945752790634359
    0.905 0.959683274344314
    0.95 0.971452718032489
};
\addlegendentry{Variance Scaling}

\addplot [semithick, color3, opacity=1.0, dashed, mark=square*, mark size=1.5, mark options={solid}]
table {%
    0.05 0.0861643339221815
    0.095 0.162805181863478
    0.14 0.237266159351361
    0.185 0.309660506409385
    0.23 0.380214703084658
    0.275 0.446041128776555
    0.32 0.508708157811297
    0.365 0.566981473932147
    0.41 0.621302713230964
    0.455 0.670290347420392
    0.5 0.715892104905558
    0.545 0.756698147393215
    0.59 0.793642705077683
    0.635 0.826578565928342
    0.68 0.85479797979798
    0.725 0.880231689088191
    0.77 0.902766453775422
    0.815 0.922589119898537
    0.86 0.939212755356253
    0.905 0.953945282420619
    0.95 0.967347239208226
};
\addlegendentry{GP-Beta}

\addplot [semithick, color4, opacity=1.0, dash pattern=on 3pt off 6pt on 6pt off 6pt, mark=triangle*, mark size=1.5, mark options={solid}]
table {%
    0.05 0.0830426274142772
    0.095 0.157605150458039
    0.14 0.231895402592246
    0.185 0.303553702600754
    0.23 0.373924388099492
    0.275 0.440971099577413
    0.32 0.505130604951927
    0.365 0.565393232933435
    0.41 0.620845741512805
    0.455 0.67279843444227
    0.5 0.718517258005048
    0.545 0.760758955160385
    0.59 0.798513854618679
    0.635 0.831906747213477
    0.68 0.861720411809751
    0.725 0.887347911171616
    0.77 0.909498284126039
    0.815 0.92837006154457
    0.86 0.945046654754814
    0.905 0.959182052809212
    0.95 0.971133611276554
};
\addlegendentry{GP-Normal}

\addplot [semithick, color5, mark=*, mark size=1.5, mark options={solid}]
table {%
    0.05 0.0669707521286929
    0.095 0.127437559288145
    0.14 0.188313348139985
    0.185 0.247257834555545
    0.23 0.306361369407033
    0.275 0.364073230444143
    0.32 0.419865149647537
    0.365 0.474663016126375
    0.41 0.52786471794464
    0.455 0.576800513499236
    0.5 0.623594835470073
    0.545 0.668134077831488
    0.59 0.709390106051225
    0.635 0.747879827546053
    0.68 0.784000863405795
    0.725 0.816049123245497
    0.77 0.84608087612967
    0.815 0.872971422404244
    0.86 0.897550086056564
    0.905 0.920248569984152
    0.95 0.942276778360324
};
\addlegendentry{GP-Normal (mv.)}

\addplot [semithick, red, dashed]
table {%
    0 0
    1 1
};
\addlegendentry{Perfect Calibration}
\end{axis}

\end{tikzpicture}
        \caption{Reliability diagram for the track $\centery$ coordinate.}
        \label{fig:tracking:evaluation:spatial:reliability:regression:cy}
    \end{subfigure}
    \vspace{0.5cm}
    \begin{subfigure}{0.5\textwidth}
\begin{tikzpicture}

\pgfplotsset{every axis/.append style={label style={font=\scriptsize}, tick label style={font=\tiny}}}

\definecolor{darkgray176}{RGB}{176,176,176}
\definecolor{lightgray204}{RGB}{204,204,204}
\definecolor{steelblue31119180}{RGB}{31,119,180}

\definecolor{color0}{rgb}{0.12156862745098,0.466666666666667,0.705882352941177}
\definecolor{color1}{rgb}{0.0,0.597656,0.296875}
\definecolor{color2}{rgb}{1.0,0.5,0.0}
\definecolor{color3}{rgb}{0.5,0.0,0.25}
\definecolor{color4}{rgb}{0.0,0.5,0.5}
\definecolor{color5}{rgb}{0.0,0.0,0.0}

\begin{axis}[
legend cell align={left},
legend style={
    fill opacity=0.8,
    draw opacity=1,
    text opacity=1,
    at={(0.03,0.97)},
    anchor=north west,
    draw=lightgray204,
    nodes={scale=0.6, transform shape},
},
legend image post style={scale=0.6},
width=\linewidth,
height=0.6\linewidth,
tick align=outside,
tick pos=left,
x grid style={darkgray176},
xlabel={Expected quantile $\quantile$},
xmajorgrids,
xmin=0, xmax=1,
xtick style={color=black},
y grid style={darkgray176},
ylabel={Observed frequency},
ymajorgrids,
ymin=0, ymax=1,
ytick style={color=black},
xticklabels={0, 0.0, 0.2, 0.4, 0.6, 0.8, 1.0},
yticklabels={0, 0.0, 0.2, 0.4, 0.6, 0.8, 1.0},
]
\addplot [semithick, color0, solid, mark=*, mark size=1.5, mark options={solid}]
table {%
    0.05 0.0463586691052016
    0.095 0.087667980597744
    0.14 0.12986334359487
    0.185 0.171450966136929
    0.23 0.21315218502573
    0.275 0.254444457066261
    0.32 0.296015040156309
    0.365 0.337983210459952
    0.41 0.379735547704786
    0.455 0.42148788494962
    0.5 0.462280333064489
    0.545 0.504078108848019
    0.59 0.54564301212073
    0.635 0.58617986845543
    0.68 0.625790914564188
    0.725 0.666248253456169
    0.77 0.707256534629846
    0.815 0.747293567038884
    0.86 0.788040576615056
    0.905 0.830230259794845
    0.95 0.876117504061069
};
\addlegendentry{Uncalibrated}

\addplot [semithick, dash pattern=on 1pt off 3pt on 3pt off 3pt, color1, mark=*, mark size=1.5, mark options={solid}]
table {%
    0.05 0.0541714798020724
    0.095 0.102607426308742
    0.14 0.150531887381252
    0.185 0.198760940903318
    0.23 0.246748619276679
    0.275 0.294299523571433
    0.32 0.340649299150015
    0.365 0.388045033706315
    0.41 0.435469503399367
    0.455 0.480629644316477
    0.5 0.525381746291731
    0.545 0.568720079538859
    0.59 0.610805560823664
    0.635 0.651988758814503
    0.68 0.692011057280622
    0.725 0.730889697304069
    0.77 0.768699390240398
    0.815 0.806089550180169
    0.86 0.841657902449958
    0.905 0.8774503887864
    0.95 0.912162433981023
};
\addlegendentry{Isotonic Regression}

\addplot [semithick, dotted, color2, mark=diamond*, mark size=1.5, mark options={solid}]
table {%
    0.05 0.0529596605862601
    0.095 0.100275660223251
    0.14 0.147546283691805
    0.185 0.194334785370723
    0.23 0.241554360649787
    0.275 0.287934476812778
    0.32 0.334496097649514
    0.365 0.381239223159996
    0.41 0.427500226880842
    0.455 0.471651238769398
    0.5 0.516726790089845
    0.545 0.560321943915056
    0.59 0.602550140666122
    0.635 0.643167483437698
    0.68 0.683336736545966
    0.725 0.722814003085579
    0.77 0.760805200108903
    0.815 0.798093066521463
    0.86 0.834246528723115
    0.905 0.870536119430075
    0.95 0.906224475905255
};
\addlegendentry{Variance Scaling}

\addplot [semithick, color3, opacity=1.0, dashed, mark=square*, mark size=1.5, mark options={solid}]
table {%
    0.05 0.04984259636726
    0.095 0.09513860578883
    0.14 0.140259093173891
    0.185 0.185498482583684
    0.23 0.230132037867464
    0.275 0.276022557412692
    0.32 0.320531548670562
    0.365 0.365221723966119
    0.41 0.410319563346469
    0.455 0.454749286587852
    0.5 0.498431625673778
    0.545 0.541400552611315
    0.59 0.58351451737102
    0.635 0.625118902024732
    0.68 0.665857000498256
    0.725 0.705519318748018
    0.77 0.745249581011913
    0.815 0.78426076912624
    0.86 0.822400009059202
    0.905 0.86042600896861
    0.95 0.899205055034651
};
\addlegendentry{GP-Beta}

\addplot [semithick, color4, opacity=1.0, dash pattern=on 3pt off 6pt on 6pt off 6pt, mark=triangle*, mark size=1.5, mark options={solid}]
table {%
    0.05 0.0531211889163051
    0.095 0.10037153634533
    0.14 0.147746674607902
    0.185 0.194804163475992
    0.23 0.241629087608837
    0.275 0.288278170112595
    0.32 0.33474573867665
    0.365 0.381235996483167
    0.41 0.42686973538671
    0.455 0.470977622734621
    0.5 0.516344763039224
    0.545 0.559794662355711
    0.59 0.60272270909555
    0.635 0.643188973028163
    0.68 0.683519101505998
    0.725 0.722476530814827
    0.77 0.760492356561445
    0.815 0.797447460222922
    0.86 0.834045208315607
    0.905 0.87081879803738
    0.95 0.906384185598003
};
\addlegendentry{GP-Normal}

\addplot [semithick, color5, mark=*, mark size=1.5, mark options={solid}]
table {%
    0.05 0.0477770140928275
    0.095 0.0906462478769874
    0.14 0.134140314802297
    0.185 0.177492374195527
    0.23 0.220407050389953
    0.275 0.26286162218044
    0.32 0.306185280067255
    0.365 0.349355569819423
    0.41 0.39173629769323
    0.455 0.43444648304146
    0.5 0.476639760972922
    0.545 0.519048890353144
    0.59 0.560651416951155
    0.635 0.601657511914432
    0.68 0.642033093435276
    0.725 0.68153390855851
    0.77 0.721352820553602
    0.815 0.761751123279579
    0.86 0.801041767255335
    0.905 0.841525274500559
    0.95 0.884172976534675
};
\addlegendentry{GP-Normal (mv.)}

\addplot [semithick, red, dashed]
table {%
    0 0
    1 1
};
\addlegendentry{Perfect Calibration}
\end{axis}

\end{tikzpicture}
        \caption{Reliability diagram for the track width.}
        \label{fig:tracking:evaluation:spatial:reliability:regression:width}
    \end{subfigure}%
    \begin{subfigure}{0.5\textwidth}
\begin{tikzpicture}

\pgfplotsset{every axis/.append style={label style={font=\scriptsize}, tick label style={font=\tiny}}}

\definecolor{darkgray176}{RGB}{176,176,176}
\definecolor{lightgray204}{RGB}{204,204,204}
\definecolor{steelblue31119180}{RGB}{31,119,180}

\definecolor{color0}{rgb}{0.12156862745098,0.466666666666667,0.705882352941177}
\definecolor{color1}{rgb}{0.0,0.597656,0.296875}
\definecolor{color2}{rgb}{1.0,0.5,0.0}
\definecolor{color3}{rgb}{0.5,0.0,0.25}
\definecolor{color4}{rgb}{0.0,0.5,0.5}
\definecolor{color5}{rgb}{0.0,0.0,0.0}

\begin{axis}[
legend cell align={left},
legend style={
    fill opacity=0.8,
    draw opacity=1,
    text opacity=1,
    at={(0.03,0.97)},
    anchor=north west,
    draw=lightgray204,
    nodes={scale=0.6, transform shape},
},
legend image post style={scale=0.6},
width=\linewidth,
height=0.6\linewidth,
tick align=outside,
tick pos=left,
x grid style={darkgray176},
xlabel={Expected quantile $\quantile$},
xmajorgrids,
xmin=0, xmax=1,
xtick style={color=black},
y grid style={darkgray176},
ylabel={Observed frequency},
ymajorgrids,
ymin=0, ymax=1,
ytick style={color=black},
xticklabels={0, 0.0, 0.2, 0.4, 0.6, 0.8, 1.0},
yticklabels={0, 0.0, 0.2, 0.4, 0.6, 0.8, 1.0},
]
\addplot [semithick, color0, solid, mark=*, mark size=1.5, mark options={solid}]
table {%
    0.05 0.049522327361952
    0.095 0.094080494371301
    0.14 0.139121445854301
    0.185 0.18298099533119
    0.23 0.227482364167168
    0.275 0.271881496291079
    0.32 0.316189751337597
    0.365 0.35912917040588
    0.41 0.402789926275971
    0.455 0.445825902238984
    0.5 0.488026945053447
    0.545 0.529018186775113
    0.59 0.570742124933262
    0.635 0.611466415240086
    0.68 0.651304654042326
    0.725 0.691364405720712
    0.77 0.730611943519896
    0.815 0.77063761629426
    0.86 0.809924912814804
    0.905 0.849996024127864
    0.95 0.893350069861753
};
\addlegendentry{Uncalibrated}

\addplot [semithick, dash pattern=on 1pt off 3pt on 3pt off 3pt, color1, mark=*, mark size=1.5, mark options={solid}]
table {%
    0.05 0.0535105716568105
    0.095 0.101797095452377
    0.14 0.150072125193244
    0.185 0.197916127882853
    0.23 0.24566817813486
    0.275 0.292799549433056
    0.32 0.340338959673109
    0.365 0.386912869318345
    0.41 0.432164962673057
    0.455 0.476411326241502
    0.5 0.519479549203175
    0.545 0.563007534352856
    0.59 0.605093015637661
    0.635 0.647132520703666
    0.68 0.687815727315046
    0.725 0.726970224651299
    0.77 0.765366114377338
    0.815 0.803256265696568
    0.86 0.840537232116688
    0.905 0.876829709832589
    0.95 0.913719878392901
};
\addlegendentry{Isotonic Regression}

\addplot [semithick, dotted, color2, mark=diamond*, mark size=1.5, mark options={solid}]
table {%
    0.05 0.0537934476812778
    0.095 0.101262591886741
    0.14 0.148947272892277
    0.185 0.196149832108177
    0.23 0.244157818313822
    0.275 0.291899219529903
    0.32 0.338455168345585
    0.365 0.38510186949814
    0.41 0.430568790271349
    0.455 0.474640393865142
    0.5 0.518025682911335
    0.545 0.560344631999274
    0.59 0.602850757782013
    0.635 0.644063662764316
    0.68 0.683711089935566
    0.725 0.722331881295943
    0.77 0.76096401669843
    0.815 0.798569516290044
    0.86 0.835250476449769
    0.905 0.872135629367456
    0.95 0.910132498411834
};
\addlegendentry{Variance Scaling}

\addplot [semithick, color3, opacity=1.0, dashed, mark=square*, mark size=1.5, mark options={solid}]
table {%
    0.05 0.0513260406758165
    0.095 0.0975166462834624
    0.14 0.143503419848711
    0.185 0.188669203243194
    0.23 0.23471259682022
    0.275 0.280002944240612
    0.32 0.326001041808217
    0.365 0.371449925261584
    0.41 0.414752910268605
    0.455 0.458322009330978
    0.5 0.501732572360375
    0.545 0.54362571907415
    0.59 0.585128187706663
    0.635 0.626132400235539
    0.68 0.667329120804457
    0.725 0.707172623091906
    0.77 0.746013951170902
    0.815 0.784855279249898
    0.86 0.823379535262943
    0.905 0.862634189427911
    0.95 0.903196765864927
};
\addlegendentry{GP-Beta}

\addplot [semithick, color4, opacity=1.0, dash pattern=on 3pt off 6pt on 6pt off 6pt, mark=triangle*, mark size=1.5, mark options={solid}]
table {%
    0.05 0.0540911540315948
    0.095 0.101488981536629
    0.14 0.149442695482005
    0.185 0.196409427380243
    0.23 0.243880994923282
    0.275 0.291142686973539
    0.32 0.337632944780056
    0.365 0.38343685300207
    0.41 0.429796647664426
    0.455 0.473654953345245
    0.5 0.517275021980204
    0.545 0.559357894438299
    0.59 0.601389716100854
    0.635 0.642706826625826
    0.68 0.682707961087949
    0.725 0.721398791797839
    0.77 0.75947701296123
    0.815 0.797419098669843
    0.86 0.834442270058708
    0.905 0.870943588870927
    0.95 0.909186307042174
};
\addlegendentry{GP-Normal}

\addplot [semithick, color5, mark=*, mark size=1.5, mark options={solid}]
table {%
    0.05 0.0488733122404812
    0.095 0.0928331638710117
    0.14 0.136526041341233
    0.185 0.180150755196056
    0.23 0.223508494890569
    0.275 0.267383142001852
    0.32 0.310905610433577
    0.365 0.353899810845967
    0.41 0.396638397700614
    0.455 0.438780552920527
    0.5 0.480303555300573
    0.545 0.521758394065221
    0.59 0.562838332945179
    0.635 0.602299385959431
    0.68 0.641857004095497
    0.725 0.681011320840457
    0.77 0.719989548245639
    0.815 0.758882571131573
    0.86 0.798786687645913
    0.905 0.838764648076934
    0.95 0.882582492175385
};
\addlegendentry{GP-Normal (mv.)}

\addplot [semithick, red, dashed]
table {%
    0 0
    1 1
};
\addlegendentry{Perfect Calibration}
\end{axis}

\end{tikzpicture}
        \caption{Reliability diagram for the track height.}
        \label{fig:tracking:evaluation:spatial:reliability:regression:height}
    \end{subfigure}
    \vspace{0.5cm}
    \begin{subfigure}{\textwidth}
\begin{tikzpicture}

\pgfplotsset{every axis/.append style={label style={font=\scriptsize}, tick label style={font=\scriptsize}}}

\definecolor{darkgray176}{RGB}{176,176,176}
\definecolor{lightgray204}{RGB}{204,204,204}
\definecolor{steelblue31119180}{RGB}{31,119,180}

\definecolor{color0}{rgb}{0.12156862745098,0.466666666666667,0.705882352941177}
\definecolor{color1}{rgb}{0.0,0.597656,0.296875}
\definecolor{color2}{rgb}{1.0,0.5,0.0}
\definecolor{color3}{rgb}{0.5,0.0,0.25}
\definecolor{color4}{rgb}{0.0,0.5,0.5}
\definecolor{color5}{rgb}{0.0,0.0,0.0}

\begin{axis}[
legend cell align={left},
legend style={
  fill opacity=0.8,
  draw opacity=1,
  text opacity=1,
  at={(0.03,0.97)},
  anchor=north west,
  draw=lightgray204,
  nodes={scale=0.8, transform shape},
},
legend image post style={scale=0.8},
width=\linewidth,
height=0.45\linewidth,
tick align=outside,
tick pos=left,
x grid style={darkgray176},
xlabel={Expected quantile $\quantile$},
xmajorgrids,
xmin=0, xmax=1,
xtick style={color=black},
y grid style={darkgray176},
ylabel={Observed frequency},
ymajorgrids,
ymin=0, ymax=1,
ytick style={color=black}
]
\addplot [semithick, color0, solid, mark=*, mark size=1.5, mark options={solid}]
table {%
    0.05 0.142279424293715
    0.095 0.229811089275369
    0.14 0.299263895673115
    0.185 0.358600947393532
    0.23 0.409611386897797
    0.275 0.454601220024764
    0.32 0.495529983755722
    0.365 0.532977019459054
    0.41 0.56727175654031
    0.455 0.599436562120162
    0.5 0.628454748895276
    0.545 0.656507366723086
    0.59 0.683690972498324
    0.635 0.709778373527507
    0.68 0.734531017482478
    0.725 0.758073860344651
    0.77 0.780781770058275
    0.815 0.803529438493258
    0.86 0.82649293998705
    0.905 0.850518567322875
    0.95 0.878128159398394
};
\addlegendentry{Uncalibrated}

\addplot [semithick, dash pattern=on 1pt off 3pt on 3pt off 3pt, color1, mark=*, mark size=1.5, mark options={solid}]
table {%
    0.05 0.262759837474067
    0.095 0.370430394878249
    0.14 0.447400332178181
    0.185 0.508542956155928
    0.23 0.558042102722367
    0.275 0.600455164566128
    0.32 0.637230392579438
    0.365 0.669580409533169
    0.41 0.699367252288754
    0.455 0.725211634282168
    0.5 0.7493031729338
    0.545 0.771262564438544
    0.59 0.791744969914312
    0.635 0.810319362309845
    0.68 0.827485733004603
    0.725 0.844060159882301
    0.77 0.860071378079688
    0.815 0.875467664350615
    0.86 0.890731768992489
    0.905 0.906174031482216
    0.95 0.924104756814538
};
\addlegendentry{Isotonic Regression}

\addplot [semithick, dotted, color2, mark=diamond*, mark size=1.5, mark options={solid}]
table {%
    0.05 0.182332788819312
    0.095 0.283487612306017
    0.14 0.362147200290407
    0.185 0.424494055721935
    0.23 0.477913150013613
    0.275 0.524418050639804
    0.32 0.565330338506217
    0.365 0.60257282875034
    0.41 0.635124557582358
    0.455 0.665033124602958
    0.5 0.693177693075597
    0.545 0.719371086305472
    0.59 0.744072737998003
    0.635 0.766312732552863
    0.68 0.787395634812596
    0.725 0.807474589345676
    0.77 0.826566612215265
    0.815 0.845806107632271
    0.86 0.864478400943824
    0.905 0.883700880297668
    0.95 0.905135447862782
};
\addlegendentry{Variance Scaling}

\addplot [semithick, color3, opacity=1.0, dashed, mark=square*, mark size=1.5, mark options={solid}]
table {%
    0.05 0.168569099062373
    0.095 0.263515196811161
    0.14 0.338066766317887
    0.185 0.399658015128867
    0.23 0.452059836028446
    0.275 0.498120215609005
    0.32 0.538948906101372
    0.365 0.576085971825882
    0.41 0.610624179009829
    0.455 0.641623635457716
    0.5 0.670403587443946
    0.545 0.697473615074512
    0.59 0.723048874394166
    0.635 0.74712370340173
    0.68 0.769647144086606
    0.725 0.790772070480591
    0.77 0.811126964714409
    0.815 0.83093264483399
    0.86 0.851095031027766
    0.905 0.87232187344295
    0.95 0.89525297821262
};
\addlegendentry{GP-Beta}

\addplot [semithick, color4, opacity=1.0, dash pattern=on 3pt off 6pt on 6pt off 6pt, mark=triangle*, mark size=1.5, mark options={solid}]
table {%
    0.05 0.181621713605037
    0.095 0.281834425253127
    0.14 0.360032899401571
    0.185 0.422927479508778
    0.23 0.476207493122323
    0.275 0.522295016875124
    0.32 0.564111290734281
    0.365 0.60054454181911
    0.41 0.633540372670807
    0.455 0.66373975438895
    0.5 0.691902776596046
    0.545 0.717955699254091
    0.59 0.742011968575399
    0.635 0.764786295697552
    0.68 0.786210612893162
    0.725 0.805887858419127
    0.77 0.825559431634476
    0.815 0.844306418219462
    0.86 0.863371054198928
    0.905 0.882793045747185
    0.95 0.904466944609887
};
\addlegendentry{GP-Normal}

\addplot [semithick, color5, mark=*, mark size=1.5, mark options={solid}]
table {%
    0.05 0.111015808278471
    0.095 0.183257879997955
    0.14 0.244599453555017
    0.185 0.297948843206644
    0.23 0.345140786267304
    0.275 0.388134986679693
    0.32 0.427692604815759
    0.365 0.465023544848819
    0.41 0.499514334240288
    0.455 0.531897731855698
    0.5 0.562293024021994
    0.545 0.591620419547053
    0.59 0.620129851687333
    0.635 0.648008770385181
    0.68 0.675064045396968
    0.725 0.701483126665038
    0.77 0.727680676183065
    0.815 0.754190642271666
    0.86 0.78134816270655
    0.905 0.810772123353423
    0.95 0.842985111930337
};
\addlegendentry{GP-Normal (mv.)}

\addplot [semithick, red, dashed]
table {%
0 0
1 1
};
\addlegendentry{Perfect Calibration}
\end{axis}

\end{tikzpicture}
        \caption{Reliability diagram for the joint multivariate quantile.}
        \label{fig:tracking:evaluation:spatial:reliability:regression:multivariate}
    \end{subfigure}
    
    \caption[Reliability diagrams (object tracking) of the spatial uncertainty for a \fasterrcnn{} before and after uncertainty calibration.]{
        Reliability diagrams (object tracking) of the spatial uncertainty for a \fasterrcnn{} before and after uncertainty calibration.
        Especially in the reliability diagrams for the single bounding box quantities, the multivariate GP-Normal leads to the best calibration properties.
        However, this seems to be detrimental for the multivariate quantile where the quantile coverage decreases.
    }
    \label{fig:tracking:evaluation:spatial:reliability:regression}
\end{figure}
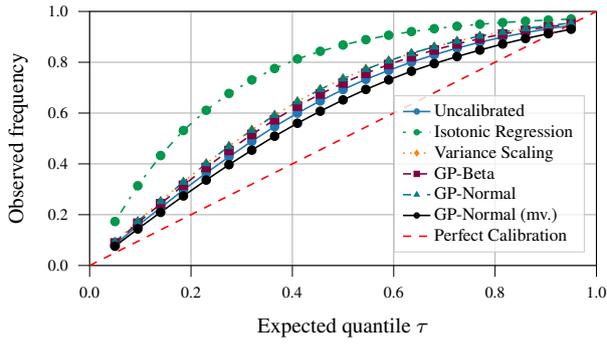
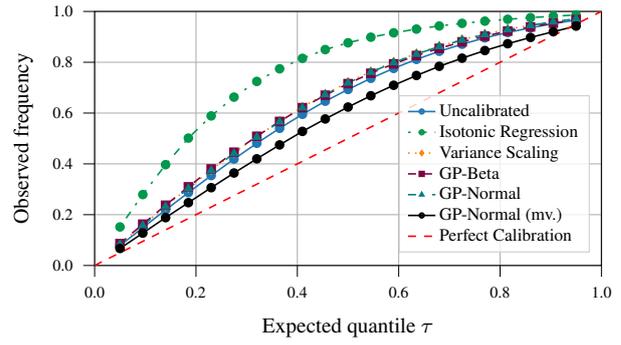
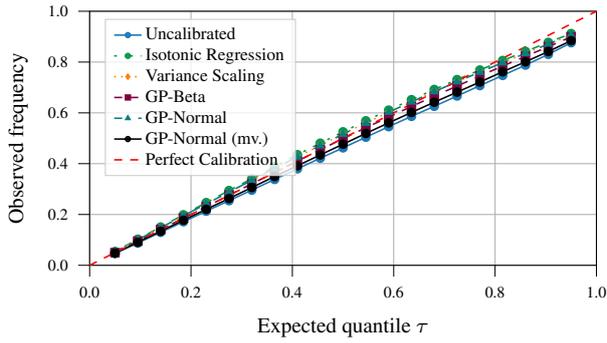
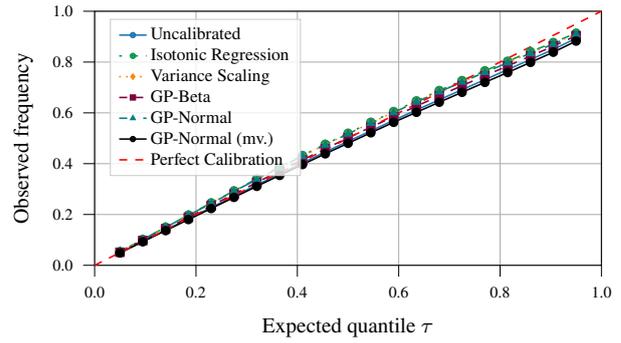
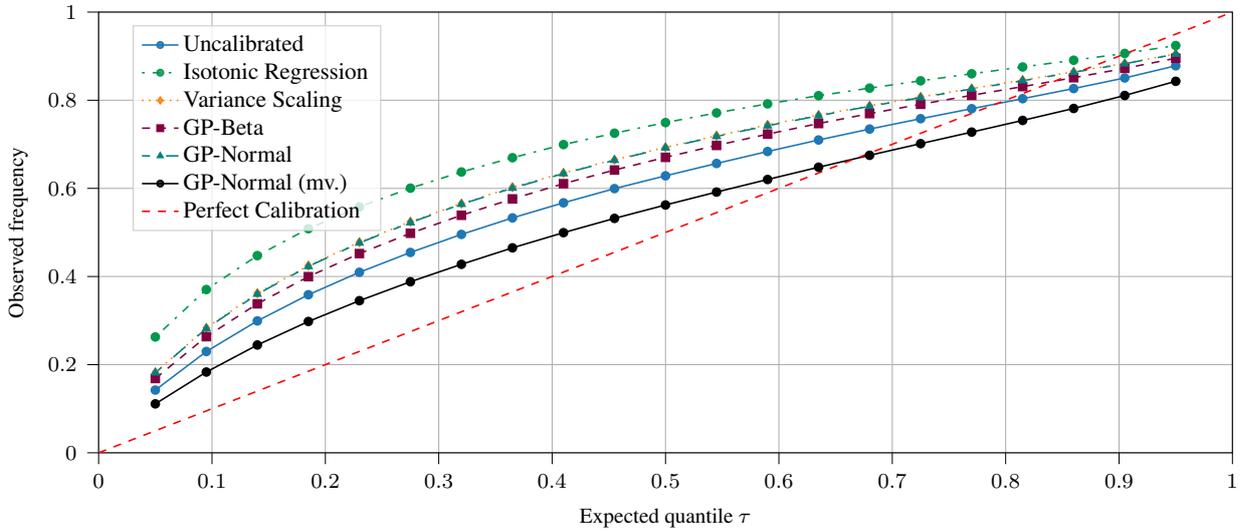
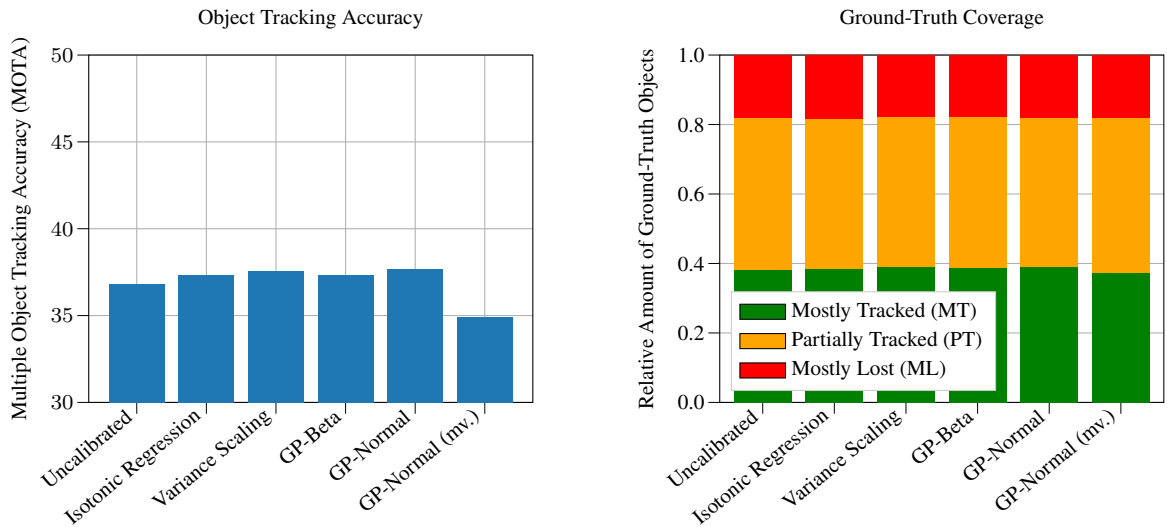
\begin{figure}[ht!]
    \centering
    \begin{subfigure}{0.5\textwidth}
\begin{tikzpicture}

\tikzstyle{every node}=[font=\scriptsize]
\pgfplotsset{every x tick label/.append style={font=\scriptsize, yshift=0.5ex}}
\pgfplotsset{every y tick label/.append style={font=\scriptsize, xshift=0.5ex}}

\definecolor{darkgray176}{RGB}{176,176,176}
\definecolor{steelblue31119180}{RGB}{31,119,180}

\begin{axis}[
width=0.9\textwidth,
height=0.75\textwidth,
tick align=outside,
tick pos=left,
title=Object Tracking Accuracy,
x grid style={darkgray176},
xmajorgrids,
xmin=-0.69, xmax=5.69,
xtick style={color=black},
xtick={0,1,2,3,4,5},
xticklabel style={rotate=40.0,anchor=east},
xticklabels={
  Uncalibrated,
  Isotonic Regression,
  Variance Scaling,
  GP-Beta,
  GP-Normal,
  GP-Normal (mv.)
},
y grid style={darkgray176},
ylabel={Multiple Object Tracking Accuracy (MOTA)},
ymajorgrids,
ymin=30, ymax=50,
ytick style={color=black}
]
\draw[draw=none,fill=steelblue31119180] (axis cs:-0.4,0) rectangle (axis cs:0.4,36.845);
\draw[draw=none,fill=steelblue31119180] (axis cs:0.6,0) rectangle (axis cs:1.4,37.315);
\draw[draw=none,fill=steelblue31119180] (axis cs:1.6,0) rectangle (axis cs:2.4,37.557);
\draw[draw=none,fill=steelblue31119180] (axis cs:2.6,0) rectangle (axis cs:3.4,37.316);
\draw[draw=none,fill=steelblue31119180] (axis cs:3.6,0) rectangle (axis cs:4.4,37.702);
\draw[draw=none,fill=steelblue31119180] (axis cs:4.6,0) rectangle (axis cs:5.4,34.917);
\end{axis}

\end{tikzpicture}
        \caption{Tracking accuracy before and after calibration.}
        \label{fig:tracking:evaluation:spatial:mota:mota}
    \end{subfigure}%
    \begin{subfigure}{0.5\textwidth}
\begin{tikzpicture}

\tikzstyle{every node}=[font=\scriptsize]
\pgfplotsset{every x tick label/.append style={font=\scriptsize, yshift=0.5ex}}
\pgfplotsset{every y tick label/.append style={font=\scriptsize, xshift=0.5ex}}

\definecolor{darkgray176}{RGB}{176,176,176}
\definecolor{green}{RGB}{0,128,0}
\definecolor{lightgray204}{RGB}{204,204,204}
\definecolor{orange}{RGB}{255,165,0}

\begin{axis}[
width=0.9\textwidth,
height=0.75\textwidth,
legend cell align={left},
legend style={fill opacity=1.0, draw opacity=1, text opacity=1, draw=lightgray204, at={(0.025, 0.32)}, anchor=north west},
tick align=outside,
tick pos=left,
title=Ground-Truth Coverage,
x grid style={darkgray176},
xmajorgrids,
xmin=-0.6, xmax=5.6,
xtick style={color=black},
xtick={0,1,2,3,4,5},
xticklabel style={rotate=40.0,anchor=east},
xticklabels={
  Uncalibrated,
  Isotonic Regression,
  Variance Scaling,
  GP-Beta,
  GP-Normal,
  GP-Normal (mv.)
},
y grid style={darkgray176},
ylabel={Relative Amount of Ground-Truth Objects},
ymajorgrids,
ymin=0, ymax=1,
ytick style={color=black},
yticklabels={0, 0.0,0.2,0.4,0.6,0.8,1.0},
]
\draw[draw=none,fill=green] (axis cs:-0.4,0) rectangle (axis cs:0.4,0.382);
\draw[draw=none,fill=green] (axis cs:0.6,0) rectangle (axis cs:1.4,0.383);
\draw[draw=none,fill=green] (axis cs:1.6,0) rectangle (axis cs:2.4,0.391);
\draw[draw=none,fill=green] (axis cs:2.6,0) rectangle (axis cs:3.4,0.388);
\draw[draw=none,fill=green] (axis cs:3.6,0) rectangle (axis cs:4.4,0.392);
\draw[draw=none,fill=green] (axis cs:4.6,0) rectangle (axis cs:5.4,0.373);
\draw[draw=none,fill=orange] (axis cs:-0.4,0.382) rectangle (axis cs:0.4,0.820);
\draw[draw=none,fill=orange] (axis cs:0.6,0.383) rectangle (axis cs:1.4,0.817);
\draw[draw=none,fill=orange] (axis cs:1.6,0.391) rectangle (axis cs:2.4,0.823);
\draw[draw=none,fill=orange] (axis cs:2.6,0.388) rectangle (axis cs:3.4,0.821);
\draw[draw=none,fill=orange] (axis cs:3.6,0.392) rectangle (axis cs:4.4,0.822);
\draw[draw=none,fill=orange] (axis cs:4.6,0.373) rectangle (axis cs:5.4,0.819);
\draw[draw=none,fill=red] (axis cs:-0.4,0.820) rectangle (axis cs:0.4,1);
\draw[draw=none,fill=red] (axis cs:0.6,0.817) rectangle (axis cs:1.4,1);
\draw[draw=none,fill=red] (axis cs:1.6,0.823) rectangle (axis cs:2.4,1);
\draw[draw=none,fill=red] (axis cs:2.6,0.822) rectangle (axis cs:3.4,1);
\draw[draw=none,fill=red] (axis cs:3.6,0.821) rectangle (axis cs:4.4,1);
\draw[draw=none,fill=red] (axis cs:4.6,0.819) rectangle (axis cs:5.4,1);

\addlegendimage{area legend, black, fill=green}
\addlegendentry{Mostly Tracked (MT)}
\addlegendimage{area legend, black, fill=orange}
\addlegendentry{Partially Tracked (PT)}
\addlegendimage{area legend, black, fill=red}
\addlegendentry{Mostly Lost (ML)}

\end{axis}

\end{tikzpicture}
        \caption{Ground-truth coverage before and after calibration.}
        \label{fig:tracking:evaluation:spatial:mota:mt}
    \end{subfigure}
    \caption[Visualization of the Multiple Object Tracking (MOT) metrics before and after spatial uncertainty calibration.]{
        Visualization of the \ac{MOT} metrics before and after spatial uncertainty calibration.
        Except for the Isotonic Regression \cite{Kuleshov2018}, each calibration method is able to keep the trajectory coverage during object tracking.
        The gain in the \ac{MOT} tracking accuracy $\text{MOTA}$ is not as large as for semantic confidence calibration in \figref{fig:tracking:evaluation:semantic:mota}.
        In this case, the standard GP-Normal is able to achieve the overall best tracking accuracy.
    }
    \label{fig:tracking:evaluation:spatial:mota}
\end{figure}

As already suggested in the evaluation of the regression uncertainty calibration methods in \secref{section:regression:experiments} and \ref{section:regression:conclusion}, the parametric calibration methods Variance Scaling and GP-Normal, that directly output normal distributions as calibration output, offer the best performance in our examinations.
Especially the (independent) GP-Normal is able to achieve a significant improvement of the overall tracking accuracy ($\text{MOTA}$).
Furthermore, the (independent) GP-Normal as well as the Variance Scaling are able to reduce the number of false positives while not degrading the amount of false negatives.
Interestingly, the multivariate GP-Normal is able to improve the tracking precision ($\text{MOTP}$) and seems to provide very good results for the calibration properties of the estimated quantiles regarding the spatial uncertainty (cf. \figref{fig:tracking:evaluation:spatial:reliability:regression}).
However, the multivariate GP-Normal also leads to a loss in the tracking accuracy.
This is mainly caused by the deterioration in the tracking $\text{IDF1}$ score.
We further discuss these observations in the next section.

\subsection{Discussion of the Effect of intermediate Calibration}
\label{section:tracking:experiments:discussion}

In our experiments, both types of uncertainty calibration offer an improvement in the tracking performance.
Especially in the case of semantic confidence calibration, we assume that the calibrated confidence leads to a significant improvement of the track management.
In our evaluations for the \ac{MOT} metrics, we filter all tracks that have an existence confidence of above $0.5$.
After semantic confidence calibration, this probability of object existence is more interpretable as it better reflects the true (observed) probability of existence.
Thus, we have a higher chance of accessing all tracks that have a true (observed) probability of being alive using a threshold of $0.5$.
However, we also observer an increase of false negatives which poses a potential safety risk that needs further investigation.
Nevertheless, we conclude that an improved confidence score leads to a better object tracking performance.
We observe that our position-dependent confidence calibration methods and especially the multivariate Histogram Binning (cf. \secref{section:confidence:methods:binning}) are able to further improve the tracking performance compared to the standard confidence-only calibration methods.
Thus, we assume that the additional position-dependency is advantageous to reliably reflect the tracking existence score which, in turn, leads to an improved track management/filtering during object tracking and evaluation.

The evaluation of the methods for spatial uncertainty calibration reveals that the non-parametric Isotonic Regression \cite{Kuleshov2018} achieves the best results regarding the ID switches and the spatial \ac{NLL}.
An advantage of the non-parametric calibration methods is that they can not only recalibrate the variance but also the mean of the calibrated distribution.
Furthermore, the Isotonic Regression has much more degrees of freedom as it offers a dynamic amount of bins that are used for the recalibration of the \ac{CDF}.
In contrast, the non-parametric GP-Beta \cite{Song2019} is also able to offer better tracking accuracy ($\text{MOTA}$), less ID switches, and an improved \ac{NLL} score.
However, although the GP-Beta is able to calibrate by means of distribution calibration, the method is still limited to a certain type of recalibration functions due to its parametrization.
At this point, it is remarkable that the simple Variance Scaling \cite{Levi2019,Laves2020}, which is nothing else but a Temperature Scaling \cite{Guo2018} for the variance of a normal distribution, is also able to achieve good tracking results although it only uses a single parameter for recalibration.
In the experiments, our GP-Normal (cf. \secref{section:regression:methods:parametric:gp}) is able to even further improve the performance of the object tracking.
As described in \secref{section:regression:methods:parametric:gp}, the GP-Normal is a combination of the Variance Scaling and the flexible parameter estimation of the GP-Beta.
As already suggested for the spatial uncertainty calibration methods in \secref{section:regression:experiments}, we can confirm in our evaluations for object tracking that a direct Gaussian fit during uncertainty calibration is advantageous if subsequent applications (such as Kalman filtering in this case) also require normal distributions.
In our experiments, we observe a superior recalibration performance of our GP-Normal as it offers a higher flexibility in calibration compared to the standard Variance Scaling.
The multivariate GP-Normal with covariance estimation offers the best scores for the spatial quantile calibration (\ac{M-QCE}), tracking precision ($\text{MOTP}$), and does not lead to a deterioration of the semantic confidence calibration metrics as opposed to the remaining spatial calibration methods.
Furthermore, the reliability diagrams in \figref{fig:tracking:evaluation:spatial:reliability:regression} show a good calibration performance of the multivariate GP-Normal.
However, it does not lead to the desired improvement in the tracking performance as initially assumed.
As already mentioned, the loss in tracking accuracy is mainly caused by the higher amount of ID switches during object tracking.
In our experiments, we observe that the multivariate GP-Normal leads to better calibrated but also narrower prediction intervals.
We assume that less observations are assigned to existing tracks, leading to intermediate losses of interrelated object IDs.
In contrast, the non-parametric Isotonic Regression leads to wider prediction intervals which might be advantageous for the track assignment on the one hand.
On the other hand, this also leads to less tracks that are mostly tracked (cf. \figref{fig:tracking:evaluation:spatial:mota}) as well as to a lower tracking precision $\text{MOTP}$ which might be a hint that the recalibrated uncertainty is also not optimal.
Since the track association does also depend on the (static) system uncertainty, we assume that an adjustment of the system noise should be done in conjunction with the new spatial calibration methods.
This is subject of future work.

Therefore, we conclude that a calibration of the semantic confidence and spatial uncertainty leads to significant improvements in object tracking.
We can confirm our initial suggestions from \secref{section:regression:experiments} that a direct Gaussian fit during spatial recalibration is advantageous for subsequent applications such as Kalman filtering that require normal distributions as an interface.
Furthermore, our position-dependent semantic confidence calibration methods (cf. \secref{section:confidence:methods}) and especially the multivariate Histogram Binning showed superior performance in our evaluations compared to the standard calibration methods.

\section{Conclusion for Calibration in Object Tracking} 
\label{section:tracking:conclusion}

The goal of this chapter is to show a possible use-case of uncertainty calibration for subsequent applications after object detection.
For example, in the context of autonomous driving, it is important to not only detect objects but also to track their position over time.
This is a crucial part of image-based environment perception.
In this chapter, we therefore applied the methods for semantic confidence calibration from \chapref{chapter:confidence} and for spatial uncertainty calibration from \chapref{chapter:regression} to the task of object tracking.
For this reason, we have introduced the concept of tracking-by-detection where an object detector is used to obtain observations of possible objects within a single frame.
These observations were used in a recursive Bayesian filtering framework to construct object trajectories that are tracked over multiple subsequent frames.
An object tracking model consists of a position tracking as well as a track management with an assessment of the belief that a track is alive, i.e., the belief that it matches a real ground-truth object.
Therefore, we have firstly derived a belief model for the object existence that consumes the (calibrated) confidence information of the object detector to make assumptions of the probability for the track existence.
Second, we have introduced the concept of Kalman filtering which is an implementation of the recursive filtering using normal distributions.
The Kalman filter is used to track the position information of an object.
Furthermore, the Kalman filter is also capable to process the uncertainty which is inherent in the observations obtained by the object detector.
Thus, we used a probabilistic object detector (cf. \secref{section:introduction:spatial}) to obtain individual uncertainty quantifications for each observation.
In this context, we have applied our spatial uncertainty calibration methods from \chapref{chapter:regression} for uncertainty recalibration.
Therefore, we have been able to evaluate the effect of intermediate uncertainty calibration on the task of object tracking.

The evaluations show that both types of uncertainty calibration lead to significant improvements in the basic tracking performance as well as in the calibration properties of the track uncertainty.
We can show that semantic confidence calibration leads to improved tracking performance with some limitations regarding the increase of false negatives.
Furthermore, our position-dependent calibration methods and especially the multivariate Histogram Binning (cf. \secref{section:confidence:methods:binning}) showed superior performance.
Thus, the calibration of the semantic confidence had a positive influence on the management of individual tracks during object tracking.
Similarly, the methods for spatial uncertainty calibration also showed an improvement of the object tracking performance as well as the spatial uncertainty representation of the tracked objects.
In this case, the non-parametric Isotonic Regression \cite{Kuleshov2018} and especially our parametric GP-Normal (cf. \secref{section:regression:methods:parametric:gp}) showed superior performance when applied to the task of object tracking.
Thus, we can confirm our initial assumptions from \secref{section:regression:experiments} that Gaussian calibration methods such as Variance Scaling \cite{Levi2019,Laves2020} or our GP-Normal are advantageous if used in conjunction with subsequent applications that use normal distributions as input (such as Kalman filtering in our case).

Therefore, we conclude that uncertainty calibration in general has the potential to influence the object tracking positively
Furthermore, our extended position-dependent confidence calibration methods (cf. \secref{section:confidence:methods}) as well as our extended distribution-aware regression calibration methods (cf. \secref{section:regression:methods:parametric:gp}) were able to further improve the tracking performance as they are a valuable contribution to better reflect the uncertainty of individual samples.
For future applications, it might be interesting to investigate the effect of intermediate Bayesian confidence calibration (cf. \secref{section:bayesian:methods}) to the task of object tracking.
In this case, the object confidence is represented by a sample distribution obtained by the Bayesian calibration methods.
A possible use case might be to adapt a particle filter framework to directly track the confidence distribution over time.
For the track management, a track might then be discarded if the expected confidence falls below a certain threshold or if the variance of the confidence distribution gets too high.
We let this open for future research as it might be a valuable contribution to the context of object tracking.

\acresetall
\chapter{Conclusion}
\label{chapter:conclusion}

In this work, we evaluated the consistency of uncertainty quantification in the context of image-based object detection which is a part of environment perception, e.g., for safety-critical applications such as autonomous driving functions.
Especially for safety-critical applications, a reliable uncertainty assessment is of special interest.
Modern detection algorithms are based on neural networks that aim to identify multiple objects with their position, size, and their class within a single image.
However, it is a known issue that modern neural networks tend to produce either overconfident \cite{Guo2018} or underconfident \cite{Schwaiger2021} uncertainty estimations, depending on their architecture and use-case.
Therefore, we started by introducing the basic concepts of object detection and uncertainty quantification in \chapref{chapter:basics}.
Furthermore, we evaluated the semantic confidence (\chapref{chapter:confidence}) as well as the spatial uncertainty (\chapref{chapter:regression}) for consistency throughout this work which both represent the uncertainty in the class label and the spatial position uncertainty, respectively.
Both types of uncertainty are estimated by probabilistic object detection models \cite{He2019,Hall2020,Feng2021}.
Thus, our target was not to evaluate or improve the baseline detection performance (e.g., precision or recall) but to assess if the predicted uncertainties reliably represent the model's uncertainty which is equivalent to the observed error.
Besides measuring for uncertainty consistency, i.e., for calibration, we focused on methods which seek to correct possibly uncalibrated uncertainty estimates without the need of retraining a detection model.
Thus, it is possible to learn and apply a post-hoc remapping either of semantic or spatial uncertainty to achieve an improved representation of the observed error.
In the last \chapref{chapter:tracking}, we evaluated the effect of uncertainty calibration on a subsequent object tracking.
In contrast to object detection, a tracking model seeks to track the same object across a sequence of frames over time.
The object tracking is based on a mathematical framework in which we can integrate our proposed calibration methods.
Therefore, we have been able to evaluate our uncertainty calibration methods for the complete perception pipeline from object detection to object tracking.

\section{Summary of Semantic Confidence Calibration}

In \chapref{chapter:confidence}, we started with our evaluations for the calibration of the semantic confidence in the context of object detection.
Similar to the task of classification, an object detector provides a confidence score but for each detection individually.
This confidence can be interpreted as the model's belief of matching a real ground-truth object.
In a first step, we extended the definition of confidence calibration to the task of object detection and further introduced a dependency on the object position to the definition of calibration.
Thus, it is possible to measure the influence of the additional position information on the calibration of the confidence score.
Besides object detection, we also provided the respective calibration definitions for instance and semantic segmentation.
On this basis, we extended common calibration methods such as Histogram Binning \cite{Zadrozny2001}, Logistic Calibration \cite{Platt1999}, and Beta Calibration \cite{Kull2017} to also include additional position information into the recalibration of detection or segmentation models.
We extended these methods in a way so that they are capable of capturing possible correlations between confidence, position information, and miscalibration.
In the scope of semantic segmentation, we could not find a major improvement in calibration as the examined models already provide qualitatively well-calibrated confidence estimates.
In contrast, we have been able to significantly improve the calibration as well as the mask quality of instance segmentation models.
Furthermore, we found the examined object detection models to be miscalibrated which could be improved using our calibration methods.
Although we could only find a minor connection between position and miscalibration, our extended methods offered a qualitatively good calibration performance.
In \chapref{chapter:tracking}, we further evaluated the effect of (position-dependent) confidence calibration on the task of object tracking.
For this reason, we derived a framework to estimate the score for the object existence using the confidence provided by the underlying detection model.
When semantic confidence calibration was applied as an intermediate step, we observed that calibration in general and especially our position-dependent calibration methods are able to significantly improve the tracking performance as well as the intrinsic track calibration properties.
Therefore, we conclude that semantic confidence calibration in general and especially our position-dependent methods are able to enhance the tracking performance as well as the consistency of the track uncertainty which is a valuable contribution especially for safety-critical applications.

\section{Summary of Bayesian Confidence Calibration}

Basically, the methods for semantic confidence calibration are obtained by standard \ac{MLE}.
This approach yields deterministic parameters for the calibration functions.
However, especially during position-dependent calibration, it might occur that a new sample during inference is out of the known distribution that has been used to learn the calibration parameters within the training phase.
In \chapref{chapter:bayesian}, we therefore proposed the term of Bayesian confidence calibration that introduces additional epistemic (model) uncertainty into a calibration mapping.
Similar to Bayesian neural networks, we can construct a calibration function whose parameters are not deterministic but rather represented by probability distributions.
During inference, we can sample from these distributions to obtain a sample distribution for each calibrated estimate that represents the epistemic uncertainty of the calibration model.
In this context, we utilized \ac{SVI} to place variational Gaussian distributions over each calibration parameter and to learn the moments of the variational distribution during calibration training.
In this way, we are able to treat the calibration functions in a Bayesian way which allows to construct probability distributions for the calibrated confidences.
In our evaluations, we showed that a calibration method learned by \ac{SVI} is able to offer the same calibration performance compared to the standard methods learned by \ac{MLE}.
Furthermore, the Bayesian calibration models provide an additional uncertainty for the calibrated confidence.
We showed that this additional uncertainty might be used as a sufficient criterion to detect a possible covariate shift.
Therefore, Bayesian confidence calibration is valuable extension especially for safety-critical applications where a reliable uncertainty assessment of each component is of great importance.

\section{Summary of Spatial Uncertainty Calibration}

After our examinations for the consistency of semantic confidence, in \chapref{chapter:regression} we focused on the calibration properties of spatial uncertainty.
Similar to the semantic confidence, the spatial uncertainty should represent the observed error during inference.
In this context, we reviewed several definitions for regression uncertainty calibration and related them to each other.
Each of these definitions comes with its own metrics to measure miscalibration that we use in our evaluations later on.
Furthermore, we presented common calibration methods in the context of spatial uncertainty calibration that are divided into parametric and non-parametric methods.
The parametric calibration methods output a parametric probability distribution (commonly Gaussian) after calibration, whereas the non-parametric methods output a probability distribution of arbitrary shape.
Subsequently, we proposed an extended parametric calibration method GP-Normal that is capable to jointly recalibrate multiple dimensions in a single forward pass while capturing possible correlations between these dimensions.
In addition, we proposed a covariance estimation scheme which allows to introduce and model correlations between dimensions that have been learned independently from each other.
The experiments for the spatial calibration methods showed that the simple non-parametric Isotonic Regression \cite{Kuleshov2018} offers the best results when applied to a probabilistic object detector.
Additionally, our extended GP-Normal method has also been able to achieve a good calibration performance.
When applied to a subsequent object tracking in \chapref{chapter:tracking}, we confirmed these observations and further observed that especially the parametric Gaussian calibration methods achieve a good performance.
In these experiments, we have been able to confirm our initial assumptions that a direct modeling of a Gaussian distribution is advantageous if a subsequent application such as Kalman filtering also requires a parametric Gaussian as well.
Therefore, similar to the task of semantic confidence calibration, we conclude that spatial uncertainty calibration is a good way to improve the uncertainty consistency of a detection as well as a state estimation within object tracking.

\section{Final Remarks}

In this work, we evaluate the uncertainty of detection and tracking models before and after the application of post-hoc calibration methods.
In our experiments, we can show that each kind of calibration is a valuable contribution towards consistent uncertainty and thus to an improved assessment of the model's reliability.
Our extended calibration metrics and methods from chapters \ref{chapter:confidence} and \ref{chapter:regression} have shown to be a good contribution for the safety-relevant context.
In the final \chapref{chapter:tracking}, we have been able to show that the task of calibration is not only relevant for the object detection itself but also advantageous for subsequent applications such as object tracking.
Therefore, we conclude that calibration in general and especially our newly proposed methods are precious contributions within a safety-relevant context.



\clearpage

\appendix 

\clearpage

\pagenumbering{Roman}

\edef\intpagevalue{\getpagerefnumber{last-roman-page}}

\setcounter{page}{\number\numexpr\expandafter\arabicnumeral\expandafter{\intpagevalue}+1}


\bibliographystyle{ieeetr}
\bibliography{bib/bibliography}

\end{document}